\pdfoutput=1
\RequirePackage{fix-cm}
\documentclass[10pt,letterpaper]{survey-worldlines}
\usepackage[utf8]{inputenc}
\usepackage{pifont}
\usepackage{array,tabularx,longtable}
\usepackage{robot-icl-worldlines-layout}
\title{In-Context Learning for Robots: Methods and Applications}
\newcommand{\surveyauthors}{%
{\normalfont\rmfamily\bfseries\fontsize{9.6}{12}\selectfont\color{ink}
\newcommand{\authorsep}{,\penalty0\hspace{.15em}}%
\mbox{Haojian Huang\textsuperscript{1,2}}\authorsep
\mbox{Zexi Li\textsuperscript{1,3}}\authorsep
\mbox{Junhao Guo\textsuperscript{1}}\authorsep
\mbox{Yehang Zhang\textsuperscript{1,2}}\authorsep
\mbox{Wenxuan Peng\textsuperscript{1,4}}\authorsep
\mbox{Bohan Zhou\textsuperscript{1,3}}\authorsep
\mbox{Weilin Ruan\textsuperscript{1,3}}\authorsep
\mbox{Leyi Wu\textsuperscript{1,2}}\authorsep
\mbox{Chenxu Wang\textsuperscript{1,5}}\authorsep
\mbox{Jianchong Su\textsuperscript{1,2}}\authorsep
\mbox{Binghui Xie\textsuperscript{1,3}}\authorsep
\mbox{Wosong Chen\textsuperscript{1,2}}\authorsep
\mbox{Yingjie Xu\textsuperscript{1,2}}\authorsep
\mbox{Tianhao Zhou\textsuperscript{1,2}}\authorsep
\mbox{Suzeyu Chen\textsuperscript{1,2}}\authorsep
\mbox{Pukun Zhao\textsuperscript{1}}\authorsep
\mbox{Jiaqi He\textsuperscript{1}}\authorsep
\mbox{Xinyi Li\textsuperscript{1,3}}\authorsep
\mbox{Runze Li\textsuperscript{7}}\authorsep
\mbox{Peiran Dong\textsuperscript{1,3}}\authorsep
\mbox{Shaoxiang Dang\textsuperscript{1}}\authorsep
\mbox{Jing Huang\textsuperscript{1}}\authorsep
\mbox{Yingbing Chen\textsuperscript{1}}\authorsep
\mbox{Yifan Chang\textsuperscript{1}}\authorsep
\mbox{Tianyi Zhang\textsuperscript{1}}\authorsep
\mbox{Shiyuan Deng\textsuperscript{1}}\authorsep
\mbox{Haozhi Wang\textsuperscript{1}}\authorsep
\mbox{Yangkai Wei\textsuperscript{1}}\authorsep
\mbox{Wenqian Li\textsuperscript{1}}\authorsep
\mbox{Han Yang\textsuperscript{1}}\authorsep
\mbox{Kaiwen Zhou\textsuperscript{1}}\authorsep
\mbox{Huaping Liu\textsuperscript{5}}\authorsep
\mbox{James Cheng\textsuperscript{3}}\authorsep
\mbox{Rui Shao\textsuperscript{6}}\authorsep
\mbox{Donglin Wang\textsuperscript{7}}\authorsep
\mbox{Yaochu Jin\textsuperscript{7}}\authorsep
\mbox{Jianye Hao\textsuperscript{8}}\authorsep
\mbox{Ying-Cong Chen\textsuperscript{1,2,*}}\authorsep
\mbox{Yinchuan Li\textsuperscript{1,*}}\par}}
\newcommand{\surveyaffiliations}{%
{\normalfont\rmfamily\fontsize{8.7}{11}\selectfont\color[HTML]{404040}
\mbox{\textsuperscript{1}Knowin AI}\quad
\mbox{\textsuperscript{2}HKUST(GZ)}\quad
\mbox{\textsuperscript{3}The Chinese University of Hong Kong}\quad
\mbox{\textsuperscript{4}Tongji University}\\
\mbox{\textsuperscript{5}Tsinghua University}\quad
\mbox{\textsuperscript{6}Harbin Institute of Technology, Shenzhen}\quad
\mbox{\textsuperscript{7}Westlake University}\quad
\mbox{\textsuperscript{8}Tianjin University}\par}}

\newcommand{\surveycorrespondence}{%
{\normalfont\itshape\fontsize{9}{11}\selectfont\color[HTML]{404040}\textsuperscript{*}\,Corresponding authors.}}

\newcommand{\surveyresourcelinks}{%
  {\normalfont\sffamily\fontsize{10}{12}\selectfont
    \href{https://jethrojames.github.io/awesome-robots-icl/}{%
      \raisebox{-.12ex}{\includegraphics[height=1em]{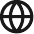}}\hspace{.45em}\textcolor{referenceblue}{\textbf{Project Page}}}%
    \hspace{2.4em}%
    \href{https://github.com/JethroJames/awesome-robots-icl}{%
      \raisebox{-.12ex}{\includegraphics[height=1em]{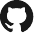}}\hspace{.45em}\textcolor{referenceblue}{\textbf{GitHub}}}%
  }%
}

\newcommand{\surveycontactdetails}{%
  {\normalfont\sffamily\fontsize{9.2}{12}\selectfont\color{ink}%
  \raisebox{-.17em}{\includegraphics[height=1.05em]{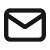}}\hspace{.5em}%
  {\fontsize{8}{10}\selectfont\bfseries CONTACT}\hspace{.9em}%
  {\normalfont\ttfamily\fontsize{9.2}{12}\selectfont
    \{\href{mailto:huanghaojian@knowin.ai}{huanghaojian},
    \href{mailto:lizexi@knowin.ai}{lizexi}\}}\hspace{.15em}%
  {\color{muted}@}\hspace{.3em}%
  \tcbox[on line,colback=ink,colframe=ink,boxrule=0pt,arc=2pt,
    boxsep=0pt,left=5pt,right=5pt,top=2pt,bottom=2pt]{%
    \normalfont\sffamily\bfseries\fontsize{9}{10}\selectfont\color{white}knowin.ai}%
  \par}}

\hypersetup{pdftitle={In-Context Learning for Robots: Methods and Applications},pdfsubject={Methods and applications of in-context learning for robots},pdfauthor={Haojian Huang, Zexi Li, Junhao Guo, Yehang Zhang, Wenxuan Peng, Bohan Zhou, Weilin Ruan, Leyi Wu, Chenxu Wang, Jianchong Su, Binghui Xie, Wosong Chen, Yingjie Xu, Tianhao Zhou, Suzeyu Chen, Pukun Zhao, Jiaqi He, Xinyi Li, Runze Li, Peiran Dong, Shaoxiang Dang, Jing Huang, Yingbing Chen, Yifan Chang, Tianyi Zhang, Shiyuan Deng, Haozhi Wang, Yangkai Wei, Wenqian Li, Han Yang, Kaiwen Zhou, Huaping Liu, James Cheng, Rui Shao, Donglin Wang, Yaochu Jin, Jianye Hao, Ying-Cong Chen, Yinchuan Li}}
\abstract{
General-purpose robots must infer what a new task requires and translate that understanding into appropriate physical action. In-context learning (ICL) for robots supports this process by using demonstrations and interaction to direct existing competence with neural parameters held fixed during deployment. We organize this literature review around the interfaces connecting contextual evidence to execution, distinguishing four families: context-conditioned policies, geometric demonstration transfer, world-model-based control, and skill- and agent-based execution. Comparing these interfaces clarifies their transfer assumptions and the roles of training, correspondence, and memory in making context useful. Across manipulation and navigation, we examine how these mechanisms preserve taught requirements as objects, environments, and execution conditions change. This analysis links method design to evaluation practices that distinguish responsiveness to teaching, physical transfer, and benefits from retained experience. The resulting agenda connects compositional task acquisition and faithful transfer with physical recursive self-improvement, in which experience improves the ability to learn subsequent tasks.
}
\begin{document}
\enlargethispage{36pt}
\vspace*{-32pt}
\maketitle
\par\noindent\begin{minipage}{\linewidth}
\setlength{\parskip}{0pt}
\noindent\makebox[\linewidth][c]{\includegraphics[width=0.87\linewidth]{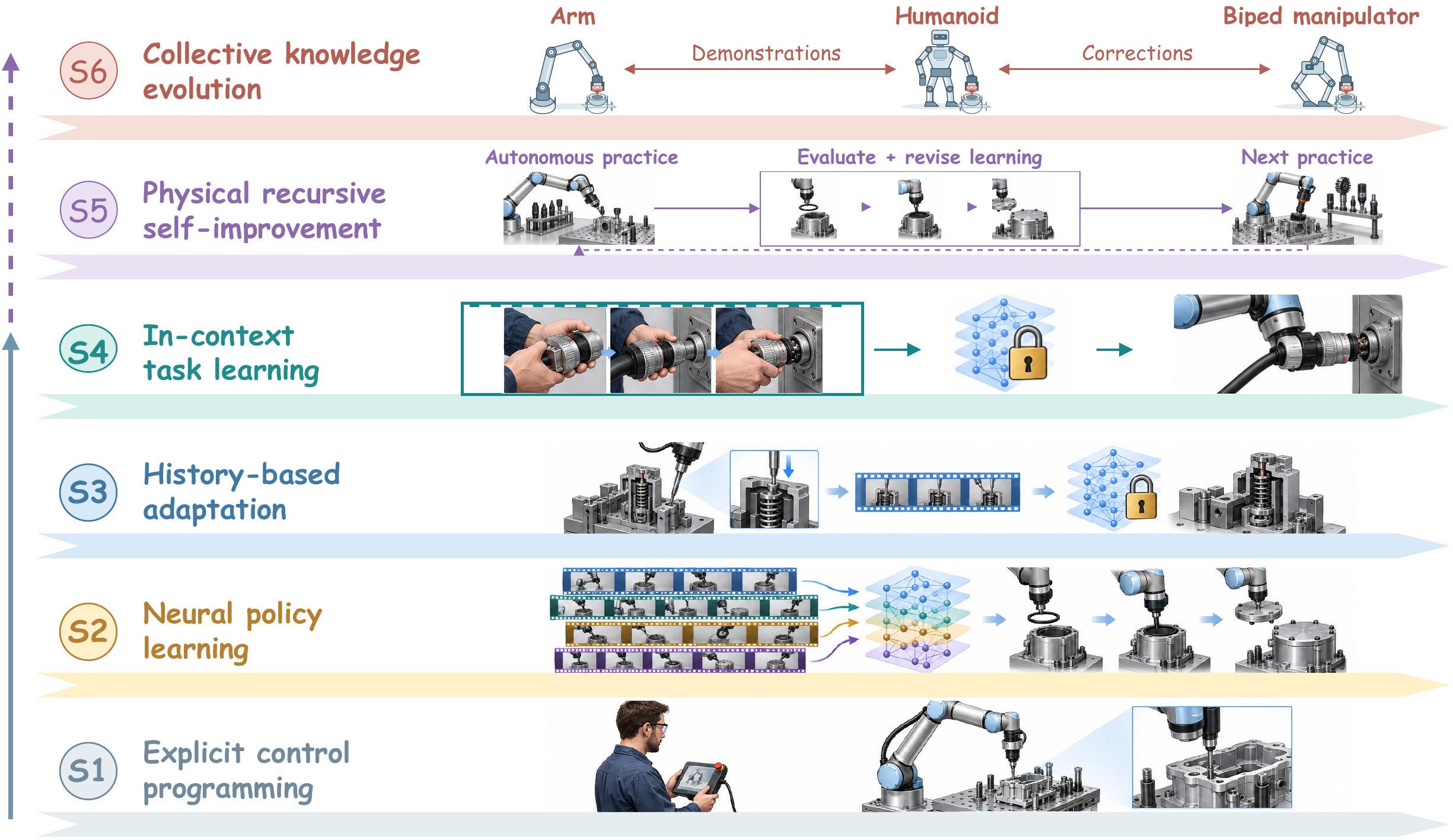}}\par
\captionsetup{type=figure}
\caption{Six learning horizons. S1 encodes control explicitly; S2 learns reusable policies. With neural parameters fixed at deployment, S3 adapts to interaction outcomes, while S4 infers task requirements from teaching. S5 improves how subsequent tasks are learned; S6 extends this process through validated knowledge exchange across embodiments.}\label{fig:cover}
\end{minipage}\par
\clearpage
\setcounter{tocdepth}{2}
\hypersetup{linktoc=all,bookmarksdepth=2}
\pdfbookmark[1]{Contents}{survey-contents}
\begingroup
\hypersetup{linkcolor=ink}
\makeatletter
\setlength{\parskip}{0pt}
\renewcommand{\l@section}[2]{%
  \addpenalty{-\@highpenalty}\addvspace{3.5pt plus 1fill}%
  {\bfseries\fontsize{10}{12}\selectfont
   \@dottedtocline{1}{0em}{1.8em}{#1}{#2}}}
\renewcommand{\l@subsection}[2]{%
  {\normalfont\fontsize{9.5}{11.7}\selectfont
   \@dottedtocline{2}{1.8em}{2.7em}{#1}{#2}}}
\vbox to \textheight{\tableofcontents\vskip0pt}
\makeatother
\endgroup
\clearpage
\pagestyle{surveybody}
\FloatBarrier
\suppressfloats[t]
\section{Introduction}
\label{sec:intro}
\nocite{duan2017oneshot,duan2016rl2,mishra2017snail,rakelly2019pearl,xu2022promptdt,laskin2022ad,finn2017mil,yu2018domainadaptive,merwe2025icpi,jiang2026robottt,arxiv260706988,mandi2021mosaic,fu2024icrt,jain2024vid2robot,dipalo2024keypoint,vosylius2024instantpolicy,sridhar2025ricl,yang2026icivla,mu2022domino,ni2023metadiffuser,goswami2025osviwm,he2026demojepa,chen2026host,zhou2026zerowam,arxiv260718840,xu2017ntp,liang2022codeaspolicies,singh2022progprompt,xu2023xskill,kim2025uniskill,chen2026showharness,argus2020flowcontrol,valassakis2022dome,dipalo2023dinobot,heppert2024ditto,dreczkowski2025mt3,wang2025odil,dong2026semancorr,kumar2021rma,liu2025locoformer,wang2026icwm,shi2025memoryvla,sridhar2025memer,torne2026mem,yin2026simplememvla,zha2023droc,lu2026aspire,chen2026architect,wang2026shaper,tziafas2024lrll,arxiv250918597,arxiv260619419,arxiv260817209,patel2026bpp,ding2026contextflow,walke2023bridge,oxe2023,khazatsky2024droid,fang2023rh20t,agibot2025,galaxea2025,chi2024umi,hoi4d2022,hot3d2025,egodex2025,ohkawa2026yubi,egoscale2026,grauman2021ego4d,damen2020epic100,grauman2023egoexo4d,goyal2017something,assembly1012022,holoassist2023,mandlekar2023mimicgen,robocasa2024,ahn2024autort,qian2026synthicl,sermanet2017tcn,zakka2021xirl,kedia2024rhyme,agia2024sentinel,xu2025faildetect,arxiv241104549,liu2023reflect,duan2024aha,ren2023knowno,dai2024racer,chen2026zeva,xiao2026enpire,arxiv260816590,wang2026instantfold,palma2026bicicle,chen2025manilong,cheng2026gptpolicyeval,arxiv260818618,tang2023saytap,wu2026adatracker,generalist2026gen15,skild2026s1,jiang2022vima,zhou2026imitator,ruoss2025lmact,dai2026robomme,sun2026memobench,gao2026gmp,zhou2025liberopro,arxiv260217659,arxiv260317300,arxiv260602277,arxiv260830536,chan2022distribution,vonoswald2022gradient,min2022metaicl,bousmalis2023robocat,ma2023eureka,ma2024dreureka,liang2024lmpc,kim2024openvla,black2024pi0,nvidia2025gr00t,brown2020fewshot,argall2009survey,ravichandar2020survey,dong2024iclsurvey,moeini2025icrlsurvey,ma2026humanvideosurvey,run2026nonstationary,hou2026worldmodel,wang2026vladata,domae2026embodimentgap,liu2022revolver,doshi2024crossformer,gupta2026kinematic,fikes1971strips,brooks1985subsumption,khatib1987operational,kaelbling2011hpn,marzinotto2014bt,levine2015visuomotor,kalashnikov2018qtopt,sharma2023medal,arxiv260829967,lesort2019continual,hospedales2022metalearning,hu2025agentmemory,yu2017uposi,zhou2019wtl,jang2022bcz,pari2021vinn,bahl2022whirl,fei2025generallevel,park2026recap,yin2024roboprompt,sridhar2024regent,park2025demodiffusion,zhang2024imop,james2018tec,dasari2020tosil,chen2025vivla,huang2026mint,she2026matchingpolicy,arxiv260805738,xu2026stellavla,chang2023awda,zhu2024orion,arxiv260322435,torabi2018bco,allu2025hrt1,vecerik2023robotap,vitiello2023pose,biza2023warping,tang2025functo,liu2024magic,defarias2025gift,zhu2026sparsedense,oh2026vlaff,wichitwechkarn2025annotationfree,survey2026wam,srinivas2018upn,li2025embodiedwm,nguyen2026iclr,son2026seetraceact,choi2026ponderpounce,huang2023voxposer,ahn2022saycan,huang2022monologue,physcap2026,feng2026regrind,fang2024kalm,haldar2025pointpolicy,li2025emp,yuan2024robopoint,qu2025spatialvla,chaumette2006visualservo,fan2026rtcf,li2020focal,yuan2022corro,li2026rememvla,cherepanov2026muvla,yang2026eventvla,oh2025rip,grigsby2023amago,elawady2024relic,shah2026halo,arxiv260826821,arxiv260829537,arxiv260708448,arxiv260821204,shi2026memoryvlapp,chen2026steerable,liu2026guava,galanti2026physicalagency,singh2024malmm,curtis2025proc3s,sarch2024ical,ju2026embodiskill,xie2026uniskillrepo,kumar2026aor,hu2026orchestrating,cui2026roboclaw,arxiv260724744,wang2026ego2robot,simpleai2026hifiumi,zhaxizhuoma2024fastumi,ha2024umilegs,luo2026omniumi,james2019rlbench,liu2023libero,gu2023maniskill2,chen2025robotwin2,vaswani2017attention,reddy2023abrupt,yu2026walloss05,dehaan2019causal,torne2025ptp,mtopt2021,bridge2021,agibot2026release,agibot2026corrections,robomind2024,robomind2025v2,robocoin2025,bharadhwaj2023roboagent,xu2026bimanualscaling,shah2025mimicdroid,li2026aceego,xiaomi2026robotics1,punamiya2026egoverse,aoe2026openaoe,cao2026acedata,wang2023mimicplay,openai2026astratraining,raventos2023diversity,ghasemipour2025selfimproving,qian2026robotok,lin2026simdex,zhou2025yoto,zhou2026bidemosyn,robosplat2025,bonardi2019humans,arxiv260509423,arxiv260831167,arxiv260304356,arxiv260519242,arxiv260401985,arxiv250706219,ross2011dagger,laskey2017dart,hoque2021thrifty,arxiv260313528,arxiv260421741,arxiv260819891,arxiv251114759,zhao2023act,octo2024,brohan2022rt1,brohan2023rt2,hpt2024,liu2024rdt,arxiv251222414,arxiv260415483,vuong2025actiontokenizer,fateh2026histat,arxiv260518746,arxiv260101075,arxiv260618960,arxiv260819059,arxiv260900619,arxiv260900950,arxiv260601247,arxiv260817129,mai2026crvlaforce,jain2025transitions,ebert2018retrying,bahety2024screwmimic,arxiv260623085,matsushima2020uncertainty,liang2024introplan,arxiv260825798,zhang2020metacure,cakmak2012questions,core2026realignment,arxiv260511951,chen2026volo,li2025iclhf,arxiv260818701,arxiv260514504,arxiv260814441,jiang2026benchmarkaudit,fei2025liberoplus,yu2019metaworld,gu2026roboreel,cherepanov2025mikasa,arxiv260813049,arxiv250617811,arxiv260816885,sato2026sail,xie2021bayesian,chen2024parallel,min2022demonstrations,garg2022functions,shen2023gradient,schaeffer2023mirage,chung2022flan,huang2026fastthinkact,duan2026rsisurvey,luo2025silvr,arxiv260212063,arxiv260206508,bi2026motus2,zala2024envgen,faldor2024omniepic,wagenmaker2025exploration,zintgraf2019varibad,arxiv260919824,arxiv260919315,arxiv260920791,arxiv260920388,arxiv260919796,arxiv260920648,arxiv260919906,arxiv260920659,arxiv260919512,arxiv260919554,arxiv260914633,simeonov2023rndf,thompson2026parttransfer,noematrix2026roborsi,chen2026eta,huang2026roboharness,zhou2024nolo,buoso2024select2plan,elnoor2024vlmgronav,anderson2018naveval,pathak2018zeroshot,kumar2018rpf,yoo2020sparsepath,savinov2018sptm,shah2022lmnav,chen2023a2nav,li2026cmmrvln,liu2026hamvln,arxiv260924411,arxiv260923432,arxiv260922966,arxiv260924778,arxiv260921740,arxiv260921122,arxiv260921229,jena2026weightsskills,lu2026wamsurvey,li2026demonstrationiclsurvey,shao2025vlasurvey,arxiv250912718,arxiv260618847,huang2026affordanceharness,wu2026robostressbench,huang2026vidpairhalluc,li2026roboharnessk1,li2026racap,pala2026membodied,you2026embodiedswe,liu2026navprobe,li2026talk2escape,zhang2026cma,wang2026tpflow,yi2026arms,zhang2026waa}
\nocite{knowin2026glow,arxiv260930249,arxiv260930134,arxiv260920646,arxiv260930092,arxiv260804933,arxiv260928798,arxiv260926408,arxiv260920820,arxiv260929166,arxiv260925636,arxiv260928952,arxiv260911561,arxiv260918016,arxiv260911308}
\nocite{arxiv260930828,arxiv260930404,arxiv260930428,arxiv260930233,arxiv260931337,arxiv260930594,arxiv260931112,arxiv260930889,arxiv260930608}

A robot may know how to move yet still need evidence about what to do. Demonstrations specify a fold or assembly order; corrections revise the procedure. Interaction reveals friction or misalignment. In navigation, earlier visits reveal locations outside the current view~\citep{elawady2024relic}. In-context learning (ICL) for robots studies how such evidence changes deployed behavior without another task-specific update to neural parameters.

Broad pretraining supplies reusable perception and control~\citep{kim2024openvla,black2024pi0,nvidia2025gr00t} from multi-task and multi-embodiment collections~\citep{oxe2023,agibot2025}. Yet the current scene can leave several procedures feasible, conceal an earlier event, or reveal little about an unfamiliar material's response. These are information gaps that broader motor competence alone cannot resolve. Fine-tuning incorporates new evidence into weights; ICL makes it available at the decision. Cross-object transfer tests whether teaching remains useful after replacing either or both interacting objects~\citep{simeonov2023rndf}. The task relation must survive while its physical realization changes.

Two historical roots motivate a broad notion of context: one-shot imitation infers intended behavior, while meta-reinforcement learning infers tasks or dynamics from outcomes~\citep{duan2017oneshot,duan2016rl2}. Sequence models retain trajectories and learning histories~\citep{xu2022promptdt,laskin2022ad}; multimodal prompting combines task specifications and sensorimotor examples~\citep{jiang2022vima,fu2024icrt}. With broader priors, teaching can direct generalist actions~\citep{generalist2026gen15,skild2026s1}, predicted futures~\citep{zhou2026zerowam}, or executable procedures~\citep{chen2026showharness}. Physical experiments supply evidence for revising execution~\citep{xiao2026enpire,chen2026zeva}. Language participates throughout as instructions, examples, corrections, and retained summaries. Figure~\ref{fig:cover} situates these capabilities within complementary horizons of adaptation and longer-term learning.

Context use depends on the relationships learned during training. GPT-3 demonstrated few-shot performance after autoregressive pretraining~\citep{brown2020fewshot}; controlled studies identify distributional conditions that promote using examples~\citep{chan2022distribution}. For robots, the critical relationship links earlier teaching or interaction to the later action it changes. Retaining a correction can extend that relationship across attempts~\citep{lu2026aspire,wang2026shaper}, provided its conditions remain valid. Broader motor competence, more informative teaching, and selective experience reuse therefore address different limits of adaptation.

Existing surveys organize learning from demonstration by teaching interfaces and learned policies, rewards, or plans~\citep{argall2009survey,ravichandar2020survey}. ICL and in-context RL reviews examine adaptation through examples and interaction~\citep{dong2024iclsurvey,moeini2025icrlsurvey}, including the validity of context under environmental change~\citep{run2026nonstationary}. In robotics, human-video reviews compare the information transferred from observation to control~\citep{ma2026humanvideosurvey}; manipulation ICL reviews organize context content, inference targets, adaptation mechanisms, and transfer~\citep{li2026demonstrationiclsurvey}. VLM-based VLA reviews distinguish monolithic and hierarchical integration of planning and action generation~\citep{shao2025vlasurvey}. Complementary reviews emphasize data and evaluation~\citep{wang2026vladata}, cross-embodiment adaptation~\citep{domae2026embodimentgap}, predictive control~\citep{hou2026worldmodel,survey2026wam}, and the acquisition and improvement of executable skills~\citep{jena2026weightsskills}.

Our organizing question is \emph{how new evidence resolves what existing competence leaves undetermined, and how that resolution survives physical execution}. The four families place this burden in different intermediates; correspondence, training, and memory determine whether those intermediates preserve the needed information. This connects the reason for using context to the mechanism and limits of transfer. Table~\ref{tab:related-surveys} in Appendix~\ref{app:related-surveys} compares related reviews, including recent world-action taxonomies~\citep{lu2026wamsurvey}.

Placing an object at the correct destination can still violate a required handle grasp. We therefore trace both the intended effect and prescribed order or contact through execution, developing three contributions:
\begin{itemize}
\setlength{\itemsep}{0pt}
\setlength{\parsep}{0pt}
\item A taxonomy of context-conditioned policies, geometric demonstration transfer, world-model-based control, and skill- and agent-based execution, with shared correspondence and memory.
\item An account of how training relationships establish context use and how object substitution, unfamiliar environments, and execution conditions limit transfer, separating broader motor competence from broader ability to learn through teaching.
\item A synthesis of reported comparisons and evaluation controls that separates context dependence, transfer, and retained-experience benefits, motivating compositional learning and improved teachability.
\end{itemize}

\begin{figure}[!tp]
\centering
\begin{minipage}{\linewidth}
\centering
\noindent\input{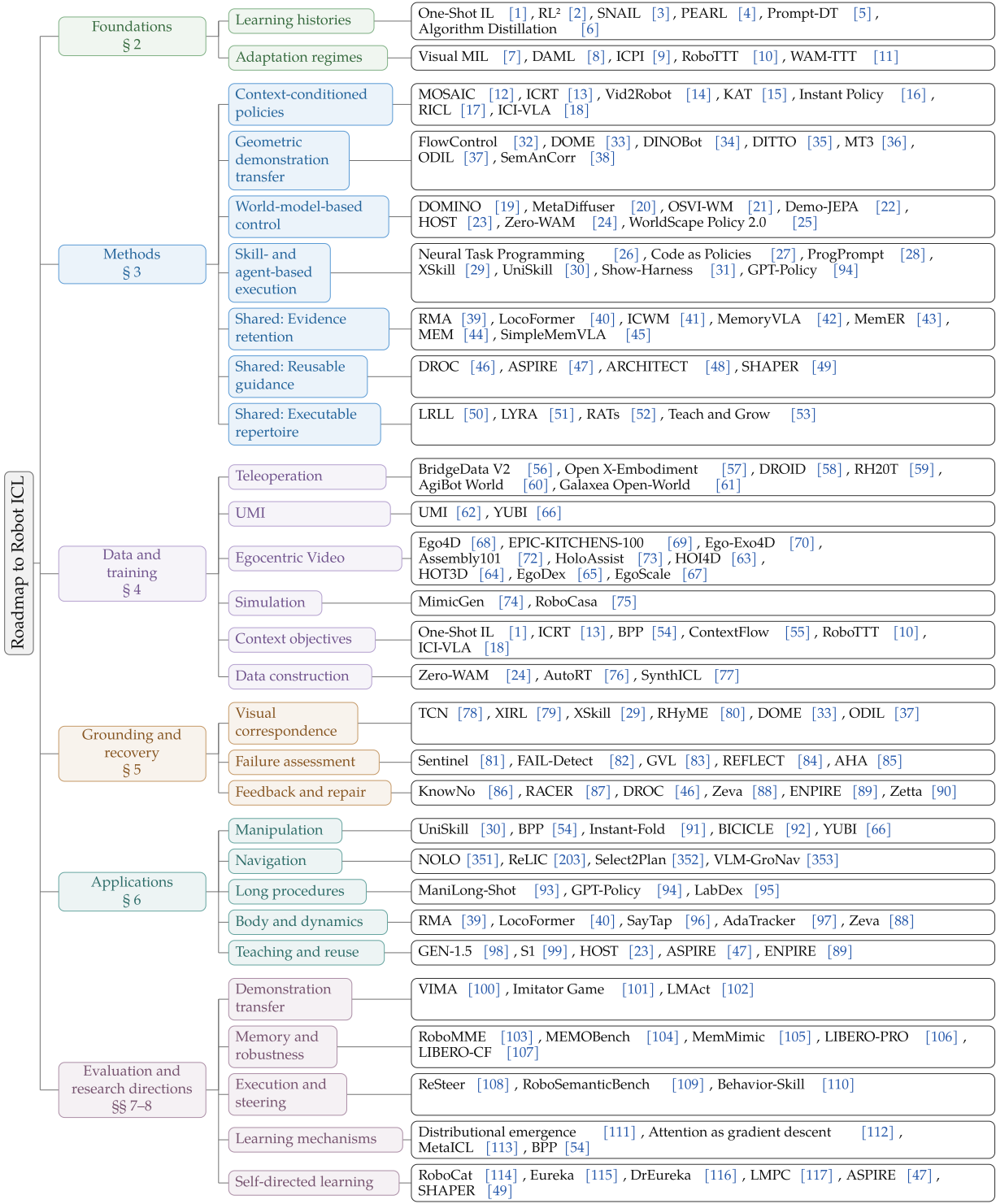}
\caption{Organization of the literature on ICL for robots. Four method families connect contextual evidence to behavior; the other branches organize foundations, data, grounding, applications, evaluation, and research directions. Correspondence and memory mechanisms connect methods across families.}
\label{fig:survey-roadmap}
\end{minipage}
\par\phantomsection\label{floatend:fig:survey-roadmap}
\end{figure}

\surveyanchor{discuss:fig:survey-roadmap}
Figure~\ref{fig:survey-roadmap} follows this argument: why a robot needs context (Section~\ref{sec:foundations}), how context changes action (Section~\ref{sec:methods}), and how that dependence is learned (Section~\ref{sec:data}). Recovery and applications expose its physical limits (Sections~\ref{sec:recovery}--\ref{sec:applications}); evaluation tests its benefit (Section~\ref{sec:evaluation}); the research agenda asks which limits must be overcome for broader transfer and improved learning (Section~\ref{sec:discussion}).

\suppressfloats[t]
\section{Foundations: Why Do Robots Need Context?}
\label{sec:foundations}
\newcommand{\CorpusTotal}{412}
\newcommand{\CorpusDated}{412}
\newcommand{\CorpusUndated}{0}
\newcommand{\CorpusBeforeRecent}{115}
\newcommand{\CorpusRecent}{297}
\newcommand{\CorpusCurrentYear}{197}
\newcommand{\CorpusPolicy}{123}
\newcommand{\CorpusGeometry}{24}
\newcommand{\CorpusWorld}{22}
\newcommand{\CorpusAgent}{91}
\newcommand{\CorpusMethods}{260}
\newcommand{\CorpusRelated}{152}

A packing robot may already grasp every part yet need a demonstration to establish their order, history to identify completed placements, and contact evidence to accommodate a tighter fit. The same visible scene can therefore require different actions. Useful context resolves this decision-relevant ambiguity through actions the robot can already execute. This distinction motivates the learning horizons, adaptation regimes, and four control interfaces below. Their history traces how robots acquired broader ways to interpret and act on new evidence.

\subsection{From competence to contextual adaptation}
\label{sec:learning-horizons}

We use six learning horizons to organize the relationship between prior competence, contextual adaptation, and learning from retained experience. S3 and S4 describe complementary sources of evidence; S5 and S6 express research objectives whose realization can combine several earlier capabilities. Together, the horizons connect distinct adaptation mechanisms to individual and collective learning objectives.

\begin{figure}[!tp]
\centering
\includegraphics[width=\linewidth]{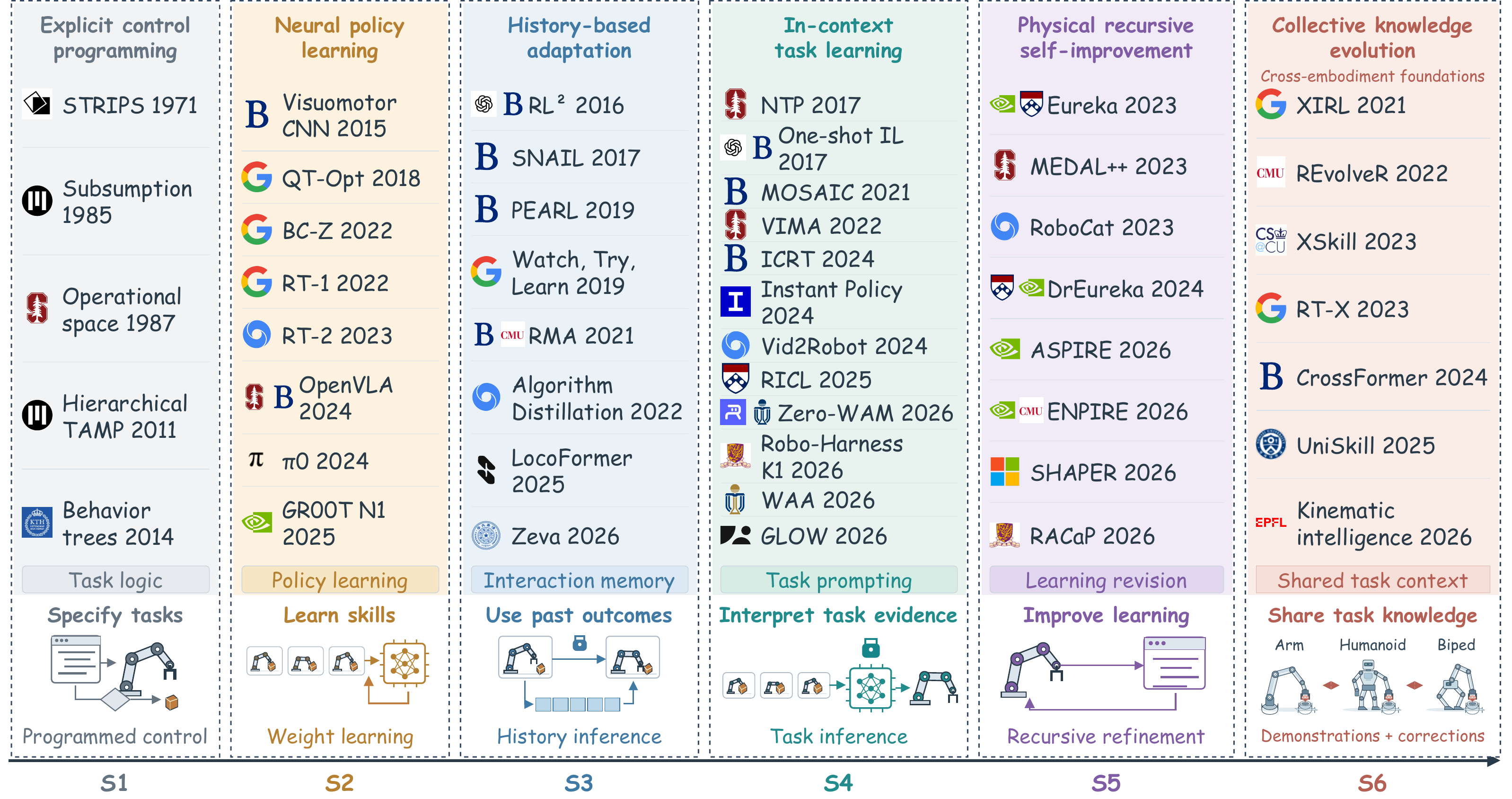}
\caption{Technical foundations for six learning horizons. S1--S2 establish programmed and learned competence. S3 and S4 adapt execution through interaction history and supplied teaching, respectively, with fixed neural parameters. The S5 and S6 columns collect mechanisms supporting improvements in subsequent learning and knowledge exchange across embodiments, including reusable procedures, shared representations, and policy transfer.}
\label{fig:stage-landscape}
\par\phantomsection\label{floatend:fig:stage-landscape}
\end{figure}

\surveyanchor{discuss:fig:stage-landscape}
The six horizons distinguish where learning changes the system (Figure~\ref{fig:stage-landscape}). Explicit control programming (S1) encodes task logic, control laws, and planning models~\citep{fikes1971strips,brooks1985subsumption,khatib1987operational,kaelbling2011hpn,marzinotto2014bt}; neural policy learning (S2) acquires reusable behavior through imitation and reinforcement~\citep{levine2015visuomotor,kalashnikov2018qtopt}. History-based adaptation (S3) uses interaction memory to infer task or dynamics information. In-context task learning (S4) infers new task requirements from supplied teaching, including instructions, demonstrations, textual examples, and corrections. Both forms of ICL for robots redirect execution with deployed neural parameters fixed.

In assembly, contact outcomes can reveal a tighter fit (S3), while a demonstration or correction can specify a new part order (S4). A retained correction may guide both interpretation and execution; these roles can share a model and input modality.

Automated practice~\citep{sharma2023medal,bousmalis2023robocat}, reward and experiment design~\citep{ma2023eureka,ma2024dreureka,xiao2026enpire}, and retained guidance~\citep{lu2026aspire,wang2026shaper} support a further objective: physical recursive self-improvement (S5), in which experience improves how the next task is learned. Collective knowledge evolution (S6) extends this learning cycle to a group of robots. Its premise is that an execution contains knowledge another body can use, even when the original motion cannot be copied. Shared progress and skill representations preserve what should be accomplished~\citep{zakka2021xirl,xu2023xskill,kim2025uniskill}; pooled policy learning supplies competence across sensing and action spaces~\citep{oxe2023,doshi2024crossformer}; morphology-aware transfer changes the physical realization~\citep{liu2022revolver,gupta2026kinematic}. These are complementary foundations for exchanging demonstrations and corrections. Whether such exchanges improve the group's subsequent learning is the S6 research objective in Figure~\ref{fig:cover}.

Physical self-improvement describes the feedback loop through which interaction changes later behavior. It can operate through retained context, executable programs, or neural updates. This is a cross-cutting process: S3 identifies history-based adaptation, S5 concerns improvements in subsequent learning, and S6 adds knowledge exchange between robots. Section~\ref{sec:icl-self-improvement} compares these update objects and their evaluation.

\begin{figure}[!tp]
\centering
\includegraphics[width=\linewidth]{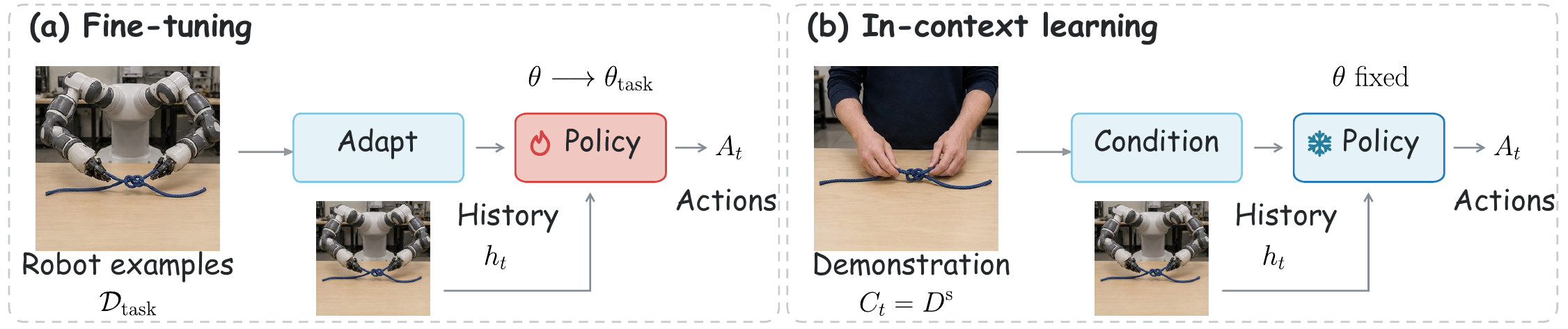}
\caption{Fine-tuning and in-context learning. (a) Task data $\mathcal{D}_{\mathrm{task}}$ update $\theta$ to $\theta_{\mathrm{task}}$. (b) A demonstration supplies $C_t=D^{\mathrm{s}}$ with $\theta$ fixed. Both policies use observation--action history $h_t$ to predict the next action block $A_t$.}
\label{fig:overview}
\par\phantomsection\label{floatend:fig:overview}
\end{figure}

\FloatBarrier
\surveyanchor{discuss:fig:overview}
New evidence changes either policy parameters or inference-time inputs (Figure~\ref{fig:overview}). Fine-tuning uses task data $\mathcal{D}_{\mathrm{task}}$ to update pretrained parameters $\theta$ to task-adapted parameters $\theta_{\mathrm{task}}$. The illustrated ICL case supplies a support demonstration $D^{\mathrm{s}}$ as context $C_t$, keeping $\theta$ fixed. Both use existing perception and motor competence; the storage of task information determines acquisition cost, persistence, and reset behavior.

\paragraph{What context enables a robot to learn.}
Fixed-parameter adaptation through supplied teaching or interaction is the review's central subject. Parameter-adaptation methods provide comparisons; pretrained policies and control mechanisms supply the competence on which adaptation operates. New evidence can specify a command, identify a familiar routine, compose known relations, or teach an unfamiliar rule. These behaviors differ in the information acquired from context and the prior competence used to realize it. Section~\ref{sec:reusable-learning-rule} develops these distinctions. The initial test is whether changing task-defining evidence changes behavior appropriately \mbox{under matched execution conditions.}

\subsection{Contextual and parameter adaptation}
\label{sec:adaptation-regimes}

Adaptation can reside in contextual state, a retained external artifact, or neural parameters. A common policy interface makes these alternatives comparable; their storage and reset behavior determine how an evaluation can identify the source of change. Table~\ref{tab:notation} collects the shared notation, separating supplied evidence, execution intermediates, and retained state.

\begin{table}[!tp]
\centering\surveytable
\caption{Core notation used throughout the survey. Groups follow the argument from contextual evidence to execution, retained knowledge, and evaluation. Superscripts $\mathrm{s}$ and $\mathrm{q}$ distinguish support and query; \mbox{other superscripts identify component roles.}}
\label{tab:notation}
\renewcommand{\arraystretch}{1.15}
\begin{tabularx}{\linewidth}{@{}>{\centering\arraybackslash}p{.255\linewidth}Y@{}}
\toprule
\rowcolor{black!12}\textbf{Notation} & \textbf{Meaning}\\
\midrule
\rowcolor{panel}\multicolumn{2}{c}{\strut\itshape Interaction and contextual evidence (Section~\ref{sec:foundations})}\\
\rowcolor{black!3}$t$, $n$ & Decision step; complete-attempt index.\\
$i$, $j$, $\ell$, $b$ & Demonstration element; memory record; skill; motion segment.\\
\rowcolor{black!3}$o_t$, $a_t$, $h_t$ & Observation; executed command; observation--action history.\\
$A_t$, $\widehat a_{t+\omega\mid t}$, $H$ & Proposed action block; action at offset $\omega$; block horizon.\\
\rowcolor{black!3}$D^{\mathrm s}$, $L$, $d_i^{\mathrm s}$ & Support demonstration; its length; its $i$th element.\\
$C_t$, $C_t^{\mathrm{task}}$, $r_t$ & Policy context; supplied task evidence; interpreted representation.\\
\rowcolor{black!3}$\theta$, $\theta_0$ & Deployed neural parameters; initialization of the adaptation phase.\\
$\bar\theta$, $\phi_t$ & Frozen backbone; adaptive parameter subset in test-time training.\\
\rowcolor{panel}\multicolumn{2}{c}{\strut\itshape Context-to-action interfaces (Sections~\ref{sec:direct-methods}--\ref{sec:structured-methods})}\\
\rowcolor{black!3}$\pi_\theta$ & Conditional distribution of the proposed action block.\\
$E_\theta^{\mathrm c}$, $E_\theta^{\mathrm m}$ & Context interpreter; demonstration motion-descriptor extractor.\\
\rowcolor{black!3}$q_\theta^{\mathrm a}$, $q_\theta^{\mathrm v}$ & Action generators conditioned on context or a predicted future.\\
$v^+$, $p_\theta^{\mathrm v}$, $p_\theta^{\mathrm d}$ & Anticipated future; context-conditioned predictor; action-conditioned dynamics.\\
\rowcolor{black!3}$g$, $f_\theta$, $p_\theta^{\mathrm g}$ & Execution specification; deterministic or stochastic specification generator.\\
$\kappa$, $k_\theta$, $\kappa_{B_n}$ & Deterministic, stochastic, and artifact-dependent executors.\\
\rowcolor{black!3}$z_\ell$, $N_g$, $\ell_t$ & Skill specification; number of skills; active skill index.\\
$\zeta_i$, $\mathcal B$ & Motion descriptor; robot kinematics and control description.\\
\rowcolor{panel}\multicolumn{2}{c}{\strut\itshape Memory and external revision (Sections~\ref{sec:adaptation-regimes}, \ref{sec:context-mechanisms}--\ref{sec:agentic-adaptation})}\\
\rowcolor{black!3}$U_\theta$ & Update of a self-contained context from an observed action transition.\\
$m_t$, $M_t$, $R_t$ & Recurrent state; external archive; retrieved records.\\
\rowcolor{black!3}$x_t$, $\xi_j$, $J_t$, $N_t$ & Executed transition; archive record; selected indices; archive size.\\
$B_n$, $\widetilde B_{n+1}$ & Retained external artifacts; proposed revision.\\
\rowcolor{black!3}$\mathcal T_n$, $\chi_n$ & Attempt trace with feedback; revision-acceptance indicator.\\
\rowcolor{panel}\multicolumn{2}{c}{\strut\itshape Training and evaluation (Sections~\ref{sec:data}, \ref{sec:evaluation})}\\
\rowcolor{black!3}$\psi$, $\tau^{\mathrm q}$, $P_{\mathrm{pair}}$ & Task specification; query trajectory; support--query pair distribution.\\
$I^{\mathrm q}$, $\ell_\theta$, $\mathcal L$ & Supervised query positions; per-target action loss; expected training loss.\\
\rowcolor{black!3}$S_n$, $\Delta_n^{\mathrm{reuse}}$ & Success/adherence rate; gain from retained evidence at attempt $n$.\\
\bottomrule
\end{tabularx}
\par\phantomsection\label{floatend:tab:notation}
\end{table}

\begin{figure}[t]
\centering
\includegraphics[width=\linewidth]{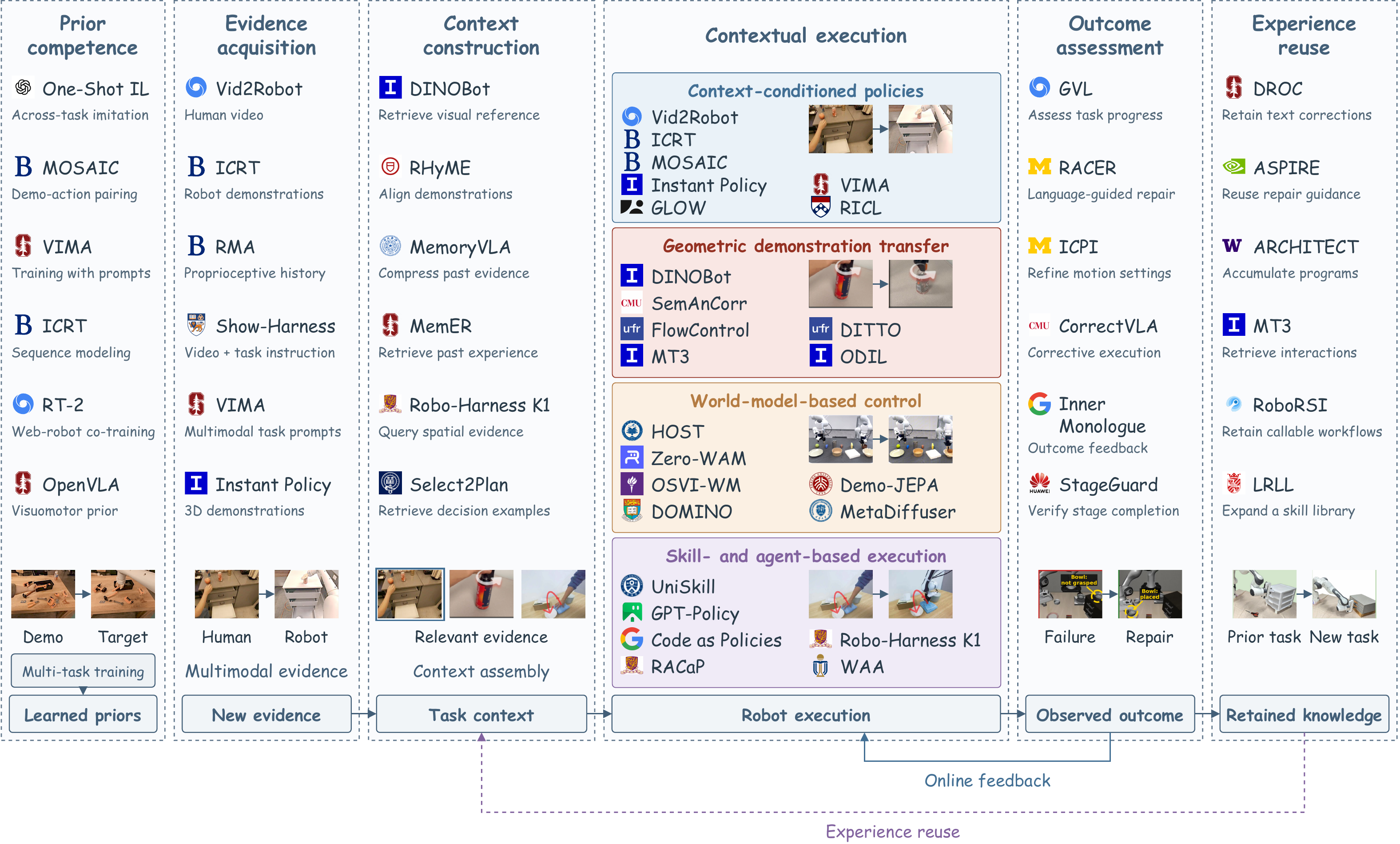}
\caption{The workflow of ICL for robots. Prior training supplies competence; deployment acquires evidence, constructs context, and generates actions through four method families. Outcome assessment provides online feedback to execution, while retained experience informs subsequent context construction.}
\label{fig:complete-chain}
\par\phantomsection\label{floatend:fig:complete-chain}
\end{figure}

At decision step $t$, the current attempt has history $h_t=(o_1,a_1,\ldots,a_{t-1},o_t)$. Observation $o_t$ includes the available images, proprioception, force, tactile or inertial measurements, and feedback messages. Action $a_t$ is the applied command; its measured physical effect appears in subsequent observations. Task-relevant context $C_t$ may contain instructions, goal references, multimodal support demonstrations $D^{\mathrm{s}}$, textual input--output examples, corrections, retrieved episodes, or memory of interaction. Superscript $\mathrm{s}$ denotes support evidence; the demonstration-only case in Figure~\ref{fig:overview} sets $C_t=D^{\mathrm{s}}$. Context can summarize the current history or retain information from earlier attempts. The policy proposes an action block $A_t=(\widehat a_{t\mid t},\ldots,\widehat a_{t+H-1\mid t})$ of horizon $H$, where $\widehat a_{t+\omega\mid t}$ is the action proposed at decision $t$ for offset $\omega$. With deployed neural parameters $\theta$,
\begin{equation}
 A_t \sim \pi_\theta\!\left(\,\cdot\mid h_t,C_t\right).
 \label{eq:policy}
\end{equation}
Only executed actions enter the subsequent history; new observations can replace the unexecuted part of a proposed block. When $C_t$ itself contains the state needed for future decisions, a memory update incorporates an action and its observed consequence:
\begin{equation}
 C_{t+1}=U_\theta(C_t,o_t,a_t,o_{t+1}).
 \label{eq:memory}
\end{equation}

Equation~\eqref{eq:memory} updates a self-contained context state. For context retrieved or compressed from experience, Equation~\eqref{eq:context-state} distinguishes the stored memory from the evidence \mbox{supplied to the policy.}

The policy interface is compatible with several architectures. Throughout the survey, $\theta$ denotes the collection of deployed neural parameters, including context interpretation, action generation, and any intervening learned model. A subscript $\theta$ on a component means that it uses the relevant subset of this collection; different components need not share weights. During fixed-parameter adaptation,
\begin{equation}
 \theta_{t+1}=\theta_t=\theta_0,
 \qquad h_{t+1}=(h_t,a_t,o_{t+1}).
 \label{eq:fixed-update}
\end{equation}
The subscript $0$ denotes the start of the deployment adaptation phase. Observations, retrieved examples, recurrent state, or stored experience change instead. Equations~\eqref{eq:memory} and \eqref{eq:fixed-update} use one executed action per update; if several actions of $A_t$ are executed before replanning, their observed transitions are incorporated in order. A reward or other feedback signal can be included when available. Updating a neural component at deployment constitutes parameter adaptation, even when the rest of the system stays frozen.

\FloatBarrier
\surveyanchor{discuss:fig:complete-chain}
Figure~\ref{fig:complete-chain} follows task information from prior competence to experience reuse. In a packing task, prior competence supplies grasps and motions. Evidence acquisition provides an instruction or demonstration specifying the part order. Context construction selects the relevant steps and relates them to the current scene. Contextual execution turns that evidence into actions; outcome assessment checks the resulting placements. Experience reuse makes verified outcomes and corrections available to later decisions. The four method families differ in how contextual execution converts evidence into control, while the surrounding operations determine which evidence it receives and whether the resulting behavior succeeds.

A corrected order changes the required actions; recorded placements identify the remaining work. Both can affect a fixed policy, but the retained information and its reset rules determine how long the change persists.

\begin{figure}[!tp]
\centering
\includegraphics[width=\linewidth]{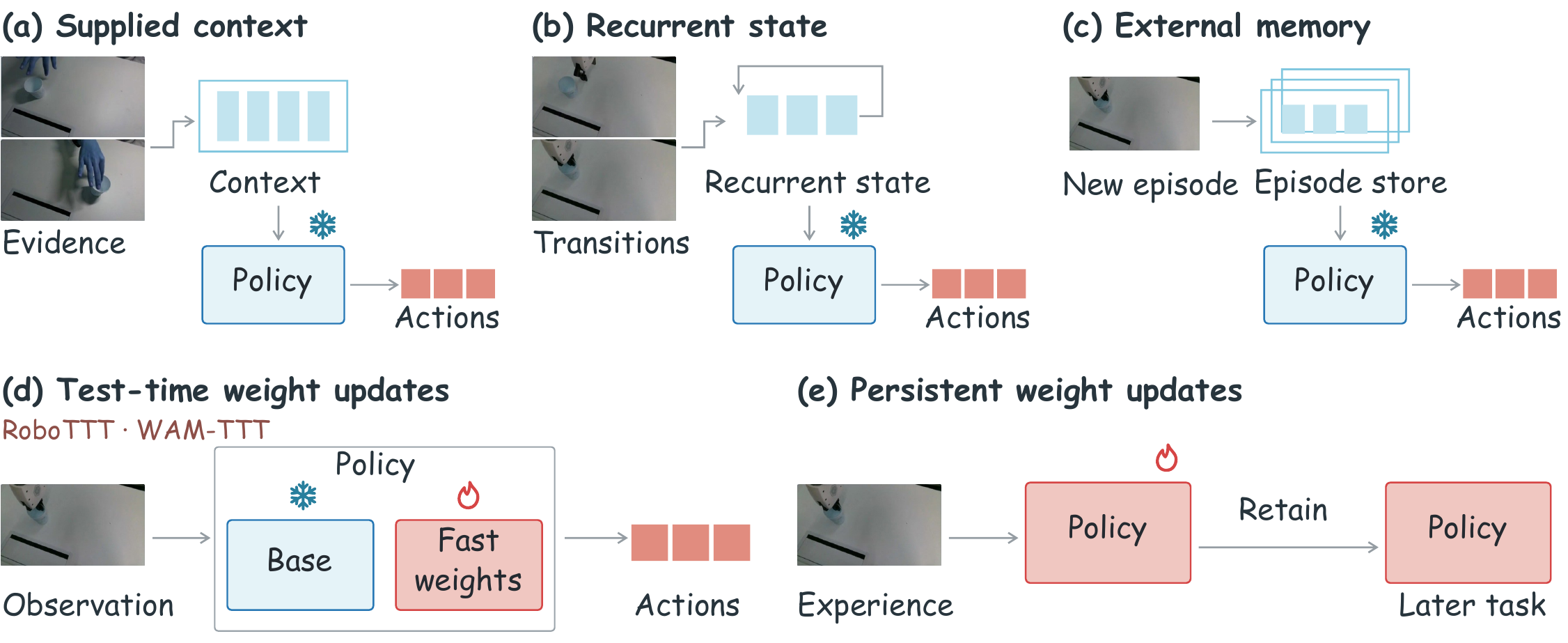}
\caption{Five locations and timescales of adaptation. In (a--c), supplied context, recurrent state, or retrieved episodes condition a policy with fixed neural parameters. In (d), fast weights change while the base remains fixed; (e) retains updated weights for later tasks. Blue borders and snowflakes mark fixed parameters; red borders mark updated parameters, with flames indicating the update stage.}
\label{fig:adaptation-state}
\par\phantomsection\label{floatend:fig:adaptation-state}
\end{figure}

A fixed model can revise a goal, skill sequence, or program after feedback; an executor turns that specification into actions. For example, a language model can use prior errors to adjust a motion primitive's settings~\citep{merwe2025icpi}. Retaining the revised program lets later attempts reuse the correction without training neural weights. This stores task information in executable content, whereas fine-tuning stores it in model parameters.

The surrounding execution system, or \emph{harness}, assembles context, exposes robot tools, executes requests, and returns feedback. Normally, this machinery remains fixed while the supplied evidence and generated requests change. A learning system can also revise its retrieval rules, context-construction code, or executable controller. We call this \emph{model-external adaptation}: experience changes the machinery connecting a model to the robot. Identifying this revised component distinguishes changes in the execution system from changes \mbox{in the neural model.}

\FloatBarrier
\surveyanchor{discuss:fig:adaptation-state}
Figure~\ref{fig:adaptation-state} distinguishes three carriers used by fixed models: supplied evidence in (a), recurrent state in (b), and retrieved episodes in (c). The lower row introduces parameter changes: fast weights adapt during deployment in (d), and updated weights persist into later tasks in (e). Storage and retention timescale are independent: an external memory can persist without training, and a test-time update can be either reset or retained. Continual learning studies acquisition and retention across changing experience~\citep{lesort2019continual}.

These distinctions determine the evaluation controls. Removing supplied evidence tests (a); clearing recurrent state tests (b); suppressing episode retrieval tests (c). Restoring pre-adaptation weights tests the contribution of parameter updates in (d--e). Section~\ref{sec:evaluation} develops these interventions under matched task conditions, so gains can be attributed to the information or parameters that actually changed.

One-shot specifies the amount of teaching, not its storage mechanism. Visual meta-imitation uses one demonstration for task-specific gradient updates~\citep{finn2017mil}. RoboTTT restricts updates to a neural subset during long-context execution~\citep{jiang2026robottt}; WAM-TTT adapts video-side fast weights from target-domain human video before rollout~\citep{arxiv260706988}. Both follow the selective-update principle in (d), despite different update timing. Thus one demonstration can support either fixed-weight conditioning or parameter adaptation. The next taxonomy distinguishes how the resulting context reaches robot actions.

\subsection{Four method families and shared design dimensions}
\label{sec:family-dimensions}

The primary taxonomy follows the \emph{deployed path from context to action}. The four families below are alternative computational interfaces, distinguished by the intermediate that execution consumes. We then separate this classification from the cross-cutting roles, modalities, and storage of evidence, before resolving \mbox{boundaries between hybrid implementations.}
\begin{itemize}
\item \textbf{Context-conditioned policies} infer actions from task evidence, including demonstrations, language, and interaction history as supported, through a learned generator or local action retrieval.
\item \textbf{Geometric demonstration transfer} reconstructs a demonstrated motion or contact reference, aligns or retargets it to the current scene, and executes that reference through tracking or replay.
\item \textbf{World-model-based control} predicts future observations, states, or transitions and uses those predictions to decode, plan, or select actions.
\item \textbf{Skill- and agent-based execution} infers skill choices, programs, or robot-tool requests for a separate executor; VLM and LLM agents form one branch.
\end{itemize}

\begin{figure}[!tp]
\centering
\includegraphics[width=\linewidth]{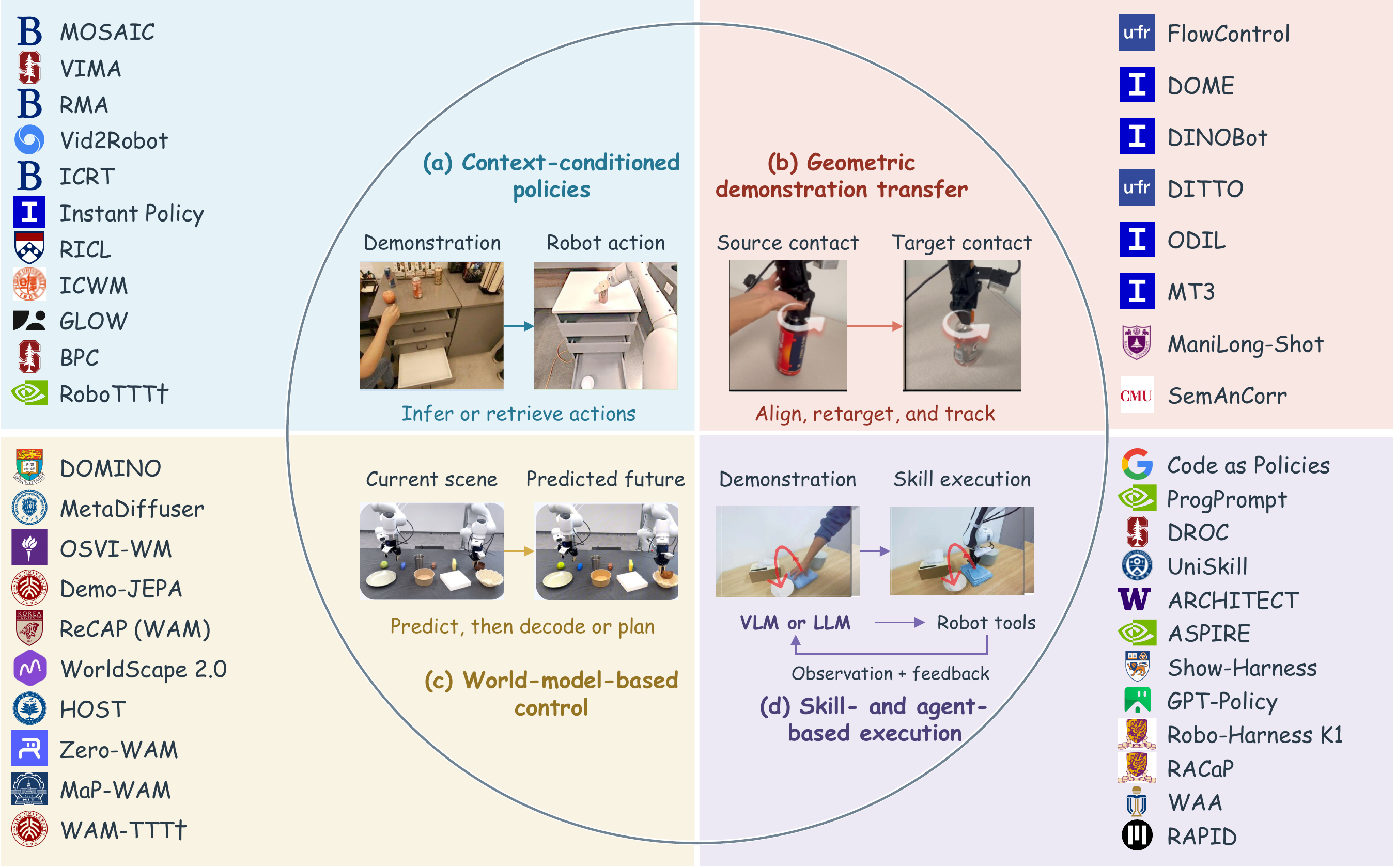}
\caption{Four context-to-control mechanisms: (a) action inference, (b) motion transfer, (c) future prediction, and (d) skill/tool selection. Surrounding works locate representative methods; $\dagger$ marks fast-weight adaptation in RoboTTT and WAM-TTT~\citep{jiang2026robottt,arxiv260706988}.}
\label{fig:design-dimensions}
\par\phantomsection\label{floatend:fig:design-dimensions}
\end{figure}

These distinctions concern deployed computation, building on the mappings, system models, and plans distinguished in learning from demonstration~\citep{argall2009survey}. They also preserve the separation between representation, optimization, and training objective emphasized in meta-learning~\citep{hospedales2022metalearning}. Transformers, diffusion models, and VLMs each support multiple interfaces. Classification follows their deployed computation: hybrid systems can combine predicted futures with skill execution, or programs with reference tracking.

Four information roles explain what this context contributes. \emph{Task specification} identifies the intended goal, relation, or procedure. \emph{Correspondence} relates demonstrated objects, contacts, phases, or commands to their executable counterparts. \emph{Physical response} describes how the current body and environment respond to action. \emph{Execution state} records the scene, relevant hidden events, and completed prerequisites. A demonstration can specify a procedure and its contact geometry; an interaction can update both physical response and execution state. These roles guide evidence selection within every method family.

Language can state a goal, illustrate a convention, revise a procedure, or summarize an earlier event. Images and sensor readings can supply the same information roles, including through numerical states serialized as text. The learning question concerns what the evidence lets the robot infer; its encoding determines how that information reaches the model. Thus modality describes how evidence is encoded, while the learning mechanism explains how that evidence changes a decision.

Data acquisition and adaptation describe two further dimensions. \emph{Teleoperation, UMI, egocentric video, and simulation} identify how records are collected (Section~\ref{sec:input-data}); visual, proprioceptive, contact, and action channels specify their sensorimotor contents. Language can accompany any source as instructions, examples, feedback, or memory; it is a cross-cutting context modality. \emph{Contextual adaptation} changes the information or external artifacts used by fixed neural components, whereas \emph{parameter adaptation} updates a neural component. Application demands describe what new evidence must resolve: task intent, unfamiliar spatial structure, physical response, or their combination. Thus a method has a control interface, uses evidence with identifiable roles, and operates under a stated update regime.

\FloatBarrier
\surveyanchor{discuss:fig:design-dimensions}
These dimensions let the same demonstration support different control computations. It can condition action inference, define a motion reference, constrain a predicted future, or specify a skill request (Figure~\ref{fig:design-dimensions}). What distinguishes the families is the intermediate required by execution. How that evidence was acquired or remembered remains a separate design choice, which is why retrieval and memory \mbox{can support several families.}

Storage adds another independent choice. A context window, recurrent state $m_t$, or external archive $M_t$ can supply the same task information with different access costs and persistence. Static API documentation establishes the command interface; task-specific examples and interaction supply changing evidence. For later comparisons with test-time training, write $\theta_t=(\bar\theta,\phi_t)$: $\bar\theta$ is the frozen backbone and $\phi_t$ the adaptive neural subset. This follows the update-regime distinction in Section~\ref{sec:adaptation-regimes}.

The boundary between policy conditioning and geometric transfer is the object being executed. Matching points to condition a diffusion policy remains context-conditioned policy inference; transferring a recorded object-relative trajectory and tracking it is geometric demonstration transfer. Similarly, estimating current dynamics for an action policy differs from predicting future consequences for control. RMA and the physical-context inference in ICWM illustrate the former~\citep{kumar2021rma,wang2026icwm}; DOMINO and Zero-WAM illustrate the latter~\citep{mu2022domino,zhou2026zerowam}. An auxiliary future-prediction head contributes to training; a deployed future predictor contributes to world-model-based control when its output guides action generation or selection.

The policy and agent interfaces can use the same VLM backbone. A VLM that emits motor-action tokens can implement a context-conditioned policy. A VLM that selects robot tools, supplies their arguments, and interprets returned execution feedback implements skill- and agent-based execution. GPT-Policy's Astra-based configuration belongs to this agent branch~\citep{cheng2026gptpolicyeval}. Demonstrations and interaction feedback supply the task-specific evidence for its tool choices and arguments.

Memory connects these interfaces across time. Alongside its form, function, and dynamics~\citep{hu2025agentmemory}, a retained fact carries physical conditions that determine when it remains applicable to the current robot and scene. Section~\ref{sec:context-mechanisms} examines access and update mechanisms, and Section~\ref{sec:recovery} examines how execution \mbox{validates or revises retained evidence.}

\emph{Intent adaptation} changes the desired relation; \emph{physical adaptation} changes its realization under new dynamics or embodiment. \emph{Joint adaptation} requires both. Navigation additionally needs spatial knowledge even when the goal and dynamics remain unchanged. Section~\ref{sec:applications} organizes applications by these information needs and their implications for long procedures and task changeover. Any of the four computational families \mbox{can serve several demands.}

\begin{table}[!tp]
\centering\surveytable
\caption{Two levels of the taxonomy. (A) Four method families differ in their execution intermediate and a key requirement for transfer. (B) Shared mechanisms establish correspondence and manage evidence within or across these interfaces. A system can combine several interfaces and shared mechanisms.}
\label{tab:methodological-taxonomy}
\begin{tabularx}{\linewidth}{@{}>{\raggedright\arraybackslash}p{.36\linewidth}Y Y@{}}
\toprule
\tablegroup{3}{A. Control interfaces: how context determines actions}
\textbf{Method family} & \textbf{Intermediate} & \textbf{Transfer requirement}\\
\midrule
Context-conditioned policies & Action distribution & Action inference preserves the taught distinction\\
Geometric demonstration transfer & Motion or contact reference & Matched interaction remains applicable\\
World-model-based control & Predicted consequences & Predicted evolution is realizable\\
Skill- and agent-based execution & Execution specification & Available skills preserve task constraints\\
\midrule
\tablegroup{3}{B. Shared mechanisms: how evidence supports those interfaces}
\textbf{Mechanism} & \textbf{Information resolved} & \textbf{Example}\\
\midrule
Geometric and functional correspondence & Where to interact & Handle contact on a new object\\
Semantic correspondence & Which roles and order & Cup as source; bowl as receiver\\
Temporal alignment & Which demonstration phase & Current grasp phase\\
Retrieval & Which experience applies & Relevant past demonstration\\
Evidence retention & What remains available & Earlier observation or outcome\\
External knowledge reuse & What guides later attempts & Repair guidance or reusable skill\\
\bottomrule
\end{tabularx}
\par\phantomsection\label{floatend:tab:methodological-taxonomy}
\end{table}

\surveyanchor{discuss:tab:methodological-taxonomy}
Table~\ref{tab:methodological-taxonomy} links each control intermediate to its transfer requirement, then separates the mechanisms shared across interfaces. Matching a handle establishes correspondence; that match may condition an action generator or define an explicit motion reference. The execution interface determines the transfer assumption carried into the final motion. Organizing methods by that interface connects shared evidence-processing mechanisms to the physical behavior they support.

\subsection{Historical development}
\label{sec:historical-development}

The historical account follows an expansion in what must be inferred from context. Task-family adaptation learned reusable responses to bounded variation; sequence interfaces made more of the preceding experience accessible; broader priors increased the behaviors that new teaching could direct. These developments overlap across the four families. The final publication census locates their recent growth within \mbox{the literature reviewed here.}

\paragraph{Task-family adaptation.}
Early contextual policies amortized adaptation across a task distribution: training learned how to interpret evidence so that deployment could change behavior without relearning the entire controller. In one-shot imitation, paired executions taught how one instance specifies actions in another~\citep{duan2017oneshot}. In meta-RL, recurrent state carried adaptation across episodes~\citep{duan2016rl2}; probabilistic task inference separated evidence interpretation from control~\citep{rakelly2019pearl} and made task uncertainty available to action selection~\citep{zintgraf2019varibad}. The resulting adaptation rule inherits the distinctions made available during training. Extending it to human demonstrations requires correspondence across bodies~\citep{yu2018domainadaptive}; extending it to changed robot dynamics requires evidence about action response~\citep{kumar2021rma}. These extensions concern different unknowns, despite sharing rapid adaptation as their objective.

\begin{figure}[!tp]
\centering
\includegraphics[width=\linewidth]{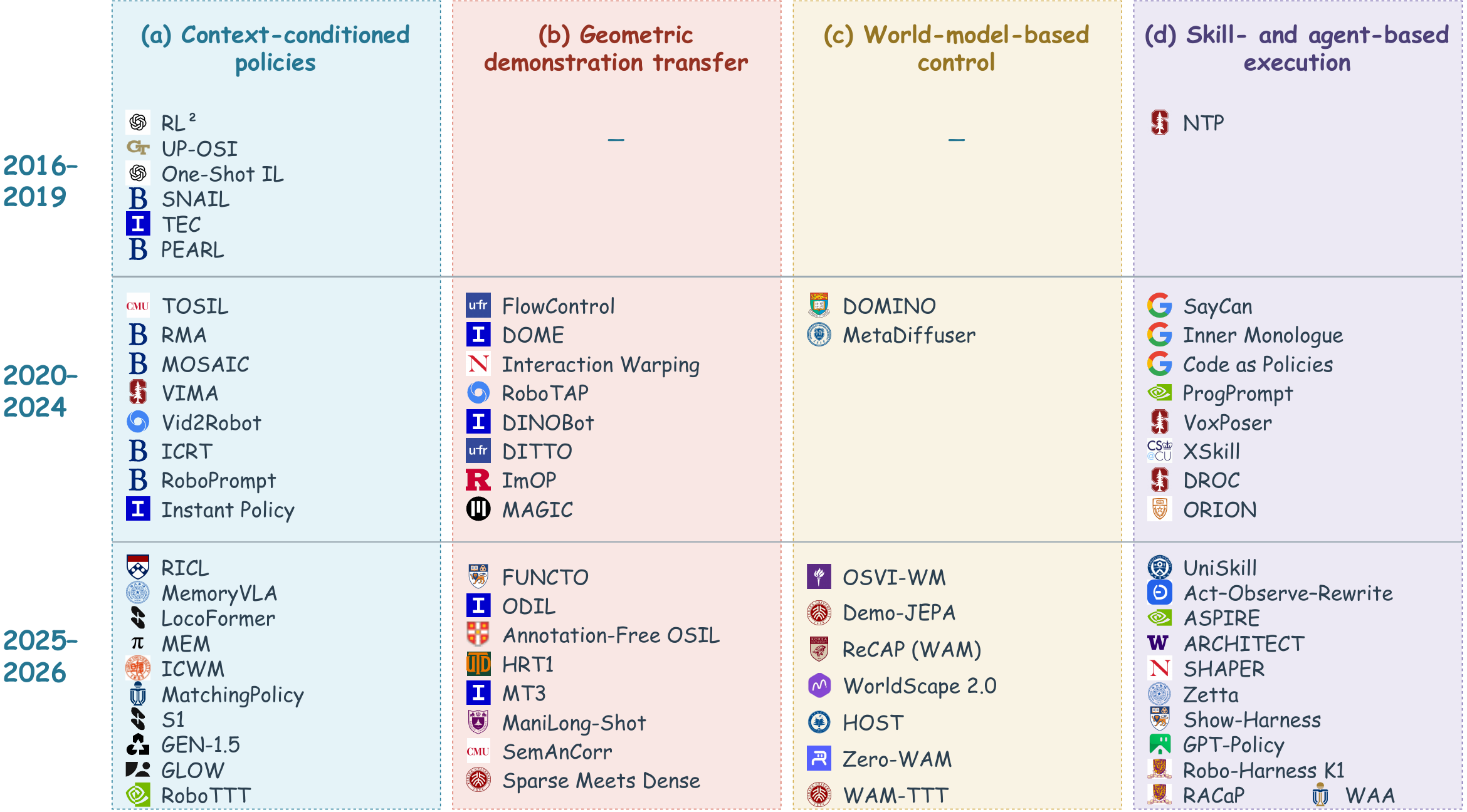}
\caption{Evolution of the four method families by first public release period. NTP: Neural Task Programming.}
\label{fig:history}
\par\phantomsection\label{floatend:fig:history}
\end{figure}

\surveyanchor{discuss:fig:history}
The four branches in Figure~\ref{fig:history} respond to different obstacles in reusing experience. Context-conditioned policies learn which actions an example implies; geometric transfer identifies a relation under which a recorded interaction remains applicable. Predictive control makes consequences available before motion, while skill and program interfaces separate a new procedure from its motor realization. Their histories overlap because the obstacles coexist. Broader priors expand the available behaviors, but also make evidence selection and retention more consequential: a capable controller still needs to follow the current teaching rather \mbox{than a familiar default.}

Online system identification provided an explicit early version of the physical-adaptation idea: recent state--action history identified dynamics parameters for a universal policy~\citep{yu2017uposi}. Later methods could learn a latent instead of naming every parameter, but the central problem remained inferring how the present body would respond. The demonstration and interaction strands were already connected by Watch--Try--Learn: one policy attempted a demonstrated task and another used the resulting feedback~\citep{zhou2019wtl}. A demonstration narrows the task hypothesis, while interaction can resolve information it omits.

\paragraph{Reusable sequence interfaces.}
A sequence interface addresses a tension between compressing experience and revisiting a decisive event. SNAIL combined temporal convolutions, which aggregate the preceding sequence, with attention, which retrieves relevant earlier information~\citep{mishra2017snail}. This made adaptation depend on both accumulated state and selective access. Transformers provide query-dependent access through attention~\citep{vaswani2017attention}. Retaining more history increases the available evidence; attention and downstream computation determine which evidence informs the current decision.

Broader teaching made a fixed task summary less sufficient: two demonstrations can have the same goal yet differ in order, contact, or tool use. Multimodal task specifications placed goals and procedures in the input \citep{jang2022bcz,jiang2022vima}, while trajectory models retained earlier observations and actions as evidence for later decisions \citep{xu2022promptdt,laskin2022ad}. The benefit of a sequence interface was access to distinctions that a single task label could erase. Its cost was an alignment problem: the robot had to recover which part of that evidence mattered at its own execution stage. Temporal video correspondence and causal sensorimotor modeling addressed this pressure through learned representations \citep{jain2024vid2robot,fu2024icrt}; spatial representations imposed geometric correspondence before action prediction \citep{vosylius2024instantpolicy}. These complementary approaches preserve temporal evidence and geometric correspondence at different stages of action generation.

\paragraph{Adaptation over broad priors.}

Broad pretraining changes what adaptation must accomplish. When perception and a useful range of motion are already available, a new example can specialize an action model \citep{sridhar2025ricl} or constrain an anticipated robot-domain future \citep{zhou2026zerowam}, rather than supply every ingredient of a new skill. Interaction histories similarly let reusable controllers account for a deployed body's response \citep{liu2025locoformer,chen2026zeva}. Broader pretraining therefore expands the behaviors available to teaching, but also strengthens familiar continuations that may conflict with a new example. Evaluation must establish whether the example \mbox{can redirect this competence.}

Geometric reuse and learned contextual inference allocate generalization differently. Visual nearest-neighbor methods rely on a representation in which nearby observations admit compatible actions~\citep{pari2021vinn}. Explicit alignment instead restores the conditions under which a recorded robot motion or human-demonstrated object trajectory remains usable~\citep{dipalo2023dinobot,heppert2024ditto}. When those conditions are insufficient, a video can initialize further interaction learning, as in WHIRL~\citep{bahl2022whirl}. The same one-example interface can therefore support replay, transformed reference execution, or new policy learning. What distinguishes these routes is the assumption that makes the demonstrated behavior executable in the new setting.

World-model-based control exposes the consequences of behavior before executing it. Contextual dynamics models first made recent transitions useful for evaluating candidate action sequences~\citep{mu2022domino,ni2023metadiffuser}. Demonstration-conditioned models then added a second role for context: specifying which future should be realized~\citep{goswami2025osviwm,zhou2026zerowam}. This development joins physical prediction with task interpretation. The controller must preserve the procedure inferred from the example while predicting how it can unfold in the robot's current scene.

Task structure offers a different way to reuse competence: preserve the procedure while delegating its physical realization. Demonstration-derived hierarchies express a new task through existing routines~\citep{xu2017ntp}. Language-model programming expands the procedure to include perception calls, arithmetic, and feedback logic~\citep{liang2022codeaspolicies}; video-derived outlines let visual teaching specify that procedure for an existing planner and executor~\citep{chen2026showharness}. The gain is compositional reuse of capabilities. Its condition is that the intermediate preserve the task's binding details, because the executor cannot recover a contact or ordering requirement that was omitted during interpretation. Section~\ref{sec:methods} develops this trade-off across the four interfaces.

These developments connect teaching, interaction, and procedural reuse. General-Level distinguishes breadth from synergy in multimodal generalists~\citep{fei2025generallevel}; for robots, the corresponding learning question is whether combining these capabilities improves subsequent acquisition. LMPC studies responsiveness to later teaching~\citep{liang2024lmpc}, while reward and learning-program search change how new skills are acquired~\citep{ma2023eureka,xiao2026enpire}. Section~\ref{sec:icl-self-improvement} develops this question for individual and shared learning.

\paragraph{Review scope and source selection.}
This review uses a mechanism-oriented narrative synthesis with targeted updates through 29 September 2026. Searches and citation chaining span visual imitation, contextual meta-RL, navigation from video and experience, system identification, example-conditioned programming, interaction memory, and robot foundation models. Sources include arXiv, official proceedings and journal articles, and author or organization releases. Technical claims are traced to these primary sources; release reports are identified as such where their evidence differs from a research paper.

Works enter the synthesis when they explain a route from new evidence to behavior, supply a relevant data or evaluation resource, or establish a training or theoretical foundation. Fixed-weight contextual methods form the central subject; parameter adaptation and prior-competence methods supply explicit comparisons. Versions of the same work share one bibliography entry, using an official publication when available. Separately cited papers and dataset releases remain distinct references. For hybrid systems, the analyzed deployment branch determines the primary family; modality, storage, and parameter \mbox{updates remain independent attributes.}

The reference inventory in Appendix~\ref{app:study-index} records this coding and the first public year, using arXiv v1 or an official release. Supplement~S1 provides the item-level records and counting rules. The resulting counts describe the composition and publication history of this reviewed corpus.

\paragraph{Publication patterns.}

\begin{figure}[!tp]
\centering
\includegraphics[width=\linewidth]{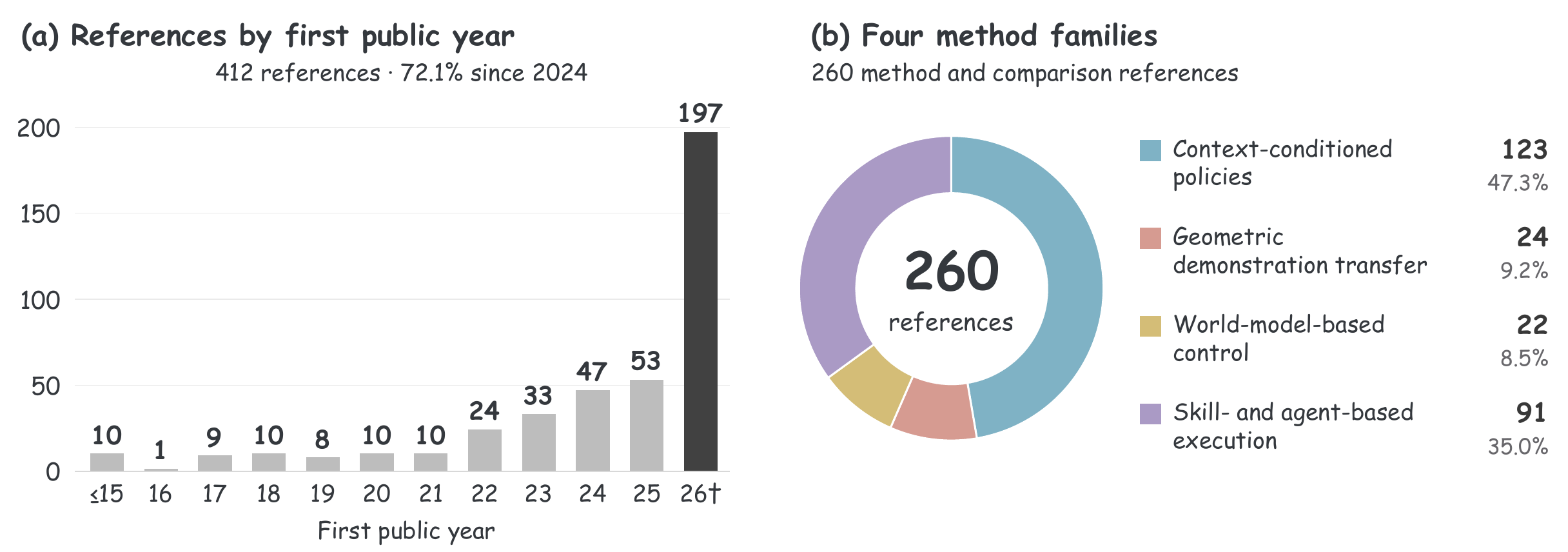}
\caption{Composition of the reviewed literature. (a) First-publication years for all \CorpusTotal{} references. (b) The four-family distribution of \CorpusMethods{} method and comparison references. Percentages in (b) use this subset, which includes supporting mechanisms and parameter-adaptation comparisons. The first bin covers publications through 2015; $\dagger$ marks the partial year ending 29 September 2026.}
\label{fig:corpus-trends}
\par\phantomsection\label{floatend:fig:corpus-trends}
\end{figure}

The \CorpusTotal{} references include \CorpusRecent{} first released since 2024 and \CorpusCurrentYear{} in the partial 2026 window. The concentration coincides with the recent expansion of generalist policies, multimodal teaching, and embodied agents. Figure~\ref{fig:corpus-trends} separates this complete reference set from the \CorpusMethods{} method and comparison references; the latter includes supporting control mechanisms and parameter-adaptation comparisons.

Within the method subset, \CorpusPolicy{} references concern context-conditioned policies, \CorpusGeometry{} geometric transfer, \CorpusWorld{} world-model-based control, and \CorpusAgent{} skill- and agent-based execution. Branch-specific coding places ReCAP\textquotesingle{}s Cosmos Policy branch under predictive world-action modeling~\citep{park2026recap}, ENPIRE under program revision~\citep{xiao2026enpire}, and GPT-Policy under contextual tool use~\citep{cheng2026gptpolicyeval}.

\surveyanchor{discuss:fig:corpus-trends}
The recent concentration of publications and the distribution across control interfaces in Figure~\ref{fig:corpus-trends} motivate a closer comparison of how methods use context. The following chapter examines their execution intermediates and the assumptions that support transfer.

\begin{takeaway}
\takepoint Context resolves what the current observation and prior competence leave undetermined: the required task, relevant history, or physical response. Its value depends on whether that information \mbox{changes a physically executable decision.}
\takepoint Demonstration following and physical adaptation have distinct historical roots. Their convergence joins inference of a new task with inference of an unfamiliar body's response.
\end{takeaway}

\FloatBarrier
\suppressfloats[t]
\section{Methods: How Does Context Change Action?}
\label{sec:methods}
Context changes action through an execution intermediate: an action distribution, a motion reference, predicted consequences, or a skill or tool specification. The four families are compared by what this intermediate preserves, why existing competence can execute it, and which changes invalidate that assumption (Table~\ref{tab:methodological-taxonomy}). Correspondence binds the evidence to the current situation; memory keeps it available; external revision carries useful lessons into later attempts. Together, these mechanisms connect evidence processing, execution, and reuse across the four families.

The factorizations use the context $C_t$, history $h_t$, proposed action block $A_t$, and fixed neural parameters $\theta$ defined in Section~\ref{sec:foundations}. Each exposes the intermediate consumed by execution, allowing stochastic or deterministic implementations. Table~\ref{tab:notation} collects the notation, with local symbols \mbox{defined alongside each equation.}

\paragraph{A shared primitive vocabulary.}\label{sec:primitive-vocabulary}
Tables~\ref{tab:methods}, \ref{tab:geometric-transfer}, and~\ref{tab:predictive-agent-comparison} use one primitive taxonomy. \emph{Source} codes demonstrations or examples (D), retained history (H), instructions or goals (I), explicit feedback (F), and scene or map evidence (E). \emph{Form} codes visual observations or features (V), state/action records (S), geometry (G), and linguistic or symbolic records (L). \emph{Context operation} separates eight operations: \textit{Retrieve} selects stored records; \textit{Attend} makes a query-dependent neural read; \textit{Align} establishes spatial or temporal correspondence; \textit{Encode} builds a learned representation; \textit{Estimate} infers hidden physical response; \textit{Parse} constructs symbolic bindings or procedures; \textit{Verify} checks candidates against feedback or constraints; and \textit{Fit} solves query-specific coefficients or geometry with network weights fixed. The tables code task-relevant contextual evidence and the principal operations that use it during deployment. Commas list evidence types; slashes list operations or carriers used together, without imposing a temporal order.

The resulting \emph{carrier} is an action output, motion reference, predicted future, learned skill, program/procedure, or tool/controller call. A carrier identifies what passes between components; the family identifies its role in the complete control path. \emph{Execute} then distinguishes six operations: \textit{Replay} returns recorded controls; \textit{Combine} blends retrieved action continuations; \textit{Decode} generates actions with a learned model; \textit{Track} follows an explicit reference; \textit{Plan} searches for an executable realization; and \textit{Invoke} runs a program or calls a tool.

\paragraph{Evidence and execution roles.}
Language contributes according to the information it carries, independently of the control interface. In direct action inference, numerical states and action examples can be serialized for a language model, as in RoboPrompt~\citep{yin2024roboprompt}; ICRT learns its sensorimotor interface without linguistic data~\citep{fu2024icrt}. For a geometric archive, language can narrow the intended task before geometric matching resolves the motion, as in MT3~\citep{dreczkowski2025mt3}. Command--program examples instead teach an executable mapping~\citep{liang2022codeaspolicies}, while textual event memory preserves information missing from recent visual history~\citep{torne2026mem}. These uses resolve task selection, action interpretation, and retention at different stages. The relevant comparison is what the text contributes to the decision, including whether it supplies a goal, a new convention, or a remembered outcome.

The separation between information and execution also applies to geometry. Matching can condition a learned action generator or supply a retargeted reference to a separate motion executor. Table~\ref{tab:correspondence-interfaces} in Appendix~\ref{app:comparisons} (p.~\pageref{tab:correspondence-interfaces}) compares these roles. Geometry can instead supervise further learning: using hand--object motion to guide residual reinforcement learning trains a new policy~\citep{feng2026regrind}. The role of geometry becomes clear by identifying what it changes: a policy input, an executable reference, or the policy parameters.

Semantic abstraction preserves object roles, prerequisites, and skill order while allowing grasps or paths to change. Object-centric plans and video outlines expose that structure at different resolutions~\citep{zhu2024orion,chen2026showharness}. Their executor must receive every binding constraint: plate-first teaching specifies order, while a handle-grasp requirement additionally restricts contact. A generic pick-and-place request can erase either distinction. This motivates tracing the taught requirement through the intermediate to the command it changes.

\subsection{Context-conditioned policies}
\label{sec:direct-methods}

Context-conditioned policies map task evidence and robot history to actions. They either reuse recorded actions selected by correspondence or generate actions from an interpreted context representation. We compare these two forms before examining three complementary demands on the representation: execution phase, spatial correspondence, and physical response. The final comparison considers how pretrained competence affects the evidence each form needs.

\Needspace*{5\baselineskip}
For a context-conditioned policy, write $r_t$ for the representation of the evidence used at the current decision:
\begin{equation}
 r_t=E_\theta^{\mathrm c}(h_t,C_t),\qquad
 \pi_\theta(A_t\mid h_t,C_t)=q_\theta^{\mathrm a}(A_t\mid h_t,r_t).
 \label{eq:direct-interface}
\end{equation}
Here $E_\theta^{\mathrm c}$ interprets context and $q_\theta^{\mathrm a}$ generates actions from its representation $r_t$; both may belong to one network. The factorization requires $r_t$ to retain the context information used by action generation. Autoregressive, diffusion, and flow-based generators can implement this interface. If an example teaches red-then-blue packing, the representation must preserve that order while the generator adapts the \mbox{reach-and-place motion to the current scene.}

\paragraph{Reusing actions through correspondence.}
Let the nonempty set $J_t$ index the demonstration records selected for the current query, and let $\bar A_j$ be the action block stored in record $j$. A nonparametric action-reuse policy has the form
\begin{equation}
 \pi_\theta(\,\cdot\mid h_t,C_t)
 =\sum_{j\in J_t}\beta_{tj}\,\delta_{\bar A_j},
 \qquad \beta_{tj}\geq 0,\quad\sum_{j\in J_t}\beta_{tj}=1.
 \label{eq:action-reuse}
\end{equation}
Here $\beta_{tj}$ is the query-dependent weight of record $j$, and $\delta_{\bar A_j}$ is a point mass at its stored action block. Nearest-neighbor replay selects one record with weight one. A policy that averages retrieved controls instead returns $A_t=\sum_{j\in J_t}\beta_{tj}\bar A_j$; this is a deterministic action combination rather than sampling a stored block. Such averaging requires compatible continuous action coordinates and separate treatment of discrete commands. Stored blocks must be expressed or retargeted in the receiving robot's action interface. Equation~\eqref{eq:direct-interface} instead uses a learned generator $q_\theta^{\mathrm a}$ to transform interpreted evidence into actions. When retrieval supplies either interface, it selects the evidence; only the replay interface directly returns recorded actions.

Action reuse assumes that proximity in a representation implies compatible control. The workflow encodes the current scene, retrieves comparable demonstration states, and returns their recorded actions \citep{pari2021vinn}. Its generalization rests on representation quality and archive coverage. A retrieved demonstration can instead condition a learned generator \citep{sridhar2025ricl}, which must infer how the example applies rather than return its action directly. This introduces a second source of generalization: transforming available \mbox{experience beyond local replay.}

Distribution shift can separate these two sources. In REGENT's unseen-MuJoCo-embodiment comparison, Retrieve-and-Play outperforms the frozen contextual transformer, while fine-tuning improves the learned agent \citep{sridhar2024regent}. The retrieved actions remain useful in a condition where the learned interpreter transfers less effectively. Archive coverage and learned transformation should therefore be compared separately: retrieval can recover an applicable motion even when a more expressive generator does not preserve its relevance.

Local continuation bridges replay and learned generation. Behavior Predictive Control (BPC) reconstructs the live observation--action history from retrieved trajectory windows, then applies the same coefficients to their action continuations~\citep{arxiv260930134}. Unlike the nonnegative mixture in Equation~\eqref{eq:action-reuse}, its regularized coefficients may be signed, permitting local extrapolation. The retrieval metric and residual correction are fitted offline; deployment solves the continuation coefficients from the current history while keeping those fitted components fixed. Reliable continuation depends on whether nearby records jointly explain the query with stable coefficients, connecting retrieval quality to local control compatibility.

The use of correspondence determines which component carries motion generalization. A retargeted human motion can serve as a proposal that a pretrained diffusion policy refines, as in DemoDiffusion~\citep{park2025demodiffusion}. The learned prior then contributes to the final motion. Alternatively, registration can directly transfer the next demonstrated end-effector pose, as in IMOP~\citep{zhang2024imop}. In that case, geometric applicability carries more of the transfer burden. Section~\ref{sec:geometric-methods} develops this second interface, where the matched evidence becomes an \mbox{explicit reference for execution.}

\begin{figure}[!tp]
\centering
\includegraphics[width=\linewidth]{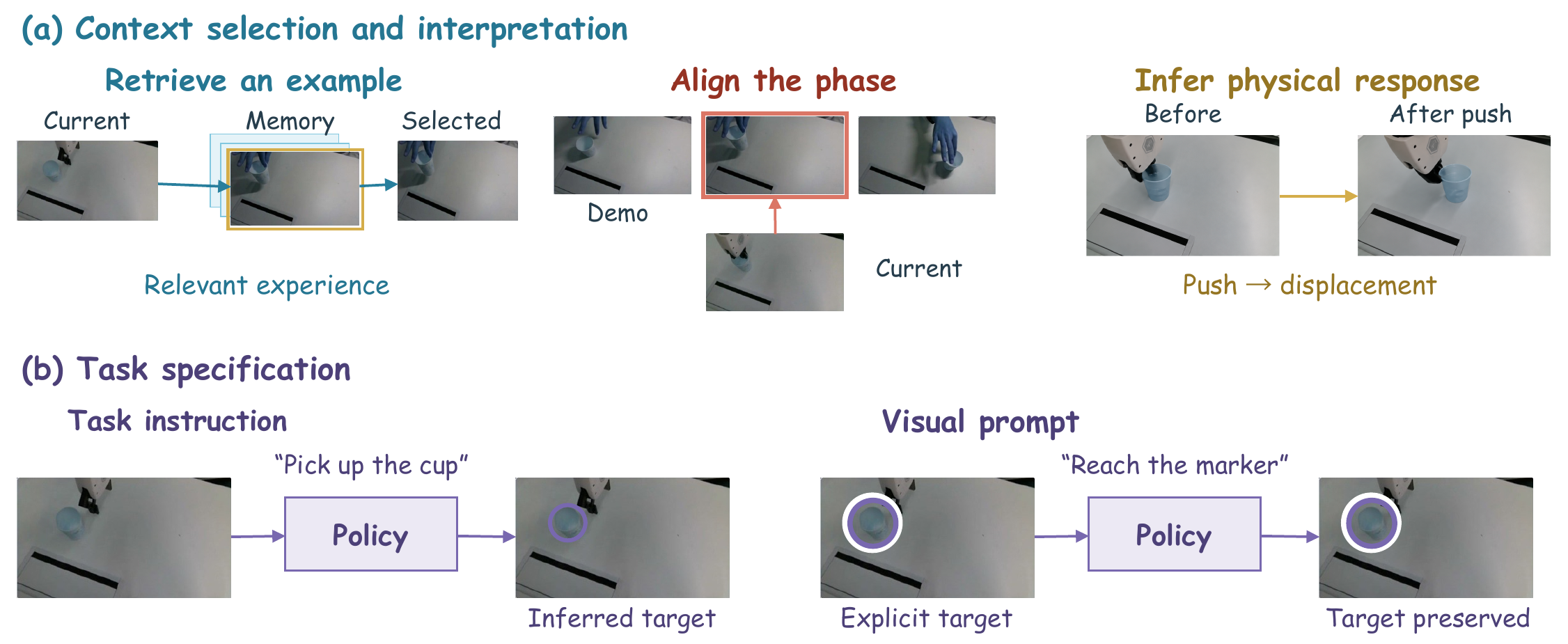}
\caption{Constructing and interpreting task context. (a) Retrieval selects relevant experience, temporal alignment identifies the current phase, and observed action effects inform physical response. (b) Language specifies an operation, while a visual prompt makes its spatial target explicit. These operations supply task evidence to the execution interfaces compared in Figure~\ref{fig:method-taxonomy}.}
\label{fig:mechanism-families}
\par\phantomsection\label{floatend:fig:mechanism-families}
\end{figure}

\surveyanchor{discuss:fig:mechanism-families}
Context interpretation can resolve three kinds of uncertainty before action generation: which experience applies, which execution phase matters, and what the observed transitions reveal about physical response (Figure~\ref{fig:mechanism-families}a). Task specification adds a complementary constraint: language can identify an operation, while a visual prompt binds it to an explicit spatial target (panel b). These shared operations organize the evidence available to each method family. Their output can condition direct action inference or supply task information to a geometric reference, a predictive model, or a skill executor. The four families differ in this subsequent context-to-action interface, illustrated in Figure~\ref{fig:method-taxonomy}.

\paragraph{From task identity to execution correspondence.}
When retrieved actions do not transfer directly, the learner must interpret what the example specifies. Early demonstration-conditioned policies compressed examples into a task representation. Task-Embedded Control Networks jointly learned a demonstration metric and an embedding-conditioned policy \citep{james2018tec}. Such a summary makes few-shot conditioning economical, but the policy must recover the relevant execution phase from its current observation. As variation extends from task identity to ordering, viewpoint, and object placement, identifying the task is no longer enough: the robot must relate what it sees now to a particular part of the example.

\surveyanchor{discuss:tab:methods}
Table~\ref{tab:methods} traces shared context operations into distinct action rules. Retrieval can return actions directly, condition a generator, or constrain a predicted future. Alignment can condition action inference or construct a motion reference. Transfer depends on what the carrier preserves and how execution realizes it; the following comparisons develop those assumptions.

\begin{table}[!tp]
\centering\surveytable
\caption{Context-conditioned policies: retrieval, demonstration interpretation, and physical adaptation (Section~\ref{sec:direct-methods}). Neural components remain fixed during the listed deployment branches. Codes: D examples, H history, F feedback; V visual, S state/action, G geometry, L symbolic. Slashes combine operations. $\dagger$: technical report; --: unspecified. Operations are defined in Section~\ref{sec:primitive-vocabulary}.}
\label{tab:methods}
\renewcommand{\arraystretch}{1.18}
\begin{tabularx}{\linewidth}{@{}>{\raggedright\arraybackslash}p{.25\linewidth}>{\centering\arraybackslash}p{.07\linewidth}>{\centering\arraybackslash}p{.10\linewidth}>{\raggedright\arraybackslash}p{.20\linewidth}>{\raggedright\arraybackslash}p{.15\linewidth}Y@{}}
\toprule
\rowcolor{black!12}\textbf{Method} & \textbf{Source} & \textbf{Form} & \textbf{Context operation} & \textbf{Carrier} & \textbf{Execute}\\
\midrule
\rowcolor{panel}\multicolumn{6}{c}{\strut\itshape Retrieval and trace guidance}\\
\rowcolor{black!3}VINN~\citep{pari2021vinn} & D & V,S & Retrieve & Actions & Replay\\
BPC~\citep{arxiv260930134} & D & V,S & Retrieve / Fit & Actions & Combine\\
\rowcolor{black!3}RICL~\citep{sridhar2025ricl} & D & V,S & Retrieve / Encode & Actions & Decode\\
TraceFlow~\citep{arxiv260920646} & H,F & V,S & Retrieve / Align & Actions & Decode\\
\rowcolor{panel}\multicolumn{6}{c}{\strut\itshape Demonstration interpretation}\\
\rowcolor{black!3}TOSIL~\citep{dasari2020tosil,goswami2025osviwm} & D & V & Attend & Actions & Decode\\
MOSAIC~\citep{mandi2021mosaic} & D & V & Attend & Actions & Decode\\
\rowcolor{black!3}RoboPrompt~\citep{yin2024roboprompt} & D & G,S & Parse & Actions & Decode\\
BiCICLe~\citep{palma2026bicicle} & D & S & Parse & Actions & Decode\\
\rowcolor{black!3}Vid2Robot~\citep{jain2024vid2robot} & D & V & Attend & Actions & Decode\\
ICRT~\citep{fu2024icrt} & D & V,S & Attend & Actions & Decode\\
\rowcolor{black!3}KAT~\citep{dipalo2024keypoint} & D & G,S & Encode & Actions & Decode\\
Instant Policy~\citep{vosylius2024instantpolicy} & D & G,S & Encode & Actions & Decode\\
\rowcolor{black!3}MatchingPolicy~\citep{she2026matchingpolicy} & D & G,S & Align / Encode & Actions & Decode\\
StellaVLA~\citep{xu2026stellavla} & D & L & Encode & Actions & Decode\\
\rowcolor{black!3}SynthICL~\citep{qian2026synthicl} & D & V & Encode & Actions & Decode\\
BPP~\citep{patel2026bpp} & D & S & Encode & Actions & Decode\\
\rowcolor{panel}\multicolumn{6}{c}{\strut\itshape Physical-response inference}\\
\rowcolor{black!3}RMA~\citep{kumar2021rma} & H & S & Estimate & Actions & Decode\\
Self-Adaptive VLA~\citep{arxiv260930092} & H & V,S & Estimate & Actions & Decode\\
\rowcolor{panel}\multicolumn{6}{c}{\strut\itshape Generalist demonstration interfaces}\\
\rowcolor{black!3}GEN-1.5$^{\dagger}$~\citep{generalist2026gen15} & D & S & -- & Actions & Decode\\
S1$^{\dagger}$~\citep{skild2026s1} & D & V & -- & Actions & Decode\\
\rowcolor{black!3}GLOW$^{\dagger}$~\citep{knowin2026glow} & D & V & Encode & Actions & Decode\\
\bottomrule
\end{tabularx}
\par\phantomsection\label{floatend:tab:methods}
\end{table}

An attention-based interpreter implements a query-dependent read over demonstration elements. Let $D^{\mathrm{s}}=(d_i^{\mathrm{s}})_{i=1}^{L}$ be a support demonstration with $L$ elements. Its $i$th element $d_i^{\mathrm{s}}$ may contain an image or a synchronized observation--action record; $k$ is a summation index over the same elements. One representative alignment is
\begin{equation}
 \alpha_{ti}=\frac{\exp s_\theta(h_t,d_i^{\mathrm{s}})}
 {\sum_{k=1}^{L}\exp s_\theta(h_t,d_k^{\mathrm{s}})},
 \qquad r_t=\sum_{i=1}^{L}\alpha_{ti}e_\theta(d_i^{\mathrm{s}}).
 \label{eq:demo-alignment}
\end{equation}
The score $s_\theta$ matches the robot history to an encoded demonstration element $e_\theta(d_i^{\mathrm{s}})$. Normalized weights $\alpha_{ti}$ select evidence relevant at decision $t$, allowing the read to change with execution progress. This inference-time read supports image-only and action-bearing demonstrations alike; its training objective is a separate choice considered in Section~\ref{sec:context-training}. Richer attention or geometric representations can preserve several relationships without imposing a monotone frame match.

Execution-dependent attention addresses the loss of phase information in a single task summary: each action can consult the relevant spatial and temporal evidence~\citep{dasari2020tosil}. Cross-domain contrastive supervision makes that read more reliable when demonstration and query differ visually~\citep{mandi2021mosaic}. Video cross-attention and causal sensorimotor sequences expose different evidence to the read, respectively visual progression and observations with recorded actions~\citep{jain2024vid2robot,fu2024icrt}. Execution mismatch creates a further problem when frame similarity is misleading. RHyME constructs semantically compatible training videos through clip retrieval and composition, then uses learned alignment to condition its policy~\citep{kedia2024rhyme}. These designs address complementary parts of correspondence: when to read, what to compare, and how training makes the comparison meaningful.

The context representation determines which distinctions action generation can recover. RGB retains appearance and temporal evidence but leaves motion implicit; compatible sensorimotor records expose the action itself. Keypoints and spatial graphs supply correspondence before generation~\citep{dipalo2024keypoint,vosylius2024instantpolicy}, reducing the need to infer task geometry from appearance. Motion latents pursue a different compression: cross-domain decoding ties human and robot movement together~\citep{chen2025vivla}, while hierarchical tokens separate persistent demonstration structure from finer execution decisions~\citep{huang2026mint}. Continuous flow matching generates actions from encoded images, proprioception, and commands~\citep{ding2026contextflow}. TP-Flow additionally extracts phase-level prototypes from support demonstrations and conditions both the initial flow prior and velocity field on them~\citep{wang2026tpflow}. Episodic training teaches this dependence; deployment recomputes prototypes with fixed weights. Representation and generator are therefore coupled choices: richer action generation depends on preserving the taught distinction during context encoding.

Temporal selection addresses a different limit from action encoding. A complete trajectory may contain many locally irrelevant stages, so ICI-VLA retrieves phase-aligned micro-demonstrations using a retriever trained with dynamic-time-warping supervision \citep{yang2026icivla}. The deployed policy reads examples selected from observable query inputs; target-action masking during training reduces reliance on exact action-prefix continuation. Phase selection determines which evidence reaches the generator, whereas token prediction or flow matching determines how that evidence becomes an action block.

\paragraph{Spatial correspondence as conditioning.}
Temporal phase identifies when an example applies; spatial correspondence identifies where its action should be realized. MatchingPolicy makes this second requirement explicit \citep{she2026matchingpolicy}. Foundation-model features establish demonstration--query point matches; tracking propagates them through time; RGB-D lifts them into 3D. A graph diffusion policy then generates gripper motion from these relationships. This workflow assumes that matched task geometry remains informative about the required action, without fixing a trajectory for the robot to replay. Section~\ref{sec:geometric-methods} considers the complementary case in which correspondence transforms an explicit executable reference.

Navigation extends this correspondence from object interactions to places. A policy can read successive visual subgoals~\citep{pathak2018zeroshot} or attend to an action-bearing path memory~\citep{kumar2018rpf} to follow new teaching. A scene preview supplies a different spatial reference: NOLO~\citep{zhou2024nolo} combines a video with optical-flow-derived action labels, the current view, and a goal image to predict actions. Its offline reinforcement learning trains the interpreter; deployment reads a new scene video with weights fixed. The shared computational interface is contextual action inference, while the context can specify a route or reveal an environment.

\paragraph{Identifying physical context from interaction.}
Temporal and spatial correspondence can identify a desired motion while leaving its physical effect uncertain. Interaction history supplies the complementary evidence when the goal is known but the current body's response is not. System identification makes this unknown explicit through physical parameters, whereas a learned adaptation latent can retain the response information needed by a policy~\citep{yu2017uposi,kumar2021rma}. ICWM extends physical-context inference to changes in observation and embodiment~\citep{wang2026icwm}. These approaches require transitions that reveal how commands affect the present system; a task demonstration alone may leave that uncertainty unresolved. Their representation conditions an action policy directly. Predicting and evaluating candidate futures introduces the additional computational stage considered in Section~\ref{sec:predictive-methods}.

The observation channels determine whether such a response is identifiable. Self-Adaptive VLA pairs rollouts under injected hardware shifts with compensated expert targets during post-training, then aggregates rollout-derived context tokens to modulate a fixed deployed policy~\citep{arxiv260930092}. Under its actuation-bias model, proprioception exposes a mismatch between commands and realized joints; under joint-encoder offsets, readings can remain consistent with commands while the physical pose is wrong, requiring visual evidence. Additional history helps only when it contains a measurement that distinguishes the possible causes. The inferred context calibrates physical response for the specified task.

\paragraph{Connecting pretrained competence to new evidence.}
Broader robot and language priors change what must be learned for demonstration following. An existing action generator can be conditioned on retrieved executions \citep{sridhar2025ricl}; grounded spatial descriptions can supply scene information missing from the current visual input \citep{arxiv260805738}. These inputs resolve different unknowns. A demonstration may change the requested procedure, while a scene description may clarify how to execute an already known procedure. The contribution of each channel depends on the decision its information resolves.

Observation-only teaching leaves action correspondence implicit. Behavioral Cloning from Observation estimates actions through inverse dynamics before training a policy~\citep{torabi2018bco}. Fixed-weight teaching uses an existing action interface learned from demonstrations or constructed through \mbox{pretrained reasoning and geometry.}

Textual action interfaces reveal how much of this burden can be handled outside a visuomotor backbone. Object poses and keyframe end-effector actions can be supplied as examples to a frozen language model \citep{yin2024roboprompt}. For bimanual tasks, conditioning one arm's trajectory on the other's prediction preserves inter-arm dependence while reducing the joint prediction problem \citep{palma2026bicicle}. BiCICLe's main TWIN experiments use ten demonstrations per task, with iterative refinement and candidate selection adding inference computation. These systems reuse pretrained reasoning, but rely on perception, coordinate conventions, and action representations that make its outputs physically meaningful.

Teaching budget can change the preferred adaptation mechanism. KAT compares a text-pretrained model reading keypoint--action examples with KeyAct-DP, a diffusion policy trained on similar representations~\citep{dipalo2024keypoint}. Contextual inference is competitive with few examples; policy training becomes stronger with more. The crossover differs against an image-based diffusion policy, indicating that both representation and adaptation mechanism determine the useful demonstration budget.

Generalist systems expose different teaching representations. BPP and GEN-1.5 accept sensorimotor prompts~\citep{patel2026bpp,generalist2026gen15}; S1 accepts video prompts~\citep{skild2026s1}. The GLOW technical report~\citep{knowin2026glow} describes a Skill Demo Encoder that converts one human video into reusable skill context. A fixed-weight autoregressive model combines it with current observations, language, robot state, and history to generate end-effector actions. Its teaching pathway belongs to context-conditioned policies (Figures~\ref{fig:complete-chain} and~\ref{fig:design-dimensions}); Harness manages tools and feedback, while KnowinWorld supplies candidate rollouts. The primary family follows the teaching-to-action pathway, with prediction and tool management providing complementary support.

The defining trade-off is between compatibility and transformation. Retrieved actions retain an explicit link to experience but need compatible control coordinates and archive coverage. A learned generator can adapt the example more flexibly, provided its representation preserves the taught distinction. REGENT's comparison locates this trade-off in the transfer of the learned interpreter under distribution shift~\citep{sridhar2024regent}. Transfer depends on correspondence identifying usable evidence and on the generator preserving its relevance. Geometric transfer makes that dependency explicit by turning the evidence into a reference \mbox{that execution must track.}

\subsection{Geometric demonstration transfer}
\label{sec:geometric-methods}

Geometric demonstration transfer preserves a motion or contact reference and adapts its realization to the query scene. Perception establishes correspondence; alignment or retargeting transforms the reference; tracking or replay executes it. This explicit reference is useful when contact geometry matters, such as preserving a handle grasp. We follow the mechanism from reference construction through spatial alignment and shape changes to reuse across a task repertoire.

IMOP makes the reference local to the next demonstrated action: learned invariant-region correspondences support analytic pose transfer, and a state router selects the relevant demonstration transition~\citep{zhang2024imop}. The reference can therefore be a target pose rather than a complete trajectory.

\paragraph{Reconstructing an executable reference.}
Let $D^{\mathrm{s}}$ denote the supplied demonstration. A geometric transfer computation can be written as
\begin{equation}
 \zeta_{1:L}=E_\theta^{\mathrm m}(D^{\mathrm{s}}),\qquad
 g=\operatorname{Retarget}(\zeta_{1:L},o_t,\mathcal B),\qquad
 A_t=\kappa(g,o_t).
 \label{eq:retarget-grounding}
\end{equation}
Here $E_\theta^{\mathrm m}$ is the motion-descriptor extractor, with $\mathrm m$ identifying its role. It produces one descriptor $\zeta_i$ for each of the $L$ demonstration elements, such as hand poses, object-relative poses, or contact locations. The operator $\operatorname{Retarget}$ retargets that sequence to the current scene $o_t$ and robot description $\mathcal B$, which specifies kinematics and control constraints. Its output $g$ is a robot motion reference consumed by executor $\kappa$. Object-centric transfer and hand-based retargeting differ in what $\zeta_i$ preserves: the motion of a manipulated object, or the demonstrator's grasp and hand motion~\citep{heppert2024ditto,allu2025hrt1}. Fitting a pose, grasp, or trajectory changes these inferred variables while the estimator's neural parameters can remain fixed.

For rigid object-relative transfer, this relationship has a particularly simple form:
\begin{equation}
 {}^{W}T_{E,b}^{\mathrm q}(\sigma)
 ={}^{W}T_{O_b}^{\mathrm q}
 \bigl({}^{W}T_{O_b}^{\mathrm s}\bigr)^{-1}
 {}^{W}T_{E,b}^{\mathrm s}(\sigma).
 \label{eq:object-relative-transfer}
\end{equation}
Here ${}^{W}T_X\in SE(3)$ is the rigid pose of frame $X$ in world frame $W$; $E$ denotes the end effector, and $O_b$ the reference object for motion segment $b$. The group $SE(3)$ contains three-dimensional rigid transformations, and the exponent $-1$ denotes the inverse transform. Superscripts $\mathrm s$ and $\mathrm q$ distinguish demonstration and query scenes, and $\sigma\in[0,1]$ is normalized progress within that segment. The middle and right factors express the demonstrated end-effector motion relative to its reference object; the left factor places it in the query scene. This form assumes that the reference remains fixed during the segment and that its relative motion remains applicable. Shape change, moving references, and embodiment differences require the richer correspondence in Equation~\eqref{eq:retarget-grounding}. Choosing $O_b$ is therefore part of task interpretation, not simply a coordinate convention.

\surveyanchor{discuss:tab:geometric-transfer}
Table~\ref{tab:geometric-transfer} traces four ways to preserve an interaction: align its visual reference, transform object-relative motion, match functional parts, or switch references across stages. A richer correspondence is useful when it preserves a constraint that a global pose transform would lose.

\begin{table}[!tp]
\centering\surveytable
\caption{Geometric demonstration transfer, grouped by the reference preserved (Section~\ref{sec:geometric-methods}). R-NDF lists descriptor alignment. Codes: D demonstration; V visual, S state/action, G geometry. Slashes combine operations. Tracking, replay, and planning denote different stages of realizing the reference.}
\label{tab:geometric-transfer}
\renewcommand{\arraystretch}{1.18}
\begin{tabularx}{\linewidth}{@{}>{\raggedright\arraybackslash}p{.25\linewidth}>{\centering\arraybackslash}p{.07\linewidth}>{\centering\arraybackslash}p{.10\linewidth}>{\raggedright\arraybackslash}p{.20\linewidth}>{\raggedright\arraybackslash}p{.15\linewidth}Y@{}}
\toprule
\rowcolor{black!12}\textbf{Method} & \textbf{Source} & \textbf{Form} & \textbf{Context operation} & \textbf{Carrier} & \textbf{Execute}\\
\midrule
\rowcolor{panel}\multicolumn{6}{c}{\strut\itshape Visual alignment}\\
\rowcolor{black!3}FlowControl~\citep{argus2020flowcontrol} & D & V,G & Align & Motion & Track\\
DOME~\citep{valassakis2022dome} & D & V,S & Align & Motion & Track / Replay\\
\rowcolor{black!3}DINOBot~\citep{dipalo2023dinobot} & D & V,S & Retrieve / Align & Motion & Track / Replay\\
\rowcolor{panel}\multicolumn{6}{c}{\strut\itshape Object-relative motion}\\
MT3~\citep{dreczkowski2025mt3} & D & G,S & Retrieve / Align & Motion & Track / Replay\\
\rowcolor{black!3}Pose-based OSIL~\citep{vitiello2023pose} & D & G,S & Align & Motion & Track\\
IMOP~\citep{zhang2024imop} & D & G,S & Align & Motion & Track\\
\rowcolor{black!3}DITTO~\citep{heppert2024ditto} & D & V,G & Align & Motion & Plan / Track\\
HRT1~\citep{allu2025hrt1} & D & G & Align / Fit & Motion & Track\\
\rowcolor{panel}\multicolumn{6}{c}{\strut\itshape Functional correspondence}\\
\rowcolor{black!3}R-NDF~\citep{simeonov2023rndf} & D & G & Align / Fit & Motion & Track\\
Interaction Warping~\citep{biza2023warping} & D & G & Align / Fit & Motion & Track\\
\rowcolor{black!3}Part decomposition~\citep{thompson2026parttransfer} & D & G & Align & Motion & Track\\
SemAnCorr~\citep{dong2026semancorr} & D & G & Align & Motion & Plan / Track\\
\rowcolor{panel}\multicolumn{6}{c}{\strut\itshape Staged and coordinated execution}\\
\rowcolor{black!3}ODIL~\citep{wang2025odil} & D & V,S & Align & Motion & Track / Replay\\
ManiLong-Shot~\citep{chen2025manilong} & D & G,S & Align & Motion & Track\\
\rowcolor{black!3}Annotation-free OSIL~\citep{wichitwechkarn2025annotationfree} & D & V,S & Align & Motion & Track / Replay\\
\bottomrule
\end{tabularx}
\par\phantomsection\label{floatend:tab:geometric-transfer}
\end{table}

\paragraph{From image alignment to object-relative transfer.}
Geometric reuse begins with an applicability assumption: restoring the relevant relative configuration makes the demonstrated interaction useful again. Image-based approaches enforce this through visual feedback. FlowControl aligns live RGB-D observations to successive masked reference frames \citep{argus2020flowcontrol}; DOME separates a learned visual servo to a bottleneck pose from subsequent velocity replay \citep{valassakis2022dome}. The difference is where feedback acts: throughout reference following or primarily before the interaction segment. Tracking active points and gripper events allows RoboTAP to change the visual reference as a task advances \citep{vecerik2023robotap}. POIL closes the loop on the transferred object points themselves: functional-part correspondences establish a reference, and point-set feedback converts tracking errors into a rigid-body twist~\citep{arxiv260930404}. Using the same points for transfer and execution ties disturbance recovery to the demonstrated interaction.

An object-relative invariant makes applicability more explicit than image agreement. Pose-based transfer preserves a recorded end-effector trajectory~\citep{vitiello2023pose}; category-level shape fitting extends correspondence to interaction keypoints~\citep{biza2023warping}. DITTO reconstructs manipulated-object motion from an RGB-D human demonstration, transfers that trajectory to the detected objects in the robot scene, and realizes it through grasping and motion planning~\citep{heppert2024ditto}. Hand-based retargeting instead reconstructs the demonstrator's grasp and movement before converting them into robot configurations~\citep{allu2025hrt1}. The embodiment gap thus appears in different places: realizing an object effect in DITTO, or translating a demonstrated hand configuration. In either case, successful transfer requires both a preserved interaction and a feasible realization.

\paragraph{Semantic anchors and functional geometry.}
Rigid pose transfer preserves a motion only when the source and target support the same local interaction. A different handle or articulated part requires correspondence within the object. SemAnCorr~\citep{dong2026semancorr} selects semantically consistent surface anchors and propagates them through a functional map. Its manipulation pipeline transfers contact regions and local frames, selects a feasible grasp, and reuses motion relative to that grasp. Contact keyframes and relative motions form its execution specification. Semantic correspondence supplies the local geometry needed to extend the rigid transfer in Equation~\eqref{eq:object-relative-transfer} across object shapes.

Shape variation makes whole-object similarity too restrictive for some interactions. What must remain compatible may be a grasp role and a contact role, rather than the complete tool geometry~\citep{tang2025functo}. A complementary approach starts from global shape and refines local curvature correspondences to preserve the contact used by a motion~\citep{liu2024magic}. Surface interaction functions and screw interpolation connect such functional geometry to a transferred trajectory~\citep{defarias2025gift}. These routes differ in how they identify the invariant, but share a control assumption: preserving the relevant interaction structure makes the adapted motion useful. This shifts the transfer problem from matching objects as wholes to matching the parts of their geometry that \mbox{the task actually uses.}

When both interacting objects change, correspondence must preserve their relation as well as locate each part. R-NDF infers local frames on task-relevant parts and aligns them to reproduce a relation on a new object pair~\citep{simeonov2023rndf}. Part decomposition further separates each part's geometry from its arrangement within the object, selecting the part relationships that explain the demonstration~\citep{thompson2026parttransfer}. This reduces interference from irrelevant shape variation while retaining the constraints needed for placement. These object-pair representations extend reference transfer beyond relocating a trajectory around one object; Section~\ref{sec:context-mechanisms} compares \mbox{the resulting transfer scope.}

Deformable manipulation requires correspondence to remain meaningful as the object changes shape. Sparse Meets Dense extracts task-relevant contact pairs from a demonstration and uses an offline learned dense representation to track them during constrained execution~\citep{zhu2026sparsedense}. Its final-state relative-pose annotation supplies interaction geometry; sparse constraints express the task, and dense correspondence maintains that geometry through deformation. This extends the reference from a rigid frame to a changing \mbox{set of contact relations.}

Affordance prediction can supply a related execution interface from a learned prior. VLAff predicts interaction heatmaps, grasp poses, and motion trajectories from an observation and language instruction~\citep{oh2026vlaff}. Its human videos provide offline affordance supervision. These outputs can ground a contextual planner's specification, while a demonstration-conditioned system additionally uses the newly supplied example to determine \mbox{which interaction should transfer.}

\paragraph{From one demonstration to a reusable repertoire.}

A reusable geometric repertoire separates two uncertainties: which recorded interaction applies, and how to enter its starting configuration. DINOBot couples visual retrieval with wrist-view alignment before replay~\citep{dipalo2023dinobot}. MT3 scales this decomposition using language filtering, geometric retrieval, pose alignment, and recorded interaction motion~\citep{dreczkowski2025mt3}. Its comparison of retrieval and behavioral cloning treats alignment and interaction as separate phases; its repertoire reaches 1,000 single-interaction tasks in less than 24 hours of human demonstration time. Multi-stage tasks use a demonstration for each stage. The scalability comes from adding applicable references without retraining a task policy, while success still depends on restoring the conditions under which each interaction can be reused.

\FloatBarrier
Independent object motion and arm coordination expose the limit of one global alignment. A grasp and a later insertion must refer to different objects when the part and receptacle move independently; dual-arm interaction must additionally preserve relative coordination. Multi-stage visual servoing establishes a coordinated reference before ODIL's replay \citep{wang2025odil}. Interaction segmentation with invariant-region matching, and VLM-based keyframe selection with repeated alignment, extend the reference-switching principle to longer demonstrations \citep{chen2025manilong,wichitwechkarn2025annotationfree}. Table~\ref{tab:geometric-transfer} compares how individual systems bind and execute these references, while Table~\ref{tab:correspondence-interfaces} in Appendix~\ref{app:comparisons} contrasts the computational roles of correspondence. The key decision is when a change of interaction requires a different reference.

Geometric transfer makes its assumption inspectable: the matched interaction must remain applicable after alignment. It shifts the burden from task-specific action generation to correspondence, reference switching, and feasible tracking. Shape variation, lost contact, and independent object motion challenge different parts of this assumption. Predictive control instead represents the consequences execution should produce.

\subsection{World-model-based control}
\label{sec:predictive-methods}

World-model-based control uses contextual evidence to predict task-relevant consequences. These predictions can specify a desired evolution for action decoding or evaluate candidate actions under the current dynamics. We compare these complementary uses, then examine demonstration conditioning and the distinction between deployment-time prediction and auxiliary training supervision.

\Needspace*{5\baselineskip}
A representative factorization is
\begin{equation}
\begin{aligned}
 v^+ &\sim p_\theta^{\mathrm v}(\,\cdot\mid C_t,h_t),\\
 A_t &\sim q_\theta^{\mathrm v}(\,\cdot\mid v^+,C_t,h_t),
\end{aligned}
\label{eq:wam}
\end{equation}
where $v^+$ denotes anticipated robot observations or their latent representation; the superscript $+$ identifies a future relative to the current decision. The future distribution $p_\theta^{\mathrm v}$ predicts that evolution, and the future-conditioned decoder $q_\theta^{\mathrm v}$ generates actions from it; $\mathrm v$ identifies the future-prediction interface. All neural components use fixed parameters from $\theta$ in the ICL regime. The two distributions may share a backbone; joint visual--action generation couples them within world-model-based control \citep{survey2026wam}. For the packing example, $v^+$ can show the red item placed before the blue one in the robot scene; action decoding must realize that particular sequence.
At the distribution level, the law of total probability combines the future-conditioned action distributions. Each is weighted by the probability of its corresponding future:
\begin{equation}
 \pi_\theta(A_t\mid h_t,C_t)=
 \int q_\theta^{\mathrm v}(A_t\mid v^+,C_t,h_t)\,
 p_\theta^{\mathrm v}(v^+\mid C_t,h_t)\,\mathrm{d}v^+.
 \label{eq:predictive-policy}
\end{equation}
The integral averages the action distribution over possible futures; discrete futures give a sum. Implementations may sample a future and then an action, or select one future as an approximation. This dependence on a predicted consequence, rather than the choice of sampling algorithm, defines the interface.

\paragraph{Generating a desired future or evaluating an action's future.}
World-model-based control has two complementary forms. In Equations~\eqref{eq:wam}--\eqref{eq:predictive-policy}, context specifies a desired evolution and an action decoder realizes it. In model-based selection, candidate actions specify possible evolutions, and context determines which consequence is preferred. One planning form is
\begin{equation}
 A_t^\star\in\arg\max_{\widehat A_t\in\mathcal A_t}
 \mathbb E_{v^+\sim p_\theta^{\mathrm d}(\cdot\mid h_t,\widehat A_t,C_t)}
 \bigl[\mathcal R(v^+,C_t)\bigr].
 \label{eq:predictive-selection}
\end{equation}
The set $\mathcal A_t$ contains admissible candidate action blocks, $\widehat A_t$ denotes a candidate, and $A_t^\star$ is a selected maximizer. The action-conditioned dynamics distribution $p_\theta^{\mathrm d}$ estimates its future $v^+$; $\mathrm d$ distinguishes forward dynamics from the context-conditioned future generator $p_\theta^{\mathrm v}$. The score $\mathcal R$ measures how that future satisfies the contextual task requirement. The expectation averages over predicted futures. This form makes the two roles explicit: learned dynamics predicts what an action may cause, whereas teaching determines which consequence is desirable. Optimizing $\widehat A_t$ leaves the neural parameters $\theta$ unchanged. 

Unknown dynamics changes which candidate trajectories are credible. One response is to infer dynamics context from recent transitions and evaluate actions with a context-conditioned world model, as in DOMINO~\citep{mu2022domino}. Another is to condition trajectory generation on those transitions and use reward and dynamics guidance during sampling, as in MetaDiffuser~\citep{ni2023metadiffuser}. Both specialize prediction through new evidence while leaving the learned parameters fixed. Their difference is where feasibility and preference enter the computation: in evaluation of candidates or in their generation and refinement.

Online fitting offers a complementary response to uncertain dynamics. MetaPusher updates a meta-learned pushing model from task transitions and repairs its planning tree as predictions change~\citep{arxiv260921122}. Sandwich-Residuals instead fits small corrections around a frozen predictor using latent prediction error~\citep{arxiv260921740}. Both adapt neural parameters at deployment. Their comparison with contextual prediction therefore turns on what changes: model parameters or the evidence supplied to an unchanged model.

\paragraph{From goal-reaching to demonstration-conditioned futures.}
Latent planning supplied an early way to connect an intended state to control. Universal Planning Networks learned a differentiable planner that optimizes a candidate action sequence \citep{srinivas2018upn}. The optimized variables are actions, so this search can run with unchanged model weights. Demonstration-conditioned prediction adds a preceding inference problem: determine which state evolution the new example calls for, rather than assume that the \mbox{goal is already supplied.}

This inference can remain in a latent space. OSVI-WM predicts demonstration-conditioned latent states and decodes them into attributed waypoints \citep{goswami2025osviwm}. Demo-JEPA instead passes a predicted latent goal to a cross-entropy-method planner, retaining that goal until an achievement threshold permits progression \citep{he2026demojepa}. Both combine world-model-based control with an explicit waypoint or goal executor: the predicted future supplies targets for a separate execution stage, rather than only features inside an action head. In Demo-JEPA's real-robot comparison, the diffusion action head has higher mean success on familiar behavior grounding, while planning has higher mean success on unseen tasks. The familiar-task advantage does not persist in its simulation comparison. The relative value of search and learned decoding depends on both task novelty and the execution setting, even when the two share \mbox{a future representation.}

Prediction can also supply a route or a discrete transition model. TADreamer reconstructs navigation waypoints from generated video and registers them to measured geometry before planning~\citep{arxiv260919824}; metric calibration connects imagined motion to feasible execution. GAVEL instead rolls plans through an explicit graph of action preconditions, effects, and location beliefs, repairing model-detectable violations before invoking semantic replanning~\citep{arxiv260919315}. Their shared predictive interface does not require \mbox{a common visual representation.}

Human-video world-action systems broaden the intermediate toward robot-domain visual evolution. Two dependencies become central: current robot progress must determine which part of the demonstration remains relevant \citep{chen2026host}, and the demonstrated procedure must change the predicted future even when the current scene admits a familiar continuation \citep{zhou2026zerowam}. These dependencies connect the older goal-reaching formulation to one-shot task transfer. The predicted evolution must preserve the procedure selected by the example.

The two dependencies operate over different timescales. A demonstration supplies a procedure that should persist, whereas recent robot observations determine the next physically relevant transition. WorldScape Policy 2.0 combines persistent demonstration context, event-level memory, and short-term visual dynamics within a joint video--action model \citep{arxiv260718840}. This moves the design problem beyond choosing RGB or latent futures: the model must keep the taught procedure stable while updating its account of the scene. Memory preserves the procedure while prediction tracks its current physical realization.

The planning and control clocks can be separated explicitly. MaP-WAM uses sparse observed segment histories to generate a language subgoal and visual plan; its executor reuses the cached plan until progress estimates trigger another planning step~\citep{arxiv260911561}. Matching the current observation to nearby plan frames calibrates that progress. At transition, actual execution frames replace the imagined plan in episodic memory. This separation matters: a forecast can guide action, but recording it as an achieved event would let prediction errors become historical evidence. The executor's auxiliary future-video head can be omitted at inference, while the upstream visual plan remains part of deployed control.

Retrieval can constrain prediction before detailed motion is generated. The retrieval-augmented policy ReCAP supplies trajectories to a frozen residual predictor~\citep{park2026recap}; its Cosmos Policy branch jointly generates future images and actions. The retrieved motion narrows the continuation, while prediction relates it to the evolving scene. Other ReCAP backbones route retrieved evidence through different action interfaces, so the predictive classification applies to its Cosmos Policy branch.

The gain in inspectability comes with a new failure mode. A predicted scene can express the correct intention yet be unreachable, miss a necessary contact, or accumulate error during execution. Pixel prediction retains appearance detail at a computational cost; latent prediction offers compression but makes the retained task detail less visible. Action-conditioned simulation, goal-conditioned generation, and joint action--world policies consequently use related representations for different purposes \citep{li2025embodiedwm}. The value of the intermediate depends on how reliably its task information survives conversion into executable motion.

\paragraph{Learning from futures without executing through them.}

\begin{figure}[!tp]
\centering
\includegraphics[width=\linewidth]{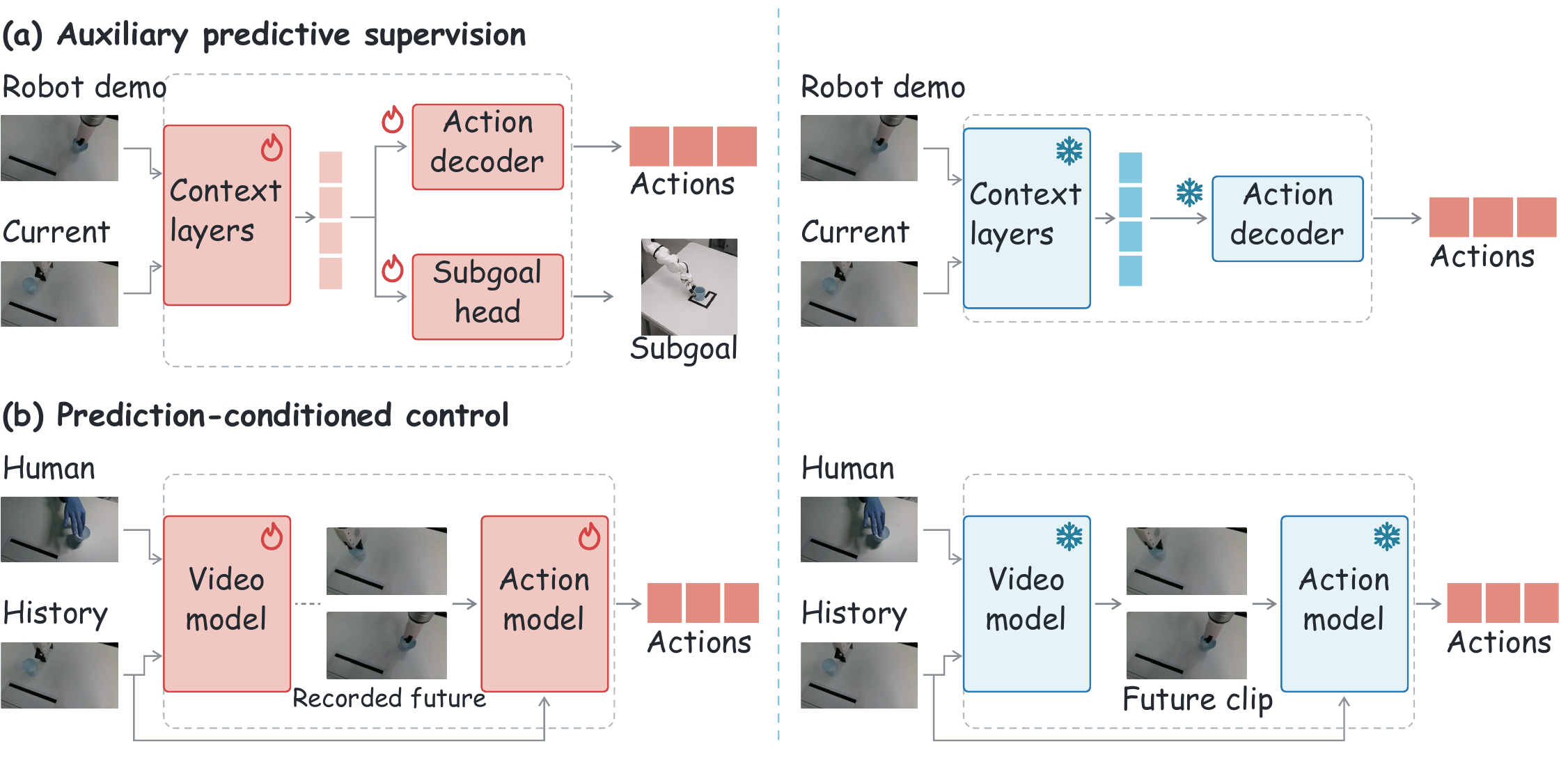}
\caption{Two roles of visual prediction: training (left) and deployment (right). Flames mark trainable modules; snowflakes mark fixed modules. (a) SynthICL trains the context layers and both heads, then removes the subgoal head; its frozen DINOv3 backbone is omitted~\citep{qian2026synthicl}. (b) Zero-WAM trains the video and action branches, then freezes both and replaces recorded future video with generated video~\citep{zhou2026zerowam}.}
\label{fig:prediction-role}
\par\phantomsection\label{floatend:fig:prediction-role}
\end{figure}

Future supervision can improve an action representation without requiring a future to be generated online. Let $\mathcal{L}_a$ be an action-training loss and $\mathcal{L}_v$ an auxiliary future-prediction loss. The training and deployment interfaces can then be written as
\begin{equation}
 \theta^\star\in\arg\min_\theta
 \bigl[\mathcal{L}_a(\theta)+\lambda\mathcal{L}_v(\theta)\bigr],
 \qquad A_t\sim\pi_{\theta^\star}(\cdot\mid h_t,C_t).
 \label{eq:auxiliary-future}
\end{equation}
Here $\lambda\geq0$ weights the auxiliary loss. Optimization acts on the designated trainable components, with any frozen backbone held fixed. The retained components of $\theta^\star$ initialize the deployed parameters $\theta_0$; the auxiliary prediction head can be removed when action generation does not depend on its output. Equation~\eqref{eq:predictive-policy}, by comparison, retains a future-dependent action computation at deployment.

A geometric intermediate can improve training without becoming an online execution dependency. Gripper traces provide motion-specific supervision, while reasoning dropout or removal of the future-trace head permits execution without generating those traces~\citep{nguyen2026iclr,son2026seetraceact}. The learned action representation then carries the benefit, avoiding a separate prediction stage at each decision. Internal semantic reasoning can play a similar role: PonderPounce conditions action inference on an asynchronous cognition token while keeping subgoal text inside the model~\citep{choi2026ponderpounce}. The intermediate shapes \mbox{the representation used for action inference.}

\surveyanchor{discuss:fig:prediction-role}
The distinction is therefore between two computational roles of prediction (Figure~\ref{fig:prediction-role}). In auxiliary predictive supervision, future targets shape the policy during training, but actions do not depend on an explicit future prediction at deployment. In prediction-conditioned control, the predicted future remains an input to action generation. In panel (a), context layers denote the trainable context and state--context transformers after the frozen visual backbone. In panel (b), the video and action blocks denote the trainable prediction branches, not the input frames. Auxiliary prediction strengthens the representation used by a context-conditioned policy. Deployment-time predictions guide execution directly, coupling control quality to the generated future.

Predictive control separates task interpretation from its physical realization through an anticipated consequence. The anticipated consequence exposes an intermediate for inspection or planning. Its contribution to control depends on both forecast accuracy and the action interface that realizes it, making prediction and decoding distinct sources of error. Skill- and agent-based execution makes a different abstraction: it specifies operations and delegates their consequences to available skills or tools.

\subsection{Skill- and agent-based execution}
\label{sec:structured-methods}

Skill- and agent-based execution assumes that existing skills or robot tools can realize a task once context has specified their selection, order, and arguments. A video showing a red block placed in the left bin before a blue block enters the right bin can become an ordered procedure. The executor supplies the available motor competence; contextual inference determines how to use it. Two branches make this separation concrete: learned skill composition and VLM and LLM agents that generate programs or tool requests. After defining their common specification--executor interface, we compare these branches and use feedback-driven revision to show when a changed specification suffices and when execution itself must be learned.

The specification can take the form of code, spatial targets, or an ordered sequence of learned skills. Its defining role is to govern an execution stage that can track or advance it, as described in Section~\ref{sec:foundations}. At that action interface, the deterministic form is
\begin{equation}
 g=f_\theta(C_t,h_t), \qquad
 A_t=\kappa(g,o_t),
 \label{eq:structured}
\end{equation}
where $f_\theta$ extracts the execution specification $g$ from context $C_t$ and robot history $h_t$, and $\kappa$ realizes it as the next action block $A_t$ using observation $o_t$. For stochastic specification generation and execution, \mbox{the same interface becomes}
\begin{equation}
 \pi_\theta(A_t\mid h_t,C_t)=
 \int k_\theta(A_t\mid g,h_t)\,
 p_\theta^{\mathrm g}(g\mid C_t,h_t)\,\mathrm{d}g.
 \label{eq:structured-policy}
\end{equation}
Here $p_\theta^{\mathrm g}$ generates execution specifications and $k_\theta$ is the executor's action distribution. Discrete programs or skills replace the integral with a sum; deterministic generation and execution recover Equation~\eqref{eq:structured}. In this family, $g$ specifies operations, their arguments, or their composition for a separate executor. A transferred geometric reference belongs to the geometric family when reference tracking is the primary dependency; hybrids can combine both interfaces. A predicted future can also be decoded into $g$, composing this interface with world-model-based control. Neural components of the executor belong to $\theta$; numerical arguments and generated code belong to $g$. Revising these arguments or code can change behavior without training \mbox{the executor's neural weights.}

Skill composition and program execution instantiate this interface as
\begin{equation}
 \begin{aligned}
 \text{Skill composition:}\quad
 g&=(z_1,\ldots,z_{N_g}), & A_t&=\kappa(z_{\ell_t},o_t),\\
 \text{Program synthesis:}\quad
 g&=f_\theta(C_t,h_t), & A_t&=\operatorname{Run}(g,h_t).
 \end{aligned}
 \label{eq:structured-subfamilies}
\end{equation}
In the first row, $\ell$ indexes skills, $z_\ell$ is a skill specification, $N_g$ is the number of skills, and $\ell_t\in\{1,\ldots,N_g\}$ identifies the active skill. In the second, $f_\theta$ generates program $g$, and $\operatorname{Run}$ executes it until it produces the next action block, including permitted perception and control calls. Tool-using agents can generate one request at a time: the tool name and arguments form $g$, the robot adapter implements $\kappa$, and its returned observation updates the next decision. A generated program may call a geometric tracker, thereby composing this family with geometric demonstration transfer.

\paragraph{Composing reusable skills.}
An early insight was that rearranging existing motor capabilities can express a new task. Neural Task Programming~\citep{xu2017ntp} made this explicit in 2017: a demonstration instantiated a hierarchy whose leaves invoked robot routines. The transferable knowledge was the decomposition and its arguments, while execution reused the available primitives. This separated generalization in task length, ordering, and object assignments from learning \mbox{the physical skills themselves.}

Human-video teaching requires recognizing behavior across bodies, scenes, and timing. XSkill learns shared human--robot skill prototypes~\citep{xu2023xskill}; UniSkill scales skill representation learning to unaligned cross-embodiment video~\citep{kim2025uniskill}. Both use human examples to guide a robot-side skill-conditioned policy, extending the interface from reordering known routines to recognizing transferable behavior in another body's demonstration.

Skill abstraction must preserve the taught distinction: attributed waypoints retain motion targets~\citep{chang2023awda}, object-relation graphs retain dependencies~\citep{zhu2024orion}, and interaction-aware procedures retain stage changes~\citep{chen2025manilong}. The executor's interface determines which of these constraints reaches motion. Semantic subtasks can alternatively condition a continuous action predictor internally, as in StellaVLA~\citep{xu2026stellavla}. An abstraction's computational role therefore depends on whether it supplies a separate executor or the policy's own action inference.

A composed procedure also needs evidence for when to advance. StageGuard distills demonstration-grounded completion reasoning into a lightweight monitor that decides whether to continue, advance, or skip a skill~\citep{arxiv260920791}. Navi-Agent verifies local navigation subgoals against a persistent topology of visual places and executed transitions~\citep{arxiv260920388}. Observed progress governs advancement through the composed procedure.

Navigation illustrates two levels of example-conditioned execution. Instruction--landmark examples in LM-Nav~\citep{shah2022lmnav} and instruction--subtask examples in A$^2$Nav~\citep{chen2023a2nav} guide a language parser that specifies what pretrained navigators should execute. Select2Plan~\citep{buoso2024select2plan} retrieves visual decision examples for selecting marked movements or waypoints. The reasoning output is an executable choice, while motion realization belongs to the available navigator. Terrain feedback can further change the costs attached to those choices, as in VLM-GroNav~\citep{elnoor2024vlmgronav}. Section~\ref{sec:navigation-demands} compares the information supplied by these contexts.

\paragraph{VLM and LLM agents: programs, tools, and feedback.}
Language-model pretraining expanded adaptation to program generation in 2022. Code as Policies~\citep{liang2022codeaspolicies} uses command--program examples to recombine perception, control, arithmetic, and feedback logic; program examples also specify available objects and interfaces~\citep{singh2022progprompt}. Multimodal reasoning extends this route to demonstrated procedures. Show-Harness converts video frames into a textual outline for an unchanged planner and an embodiment-specific action interpreter~\citep{chen2026showharness}. The outline carries procedural structure into execution; spatial and contact constraints must also appear in this representation to guide the action interpreter.

Preserving contact structure requires a richer executable abstraction. RAPID derives local trajectory-optimization primitives, relational composition constraints, and a reconstructed verification environment from a human demonstration~\citep{arxiv260930249}. The inferred predicates and simulated scene variants support iterative program revision before the resulting primitives and strategy are frozen for transfer. Compared with an outline over supplied skills, teaching here helps construct both the available operations and their checks. Transfer depends on whether the inferred relations preserve the demonstrated interaction and whether reconstruction makes verification physically informative. Its nonprehensile experiments support this demonstration-conditioned pathway; the separate LIBERO-Pro experiment instead constructs programs without demonstrations. Reconstruction and program search contribute to the time and interaction needed to acquire the new task.

The command interface determines which decisions contextual inference must resolve. GPT-Policy uses goal images or demonstrations, available state and action records, and interaction feedback to request Cartesian targets or gripper commands~\citep{cheng2026gptpolicyeval}. RoboDawn instead grounds discrete translation, rotation, and gripper commands through in-context examples~\citep{arxiv260922966}. In both cases, fixed models request actions from existing motor competence and use returned feedback to revise subsequent requests.

Tools expand the evidence and behavior available to this inference. Robo-Harness K1 exposes calibrated depth, spatial measurements, persistent visual anchors, and grasp hypotheses to ground generic motion requests~\citep{li2026roboharnessk1}. The agentic framework RACaP exposes typed Policy APIs: a ReAct agent combines working memory, retrieved experience, and visual feedback to choose calls, arguments, and recovery steps~\citep{li2026racap}. Both retain fixed neural weights during deployment; RACaP also freezes its generated source code.

The memory--executor interface sets another limit. In 2AM, the agent retains task history but sends only subtask language and optional 2D grasp, place, and move hints to an episodically stateless action model~\citep{arxiv260911308}. These cues bind remembered intent to current-image instances without asking the agent to produce robot trajectories. Their validity differs: a grasp cue expires after acquisition, whereas a destination can persist during transport. The action model is trained to interpret noisy or incomplete hints. External task memory works through a learned steering interface. Its expressiveness determines which temporal and contact-dependent requirements can reach the action model.

Previewable actions add a further use of context: a candidate can be revised before it changes the scene. World Action Agent (WAA) selects interaction-centered views, renders proposed poses with kinematic planning feedback, and converts in-view corrections into robot motion~\citep{zhang2026waa}. Its main agent consults a separate Skill Agent for demonstration-derived procedures and visual checks. Tool requests form the execution specification, while geometric planning checks their feasibility before control.

After execution, the interface determines which feedback reaches the next decision. ETA, implemented in OpenETA~\citep{chen2026eta}, requires a fresh observation after each world-changing tool call. RoboHarness addresses transitions between pretrained policies with different operating conditions: execution memory selects a policy and retrieves states that guide entry into its operating region~\citep{huang2026roboharness}. Feedback thus supports both revising an action and selecting the competence needed to execute it.

Generated programs can also construct spatial affordance and constraint maps for geometric motion planning~\citep{huang2023voxposer}. The interface allocates competence: a named pick-and-place routine supplies perception and grasp synthesis, whereas pose-level control exposes those decisions to the agent. Improved success may consequently reflect a stronger supplied executor as well as better contextual inference.

A plausible procedure can still request an action that the current scene cannot support. Skill-affordance estimates address this gap before execution by coupling linguistic relevance to physical feasibility~\citep{ahn2022saycan}. Scene and execution feedback address it after a transition by revising what should happen next~\citep{huang2022monologue}. When feasibility depends on an unknown property, targeted measurements such as proprioceptive estimates of mass or stiffness can change the manipulation strategy~\citep{physcap2026}. These are complementary uses of evidence: predict applicability, observe the outcome, and acquire information missing from either. They connect program-level reasoning to the actual conditions under which a skill works.

\begin{figure}[!tp]
\centering
\includegraphics[width=\linewidth]{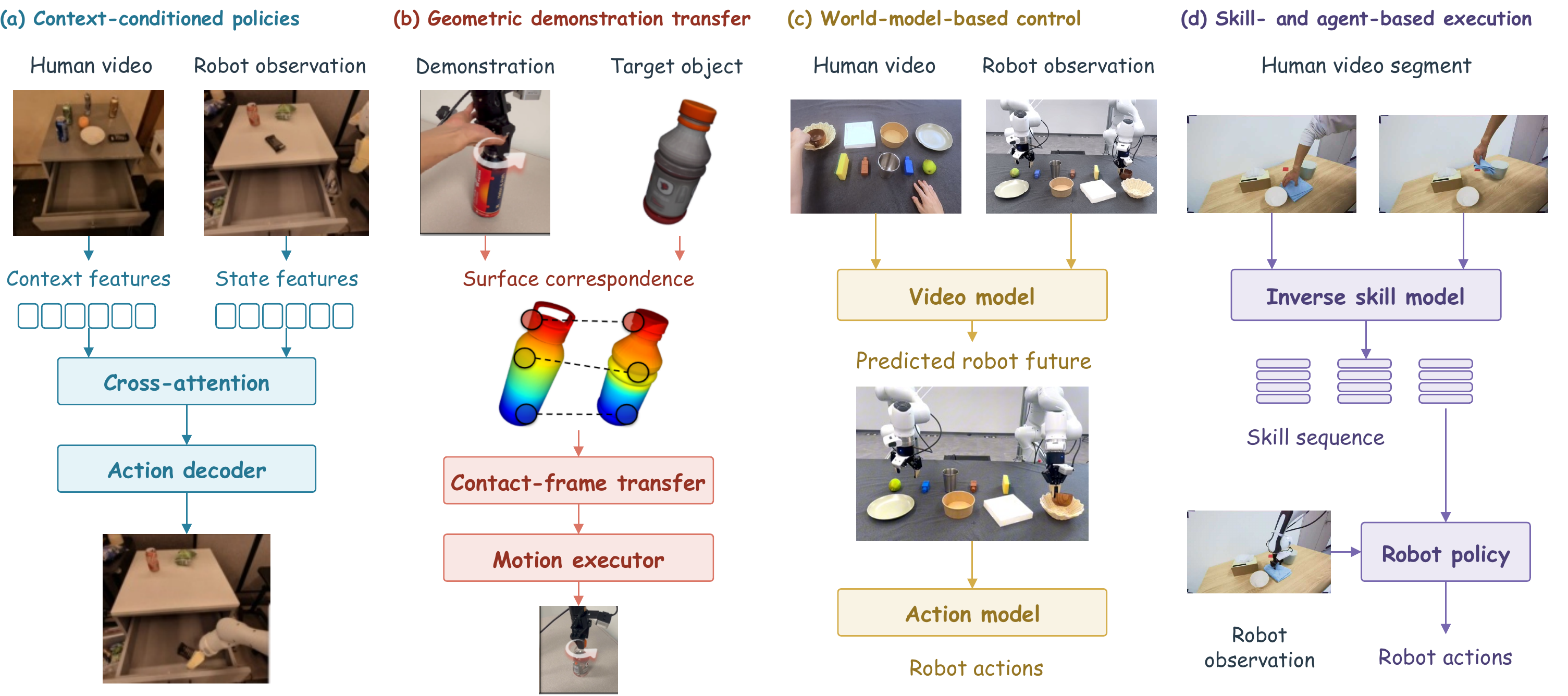}
\caption{Four technical paradigms convert context into actions through (a) direct action inference, (b) geometric motion-reference transfer, (c) prediction-conditioned control, and (d) skill or tool execution. Representative implementations are Vid2Robot~\citep{jain2024vid2robot}, SemAnCorr~\citep{dong2026semancorr}, Zero-WAM~\citep{zhou2026zerowam}, and UniSkill~\citep{kim2025uniskill}, respectively.}
\label{fig:method-taxonomy}
\par\phantomsection\label{floatend:fig:method-taxonomy}
\end{figure}

Feedback-driven revision changes either the procedure or the learned capability used to execute it. A frozen language model can adjust a primitive's direction, distance, or duration from outcome errors~\citep{merwe2025icpi}; these values remain arguments of $g$. CaP-X contrasts feedback-driven code generation with reinforcement learning of the coding policy~\citep{arxiv260322435}. ENPIRE exposes both choices within physical experimentation: revise perception and control programs, or collect experience to train a neural policy~\citep{xiao2026enpire}. VoxPoser's online dynamics-learning extension similarly changes a learned component~\citep{huang2023voxposer}. The practical distinction is what the correction must supply. Recombining available operations changes their use; acquiring a missing predictive or motor capability requires a learning process for that component.

Across these forms, explicit structure exposes task order, constraints, and intervention points. Its central trade-off is \emph{portability versus physical detail}: a compact procedure can cross scenes and bodies while omitting a required grasp or compliance condition. The executor can recompute only details left open by the teaching. DualManip separates the corresponding update rates: live geometric correspondence revises grasp contacts while semantic reasoning is invoked when geometric validation fails~\citep{arxiv260931112}. This hybrid preserves task intent across motion without requiring full replanning for every observation.

\surveyanchor{discuss:tab:predictive-agent-comparison}
Table~\ref{tab:predictive-agent-comparison} separates context operations from their carriers. \textit{Encode} commonly supports learned future prediction or skill inference; \textit{Parse} exposes procedures and calls; \textit{Retrieve} and \textit{Verify} can augment either route. A video can become a future, a motion reference, a skill, or a program, each preserving different task constraints for its executor. The execution column exposes a second difference: a predicted future can guide planning or condition action decoding, while a motion reference can support tracking or replay. The carrier and execution operation jointly specify the transfer assumption.

\begin{table}[!tp]
\centering\surveytable
\caption{Predictive and skill/agent interfaces (Sections~\ref{sec:predictive-methods}--\ref{sec:structured-methods}). Carriers identify intermediate representations; families follow the full control path. Listed deployment branches keep neural weights fixed. Codes: D examples, H history, I instruction/goal, F feedback, E scene/map; V visual, S state/action, G geometry, L symbolic. Slashes combine entries. }
\label{tab:predictive-agent-comparison}
\renewcommand{\arraystretch}{1.18}
\begin{tabularx}{\linewidth}{@{}>{\raggedright\arraybackslash}p{.25\linewidth}>{\centering\arraybackslash}p{.07\linewidth}>{\centering\arraybackslash}p{.10\linewidth}>{\raggedright\arraybackslash}p{.20\linewidth}>{\raggedright\arraybackslash}p{.15\linewidth}Y@{}}
\toprule
\rowcolor{black!12}\textbf{Method} & \textbf{Source} & \textbf{Form} & \textbf{Context operation} & \textbf{Carrier} & \textbf{Execute}\\
\midrule
\rowcolor{panel}\multicolumn{6}{c}{\strut\itshape I. World-model-based control}\\
\rowcolor{black!3}OSVI-WM~\citep{goswami2025osviwm} & D & V & Encode & Future & Plan\\
Demo-JEPA~\citep{he2026demojepa} & D & V & Encode & Future & Plan\\
\rowcolor{black!3}HOST~\citep{chen2026host} & D & V & Encode & Future & Decode\\
Zero-WAM~\citep{zhou2026zerowam} & D & V & Encode & Future & Decode\\
\rowcolor{black!3}WorldScape Policy 2.0~\citep{arxiv260718840} & D,H & V,L & Encode & Future / Actions & Decode\\
ReCAP (WAM)~\citep{park2026recap} & D & V,S & Retrieve / Encode & Future / Actions & Decode\\
\rowcolor{black!3}MaP-WAM~\citep{arxiv260911561} & H & V & Align / Encode & Future & Decode\\
FARE~\citep{arxiv260918016} & H & V,S & Verify & Future / Actions & Decode\\
\rowcolor{panel}\multicolumn{6}{c}{\strut\itshape II. Skill- and agent-based execution: learned skills}\\
\rowcolor{black!3}XSkill~\citep{xu2023xskill} & D & V & Encode & Skill & Decode\\
AWDA~\citep{chang2023awda,goswami2025osviwm} & D & V & Encode & Motion & Track\\
\rowcolor{black!3}ORION~\citep{zhu2024orion} & D & V,G & Parse / Align & Program & Invoke\\
UniSkill~\citep{kim2025uniskill} & D & V & Encode & Skill & Decode\\
\rowcolor{panel}\multicolumn{6}{c}{\strut\itshape II. Skill- and agent-based execution: programs and tool calls}\\
\rowcolor{black!3}Code as Policies~\citep{liang2022codeaspolicies} & D,I & L & Parse & Program & Invoke\\
Show-Harness~\citep{chen2026showharness} & D & V & Parse & Program & Invoke\\
\rowcolor{black!3}GPT-Policy~\citep{cheng2026gptpolicyeval} & D,H,I & V,S,L & Parse & Calls & Invoke\\
WAA~\citep{zhang2026waa} & D & V,G & Parse / Verify & Calls & Invoke\\
\rowcolor{black!3}RAPID~\citep{arxiv260930249} & D & V,G & Parse / Verify & Program & Invoke\\
Robo-Harness K1~\citep{li2026roboharnessk1} & I,E & V,G,L & Parse / Align & Calls & Invoke\\
\rowcolor{black!3}RACaP~\citep{li2026racap} & H,I,F & L & Retrieve / Parse & Calls & Invoke\\
2AM~\citep{arxiv260911308} & H,I & V,L & Parse & Calls & Decode\\
\rowcolor{black!3}Mimir~\citep{arxiv260804933} & H,I & L & Parse / Verify & Calls & Invoke\\
SparseNav~\citep{arxiv260926408} & I,E & G,L & Parse / Align & Motion & Track\\
\rowcolor{black!3}DROC~\citep{zha2023droc} & H,F & L & Retrieve / Parse & Program & Invoke\\
\bottomrule
\end{tabularx}
\par\phantomsection\label{floatend:tab:predictive-agent-comparison}
\end{table}

\paragraph{Choosing an interface by the constraint it preserves.}
A useful interface preserves the task distinction most likely to be lost before execution. Direct action inference can exploit learned correspondence when the teaching resembles its training relationships. Geometric transfer exposes a binding contact or relative motion for inspection and tracking. Predictive control makes consequences available for selection, while skill and program interfaces expose ordering and preconditions. These choices relocate the generalization burden: to the learned interpreter, the matched interaction, the future model, or the supplied executor. Hybrids compose these assumptions. A predicted future can yield waypoints, and a program can invoke a geometric tracker. Their placement follows the intermediate that carries the demonstrated requirement into execution, allowing a hybrid to combine several families.

\subsection{Shared mechanisms: correspondence and memory}
\label{sec:context-mechanisms}

All four interfaces face three linked requirements. A taught task must first be bound to the objects in the current scene, including replacements for those in the demonstration. Its evidence must remain available at the decision that needs it, and its interpretation must express a change the controller can execute. The discussion therefore moves from transfer across objects to access across time and then to the command interface that realizes a correction.

\surveyanchor{discuss:fig:method-taxonomy}
Figure~\ref{fig:method-taxonomy} shows the intermediates passed to control: attended demonstration features, transferred contact frames, a predicted robot future, and a skill sequence. Evidence selection and interpretation (Figure~\ref{fig:mechanism-families}) precede these computations. Correspondence binds the evidence to the current scene; memory preserves it until the relevant decision.

\paragraph{Cross-object transfer of demonstrated tasks.}

A demonstration can specify a relation that remains meaningful after its objects are replaced. Consider a robot shown how to pour from a jug into a cup. It may then use a bottle with the same cup, keep the jug but pour into a bowl, or use both the bottle and the bowl. Table~\ref{tab:object-transfer} contrasts these three settings using this illustrative task. The held vessel is the manipulated object; the cup or bowl is the receiving object, a role that can also be a support or contact surface in other tasks. Each role can change within a category or through a functionally compatible substitute. Novelty relative to the demonstration and novelty relative to training remain separate properties. We first examine which relations transfer between new instances, then consider functional substitution and the control needed to realize the transferred requirement.

\begin{table}[!tp]
\centering\surveytable
\caption{Object substitution illustrated by pouring. The demonstration uses a jug and a cup; all three settings preserve the goal of pouring into the receiver. Arrows indicate object replacement; the last column \mbox{gives example motion adjustments.}}
\label{tab:object-transfer}
\begin{tabularx}{\linewidth}{@{}>{\raggedright\arraybackslash}p{.18\linewidth}>{\raggedright\arraybackslash}p{.20\linewidth}>{\raggedright\arraybackslash}p{.20\linewidth}Y@{}}
\toprule
\textbf{What changes} & \textbf{Held vessel} & \textbf{Receiver} & \textbf{Motion adjustments}\\
\midrule
Held vessel & Jug $\rightarrow$ bottle & Same cup & Grasp; tilt angle\\
\addlinespace[3pt]
Receiver & Same jug & Cup $\rightarrow$ bowl & Pouring position; height\\
\addlinespace[3pt]
Both objects & Jug $\rightarrow$ bottle & Cup $\rightarrow$ bowl & Grasp; tilt; pouring position\\
\bottomrule
\end{tabularx}
\par\phantomsection\label{floatend:tab:object-transfer}
\end{table}

\surveyanchor{discuss:tab:object-transfer}
In this example, the robot must still direct the liquid into the receiver. A bottle can require a different grasp and tilt angle from the demonstrated jug; a bowl can require a different pouring position and height from the cup. Replacing both couples these adjustments: the new vessel must be grasped and oriented so that its outlet directs the flow into the new opening. The transferred requirement is thus preserved while the motion changes. Each setting isolates a different set of coupled geometric adjustments.

\paragraph{From object identity to task-relevant relations.}
An unchanged object in a new pose often permits relocation of a reference. A new instance can change the grasp, contact surface, or clearance even when its role stays fixed. R-NDF transfers relations such as placing a bowl on a mug or a bottle in a container to previously unseen object pairs~\citep{simeonov2023rndf}. Its original setting uses five to ten demonstrations with pretrained category representations and a marked 3D keypoint. Descriptor alignment recomputes placement with those representations fixed; the paper also studies a learned energy-based refinement, which requires training an additional neural component. A part-based representation supports one-demonstration transfer across larger shape variation~\citep{thompson2026parttransfer}. Its real-robot tests include new mug--rack and bowl--mug pairs, using learned part models and segmentation assistance. These results connect the relation shown in a demonstration to execution on new objects while exposing the prior geometry and human input that make the transfer possible.

\paragraph{From new instances to functional substitutes.}
Category changes make the preserved role more important than whole-object similarity. FUNCTO transfers tool use from an actionless human RGB-D video and task description, distinguishing spatial, instance, and category generalization~\citep{tang2025functo}. Its functional keypoints locate how a replacement tool should be grasped and used; the experiments emphasize tool variation, with less emphasis on changes to the receiving object. SemAnCorr transfers a pouring motion from a kettle to a toy cup and a cutting motion from scissors to a nail clipper, using recorded contact and motion from kinesthetic teaching~\citep{dong2026semancorr}. These different evidence interfaces support functional correspondence without requiring a common demonstration format. Cross-category matching can also be supplied: the part-decomposition study evaluates pre-pouring alignment for a watering-can substitution using provided part equivalencies~\citep{thompson2026parttransfer}. This protocol isolates geometric alignment with the cross-object correspondence already specified.

Functional substitution must preserve the requirement actually taught. A general request to transfer contents can admit another vessel; a specified vessel or handle contact can remove that freedom. At a higher level of abstraction, preserving a purpose may require a different procedure rather than a transformed motion. The Imitator Game distinguishes category-preserving substitution from this affordance-adapted imitation~\citep{zhou2026imitator}. The invariant is therefore a task-dependent choice: object relation, contact constraint, process, or intended effect. Section~\ref{sec:context-transfer-evaluation} tests whether the demonstration determines that choice.

\paragraph{Realizing a transferred requirement.}
The four control interfaces consume this requirement differently. A context-conditioned policy can bind it within action features; geometric transfer exposes a reference; predictive control can express a desired future; and a skill or program can bind its arguments to new objects. Each interface preserves the requirement in a different form for its executor. Geometry used to train a policy~\citep{fang2024kalm,haldar2025pointpolicy} also differs from a reference supplied at deployment or a demonstration-fitted motion system~\citep{li2025emp}. Likewise, spatial affordance prediction and spatial action representations provide useful priors~\citep{yuan2024robopoint,qu2025spatialvla}; a newly supplied demonstration must still determine which requirement \mbox{the prior should realize.}

Relational skill guidance provides an agent-based route to transfer. WAA's frozen LIBERO-derived skills transfer to robosuite lifting, stacking, and restacking, where objects, scenes, and viewpoints change; grounding resolves poses from the current scene~\citep{zhang2026waa}.

When interpretation produces a desired pose or feature configuration, a separate controller can track it. For a stationary target, a local visual-servo rule is
\begin{equation}
 \boldsymbol e_t=\boldsymbol s_t-\boldsymbol s_t^\star,
 \qquad
 \dot{\boldsymbol q}_t=-k_{\mathrm{vs}} J_{\mathrm{vis},t}^{\dagger}\boldsymbol e_t.
 \label{eq:visual-servo}
\end{equation}
Here $\boldsymbol s_t$ and $\boldsymbol s_t^\star$ are current and desired visual features, $\boldsymbol e_t$ their error, $\boldsymbol q_t$ the joint configuration, and $J_{\mathrm{vis},t}$ the feature Jacobian. Its pseudoinverse and positive gain $k_{\mathrm{vs}}$ map feature error to joint velocity~\citep{chaumette2006visualservo}. Context changes the reference while feedback corrects its realization. New contact or dynamics can still invalidate that realization, even with correct correspondence. The final part of this subsection examines this execution boundary. Transfer also requires the defining evidence to remain accessible when needed, motivating \mbox{the memory mechanisms below.}

\paragraph{Accessing and updating evidence across time.}
Alignment selects information within a demonstration, retrieval selects records from an archive, and memory preserves evidence across decisions. These operations can expose policy examples, motion references, predicted rollouts, or programs. The retrieved object's use in control determines its method family; the access mechanism itself does not.

A useful memory must answer three questions: which fact matters, how it can be recovered, and when it stops being valid. Task requirements, physical-response estimates, and completed-step records need different update rules (Table~\ref{tab:memory-roles}). A moved part can invalidate its remembered pose while leaving the taught order intact. The read--update decomposition separates stored evidence from the policy input; subsequent comparisons relate selection, compression, and reset timing to the decision they support. Sections~\ref{sec:recovery} and~\ref{sec:last-mile} develop validity during execution and reuse.

\paragraph{Carrying evidence between decisions.}
A recurrent state $m_t$ and an external archive $M_t$ provide complementary ways of retaining experience. Let $x_t=(o_t,a_t,o_{t+1})$ be an executed transition, augmented with feedback when available. A common read--update decomposition is
\begin{equation}
 \begin{aligned}
 m_{t+1}&=u_\theta(m_t,x_t), &
 M_{t+1}&=w_\theta(M_t,x_t),\\
 R_t&=\rho_\theta(h_t,M_t), &
 C_t&=b_\theta(C_t^{\mathrm{task}},m_t,R_t).
 \end{aligned}
 \label{eq:context-state}
\end{equation}
The maps $u_\theta$ and $w_\theta$ update recurrent and external memory, $\rho_\theta$ retrieves records $R_t$, and $b_\theta$ constructs the policy input. The supplied task evidence $C_t^{\mathrm{task}}$ includes instructions, goals, examples, and newly received corrections in any supported modality; it reduces to $D^{\mathrm{s}}$ in a demonstration-only configuration. The retained state $(m_t,M_t)$ and the selected input $C_t$ are distinct. An implementation may omit either memory component, and its maps may combine analytic operations with fixed neural components. This decomposition generalizes the self-contained update in Equation~\eqref{eq:memory} without restricting context to demonstrations.

\paragraph{Retaining the information that changes control.}
Early latent task inference compressed transitions to identify reward or dynamics~\citep{li2020focal,yuan2022corro}. Selective retention matters because a demonstrator's collection style can correlate with task identity in training logs without identifying the task reliably at deployment.

Physical adaptation depends on when evidence is acquired and how long it remains relevant. RMA uses recent task transitions; ICWM collects task-agnostic probes; LocoFormer carries interaction across trials~\citep{kumar2021rma,wang2026icwm,liu2025locoformer}. RopeFormer extends cross-trial action--response conditioning to dynamic ropes, retaining policy context while resetting the physical trial~\citep{arxiv260923432}. Zeva and Zeva-Ego instead encode action effects at two timescales; the latter combines recent within-attempt traces with phase-matched memory across attempts on the same task instance~\citep{chen2026zeva,arxiv260924411}. These fixed-weight interfaces trade the cost of collecting evidence against its \mbox{relevance to subsequent control.}

Event memory addresses a different representation need: an earlier observation may establish that a button was pressed, even when the current scene no longer reveals that event. Perceptual retrieval and semantic event records preserve such evidence with different detail and processing costs~\citep{shi2025memoryvla,sridhar2025memer,torne2026mem}. Recurrent designs compress visual history at frame or action-chunk timescales~\citep{li2026rememvla,cherepanov2026muvla}; event-triggered storage retains selected raw keyframes~\citep{yang2026eventvla}. MemBodied combines a fixed-size associative interaction state with an initial-scene anchor, preserving early evidence alongside later updates~\citep{pala2026membodied}. ARMS addresses concurrent execution by asynchronously supplying perception, embodied state, and records of which arm acted and when~\citep{yi2026arms}. These designs retain different temporal structure: a stable reference, accumulated events, or actor-specific action history. Their deployed memory updates change the evidence read by a trained policy. HIRE retains ordered wrench traces to distinguish visually aliased interaction states, coupling this history-conditioned action inference to a faster contact executor~\citep{arxiv260930828}. Its shuffled-history comparison tests whether temporal order, rather than force availability alone, supplies the missing state evidence.

Explicit memory can expose the binding that a compressed state leaves implicit. Mimir separates world facts from task progress, retrieving candidates for the active goal and marking completion only when feedback supports its postcondition~\citep{arxiv260804933}. OCC4M makes a complementary distinction between persistent object identity and a remembered place~\citep{arxiv260928798}. A selected object's target follows its updated track; a historical-location target remains fixed and is reprojected into the current view. Together, these designs show why memory must preserve the intended referent and its update rule. Object motion updates a tracked referent, while task progress advances after the required postcondition is observed. OCC4M evaluates moving views in simulation with calibrated support geometry and uses a fixed camera in its hardware study.

Past evidence and imagined futures answer different temporal questions. An observation window establishes what happened, as in SimpleMemVLA~\citep{yin2026simplememvla}; combining retrieved events with imagined states, as in MemoryVLA++, adds what could happen next~\citep{shi2026memoryvlapp}. Acquiring a new rule for what to remember poses a further generalization problem. In $\mu$VLA, recurrence helps held-out tasks with familiar memory requirements but yields little improvement when the required memory structure changes~\citep{cherepanov2026muvla}. The distinction is between recalling evidence under a learned rule and inferring which evidence an unfamiliar task makes relevant.

Where selection is performed changes both cost and flexibility. Workspace Models move VLM-based saliency selection into training: an encoder learns latent workspace tokens by reconstructing selected historical image patches, and a downstream policy learns to act from those tokens~\citep{arxiv260920820}. This amortizes online reasoning but also inherits the saliency supervision's information choices. A comparison with runtime retrieval can vary decision latency and memory demand together, testing both familiar event recall and \mbox{the retention of new teaching.}

History can also reconstruct an unobserved part of the present, rather than a past event or future consequence. LIFD combines recurrent evidence with geometry-anchored scene-token completion before action prediction~\citep{arxiv260919796}. This distinction matters for classification: generating missing current state supports context-conditioned inference, whereas world-model-based control requires predicted consequences \mbox{on the execution path.}

\paragraph{Selecting, compressing, and using history.}
The benefit of memory depends on which event survives compression and whether the current decision needs it. Selection and ordering already matter in language ICL \citep{dong2024iclsurvey}; robotics adds short physical events whose removal changes the meaning of an otherwise similar trajectory. Robust Instant Policy's downsampling study identifies the loss of brief events such as a grasp transition~\citep{oh2025rip}; its full system additionally fits a policy at deployment.

The useful horizon follows the unknown. AMAGO's MetaWorld ML-1 analysis finds that short histories can identify hidden goals, while longer contexts can reduce sample efficiency~\citep{grigsby2023amago}. ReLIC instead reuses experience across episodes to locate objects in an unfamiliar home, with deployed neural parameters fixed~\citep{elawady2024relic}. Section~\ref{sec:navigation-demands} relates this longer horizon to acquisition of spatial knowledge.

Retrieval faces two decisions: which event answers the current uncertainty, and whether that uncertainty requires history at all. Question-answering supervision can teach relevance, as in HALO~\citep{shah2026halo}; a calibrated history gate can give current observations precedence when they already suffice~\citep{gao2026gmp}. The decisions expose different errors: missing a relevant event or allowing an irrelevant record to override the current scene.

For an archive $M_t=\{\xi_j\}_{j=1}^{N_t}$, $\xi_j$ is its $j$th record and $N_t$ is the number of records available at decision $t$. Let $J$ be a candidate set of record indices and $J_t$ the selected set. A budgeted retrieval rule is
\begin{equation}
 J_t\in\arg\max_{J\subseteq\{1,\ldots,N_t\},\,|J|\leq K}
 \sum_{j\in J}\bigl[s_\theta(h_t,\xi_j)-\eta\bigr],
 \qquad R_t=(\xi_j)_{j\in J_t}.
 \label{eq:memory-retrieval}
\end{equation}
Here $|J|$ counts selected records, $K$ is the maximum record count, and $\eta$ is the relevance threshold. The compatibility score $s_\theta$ is used as in Equation~\eqref{eq:demo-alignment}, with an archive record as its second argument. The retrieved sequence $R_t$ contains the records indexed by $J_t$. A fixed ordering rule arranges selected records in $R_t$; a fixed tie rule resolves equal scores. The empty set permits a read with no useful history. When retrieval supplies the action-reuse policy in Equation~\eqref{eq:action-reuse}, a nonempty selection yields its normalized mixture; an empty read instead requires the system\textquotesingle{}s fallback policy or an information-acquisition step. This additive form describes thresholded selection; sequence-aware retrieval can score combinations when the value of one \mbox{event depends on another.}

\begin{table}[!tp]
\centering
\surveytable
\caption{Four roles of contextual evidence and when to recheck their applicability. One episode can supply several roles, but each can remain valid for a different duration. Representative mechanisms illustrate each role; Table~\ref{tab:context-memory} compares storage and access.}
\label{tab:memory-roles}
\begin{tabularx}{\linewidth}{@{}>{\raggedright\arraybackslash}p{0.25\linewidth}>{\raggedright\arraybackslash}p{0.32\linewidth}Y@{}}
\toprule
\textbf{Information role} & \textbf{Evidence and mechanisms} & \textbf{Recheck after} \\
\midrule
Task specification & Plate-first order; corrections and relational programs~\citep{zha2023droc,arxiv260930249} & A new task or correction \\
\addlinespace[2pt]
Correspondence & Matched target or phase; object tracks and trace retrieval~\citep{arxiv260928798,arxiv260920646} & Object or execution-phase changes \\
\addlinespace[2pt]
Physical response & Action effects; trial context and hardware-shift tokens~\citep{chen2026zeva,arxiv260930092} & Tool, material, or contact changes \\
\addlinespace[2pt]
Execution state & Verified completion; task agendas and observed segment records~\citep{arxiv260804933,arxiv260911561} & New action, observation, or scene reset \\
\bottomrule
\end{tabularx}
\par\phantomsection\label{floatend:tab:memory-roles}
\end{table}

Motion-sensitive features can retain executed change and temporal order~\citep{arxiv260826821}. The next computation determines whether this evidence supports progress estimation or physical-response inference. The memory mechanism need not change when its information serves a different control purpose.

\paragraph{Stage-dependent retrieval and action refinement.}
Relevant experience must match execution progress as well as object appearance. RTCF aligns the growing history to successful trajectories and transfers only bounded low-frequency components from the selected action chunk \citep{fan2026rtcf}. The frozen policy supplies the remaining motion components and gripper decisions. Temporal correspondence thus selects evidence, and selective residual transfer limits how it changes control. Its fixed evaluation bank isolates the use of retained trajectories from online archive growth.

Failure traces add evidence about continuations to avoid. TraceFlow builds a bounded guidance field from progress-aligned successful and failed action traces, steering a frozen flow-matching expert without changing its weights~\citep{arxiv260920646}. This extends the role of retrieval from selecting an example to shaping action generation through outcome contrast. Reported ordering gains coexist with unresolved counting and occlusion deficits. These results expose two linked requirements: retained event evidence identifies the relevant stage, and action-space guidance favors a better continuation within it.

\paragraph{Persistence and update timing.}
Persistence is determined by the state that survives a boundary between decisions or attempts. Recurrent state and verified subgoal records retain information without changing the action generator~\citep{arxiv260829537}; reference traces and task memory can also organize repeated calls to fixed primitives~\citep{arxiv260708448}. Fast-weight methods encode evidence in an optimized parameter subset, either during the sequence~\citep{jiang2026robottt} or before rollout~\citep{arxiv260706988}. Learning a value function from deployment experience changes a further component while leaving the actor fixed~\citep{arxiv260821204}. These choices differ in reset behavior and acquisition cost. Clearing observations, resetting adaptive weights, and restoring a critic to its initial state isolate these different adaptation components even when the actor remains unchanged.

\FloatBarrier
\surveyanchor{discuss:tab:memory-roles}
Table~\ref{tab:memory-roles} makes the information needs concrete: recalling a completed placement preserves progress, whereas retaining an insertion response supports physical adaptation. The same storage mechanism can serve either purpose; Table~\ref{tab:context-memory} in Appendix~\ref{app:comparisons} (p.~\pageref{tab:context-memory}) compares representation and update timing. Together the tables distinguish the required information, its storage and access, and the conditions for its reuse.

Memory capacity, access cost, and adaptation speed are separate constraints. Caching keys and values avoids repeatedly encoding the prefix, but attention over an expanding cache still accesses more history. Segment recurrence bounds direct access to earlier states; fast-weight memory summarizes the stream in adaptive parameters; external retrieval keeps a larger archive but must select the relevant record. Compression trades access cost against recoverable detail, while fast-weight adaptation before rollout adds preparation time to a subsequently fixed per-action cost. The useful comparison is how long a decisive event remains recoverable and how quickly new evidence can change a command. Section~\ref{sec:evaluation} connects these questions to context-scaling experiments and feedback-to-action delay.

\paragraph{Exposing changes the controller can execute.}
Memory makes evidence available; the command interface determines whether the inferred change can affect execution. ReSteer exposes the state dependence of this interface: policies that complete two instructions from their usual starting states can fail to switch between them midway \citep{arxiv260317300}. Transition data and policy refinement improve switching. The implication for contextual control is that a new instruction or correction must be actionable from the state reached \mbox{under the previous one.}

Command granularity determines what the context interpreter can express. Steerable Policies lets an off-the-shelf VLM use history to choose subtask descriptions, motion instructions, or pixel targets \citep{chen2026steerable}. With the upper-level model unchanged, the richer interface improves progression over subtask-only steering. A spatial cue can revise a grasp where repeating an object name cannot. This fixed-model branch connects contextual inference to a responsive executor; Section~\ref{sec:recovery} examines which correction the failure evidence justifies.

The path from evidence to correction has three dependencies: the relevant event must remain available, the command interface must express the required change, and the controller must execute it from the current state. This applies to all four method families. Agentic systems expose an additional level of adaptation by revising the records, programs, or context-construction procedures that support later decisions.

\subsection{External knowledge reuse and revision}
\label{sec:agentic-adaptation}
\begin{table}[!tp]
\centering
\surveytable
\caption{External artifacts used with fixed neural weights. Plans can guide the current task; retained guidance and callable routines can support later tasks. Table~\ref{tab:agentic-mechanisms} compares their update scopes.}
\label{tab:knowledge-roles}
\begin{tabularx}{\linewidth}{@{}>{\raggedright\arraybackslash}p{0.25\linewidth}>{\raggedright\arraybackslash}p{0.32\linewidth}Y@{}}
\toprule
\textbf{Artifact} & \textbf{Execution content} & \textbf{Examples} \\
\midrule
Plans and programs & Procedure; controller & Code as Policies~\citep{liang2022codeaspolicies}; ProgPrompt~\citep{singh2022progprompt}; Show-Harness~\citep{chen2026showharness} \\
\addlinespace[2pt]
Reusable guidance & Corrections; constraints; patterns & DROC~\citep{zha2023droc}; ASPIRE~\citep{lu2026aspire}; ARCHITECT~\citep{chen2026architect}; WAA~\citep{zhang2026waa} \\
\addlinespace[2pt]
Executable repertoire & Callable skills; composition & LRLL~\citep{tziafas2024lrll}; RATs~\citep{arxiv260619419}; RoboRSI~\citep{noematrix2026roborsi} \\
\addlinespace[2pt]
Execution oversight & Context; verification; intervention & SHAPER~\citep{wang2026shaper}; Zetta~\citep{arxiv260816590} \\
\bottomrule
\end{tabularx}
\par\phantomsection\label{floatend:tab:knowledge-roles}
\end{table}

External knowledge reuse retains a lesson or executable artifact beyond the attempt that produced it. Three forms have different consequences: guidance changes later inference, callable routines enlarge the available repertoire, and revised context or verification code changes how the system uses evidence. Table~\ref{tab:agentic-mechanisms} in Appendix~\ref{app:comparisons} records these update objects. The analysis first separates acquisition from reuse, then compares the three forms before considering consolidation into neural parameters. This progression connects immediate contextual adaptation to reuse and then to improvements in subsequent learning.

To separate within-attempt inference from persistent adaptation, let $B_n$ collect the external artifacts retained before attempt $n$: memory records, skill code, controller code and its numerical settings, or context-construction code. Let $\mathcal{T}_n$ be the executed trace and feedback from that attempt. A proposal-and-validation \mbox{loop has the form}
\begin{equation}
 \begin{aligned}
 \widetilde B_{n+1}&\sim F_\theta(\cdot\mid B_n,\mathcal{T}_n),\\
 B_{n+1}&=
 \begin{cases}
 \widetilde B_{n+1},&\chi_n=1,\\
 B_n,&\chi_n=0,
 \end{cases}
 \qquad \theta=\theta_0.
 \end{aligned}
 \label{eq:agentic-update}
\end{equation}
The fixed model $F_\theta$ generates a proposed artifact collection $\widetilde B_{n+1}$; the tilde marks a candidate awaiting validation. The acceptance indicator $\chi_n\in\{0,1\}$ is one for acceptance and zero for rejection under the system's validation procedure, potentially including additional rollouts. Writing the effective policy as $\pi_\theta(A_t\mid h_t,C_t;B_n)$ makes this external dependence explicit. The next attempt uses $B_{n+1}$ to construct \mbox{context or execute commands.} 

In skill- and agent-based execution, accepted artifacts can select an executor $\kappa_{B_n}$ or determine the program $g$ and its numerical arguments. These generated code and settings belong to the external artifacts; learned neural parameters of any executor remain part of $\theta$. Systems without a separate acceptance gate use $\chi_n=1$. The attempt index $n$ differs from the decision index $t$: one artifact revision can involve several model calls and robot transitions. Section~\ref{sec:agentic-evaluation} accounts for the trials needed to acquire and validate the \mbox{artifact before measuring reuse.}

\surveyanchor{discuss:tab:knowledge-roles}
Table~\ref{tab:knowledge-roles} groups external artifacts by their role in execution; Table~\ref{tab:agentic-mechanisms} (p.~\pageref{tab:agentic-mechanisms}) further separates plans from executable programs. A \emph{procedural plan} specifies an explicit step sequence or outline. \emph{Reusable guidance} stores acquired rules, applicability conditions, or annotated lessons. An \emph{executable repertoire} contains separately addressable skills, routines, or motion primitives; a \emph{program or its settings} specifies the task controller. \emph{Execution oversight} uses code that constructs model inputs or implements a separate verifier, monitor, or intervention mechanism. Checks embedded in the task program belong to that program. These objects can coexist: a stored lesson guides subsequent synthesis, while a callable function directly \mbox{extends the execution vocabulary.}

Two uses of the retained artifacts explain how this update reaches control:
\begin{equation}
 \begin{aligned}
 \text{Guidance reuse:}\quad
 C_t&=b_\theta(C_t^{\mathrm{task}},m_t,R_t;B_n),
 & A_t&\sim\pi_\theta(\cdot\mid h_t,C_t),\\
 \text{Executable reuse:}\quad
 g&=f_\theta(C_t,h_t;B_n),
 & A_t&=\kappa_{B_n}(g,o_t).
 \end{aligned}
 \label{eq:artifact-use}
\end{equation}
The context builder $b_\theta$ and memory inputs are those of Equation~\eqref{eq:context-state}; the semicolon makes dependence on the retained artifacts $B_n$ explicit. Guidance reuse changes the evidence supplied to otherwise unchanged control. Executable reuse makes the retained library or controller implementation available to specification generation and execution; $\kappa_{B_n}$ denotes that artifact-dependent executor. Both may occur in one system. Together, Equations~\eqref{eq:agentic-update} and~\eqref{eq:artifact-use} connect an artifact's acquisition to its later use under fixed neural parameters $\theta$.

\paragraph{Grounding an interface and learning through it.}
Show-Harness makes the distinction concrete. Its VLM selects semantic motion commands, while a body-specific interpreter supplies calibrated execution. Video can specify an otherwise missing task order. Yet the action-representation experiment shows that reading explicit command conventions is substantially easier than discovering arbitrary command meanings through probing \citep{chen2026showharness}. The former supplies a correspondence; the latter requires inferring it from action effects. The interpreter supplies the low-level control in both conditions.

High-level commands leave grasp synthesis and feedback control to the executor; lower-level commands require the model to resolve more geometry and timing. Guava compares this allocation through workflows and action abstractions~\citep{liu2026guava}, while compositions of frozen motor capabilities, planning, and verification expose its system-level effect~\citep{galanti2026physicalagency}. This allocation determines how existing competence is used; context interventions identify the additional task information supplied by examples or interaction.

SkipVLA makes this allocation explicit within a hybrid policy: a classical planner handles free-space transit, while the instruction-conditioned VLA handles contact-rich motion~\citep{arxiv260920648}. Its learned target predictor connects the two. The routing design improves execution efficiency by assigning transit and contact to interfaces suited to their respective demands.

The first extension is from open-loop synthesis to outcome-conditioned revision. Textual feedback can revise a plan~\citep{huang2022monologue}; intermediate observations can revise the plan and Python controller separately~\citep{singh2024malmm}; program checks can make some responses executable without another model call~\citep{singh2022progprompt}. Constraint violations narrow revisions to code and sampling bounds~\citep{curtis2025proc3s}, while primitive-level errors narrow them to control settings~\citep{merwe2025icpi}. These interfaces determine the available edit space. Section~\ref{sec:recovery} addresses the separate question of which edit the observed failure justifies.

\paragraph{Retaining guidance and executable routines.}
Reuse asks which part of a successful repair remains valid later. Constraints attached to plan or code revisions preserve its scope~\citep{zha2023droc}; causal annotations explain its effect~\citep{sarch2024ical}; distilled rules and code patterns support later synthesis~\citep{chen2026architect}. Attribution determines the appropriate change: failure to follow valid guidance calls for execution repair, while a defective skill may require revising guidance or code~\citep{ju2026embodiskill}. Retaining the cause helps avoid inappropriate reuse.

Visual evidence makes retained guidance more explicit. WAA turns expert demonstrations and human corrections into procedures with applicability conditions, reference images, and outcome checks~\citep{zhang2026waa}. In its one-demonstration stove study, review replaces knob motion as the success criterion with an observed activation signal. Fixed agents then interpret the acquired guidance.

Executable reuse makes recurring behavior callable. Function synthesis supplies missing operations~\citep{liang2022codeaspolicies}; program clustering and human feedback refine reusable candidates~\citep{tziafas2024lrll,arxiv250918597}. Video-grounded retrieval supplies implementation details such as contacts and target poses~\citep{xie2026uniskillrepo}, while repeated successful code can enter a library~\citep{arxiv260322435}. Transfer then depends on scene-specific arguments and valid execution preconditions.

Autonomous trials can supply the evidence previously provided by a teacher's correction, but the retained object determines what later tasks gain. Rewriting a controller improves the current solution~\citep{kumar2026aor}; task-agnostic play can accumulate code skills and failure records for later use~\citep{arxiv260619419}. Distilling a validated repair into a failure signature, applicability condition, and strategy instead retains guidance for constructing a future solution~\citep{lu2026aspire}. These routes share outcome-driven acquisition while transferring different knowledge: a specific implementation, a callable operation, or a rule for choosing a repair.

A structured library also supports failure attribution. RoboRSI organizes diagnosis and revision through a task--skill tree, retaining successful workflows as callable code~\citep{noematrix2026roborsi}. Its code-on/code-off comparison reports 174/600 versus 129/600 successes over 120 tasks and five layouts each. This tests retained routines; its separate policy-training case consolidates corrective trajectories into neural parameters. The retained object distinguishes these two forms of reuse.

Separating semantic effects from geometric realization supports reuse across scenes. Teach and Grow's Skill Blocks specify effects, scope, executors, verification, and recovery~\citep{arxiv260817209}. Its simulation pilot evaluates local library additions with deterministic decomposition and a fixed executor; agent-based induction and Experience Memory updates remain proposed. The evaluated contribution expands a repertoire, while the proposed extension would learn how to build it.

Reuse can encompass a complete closed-loop policy. Programs refined through demonstrations and simulation feedback become examples for new policy generation~\citep{arxiv260919906}. Generalized TAMP experiments similarly separate simulator-assisted code synthesis from evaluation of frozen programs on unseen instances~\citep{arxiv260930233}. RIVET makes the interface explicit: generated perception and planning programs share object poses and a relation graph, then operate without runtime VLM regeneration~\citep{arxiv260931337}. HuGo instead refines high-level policy code from rollout trajectories and video while preserving a frozen whole-body controller~\citep{arxiv260930594}. Together, these designs locate adaptation in program construction and revision; subsequent execution tests whether the acquired logic transfers beyond its synthesis trials.

\paragraph{Adapting how evidence reaches control.}
Retained guidance improves how available routines are invoked. Parameterized call sequences preserve useful orderings, while global heuristics describe the operating conditions and failures of fixed primitives~\citep{arxiv260708448}. Hi-VLA's tabletop study likewise finds greater benefits from prior-episode affordance summaries than longer within-episode history~\citep{hu2026orchestrating}. Such memory captures conditions of use, connecting agent reasoning to system identification of available capabilities.

Repeated failures in evidence selection or verification motivate adapting the surrounding inference procedure. Revising history-selection code changes what the planner can see; revising a critic or recovery tool changes how execution is checked and corrected. SHAPER targets procedural guidance and context construction with the planner and executor fixed~\citep{wang2026shaper}; Zetta targets critics, recovery tools, and governance around a frozen policy~\citep{arxiv260816590}. RACaP jointly evolves Policy APIs, the ReAct harness, and experience memory before freezing the resulting system for deployment~\citep{li2026racap}. This separates two timescales: external-artifact revision improves the available interface, whereas runtime context selects how the current task uses it.

Compiling a repair also changes when reasoning is paid for. HarnessPAI retains a task program during each rollout, including observation-conditioned checks, and revises it between rollouts using stage-indexed failure-to-repair memory~\citep{arxiv260929166}. Its program-level open loop therefore still permits sensor feedback during execution. This amortizes synthesis across repeated tasks, while a changed instruction or operating condition can require renewed validation. Its skill-library comparison shows faster program convergence on three representative development tasks. Held-out acquisition tests would measure how this benefit transfers to subsequent learning. In its LIBERO evaluations, the reported 50 seeds include 15 used for program evolution, a distinction relevant to interpreting transfer.

\paragraph{Connecting contextual acquisition to neural consolidation.}
Neural consolidation turns acquired experience into future model competence. Distilling harness interaction traces~\citep{liu2026guava,zhang2026waa} or combining orchestration with autonomous data collection and policy refinement~\citep{cui2026roboclaw} transfers work initially performed through inference into trained parameters. EmbodiedSWE diversifies verified agent solutions into demonstrations for VLA training~\citep{you2026embodiedswe}, supplying variations beyond the solved task instances. The training target determines what is consolidated: successful trajectories teach execution, whereas successful teaching exchanges, as in LMPC, teach responsiveness to later feedback~\citep{liang2024lmpc}. Section~\ref{sec:icl-self-improvement} develops this distinction. The next chapter examines how training data connect task evidence to execution.

\begin{takeaway}
\takepoint Context determines an action distribution, a geometric reference, predicted consequences, or an execution specification. Equations~\eqref{eq:direct-interface}, \eqref{eq:retarget-grounding}, \eqref{eq:wam}, and~\eqref{eq:structured} identify these four interfaces.
\takepoint Context interpretation and motor competence limit transfer separately. Retrieval and memory support transfer by supplying usable evidence to a controller that responds to later corrections.
\takepoint With neural weights fixed, context can revise task outlines, executable routines, and programs. Retaining these artifacts supports reuse, and successor comparisons measure their contribution to learning subsequent tasks.
\end{takeaway}

\FloatBarrier
\suppressfloats[t]
\section{Data and Training: How Is Context Use Learned?}
\label{sec:data}
\begin{figure}[tp]
\centering
\includegraphics[width=\linewidth]{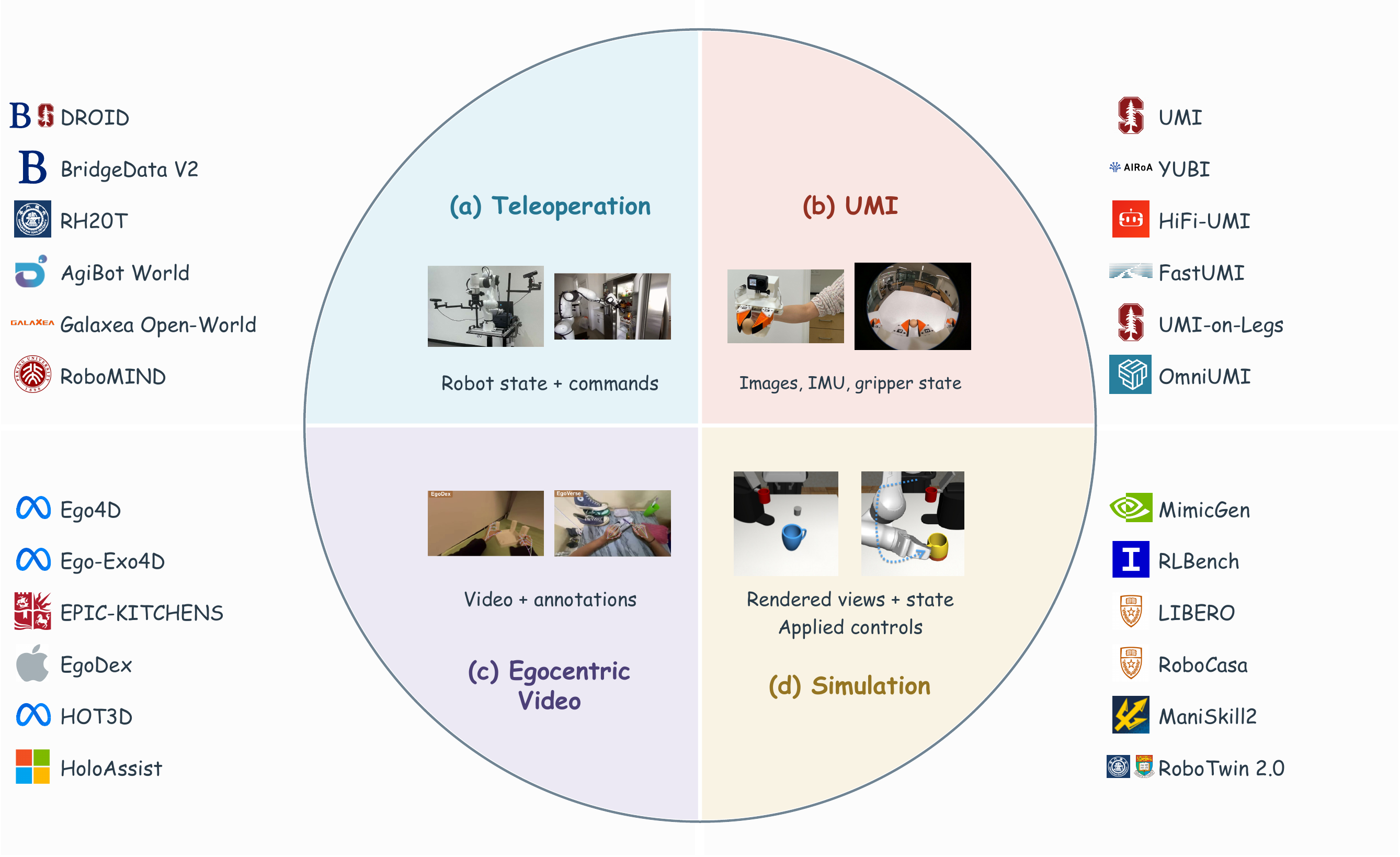}
\caption{Four sources of contextual evidence for robots: (a) teleoperation, (b) UMI, (c) egocentric video, and (d) simulation. Representative resources include DROID~\citep{khazatsky2024droid}, UMI~\citep{chi2024umi}, Ego4D~\citep{grauman2021ego4d}, and MimicGen~\citep{mandlekar2023mimicgen}, respectively. Their recordings differ in visual, state, action, and contact channels; language can accompany each as \mbox{instructions, examples, or feedback.}}
\label{fig:input-data-taxonomy}
\par\phantomsection\label{floatend:fig:input-data-taxonomy}
\end{figure}

An input channel makes context accessible; training determines whether the policy uses it. When the current scene already identifies every target action, a model can fit demonstrations without learning to consult earlier teaching. The central data question is therefore which relationships make earlier evidence informative about a later decision. Acquisition sources determine what is measured, training units establish these relationships, and construction at scale supplies their missing variations. This connects the transfer assumptions of Section~\ref{sec:methods} to the experience needed to support them.

Source taxonomies, such as the Data Pyramid survey~\citep{arxiv260724744}, describe where embodied data originate and how closely they match robot execution. ICL for robots additionally asks what earlier teaching or interaction reveals about a later decision. A record\textquotesingle{}s source, channels, and role in learning are therefore separate dimensions. Explicit pairs, continuous streams, and linked attempts organize these dimensions differently; pretrained interpretation and geometric construction can also connect evidence to execution without jointly \mbox{training the entire path.}

\subsection{Input sources and context modalities}
\label{sec:input-data}

An acquisition source determines what is measured directly and what must be inferred. Teleoperation records robot commands, UMI estimates motion through a robot-like device, egocentric video records human activity, and simulation provides modeled transitions. The four sources are compared on this common basis.

Language can accompany every acquisition source. Instructions specify constraints, command--program examples establish conventions, corrections revise decisions, and summaries retain earlier events. Code as Policies, DROC, and MEM illustrate the latter three roles~\citep{liang2022codeaspolicies,zha2023droc,torne2026mem}. Records should preserve each statement's source, time, and referent: a teacher's requested change and a model's interpretation of an \mbox{event carry different evidence.}

\surveyanchor{discuss:fig:input-data-taxonomy}
Figure~\ref{fig:input-data-taxonomy} contrasts the directly observed interaction: robot execution in teleoperation, instrumented hand motion in UMI, human activity in egocentric video, and modeled execution in simulation. These sources leave different correspondence problems for the learner, as the following parallel comparisons explain.

\paragraph{Teleoperation: measurements and commanded interventions.}
A real robot provides synchronized scene observations, measured robot state, and issued commands. DROID and BridgeData V2 supply such robot demonstrations~\citep{khazatsky2024droid,walke2023bridge}; RH20T additionally records force--torque measurements and, in one hardware configuration, fingertip tactile signals~\citep{fang2023rh20t}. These records can connect an attempted intervention to the body's response. A gripper-close command specifies an action, while measured aperture and contact help establish whether the object was grasped. Joint torque or motor current also requires interpretation through the robot's mechanics before it can describe external contact.

\begin{table}[!tp]
\centering
\surveytable
\caption{Acquisition interfaces compared by state, actions, and contact evidence. UMI pose and width are estimated; extra sensing and simulated contact records depend on the setup. Table~\ref{tab:data} lists resources.}
\label{tab:data-families}
\begin{tabularx}{\linewidth}{@{}>{\raggedright\arraybackslash}p{0.18\linewidth}>{\raggedright\arraybackslash}p{0.26\linewidth}>{\raggedright\arraybackslash}p{0.27\linewidth}Y@{}}
\toprule
\textbf{Source} & \textbf{State} & \textbf{Actions} & \textbf{Contact} \\
\midrule
Teleoperation & Images; joints; gripper & Recorded commands & Force; tactile (optional) \\
\addlinespace[2pt]
UMI & Images; IMU; gripper pose & Mapped gripper targets & Visual; force (optional) \\
\addlinespace[2pt]
Egocentric video & Video; optional sensing & Inferred or paired actions & Visual interaction \\
\addlinespace[2pt]
Simulation & Rendered views; state & Simulator controls & Modeled contact \\
\bottomrule
\end{tabularx}
\par\phantomsection\label{floatend:tab:data-families}
\end{table}

\paragraph{UMI: tracked motion without the robot arm.}
UMI reduces collection dependence on a robot arm by preserving a gripper-like interface in the demonstrator's hand. Images and inertial measurements support device-motion estimation, while visual tracking estimates gripper width~\citep{chi2024umi}. The resulting trajectory can supply end-effector targets without measuring arm-joint motion. Sharing the end effector between collection and execution reduces a further source of correspondence error~\citep{ohkawa2026yubi}; synchronized tracking improves the fidelity of the recovered motion~\citep{simpleai2026hifiumi}. Extensions address acquisition, whole-body realization, and richer interaction sensing~\citep{zhaxizhuoma2024fastumi,ha2024umilegs,luo2026omniumi}. Their common trade-off is deliberate: an instrumented interface restricts how a human demonstrates, but reduces the inference needed to turn that \mbox{demonstration into robot-compatible evidence.}

\paragraph{Egocentric video: visual procedure with optional human sensing.}
Egocentric recordings expose object interactions and their order without requiring a robot-compatible device. Ego4D and EPIC-KITCHENS-100 provide egocentric activity, and Ego-Exo4D adds synchronized external views~\citep{grauman2021ego4d,damen2020epic100,grauman2023egoexo4d}. Some collections also provide audio, gaze, inertial measurements, depth, or tracked hands. They measure human motion and the surrounding environment. Robot action targets must come from paired execution, reconstruction and retargeting, or a controller that grounds the observed procedure. An action-free video supplies visual observations without recorded robot commands.

\paragraph{Simulation: rendered observations and modeled responses.}
Simulation makes action consequences and privileged state available under a specified dynamics model. Demonstration environments such as RLBench, LIBERO, and ManiSkill2 support task execution with synchronized observations and controls~\citep{james2019rlbench,liu2023libero,gu2023maniskill2}; trajectory or task-program construction produces additional executions~\citep{mandlekar2023mimicgen,chen2025robotwin2}. This control over the generating process is useful for contextual learning because a procedure can be held fixed while scene geometry changes, or varied while the scene remains comparable. Privileged object and contact states can verify those relationships during construction, even when the deployed policy receives cameras and proprioception. Section~\ref{sec:data-construction} examines how synthesis preserves the assigned teaching condition.

\paragraph{Channels and policy inputs.}

\FloatBarrier
\surveyanchor{discuss:tab:data-families}
Table~\ref{tab:data-families} separates measured state, action information, and contact evidence. A channel used during collection need not be a direct policy input: UMI uses IMU measurements for pose estimation, while its policy observes images, end-effector pose, and gripper width~\citep{chi2024umi}. For ICL, the available fields must be specified independently for the support demonstration, the current query, and the supervision target. A human support may contain images alone while the executing robot queries its joint state and contact sensors. Conversely, recorded support commands can enter context, whereas future query commands serve as training targets. In the notation of Section~\ref{sec:foundations}, available sensory measurements enter $o_t$, issued commands are $a_t$, and their time-ordered records form $h_t$; the selected evidence enters $C_t$. Textual instructions, examples, and feedback can accompany these records and enter $C_t$ through \mbox{the model's context interface.}

\begin{table}[!tp]
\centering
\surveytable
\caption{Context units, retained evidence, and adaptation roles. Support actions may enter context; future query actions are training targets.}
\label{tab:context-organization}
\begin{tabularx}{\linewidth}{@{}>{\raggedright\arraybackslash}p{.24\linewidth}>{\raggedright\arraybackslash}p{.42\linewidth}Y@{}}
\toprule
\textbf{Context unit} & \textbf{Evidence} & \textbf{Role}\\
\midrule
Support example & Demonstrations; paired examples & Task; procedure; contact\\
\addlinespace[3pt]
Scene preview & Exploration views; spatial records & Layout; locations; routes\\
\addlinespace[3pt]
Query history & Observations; actions; dialogue & Progress; recent response\\
\addlinespace[3pt]
Transition sequence & Observation--action--outcome & Physical response\\
\addlinespace[3pt]
Cross-attempt record & Episodes; summaries; corrections & Reuse across resets\\
\bottomrule
\end{tabularx}
\par\phantomsection\label{floatend:tab:context-organization}
\end{table}

\subsection{Training units and context dependence}
\label{sec:context-training}

\surveyanchor{discuss:tab:training-organization}
Channels become useful training evidence through temporal and task relationships. Table~\ref{tab:training-organization} compares paired episodes, continuous streams, and linked attempts. We define the temporal units first, then use support--query supervision to distinguish input channels, action targets, and pairing criteria. Continuous experience and corrective records extend this dependency beyond explicit pairs.

\begin{table}[!tp]
\centering\surveytable
\caption{Three training organizations compared by evidence, supervision, and record relationship. Support can omit robot actions; linked attempts can use predictive, corrective, or reward-based supervision.}
\label{tab:training-organization}
\begin{tabularx}{\linewidth}{@{}>{\raggedright\arraybackslash}p{.21\linewidth}YYY@{}}
\toprule
\textbf{Organization} & \textbf{Earlier evidence} & \textbf{Learning signal} & \textbf{Link between records}\\
\midrule
Paired episodes & Support demonstration & Query actions & Shared task or environment\\
\addlinespace[3pt]
Continuous stream & Observation/action history & Subsequent actions & Temporal continuation\\
\addlinespace[3pt]
Linked attempts & Trials; outcomes; corrections & Selected actions or feedback & Consequences across resets\\
\bottomrule
\end{tabularx}
\par\phantomsection\label{floatend:tab:training-organization}
\end{table}

\paragraph{Temporal units of contextual evidence.}

A contextual training unit must identify what was observed, what was commanded, and what the resulting feedback established. Language can specify the intended task or annotate the outcome. The sequence must preserve when each item became available, so that a later consequence can supervise learning without appearing as evidence for an earlier decision.

Temporal alignment preserves the relation between an intervention and its consequence through sensor timestamps, command application times, coordinate conventions, and episode boundaries. Measured motor state must remain distinguishable from its command or setpoint. Aligned windows or timestamp-aware encoding accommodate different sampling rates while retaining brief contact events. Missing channels and reconstructed motion must also remain identifiable, because they supply different \mbox{evidence from direct measurements.}

\surveyanchor{discuss:tab:context-organization}
Table~\ref{tab:context-organization} groups records by what they contribute to learning. Support can specify behavior or reveal an environment; query history locates execution; transitions reveal action consequences. These units overlap: history contains transitions, and retrieved segments can come from demonstrations or earlier attempts. Their role is independent of modality. Images or synchronized robot records can supply a demonstration, while state--action history can support physical adaptation without one, as in RMA~\citep{kumar2021rma}.

\begin{figure}[t]
\centering
\includegraphics[width=\linewidth]{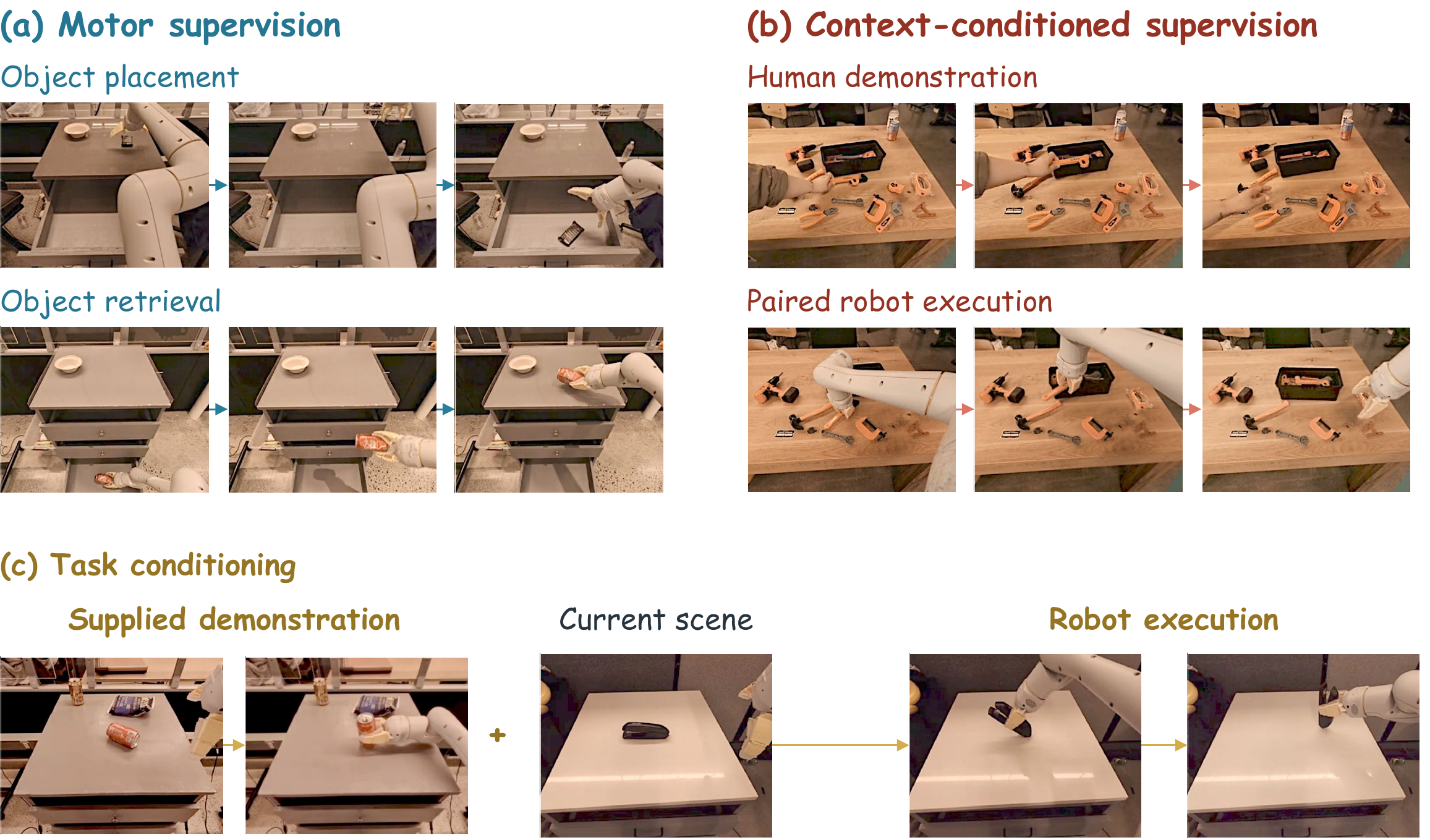}
\caption{Demonstrations serve three roles: (a) supervise motor actions, (b) train context use through paired episodes, and (c) specify a new task to a fixed policy. Panel (c) illustrates a can-uprighting demonstration specifying stapler reorientation; it is a conceptual example of cross-object transfer.}
\label{fig:data-roles}
\par\phantomsection\label{floatend:fig:data-roles}
\end{figure}

\paragraph{Supervising a query from a support episode.}
Following the paired-execution principle of one-shot imitation~\citep{duan2017oneshot}, let support demonstration $D^{\mathrm{s}}$ and query observation--action trajectory $\tau^{\mathrm{q}}$ share task specification $\psi$. The matched executions may differ in objects, timing, or embodiment. A common supervised objective predicts query actions from their history and support:
\begin{equation}
 \mathcal{L}(\theta)=
 \mathbb{E}_{(D^{\mathrm{s}},\tau^{\mathrm{q}})\sim P_{\mathrm{pair}}}
 \left[\frac{1}{|I^{\mathrm{q}}|}\sum_{t\in I^{\mathrm{q}}}
 \ell_\theta\!\left(A_t^{\mathrm{q}};h_t^{\mathrm{q}},D^{\mathrm{s}}\right)\right].
 \label{eq:training}
\end{equation}

Here $P_{\mathrm{pair}}$ samples executions matched by specification $\psi$. Query history $h_t^{\mathrm{q}}$ ends at the current observation, while $A_t^{\mathrm{q}}=a_{t:t+H-1}^{\mathrm{q}}$ is the recorded target block. The nonempty set $I^{\mathrm{q}}$ contains positions with complete targets; averaging over it normalizes each episode's contribution. The loss $\ell_\theta$ may use likelihood, regression, denoising, or flow matching. These objectives differ in action modeling but share the same supervisory dependency: the support must explain which query behavior is appropriate.

Equation~\eqref{eq:training} specializes the general context interface to $C_t=D^{\mathrm{s}}$. In instruction-bearing episodes, the same query-action loss conditions on $C_t$ containing both the demonstration and its associated text. Instructions specify goals and constraints; demonstrations can convey remaining choices of order or contact. A correction enters context when received, preserving what was available before each supervised action.

\surveyanchor{discuss:fig:data-roles}
Figure~\ref{fig:data-roles} distinguishes motor supervision, training to use teaching, and deployment-time task specification. Uprighting a differently shaped object illustrates the third role: the operation must survive changed appearance. Paired supervision learns this dependency explicitly; continuous streams can expose it through \mbox{the sequence of experience.}

\paragraph{Support channels and action supervision.}
Equation~\eqref{eq:training} supervises query actions while leaving support channels open. Vid2Robot conditions on video without recorded support commands~\citep{jain2024vid2robot}; ICRT supplies support actions in a sensorimotor prefix~\citep{fu2024icrt}. Both supervise the query execution. Support content, the action loss, and the architecture that combines evidence are therefore separate choices.

\paragraph{Task identity and support--query pairing.}
The shared specification $\psi$ determines the granularity of teaching. Pairing all executions labeled ``put the object away'' can preserve a destination while discarding grasp or route preferences. BPP makes this dependence explicit in its task-grouped demonstrations \citep{patel2026bpp}. The pairing rule must retain the distinction the deployment example is expected to convey. For cross-object transfer (Section~\ref{sec:context-mechanisms}), support and query can change the manipulated object, the receiver, or both while preserving the same requirement. Conversely, retaining comparable objects while varying the required relation makes the support informative. Together, these pairings separate task meaning from object identity and provide examples of which substitutions preserve the intended function.

Pairing requires both semantic agreement and compatible action meaning. Additional demonstrations can distinguish task interpretations left ambiguous by one example, as Instant Policy illustrates with differently shaped boxes~\citep{vosylius2024instantpolicy}. The resulting query target must still express the same intervention in consistent coordinates, timing, and gripper semantics. WALL-OSS-0.5 addresses this physical side through synchronized video and actions, camera repair, and end-effector calibration~\citep{yu2026walloss05}. These requirements are complementary: semantic pairing determines which behavior is intended, while calibration determines what command realizes it. With shared action coordinates, each body retains its own reachability and contact mechanics.

A shared environment supplies a different pairing relation from a shared procedure. Route-following training pairs a demonstrated path with a perturbed execution, as in RPF~\citep{kumar2018rpf}. Scene-conditioned training instead pairs a preview with navigation to goals within that environment, as in NOLO~\citep{zhou2024nolo}; ReLIC~\citep{elawady2024relic} links episodes in one home while changing the target and start. The shared variable is therefore the route in the first case and the environment in the latter two. Scene identity and memory-reset boundaries determine which spatial knowledge remains available across examples or episodes.

\paragraph{From paired episodes to continuous experience.}
Explicit pairing makes the teaching--execution relationship part of the training unit. S1 supplies an in-context demonstration in its pretraining episodes~\citep{skild2026s1}. GEN-1.5 instead reports next-action pretraining on randomly sampled continuous physical spans, without dedicated demonstration packing or an ICL-specific objective, and inserts sensorimotor examples as prompts at deployment~\citep{generalist2026gen15}. A stream exposes earlier observations and actions without assigning them separate support and query roles. Both systems later accept task examples, acquired through different organizations of training experience. Action generation is a further choice: ContextFlow uses explicit robot support--query supervision with continuous flow matching~\citep{ding2026contextflow}. The action loss and the organization of experience play different roles: one fits the action model, while the other links preceding evidence to subsequent behavior.

\paragraph{Behavioral diversity and contextual identifiability.}
\begin{figure}[!tp]
\centering
\includegraphics[width=0.85\linewidth]{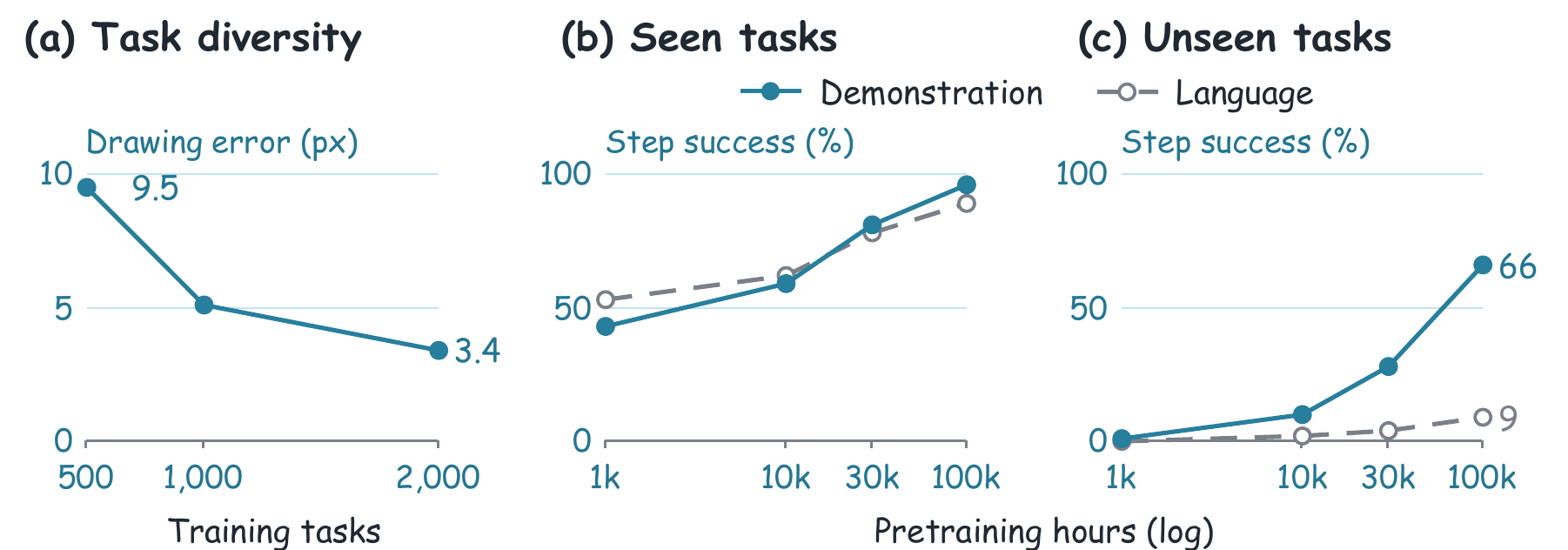}
\caption{Task diversity and prompting affect transfer differently. (a) BPP drawing error with 10,000 demonstrations, averaged over three seeds~\citep{patel2026bpp}. (b,c) S1 compares demonstration and language prompting on seen and unseen tasks; its cumulative per-step success measure includes human recovery interventions~\citep{skild2026s1}.}
\label{fig:data-scaling}
\par\phantomsection\label{floatend:fig:data-scaling}
\end{figure}

Access to earlier evidence becomes useful when it resolves uncertainty about the continuation. If the current scene already determines every training action, the loss can be minimized without reading a demonstration. Comparable scenes with different valid procedures instead give teaching predictive value; repetitions of one procedure across scenes teach which variations leave it unchanged. The practical consequence is to count distinguishable teaching problems as well as trajectories. The comparisons below examine this dependency, while Section~\ref{sec:reusable-learning-rule} asks how far \mbox{the resulting inference transfers.}

Context can also support shortcuts. Correlations with previous actions can improve imitation fit while impairing deployed control \citep{dehaan2019causal}; a prefix may identify collection habits rather than task requirements. Varying demonstrators, procedures, and physical conditions independently helps separate these explanations. Holding out task relations, alongside trajectories, tests transfer beyond familiar procedures. Multiple valid procedures in comparable scenes support both contextual training and tests in which the supplied example changes \mbox{the correct action (Section~\ref{sec:evaluation}).}

\surveyanchor{discuss:fig:data-scaling}
The evidence distinguishes behavioral diversity from visual variety. ICRT's selected DROID-only training condition fails its test tasks, whereas its multi-task scenes require different actions from similar initial observations \citep{fu2024icrt}. BPP supplies a controlled allocation comparison: with 10,000 drawing demonstrations, mean Chamfer error is 9.5 pixels for 500 tasks with 20 demonstrations each, 5.1 for 1,000 tasks with ten each, and 3.4 for 2,000 tasks with five each (Figure~\ref{fig:data-scaling}a). Its less diverse folding setting can instead favor language conditioning \citep{patel2026bpp}. Demo-JEPA separately varies task classes and trajectories per task \citep{he2026demojepa}. Across these studies, useful diversity adds decisions that an example must resolve, supported by enough executions to learn the relationship.

S1 separates scaling gains on seen and unseen tasks under matched data, compute, and architecture apart from the prompt embedding \citep{skild2026s1}. Across 1k--100k training hours, both prompting methods improve on seen tasks. On unseen tasks, demonstration prompting rises from 1\% to 66\%, while language prompting rises from 0\% to 9\% (Figure~\ref{fig:data-scaling}b,c). Thus similar seen-task performance can conceal different returns to new-task teaching. The cumulative per-step metric includes human recovery, mainly for the language baseline. The S1 comparison concerns combined data scaling and prompting, while BPP directly isolates task diversity at a fixed demonstration budget. Matching the procedural information conveyed by text and video would further isolate modality from teaching detail (Section~\ref{sec:context-transfer-evaluation}).

\paragraph{Linked attempts and corrective supervision.}
Learning from a failed action or a diagnostic probe extends context dependence across attempts. The training unit must link what was tried to what happened next, preserving reset boundaries and any change in the task or physical condition. Algorithm Distillation models such learning histories \citep{laskin2022ad}; LocoFormer retains experience over varied morphologies and dynamics \citep{liu2025locoformer}. Reward-based meta-training can value an informative action through its benefit to later attempts \citep{duan2016rl2}. A short probe gains learning value when its outcome informs the next decision. Linked attempts expose \mbox{this dependency during training.}

In recovery-conditioned supervision, the failed attempt and its feedback enter context while a later continuation supplies corrective targets. RoboTTT's DAgger Distillation uses the full rollout as context but applies action loss only to human corrections, also improving its GDN comparator \citep{jiang2026robottt}. The failed actions record what was tried; the selected corrections specify what to imitate. This training organization can \mbox{support different memory mechanisms.}

Equation~\eqref{eq:training} extends to this case by replacing support-only conditioning with $C_t$. For block-level losses, $I^{\mathrm{q}}$ includes only complete blocks composed entirely of actions selected for imitation. An objective supporting action-wise masking can instead suppress unselected positions within a block. Other executed actions remain in history. The context mask controls available evidence, whereas the target mask controls desired behavior. Runtime context contains only evidence available before the decision; future outcomes may supply supervision. Attempt boundaries must separately record physical resets, memory resets, and changes of tool or dynamics.

Long-context control requires representations that preserve decisive events. Past-Token Prediction adds reconstruction of earlier actions to future-action prediction, first learning short-context visual features and then caching them for longer-context training \citep{torne2025ptp}. Caching reduces repeated computation; the cached visual features determine which task details remain available to the controller.

The supervisory unit determines what evidence can explain a later action: pairs link teaching to realization, streams link earlier events to current control, and linked attempts connect outcomes to revised decisions. Their common requirement is a decision that earlier evidence helps disambiguate. The remaining analysis examines acquisition sources that preserve this dependency with usable action correspondence.

\subsection{Data coverage and action correspondence}
\label{sec:data-coverage}

Data coverage has two distinct meanings: the range of behavior recorded and the range of teaching--execution relationships recoverable from it. Robot records, motion-bearing human demonstrations, and ordinary video trade direct action correspondence against collection breadth. The comparison below follows that trade-off, asking which annotations make each source usable for contextual training. Simulation supplies controlled counterparts in Section~\ref{sec:data-construction}; Table~\ref{tab:data} lists resource-level details.

\paragraph{From reusable experience to annotated procedures.}
Motor coverage supplies reusable behavior before teaching can specialize it. Autonomous multi-task experience and shared demonstration collections addressed this need~\citep{mtopt2021,bridge2021}; Bridge Data included 7,200 demonstrations across 71 tasks and ten environments. Later collections broadened scenes and embodiments~\citep{walke2023bridge,oxe2023,khazatsky2024droid}. Contextual training additionally needs relationships between these executions that preserve what a new example should teach.

Procedure annotations identify why a continuation changes. AgiBot World's March 2025 release spans 1,001,552 trajectories, 2,976.4 hours, 217 tasks, and 106 scenes, including subtask and recovery records~\citep{agibot2025}. Task intervals, step instructions, and later corrections connect demonstrations, autonomous attempts, and human interventions~\citep{agibot2026release,agibot2026corrections}. Linking an attempt to its correction preserves the evidence that makes \mbox{a different continuation appropriate.}

Broader embodiment coverage makes shared task meaning and shared action meaning increasingly separate problems. Galaxea's common platform and subtask timing provide a comparatively consistent collection interface~\citep{galaxea2025}; RoboMIND 2.0 and RoboCOIN broaden the range of platforms~\citep{robomind2024,robomind2025v2,robocoin2025}. Procedural metadata identifies which executions teach the same requirement, while kinematic metadata describes how their realization differs. A cross-body pair needs both. This explains why adding embodiments can expand motor coverage while increasing the work needed to construct reliable contextual pairs.

\paragraph{Robot records supply executable supervision.}
Robot records become contextual supervision when a query execution is associated with an example of its intended procedure \citep{oxe2023,khazatsky2024droid,agibot2025,walke2023bridge}. Cross-scene retrieval can locate candidate executions \citep{bharadhwaj2023roboagent}, but matching must distinguish alternatives that begin in similar scenes.

Human--robot pairing exposes an asymmetry between teaching and supervision. A human video can specify the intended interaction, while robot measurements reveal the commands and contact response needed to realize it. RH20T provides corresponding human videos together with robot images, joints, gripper state, force--torque signals, audio, and tactile sensing in one configuration~\citep{fang2023rh20t}. These richer query records can supervise a policy whose eventual human support contains fewer channels. Task-level human--robot organization helps define the transfer setting~\citep{zhou2026imitator}, but a shared label such as \emph{fold} can still merge different procedures. Matching ordered events and terminal outcomes, as in WorldScape's robot and UMI pairs, preserves a finer teaching relation~\citep{arxiv260718840}. Pairing granularity therefore determines whether the model learns an activity category, a procedure, or an interaction detail.

Mixed capture can reduce pairing cost. XR-2 combines 531.7 hours of robot teleoperation with roughly 1,000 hours of dual-UMI demonstrations~\citep{xu2026bimanualscaling}. Robot views, state, and actions supply executable targets; egocentric views and gripper motion supply candidate support. Shared laundry tasks narrow the search, while procedure-level matching identifies compatible executions.

\paragraph{Motion-bearing demonstrations reduce correspondence uncertainty.}
Instrumented teaching trades restrictions on collection for less uncertainty in transfer. Handheld grippers make the demonstrator's observations and motion closer to a robot-compatible interface \citep{chi2024umi}. Preserving the end effector across collection and execution further reduces what must be inferred: YUBI reports 8,434 hours over 119 tasks and transfers the same gripper between arms through per-robot inverse kinematics \citep{ohkawa2026yubi}. Sharing the gripper reduces hand-to-robot correspondence uncertainty, leaving each arm's reachability and control limits to \mbox{be handled during execution.}

Bimanual teaching makes relative motion fidelity more consequential than the visual plausibility of either hand alone. Stereo-inertial tracking, synchronized views, and inter-gripper reconstruction preserve this coordination~\citep{simpleai2026hifiumi}. HiFi-UMI's reported 2K release also includes subtask boundaries, and its policy studies use UMI post-training without target-task teleoperation. Calibration preserves the physical relation between hands; boundaries identify which procedure that relation instantiates. Modular tracking can simplify acquisition~\citep{zhaxizhuoma2024fastumi}, while a simulation-trained whole-body controller can supply realization beyond the demonstrated gripper motion~\citep{ha2024umilegs}. More accessible capture and more accurate coordination are consequently distinct improvements to the teaching interface.

Contact sensing extends what the teaching record can express. OmniUMI synchronizes vision and motion with tactile signals, internal grasping force, and external interaction wrench~\citep{luo2026omniumi}. Such measurements can distinguish a visually similar motion performed with different contact conditions. Their transfer still depends on compatible sensing, calibration, and execution interfaces; a measured wrench and a commanded gripper setting carry different information.

Collection can be conditioned on the current policy's weaknesses. HIL-UMI compares handheld human trajectories with policy predictions on the same observation stream and targets additional data to disagreement and low-advantage segments~\citep{arxiv260920659}. The resulting demonstrations support further policy training. Thus policy-aware collection improves which experience is acquired, while remaining distinct from using that experience as fixed-weight deployment context.

Removing the collection device shifts this burden from measurement to inference. Tracked hands and segmented point clouds can be converted into gripper poses \citep{vosylius2024instantpolicy}; motion inferred from human play can supply pseudo-actions, retrieved context--query relations, and retargeted humanoid supervision \citep{shah2025mimicdroid}. These approaches broaden the usable recording pool, but the action evidence inherits reconstruction and retargeting error. The trade-off is most consequential near contact, where a small positional error can change whether a grasp or insertion is feasible. A common action format enables joint use of these records while their reconstruction reliability determines their supervisory value.

Mixed-source training can preserve this distinction instead of forcing all trajectories to carry the same supervisory weight. Aligning camera-space actions, embodiment, and chunk duration provides a common temporal interface, while reliability-weighted human pseudo-actions supplement a robot-supervised primary objective \citep{li2026aceego}. In ACE-Ego-0, the auxiliary objective emphasizes more dependable position channels. This weighting lets human records expand coverage while measured robot actions anchor execution. Carrying reconstruction confidence into contextual pairs likewise avoids treating uncertain hand tracking as an \mbox{equally reliable robot target.}

Zeva-Ego separates this acquisition of a physical prior from deployment adaptation~\citep{arxiv260924411}. Its Action-Centric Encoder converts egocentric visual transitions into continuous action-token supervision for mid-training. A later interaction-conditioned stage trains the action model to use observed action effects; deployment freezes the weights and updates memory. Human experience expands prior competence, while robot interaction supplies evidence about the current physical conditions.

Motion annotation reduces the uncertainty left by ordinary video while retaining a human-centered collection process. Hand--object geometry~\citep{hoi4d2022,hot3d2025} and contemporaneous hand and finger tracking, as in EgoDex's 829 hours~\citep{egodex2025}, expose motion that otherwise must be reconstructed from RGB. Robot learning still requires conversion into feasible contacts and controls. EgoScale's 20,854 hours of action-labeled human data are connected to robot behavior through aligned mid-training and downstream post-training~\citep{egoscale2026}; its one-demonstration experiment uses policy updates with one robot demonstration per task and additional human trajectories. These stages show why improved action evidence and fixed-weight task acquisition are separate contributions: the former can support either contextual inference or further policy fitting.

The transition from human pretraining to robot deployment also changes the meaning of language supervision. Xiaomi-Robotics-1 combines over 100K hours of UMI and egocentric records with cross-embodiment post-training~\citep{xiaomi2026robotics1}. State-transition descriptions during pretraining and imperative instructions during post-training serve different prediction targets. For demonstration-conditioned learning, the corresponding design choice is which task details remain unspecified by the instruction and must be recovered from the example. This connects source breadth to the pairing criterion in Section~\ref{sec:context-training}.

\paragraph{Ordinary video broadens procedural coverage.}
Ordinary video reverses the trade-off of instrumented capture: activity coverage is broad, while robot-compatible action information is sparse. Egocentric activity labels help locate candidate procedures \citep{grauman2021ego4d,damen2020epic100}; paired viewpoints and short human--object changes offer cues about correspondence and outcomes \citep{grauman2023egoexo4d,goyal2017something}. These sources can identify what happened without fully specifying how a different body should reproduce it. For contextual teaching, selection must preserve the choice that distinguishes this example from another clip of the same activity.

Annotation is useful when it removes uncertainty about the relation teaching must preserve. Common capture protocols reduce incompatible recording conventions~\citep{punamiya2026egoverse}; atomic segments and reconstructed hand and camera motion expose where an interaction begins and ends~\citep{aoe2026openaoe}. Object-state and contact annotations can resolve physical events hidden in RGB~\citep{cao2026acedata}. These additions are not interchangeable: reliable boundaries help preserve order, whereas contact-sensitive transfer needs relative motion and object response. The appropriate annotation follows the task distinction the support example must convey.

\begin{figure}[!tp]
\centering
\includegraphics[width=\linewidth]{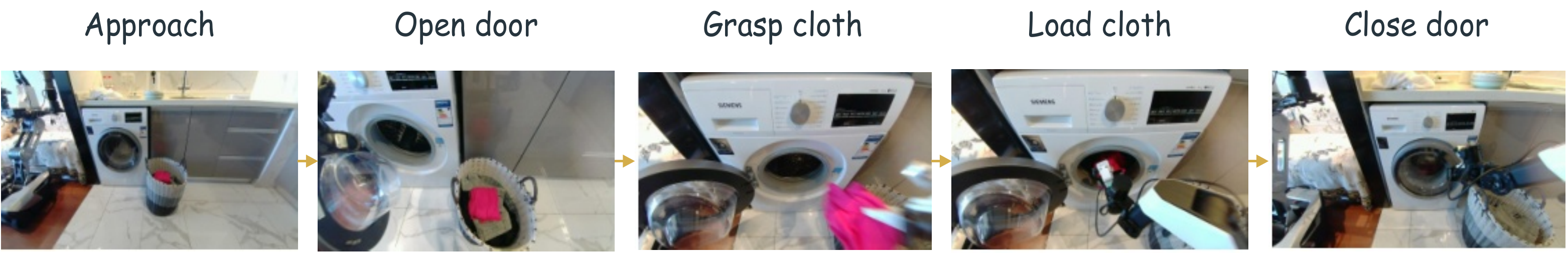}
\caption{Temporal anchors for contextual teaching: opening the door enables loading, and closing it completes the procedure. These transitions align a teaching example with progress in a later execution.}
\label{fig:data-landscape}
\par\phantomsection\label{floatend:fig:data-landscape}
\end{figure}

Procedural and conversational datasets contribute another missing dimension. Assembly101 records natural variations, mistakes, and corrections during assembly and disassembly; HoloAssist pairs a performer's egocentric activity with an instructor's spoken guidance and intervention annotations \citep{assembly1012022,holoassist2023}. Their contribution is the relationship between a partial procedure and information that changes its continuation. Transferring that relationship to robot ICL requires robot-compatible continuation targets or an executor that can realize the corrected instruction. Preserving mistakes, instructions, and corrections as linked events retains this distinctive supervision.

Human-video priors and demonstration following use this evidence differently. Learning latent plans from human play can reduce the robot experience needed for control \citep{wang2023mimicplay}, even when human and robot executions are not paired. Following a newly supplied video additionally requires preserving the demonstrated choice in a compatible robot execution. WAM-TTT learns that connection through paired human--robot meta-training, while its deployment-time fast-weight update needs only action-free human video \citep{arxiv260706988}. This is parameter adaptation under the distinction in Section~\ref{sec:foundations}. Robot supervision establishes the connection during training; the later teaching input can omit recorded robot commands.

\FloatBarrier
\surveyanchor{discuss:fig:data-landscape}
Preserving a teaching relation requires more structure than a shared task name. A mobile-manipulation episode contains identifiable substeps and transitions that can anchor retrieval or a corrected continuation (Figure~\ref{fig:data-landscape}). Human support and robot execution can traverse those stages at different speeds, so task correspondence matters more than frame equality. Constructing this usable teaching signal incurs a cost beyond recording the raw activity, motivating the acquisition comparison below.

\begin{figure}[t]
\centering
\includegraphics[width=\linewidth]{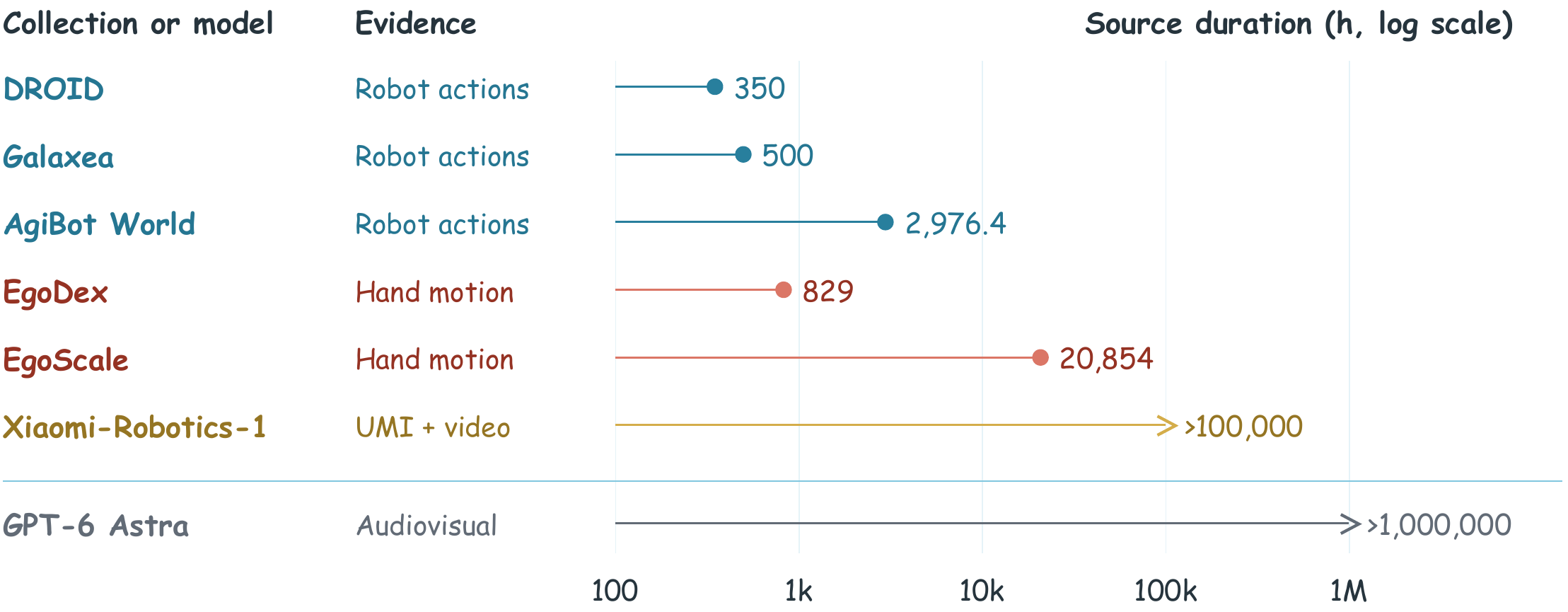}
\caption{Reported source duration across embodied collections and general-model pretraining~\citep{khazatsky2024droid,galaxea2025,agibot2025,egodex2025,egoscale2026,xiaomi2026robotics1,openai2026astratraining}. Action-aligned recordings support grounded teaching episodes; broader audiovisual corpora supply knowledge across activities and domains. The contrast motivates comparing explicit context-use training with broader pretraining. Arrowheads denote lower bounds.}
\label{fig:data-tradeoffs}
\par\phantomsection\label{floatend:fig:data-tradeoffs}
\end{figure}

\subsection{Acquisition cost and contextual capability}
\label{sec:data-tradeoffs}

Data value depends on the cost of producing usable teaching, not recording alone. We first examine correspondence costs, then compare explicit contextual training with broad priors and robot grounding.

\paragraph{Collection cost and correspondence.}
Filtering, synchronization, reconstruction, pairing, and validation contribute to acquisition cost. Egocentric video exposes hand--object interaction; UMI also measures device motion through a robot-like interface. The required conversion depends on the task: order-sensitive packing needs event boundaries and contrasting sequences, whereas tight insertion needs calibrated relative motion and contact evidence. Additional views or longer recordings improve coverage only if they preserve the distinctions needed for the task.

\surveyanchor{discuss:fig:data-tradeoffs}
Figure~\ref{fig:data-tradeoffs} compares source duration and evidence type. Xiaomi-Robotics-1 combines UMI and egocentric recordings; general video spans broader activity with less robot-action correspondence. Hours measure exposure, while annotation effort and usable support--query pairs characterize the work needed to \mbox{turn it into teaching.}

GPT-Policy makes the contribution of correspondence concrete~\citep{cheng2026gptpolicyeval}. It compares robot video with the same video augmented by aligned end-effector poses, gripper states, and commands. Plug reinsertion succeeds in zero of three video-only trials and two of three augmented-reference trials. This is a combined state--action comparison. Goal-image and occluded-goal settings provide endpoint or self-interaction evidence (Section~\ref{sec:applications}), showing how the content of the acquired evidence shapes its contribution.

\paragraph{Learning targets and acquisition routes.}
\label{sec:icl-acquisition-routes}
Pretraining can expand both executable behavior and the requirements inferred from teaching. Support--query episodes explicitly supervise the example--continuation relationship~\citep{duan2017oneshot}; GPT-3 showed that broader pretraining can also support example-based specification at inference~\citep{brown2020fewshot}. Both routes can yield fixed-parameter adaptation. Their transfer to unfamiliar rules is an empirical question distinct from the choice of training recipe.

S1 and GEN-1.5 illustrate explicit contextual pairing and continuous physical experience (Section~\ref{sec:context-training}). Their shared demonstration interface motivates comparing which training relationships produced context use, with tasks, information, motor interfaces, and training budgets matched.

\paragraph{Mechanisms of fixed-parameter context use.}
\label{sec:context-learning-mechanisms}
Fixed-parameter computations offer candidate explanations for context-dependent behavior. Bayesian accounts infer a latent rule shared within a sequence~\citep{xie2021bayesian}; distributional studies identify repeated structure that favors contextual inference~\citep{chan2022distribution,chen2024parallel}. Controlled models can bind items to labels and retrieve their association~\citep{reddy2023abrupt}, or implement gradient-descent-like updates through linear self-attention~\citep{vonoswald2022gradient}. Comparisons with LLaMA show that similarities between contextual and parameter learning depend on the setting~\citep{shen2023gradient}. These accounts distinguish what training makes inferable from the computation used at deployment. For robots, tracing a taught distinction into the representation that guides action connects the training relationship to its physical consequence.

\paragraph{General-model priors and robot grounding.}
Astra adds a different source of prior competence to this comparison. Its public training summary describes broad text, image, and audiovisual sources, including video sources above one million hours, represented partly through transcripts, captions, or metadata~\citep{openai2026astratraining}. In GPT-Policy, a fixed general VLM interprets task references and interaction feedback, proposes robot-tool requests, and receives outcomes from a constrained controller~\citep{cheng2026gptpolicyeval}. This places contextual interpretation in a general model and physical realization in an explicit robot interface. The disclosure characterizes source exposure; the causal contributions of individual training components to the observed robot behavior remain unresolved.

These acquisition routes supply complementary resources. Broad priors can interpret procedures and familiar interactions, while grounded records connect them to the receiving robot's commands. GPT-Policy's action-reference comparison illustrates this complementarity at contact-sensitive execution. The useful comparison therefore concerns the task distinctions and action correspondence obtained per unit of acquisition effort, with prior competence and the execution interface specified.

\paragraph{From exposure to usable teaching.}
The reported comparisons support different conclusions: BPP varies task diversity at fixed demonstration count, S1 varies scale and prompting, and general-model disclosures describe cross-domain exposure. Together they motivate measuring the marginal value of an additional recording: does it add a missing motion, distinguish another valid procedure, or reveal how to correct a failure? Section~\ref{sec:reusable-learning-rule} connects these contributions to selection, composition, and rule inference. Retrieval and synthesis can then target the missing relationship rather than increase volume indiscriminately.

\subsection{Scaling data construction}
\label{sec:data-construction}

The coverage and cost analysis motivates three complementary construction routes. Retrieval reuses existing evidence; spatial supervision and task-preserving synthesis construct missing counterparts; failure--correction collection adds continuations absent from successful demonstrations. A demonstration can supply an execution reference or seed new training examples. The construction route determines which teaching relationships enter the learning units defined in Section~\ref{sec:context-training}.

Autonomous collection can increase experience while changing which situations the learner encounters. Task orchestration supplies new executions~\citep{ahn2024autort}; learned reward and success signals can select experience for subsequent policy training~\citep{ghasemipour2025selfimproving}. A contextual learner additionally needs the evidence that explains each selected continuation. Retaining its instruction, demonstration, or correction preserves a candidate teaching--execution pair. Contrasting pairs are needed to establish which behavior changes with that teaching. This links acquisition policy to the learning objective: selecting only successful motion favors motor competence, while preserving the preceding uncertainty and its resolution can train responsiveness to new evidence.

\paragraph{Retrieval for training and contextual pairing.}
Training retrieval asks which external behavior can improve a robot's learned competence. Motion embeddings can locate human clips whose reconstructed trajectories provide reward guidance~\citep{qian2026robotok}; semantic, hand-pose, and motion similarity can select human samples compatible with robot demonstrations~\citep{lin2026simdex}. In both cases, relevance is defined by the supervision entering the policy's weights. This differs from contextual pairing, where the selected clip must explain a particular query procedure even if both behaviors are already within the motor repertoire. PHASE selects variable-length contact phases using tactile evidence, preserving insertion-stage compatibility in the retrieved training set~\citep{arxiv260930889}. This supplies a concrete counterpart to runtime retrieval: the selected evidence changes the policy learned offline.

Pair construction has a different criterion: the retrieved behavior must agree with the intended procedure. RHyME composes human clips for a recorded robot trajectory through sequence-level optimal transport \citep{kedia2024rhyme}. Existing video replaces newly captured support, but constituent behaviors must occur in both collections, and clip boundaries can introduce pairing errors. During deployment, retrieval must additionally match current progress and motion compatibility. Retrieval-based ReCAP supplies state--action chunks to a frozen residual predictor \citep{park2026recap}; expanding that archive changes runtime context through the mechanism in Section~\ref{sec:methods}. Training selection, pair construction, and runtime retrieval therefore require different judgments of relevance.

\paragraph{Constructing spatial supervision.}
Spatial supervision reduces ambiguity that action loss alone must resolve. Synthetic cross-scene point correspondences associate the same task geometry across executions~\citep{she2026matchingpolicy}; reconstructed contact heatmaps, hand configurations, and motion associate human interactions with affordances~\citep{oh2026vlaff}. The first teaches where an analogous relation occurs, while the second teaches which interaction geometry is plausible. Contextual training connects either spatial prior to the supplied example, so that teaching selects which relation the robot preserves. Spatial accuracy and responsiveness to teaching then become complementary training objectives.

\paragraph{One demonstration as a reference or a data seed.}
A single demonstration can be reused as an executable reference or as a seed for training data. Retrieval and alignment support the first route in MT3~\citep{dreczkowski2025mt3}. The second route generates additional trajectories by preserving bimanual coordination across object-dependent changes~\citep{zhou2025yoto,zhou2026bidemosyn}, or by varying reconstructed scenes and embodiments~\citep{robosplat2025}. These routes place generalization in different objects: a runtime reference or a policy fitted to its transformed executions. For contextual training, each generated trajectory remains associated with the demonstration whose task relation it realizes. This association turns motion augmentation into supervision for interpreting new teaching.

\paragraph{Task-preserving synthesis.}
Task-preserving synthesis needs both controlled variation and a criterion for what remains unchanged. Randomized human-arm renderings vary embodiment appearance~\citep{bonardi2019humans}; geometric pseudo-demonstrations and waypoint renderings construct motion-correspondent pairs~\citep{vosylius2024instantpolicy,qian2026synthicl}. Executing transformed segments in a simulator adds physical consequences and a success filter~\citep{mandlekar2023mimicgen}. Environment generation and compilation of task semantics into costs and transition conditions broaden the available variations~\citep{arxiv260509423,arxiv260831167}. These operations preserve different evidence. Rendering supplies correspondence, while executed validation tests realization under modeled dynamics; contextual supervision requires both to remain tied to the governing task specification.

KnowDemo makes the preserved requirement explicit~\citep{arxiv260921229}. It extracts task conditions from human video, grounds them in the target scene, and screens alternative contacts and subtask orders before planning and simulation. The verified trajectories then train a robot policy. This connects behavioral diversity to task constraints: varying an incidental grasp can expand training coverage, while varying a required order changes what the example teaches.

Simulation provides controlled task and scene variation alongside executable action records. RoboCasa expanded household environments and compositional tasks in 2024; RoboCasa365 adds 365 tasks and 2,500 kitchen scenes, with 612 hours of human-operated demonstrations in simulation and 1,615 hours generated with MimicGen \citep{robocasa2024,arxiv260304356}. These are simulated robot trajectories, including those collected by human operators. For ICL, the opportunity is to hold a procedure fixed across environments or vary the procedure within a matched environment, then verify that each generated execution preserves its assigned teaching condition.

Cross-embodiment synthesis must distinguish a changed appearance from a newly validated interaction. Generating human-video counterparts for robot tasks starts from a robot execution~\citep{zhou2026zerowam}; retargeting human motion and rendering robot arms into its scene starts from a human execution~\citep{wang2026ego2robot}. In the latter, visible object responses remain those of the human recording. Physical-consistency and forward--inverse checks can test whether a converted transition admits a compatible action~\citep{arxiv260519242,arxiv260401985}. The teaching relation adds another requirement: the conversion must preserve order, goal, and binding contacts. A physically plausible transition can still be the wrong continuation for the supplied demonstration.

Agent collection can obtain additional executions through shared interfaces. GUMI records pre-action observations and semantic commands for humans and agents \citep{chen2026showharness}. An ICL collection could use this interface to execute contrasting valid orders in comparable scenes, retaining each task specification, recording, and outcome. Pairing each successful execution with its governing specification would preserve the distinction needed for contextual supervision.

Processing must preserve intentional variation. An AgiBot-based study finds that demonstrator-specific speed variation can assign different chunks to similar observations, and that reducing this ambiguity improves downstream learning \citep{arxiv250706219}. Uniform speed conversion is unsuitable when insertion requires alignment phases or pouring requires pauses. The corresponding ICL decision is selective: normalize incidental collection style while retaining a pause, route, or contact order when it is part of what the example teaches.

\paragraph{Failure--correction collection.}
Successful demonstrations leave recovery states underrepresented. Corrective collection changes the state distribution on which supervision is obtained: DAgger queries an expert on learner-visited states~\citep{ross2011dagger}, DART uses perturbations to elicit nearby corrections~\citep{laskey2017dart}, and selective intervention concentrates human effort~\citep{hoque2021thrifty}. XR-2's separation of policy actions from human takeovers improves folding from 58\% to 93\% across three retraining rounds~\citep{xu2026bimanualscaling}. This expands the repairs represented in weights. Contextual recovery requires retaining the linked failure and feedback as well, so the learner can infer which repair the current evidence calls for.

The source of a correction determines what can be learned about its applicability. Synthetic perturbations, intervention in predicted rollouts, and planner-guided exploration expose different failure continuations~\citep{arxiv260313528,arxiv260421741,arxiv260819891}. Learning from intervention and autonomous outcomes can use these records without imitating every action~\citep{arxiv251114759}. For contextual learning, the record must retain the relation among the failed action, the corrective change, and its consequence. Action-level arguments and return values make that relation more specific than a task-level failure flag, as ASPIRE illustrates~\citep{lu2026aspire}. Richer failure records are useful when they help distinguish the causes a later correction must address.

Corrective records also need to expose mistaken task-state estimates. After a failed grasp, the gripper may be empty while a stored plan marks pickup complete. Linking the repair to observations that establish this discrepancy supplies supervision for revising context as well as motion. Retaining the repair as guidance changes the external artifacts $B_n$ in Equation~\eqref{eq:agentic-update}; using its trajectories as policy targets changes neural parameters. Section~\ref{sec:evaluation} tests whether either acquisition improves later tasks.

\subsection{Connecting teaching data to a pretrained policy}
\label{sec:policy-integration}

Data construction must be matched by an interface through which teaching can change behavior. Three dependencies organize this integration: access to the evidence, preservation of its task distinction in the control intermediate, and an action encoding that can realize it. Access comes first. An observation-conditioned policy such as ACT needs an added context channel to distinguish procedures compatible with the same scene~\citep{zhao2023act}. Language-conditioned policies expose goals and constraints~\citep{kim2024openvla,black2024pi0}; goal images specify endpoints~\citep{octo2024}; sequential demonstrations can supply order or contact preferences~\citep{fu2024icrt,ding2026contextflow}. Adding an encoder provides access, while teaching-dependent targets establish use.

The missing dependency determines which training records a method needs. Foundation policies supply perception and motor interfaces~\citep{brohan2022rt1,brohan2023rt2,octo2024,kim2024openvla,hpt2024,liu2024rdt}, but the four families consume different task-specific intermediates: actions, geometric references, futures, or executable specifications. Subtask annotations and skill representations expose higher-level structure~\citep{agibot2025,galaxea2025,robomind2025v2,kim2025uniskill}; phase-aligned examples connect that structure to a current decision~\citep{yang2026icivla}. Direct supervision of an intermediate is one way to establish this connection. Joint training, pretrained interpretation, and geometric construction can also supply it. The requirement shared by the four interfaces is that the execution intermediate preserve the distinction selected by the evidence.

Human pretraining and deployment teaching act at different stages of this connection. Richer robot pretraining can increase the benefit of joint fine-tuning on human and robot demonstrations with reconstructed motion and subtask labels \citep{arxiv251222414}. For ICL, the resulting policy must additionally respond to a new example with its parameters fixed. Conditioning on language, quality, speed, control modality, and visual subgoals, as in the $\pi_{0.7}$ recipe \citep{arxiv260415483}, illustrates how inputs can distinguish otherwise mixed executions. Demonstration conditioning can extend these inputs when a video conveys the distinction that the action generator needs.

Encoding is the final bottleneck between an informative training pair and an executable command. Regularized ICRT action tokens and hierarchical tokenization preserve different aspects of motion~\citep{vuong2025actiontokenizer,fateh2026histat}. LipVQ-VAE's improvement from a low robot-success baseline shows that relieving a representation bottleneck can help while leaving execution difficult. The teaching signal passes through successive constraints: pairing identifies the intended relation, context encoding retains it, and action representation makes its realization expressible. These three stages must preserve the same taught distinction. The next chapter follows that requirement into deployment, where the current physical state determines how it can be realized and revised after failure.

\begin{takeaway}
\takepoint Data sources determine the available measurements; teaching--execution relationships determine what context can teach. Robot data supports motor competence; alternative taught behaviors within the same scene expose the additional dependence on demonstrations.
\takepoint Retrieval can select training data, construct a demonstration--execution pair, or supply examples during control. Only the last directly changes a frozen policy's input at deployment; the first two change the experience \mbox{from which it learns.}
\takepoint Useful scaling adds missing motor competence, task-defining alternatives, or evidence that changes a later attempt. The value of added experience follows the new distinctions it makes available to the learner.
\end{takeaway}

\FloatBarrier
\suppressfloats[t]
\section{Grounding: How Does Context Guide Recovery?}
\label{sec:landscape}
\label{sec:recovery}

Objects can move, grasps can fail, and resistance can change after a task has been correctly interpreted. Recovery aims to preserve the taught requirement while revising the state estimate and continuation. It establishes the current state, diagnoses the deviation, and selects a feasible repair. Retaining the repair with its applicable conditions makes that feedback useful beyond the current attempt.

\subsection{Grounding the task in the current physical state}
\label{sec:current-grounding}

Grounding requires two complementary estimates: the task requirement that remains binding and the physical state that may have changed. An insertion can preserve its destination and prerequisite order despite a moved part. We establish these estimates before examining motion feasibility and the entry \mbox{conditions of transferred segments.}

\paragraph{Preserving the task reference during execution.}
A transferable task reference must distinguish progress toward a goal from adherence to the taught procedure. Cross-view progress representations can align executions despite appearance or embodiment changes~\citep{sermanet2017tcn,zakka2021xirl}; demonstration-conditioned skill inference selects which procedure that progress belongs to~\citep{xu2023xskill,kedia2024rhyme}. Current observations must then establish whether its prerequisites occurred and its required contacts remain valid. Terminal appearance can conceal an incorrect order, so verification must use the relations supplied by teaching~\citep{arxiv260518746}. This connects correspondence in Section~\ref{sec:context-mechanisms} to recovery: a deviation is meaningful only relative to the procedure the robot was \mbox{actually asked to follow.}

\begin{figure}[!tp]
  \centering
  \includegraphics[width=\linewidth]{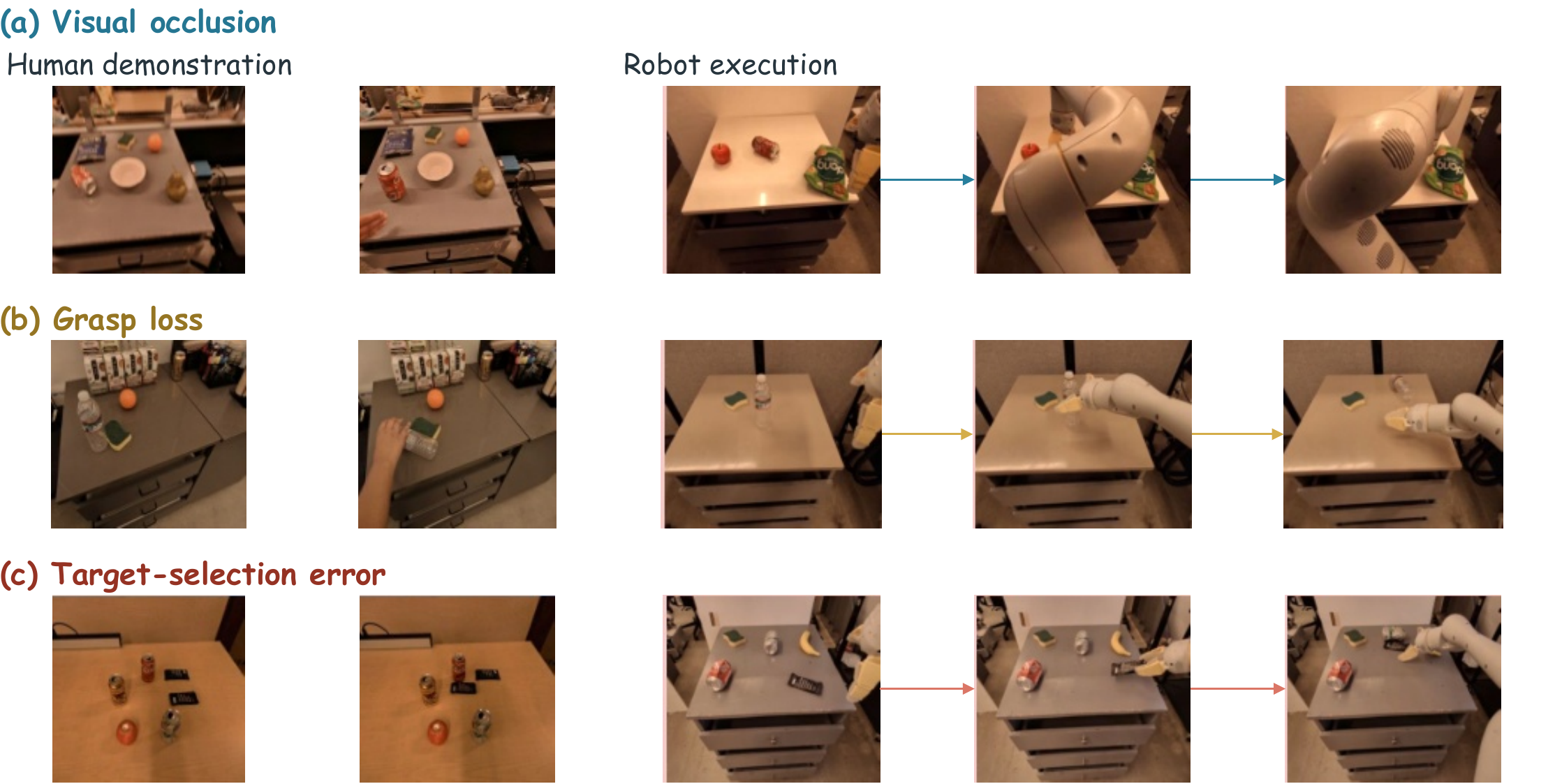}
  \caption{Three failures call for different recovery evidence: (a) visual occlusion, (b) grasp loss, and (c) incorrect target selection. Human demonstrations on the left establish the intended behavior; robot executions on the right reveal where realization departs from it.}
  \label{fig:recovery-choices}
\par\phantomsection\label{floatend:fig:recovery-choices}
\end{figure}

\surveyanchor{discuss:fig:recovery-choices}
Figure~\ref{fig:recovery-choices} distinguishes three failures in video-conditioned manipulation. An occluded interaction calls for a view that restores the missing geometry; a lost grasp calls for checking contact before resuming transport; a wrong target calls for revisiting the instruction and demonstration correspondence. Each repair addresses a different missing fact. Contact-intensive tasks add another ambiguity: a stalled insertion can reflect misalignment, unfamiliar resistance, or excessive force. A bounded probe can distinguish physical responses, whereas an irreversible operation may require inspection or clarification before another attempt.

\paragraph{Maintaining a current, task-relevant scene.}
A repair must distinguish an intended change from its observed outcome. A placement command or predicted future specifies what should happen; completion evidence supports recording the step as finished. If the part subsequently moves, its remembered pose needs revision while the earlier placement instruction can remain binding. Memory and geometry-indexed retrieval recover unavailable observations~\citep{arxiv260101075,arxiv260618960}; dynamic scene records replace spatial facts contradicted by new measurements~\citep{arxiv260819059,arxiv260900619,arxiv260900950}. When neither establishes applicability, another view or a reversible probe can distinguish the remaining explanations~\citep{arxiv260601247,arxiv260817129}. A navigation failure can similarly invalidate a remembered route or its terrain assessment. VLM-GroNav~\citep{elnoor2024vlmgronav} aligns visual terrain examples with sinkage or slip indicators before revising traversal estimates. This distinguishes physical feedback about a surface from geometric evidence that a passage is open.

Applicability can be checked before a remembered event influences control. CoreSense examines episode scope, provenance, temporal validity, and consistency~\citep{arxiv260919512}. Its recorded-data and simulated-signal evaluation measures evidence selection. Physical recovery evaluates the subsequent step: whether the selected evidence supports a repair that succeeds \mbox{on the robot.}

\paragraph{Connecting intended change to feasible motion.}
A corrected interpretation needs an executable continuation. Forward--inverse consistency checks agreement between proposed actions and modeled transitions; force-conditioned control adds measured contact feedback~\citep{arxiv260401985,mai2026crvlaforce}. These mechanisms make a context-selected correction executable. When the required motion lies outside the controller's repertoire, recovery also requires a new skill or a change in the physical setup.

\paragraph{Checking whether a transferred segment still applies.}
Geometric one-shot methods distinguish initial alignment from recovery during execution. DOME and ODIL first establish a relative starting configuration, whereas staged alignment can re-establish it at later subtask boundaries~\citep{valassakis2022dome,wang2025odil,wichitwechkarn2025annotationfree}. A transport segment also depends on the object remaining grasped: after a slip, geometric alignment can persist while the segment's entry condition fails. Its entry condition and intended completion therefore carry different information: the former establishes whether the recorded motion can begin, and the latter whether the next segment should follow. Learned subgoal transitions similarly govern progression within trained long-horizon policies~\citep{jain2025transitions}.

Failure can invalidate a reference or expose missing control competence. Robustness via Retrying recomputes a continuation through image registration and predictive control~\citep{ebert2018retrying}; ScrewMimic uses a demonstrated screw-motion representation to guide further bimanual learning~\citep{bahety2024screwmimic}. The former changes the continuation through existing components; the latter improves the component that realizes it. This distinction directs recovery toward revising the current reference or expanding the motor capability needed to realize it.

\subsection{From failure evidence to corrective execution}
\label{sec:corrective-execution}

Recovery proceeds from assessment to diagnosis and correction. Assessment establishes a deviation from the taught procedure; diagnostic evidence distinguishes its plausible causes; the controller's correction interface determines which repair can be executed. The resulting continuation may be absent \mbox{from the successful demonstration.}

\paragraph{Task-conditioned progress assessment.}
Failure assessment requires a reference for what should have happened. Action consistency, visual progress, and calibrated rollout scores can detect departures from expected execution~\citep{agia2024sentinel,xu2025faildetect,arxiv260623085}. For ICL, the reference follows the current teaching, allowing an unfamiliar but valid order to define expected progress. GVL illustrates example-conditioned assessment by using observation--progress examples to improve a frozen VLM's estimates from shuffled frames, including human examples for robot trajectories~\citep{arxiv241104549}. The taught reference makes progress assessment responsive \mbox{to the current task.}

Progress and physical acceptability can disagree. A connector can reach its apparent destination with an unsuitable contact force, just as a kit can reach the right inventory in the wrong order. Verification must therefore test both the context-defined requirement and the measurements needed \mbox{to establish acceptable execution.}

\paragraph{Diagnostic evidence acquisition.}
Diagnosis turns failure detection into a repair choice by identifying the likely cause. Multisensory diagnosis and learned failure knowledge organize clues to a stalled action~\citep{liu2023reflect,duan2024aha}, but misalignment and an unseen obstruction can still require different interventions. Uncertainty about the intended task can be reduced by clarification and reasoning over alternative plans~\citep{matsushima2020uncertainty,ren2023knowno,liang2024introplan}. Uncertainty about resistance instead requires sensing or interaction, including tactile evidence absent from RGB~\citep{arxiv260825798}. The unresolved cause determines the information source: teaching clarifies intent, while sensed action effects reveal the resistance encountered by the robot.

Task-context exploration and question-based teaching formalize the choice to acquire evidence~\citep{zhang2020metacure,cakmak2012questions}. The useful check is the one that discriminates causes relevant to the next action: a view for occlusion, contact feedback for grasp loss, or clarification for ambiguous intent. Affordance Agent Harness routes detection, segmentation, and other skills, verifies evidence, and retries before committing an actionable region~\citep{huang2026affordanceharness}. Its episodic memory supplies priors for recurring objects. This illustrates how verification can govern evidence acquisition before a motion controller uses the result.

\paragraph{Correction granularity and feasibility.}
Correction granularity should match the cause of failure. A wrong target requires changing task selection; a displaced approach point requires a spatial update; insufficient displacement may require only a magnitude adjustment. Corrective-language training makes supervisory instructions actionable~\citep{dai2024racer}, while sparse adjustments around a frozen VLA can address magnitude errors when the target is already correct~\citep{arxiv260829967}. Subtask, motion, and pixel-target interfaces expose these changes at different resolutions~\citep{chen2026steerable}. The controller must also accept the correction from the state already reached, connecting diagnostic precision to the steerability analysis in Section~\ref{sec:methods}.

\begin{table}[!tp]
\centering
\surveytable
\caption{How teaching and feedback support assessment, correction, and reuse. Each row links supplied evidence to the decision or control element it changes. ENPIRE denotes its program-refinement branch.}
\label{tab:recovery}
\begin{tabularx}{\linewidth}{@{}>{\raggedright\arraybackslash}p{0.21\linewidth}YY@{}}
\toprule
\textbf{Work} & \textbf{Available evidence} & \textbf{Affected output} \\
\tablegroup{3}{Assessment and assistance}
GVL~\citep{arxiv241104549} & Observation--progress examples & Progress estimate \\
IntroPlan~\citep{liang2024introplan} & Retrieved reasoning examples & Plan; clarification \\
\tablegroup{3}{Corrective execution}
ICPI~\citep{merwe2025icpi} & Execution errors; correction examples & Motion-primitive parameters \\
RACER~\citep{dai2024racer} & Current images; corrective language & Action continuation \\
CorrectVLA~\citep{arxiv260829967} & Human feedback & Sparse action bias \\
\tablegroup{3}{Experience reuse}
DROC~\citep{zha2023droc} & Retrieved language corrections & Plan; skill calls \\
Zeva~\citep{chen2026zeva} & Action--effect history & Subsequent actions \\
ENPIRE~\citep{xiao2026enpire} & Execution feedback & Executable program \\
\bottomrule
\end{tabularx}
\par\phantomsection\label{floatend:tab:recovery}
\end{table}

Iterative in-context policy improvement links an observed error directly to a motion primitive's settings. A frozen language model uses primitive parameters, execution errors, and correction examples to propose an adjustment; its comparison favors relative errors over raw states and targets \citep{merwe2025icpi}. The error is useful because it identifies a change the primitive can make. Neural parameters stay fixed, while the generated control settings change within the primitive's available range.

When the frozen action policy cannot continue from the failure state, CoRe uses history and counterfactual continuations to choose a rejoin state, then realigns displaced objects before returning control \citep{core2026realignment}. The teaching still specifies the task; restoration makes an existing skill usable again.

Recovery can instead revise the history that conditions prediction. FARE compares a full causal cache, a recovered prefix, and a reset cache under the same latest observation, selecting their predicted continuations with a frozen world-action model~\citep{arxiv260918016}. This changes the model's retained evidence rather than physically restoring the scene. A compatible historical prefix must therefore be re-grounded in the current state. Its ablations find that always invoking recovery search or committing an unverified prefix can underperform the base policy, supporting selective intervention with verification. A continuation must also preserve completed steps and restore failed prerequisites, as captured by task graphs and \mbox{verified subgoal memory \citep{arxiv260511951,arxiv260829537}.}

\surveyanchor{discuss:tab:recovery}
Feedback is useful when it identifies a decision the robot can revise: its estimate of progress, its next action, or a rule retained for later use (Table~\ref{tab:recovery}). A contextual repair must change that decision while preserving the task requirements outside the correction's scope. This connects immediate recovery to the longer-term validity of the lesson it produces.

\subsection{Applicability and reuse of corrections}
\label{sec:correction-reuse}

Reuse retains a conditional lesson: which intervention corrected which failure, and under what conditions. Guidance or executable artifacts preserve it through Equation~\eqref{eq:agentic-update}; corrective training can incorporate such lessons into policy weights (Section~\ref{sec:data}).

Different lessons require different reset rules. Language-correction memory retains task constraints, whereas action--effect history retains physical response~\citep{zha2023droc,chen2026zeva}. A workcell reset clears completed-step memory; a changed connector or tool can invalidate a response estimate while leaving the assembly order intact. For a jammed connector, an adjustment justified by measured resistance should not be reused when misalignment is the cause. Retaining the observation, intervention, and outcome allows later attempts to reassess that condition instead of replaying the repair unconditionally.

Program-based reuse makes repairs executable. ENPIRE's program-revision branch uses fixed neural models; Zetta retains critics and recovery routines around a frozen policy~\citep{xiao2026enpire,arxiv260816590}. Interruptible execution exposes intervention points~\citep{chen2026volo}. Later trials test the repair's benefit under applicable conditions. Task structure and irreversibility determine what can be repeated, linking recovery to the application demands below.

\begin{takeaway}
\takepoint Context specifies the recovery reference and helps interpret failure. Physical measurements establish quality; a responsive control interface makes the correction executable.
\takepoint Corrective training expands available repairs; contextual recovery selects or revises one from fresh evidence. Reuse requires retaining the correction with its outcome and applicable physical conditions.
\end{takeaway}

\FloatBarrier
\suppressfloats[t]
\section{Applications: What Must Context Resolve?}
\label{sec:applications}
The value of context depends on the uncertainty an application makes consequential. Manipulation can leave the intended interaction unspecified; navigation can hide the route or local spatial structure; a long procedure can hide a completed prerequisite; an unfamiliar body can hide its response to action. These demands determine what teaching, observation, or interaction must supply. Repeated task changeover combines them with acquisition and reuse costs. Any of the four method families can serve several demands, provided its intermediate preserves the information the application requires.

\Needspace*{5\baselineskip}
\subsection{Task specification in manipulation}
\label{sec:manipulation-demands}

In rearrangement, an instruction or demonstration can specify which object belongs in which container, how components should be ordered, or which tool serves a purpose. A moved destination requires a new path that preserves the demonstrated object--destination relation. The Imitator Game makes differences between demonstration and execution explicit through scene, object, and functional changes \citep{zhou2026imitator}. These changes test preservation of intent across altered object correspondence. We first consider the distinctions an example must convey, then how they survive replacement of the objects involved, and finally how geometry or contact response changes their realization.

\paragraph{When an example conveys more than a task label.}
The value of a demonstration depends on what the user needs to specify. A final drawing leaves stroke order, direction, and tool approach unresolved; BPP's DrawAnything study connects broader drawing-task diversity to better demonstration-conditioned generalization \citep{patel2026bpp}. In its three-task laundry study, language is more reliable than behavior demonstrations, with some demonstration-conditioned failures executing the wrong fold. These contrasting settings identify an application condition: video is especially useful when it communicates a choice that a short task label or goal image leaves open. A goal image can specify the task when only a visible endpoint matters; a demonstration additionally conveys a required path or order.

\begin{figure}[!tp]
\centering
\includegraphics[width=\linewidth]{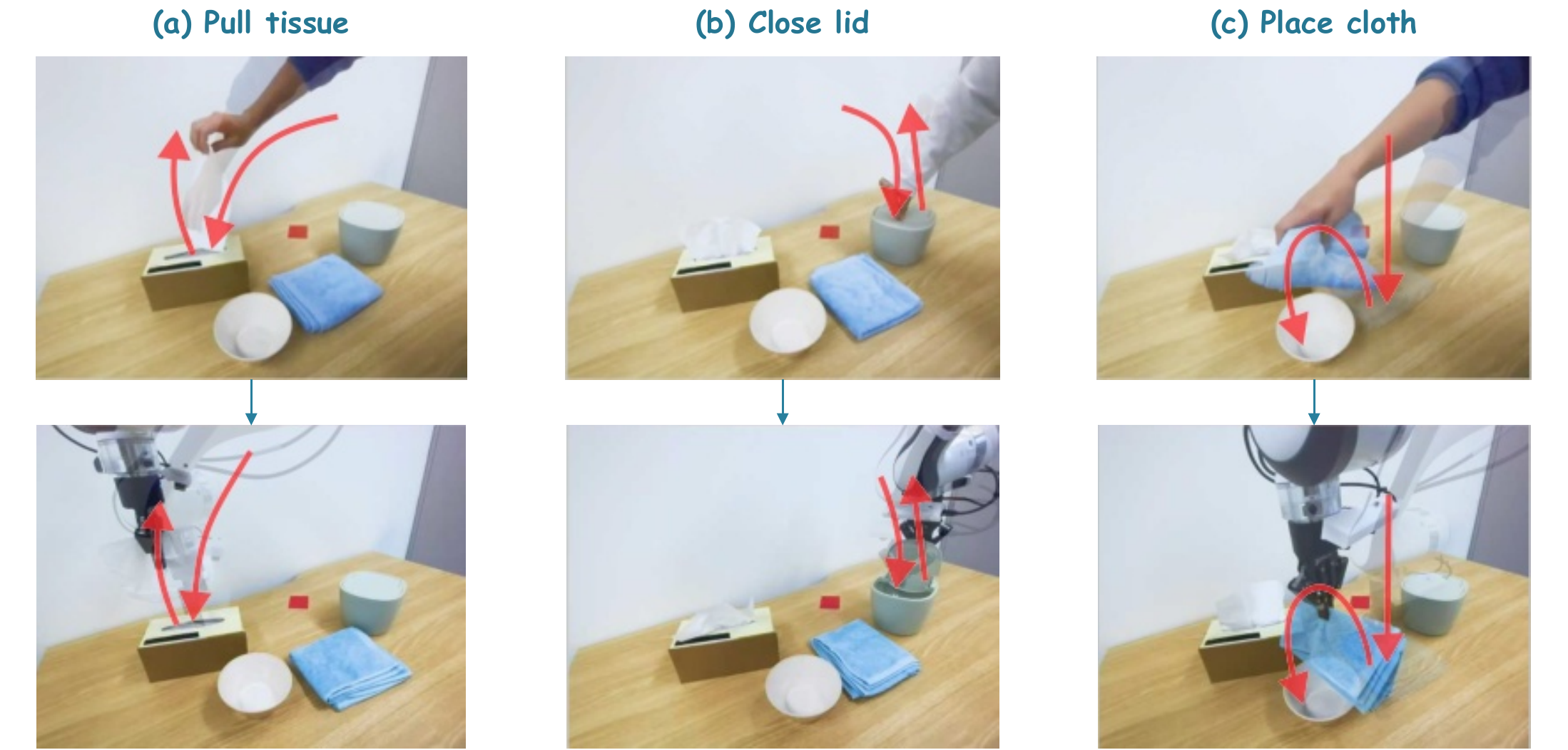}
\caption{Cross-embodiment skill transfer in UniSkill~\citep{kim2025uniskill}: (a) tissue pulling, (b) lid closing, and (c) cloth placement. Human demonstrations above and robot executions below preserve the interaction despite changes in \mbox{embodiment and object appearance.}}
\label{fig:cross-embodiment}
\par\phantomsection\label{floatend:fig:cross-embodiment}
\end{figure}

Replacing objects makes the distinction between task meaning and execution concrete. A new workpiece can preserve an assembly relation while changing its grasp; a new fixture can preserve the goal while changing the insertion geometry. Replacing both requires correspondence at both ends of the interaction, as the object-pair transfers in Section~\ref{sec:context-mechanisms} illustrate. The application must specify whether an object identity is binding or whether another object may serve the same role. Functional substitution is useful only when it preserves that requirement, including contact, order, and material constraints.

Personalized behavior adds persistent conventions to task specification. Language corrections can retain user preferences and guide subsequent plans~\citep{zha2023droc,li2025iclhf}. A request to place objects quietly should govern later placements, whereas a correction to one object's location need not. Such applications require both an acceptable endpoint and adherence to the user's continuing constraint.

Realizing these requirements across bodies adds a correspondence problem. In bimanual transfer, coordination must be preserved while each arm uses a feasible motion. Conditioning one predicted arm trajectory on the other addresses dependence within the action inference problem~\citep{palma2026bicicle}. Sharing an end effector across arms reduces a different uncertainty, the physical mismatch between demonstration and execution~\citep{ohkawa2026yubi}. Teaching specifies the intended cloth region or contact relation; action inference and embodiment design determine how coordinated motion realizes it.

\surveyanchor{discuss:fig:cross-embodiment}
Figure~\ref{fig:cross-embodiment} shows the preserved interaction across human and robot executions: pulling tissue, closing a lid, or placing cloth. The relevant object part and destination remain identifiable while the robot changes its approach and contacts. A handle-specific instruction restricts that variation; an endpoint-only request \mbox{leaves the grasp open.}

\begin{figure}[t]
\centering
\includegraphics[width=\linewidth]{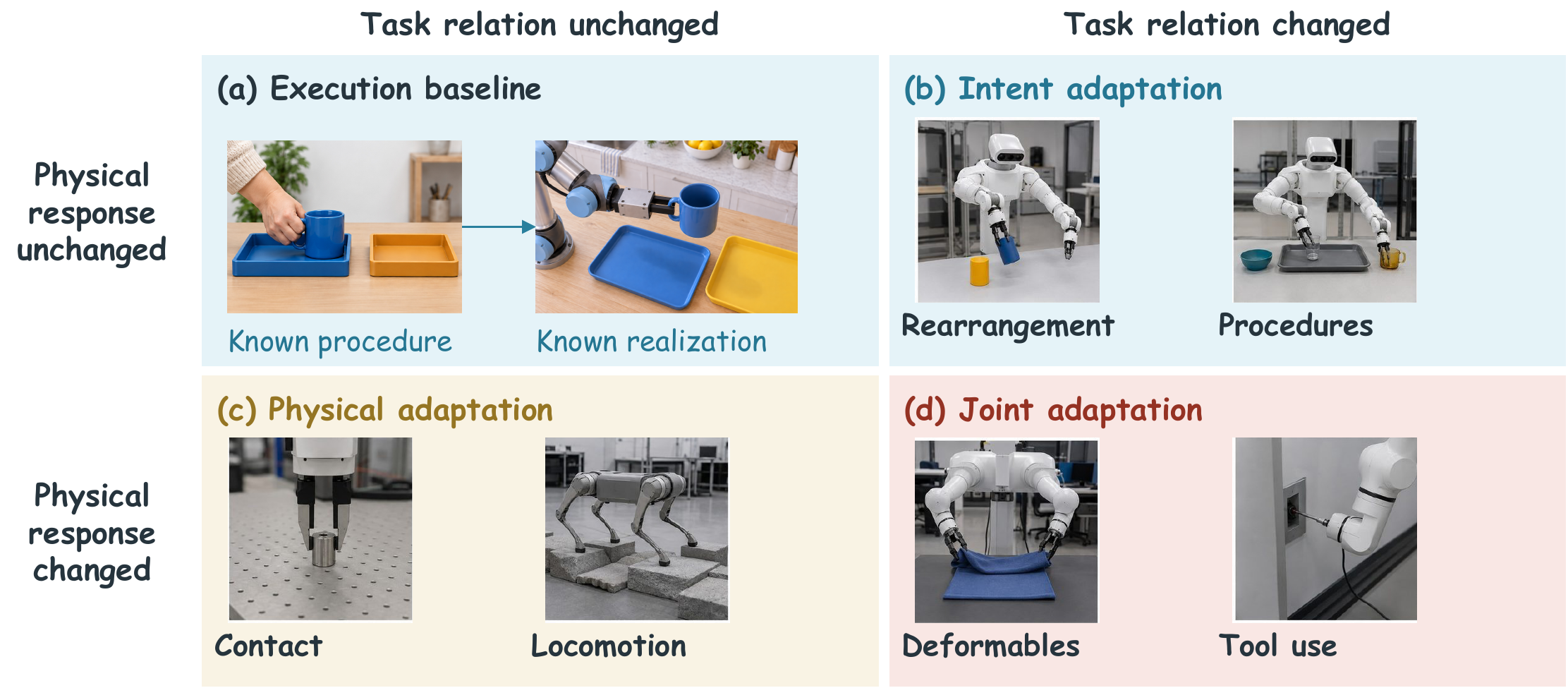}
\caption{Task meaning and physical realization vary independently: (a) an unchanged baseline, (b) a new task relation, (c) a changed physical response, and (d) both changes together. Teaching specifies the intended relation; interaction reveals how the robot can realize it.}
\label{fig:application-map}
\par\phantomsection\label{floatend:fig:application-map}
\end{figure}

Method selection follows the binding requirement. Geometric transfer exposes a contact or motion constraint; a skill interface exposes order and prerequisites; contextual policies and predictive control can encode either through their learned intermediates. The application supplies the criterion for judging these choices. Training pairs that equate executions solely by endpoint can erase a route or contact requirement that \mbox{the user values (Section~\ref{sec:context-training}).}

\paragraph{When intent and physical response change together.}
Deformable manipulation couples the procedure to a changing scene. A fold alters visible surfaces and accessible contacts, so the correct next action cannot be obtained by repeatedly matching the initial demonstration frame. Geometric and semantic information can jointly condition a folding policy \citep{wang2026instantfold}; preserving fold boundaries then provides a more useful correspondence than copying absolute hand trajectories. Instant-Fold's simulated mode variation and real garment variation examine different parts of this transfer. Its RGB-D and interaction interface, including estimated and potentially corrected human interaction points, also supplies richer information than \mbox{ordinary RGB \promptvideo{}.}

Material changes can leave the intended fold unchanged while altering the force needed to realize it. Visuo-tactile measurements distinguish completion from excessive compression \citep{arxiv260818701}. Here the demonstration supplies the procedure and action--effect evidence supplies the current material response, giving the two context sources different roles.

A new arrangement under familiar conditions primarily changes the requested relation. A familiar insertion under changed resistance primarily changes physical realization. A new fold on unfamiliar fabric couples both: the taught fold remains the reference, but each action alters the geometry and contact conditions for the next. This independent variation provides a basis for choosing teaching and feedback, even within one application.

\surveyanchor{discuss:fig:application-map}
Figure~\ref{fig:application-map} crosses the two demands: a changed task relation, changed physical response, or both. This view identifies what evidence an application needs before comparing the method used to process it. Navigation extends this question from object interaction to spatial knowledge and traversability.

\subsection{Navigation and spatial adaptation}
\label{sec:navigation-demands}
\suppressfloats[t]

Navigation requires more than a destination: a robot may need a demonstrated route, knowledge of an unfamiliar building, examples of how to interpret directions, or feedback about an unsuccessful passage. Table~\ref{tab:navigation-context} organizes this evidence into four context types. The following paragraphs compare them in parallel by what they convey and how they change navigation. This application taxonomy concerns the supplied information; the method families in Section~\ref{sec:methods} concern the computation that turns it into behavior. Several context types can be used together.

\begin{table}[!htbp]
\centering\surveytable
\caption{Navigation contexts grouped by the information they supply at deployment. Each type supports a distinct decision, and a system can combine several types. These information categories complement the four computational method families in Section~\ref{sec:methods}.}
\label{tab:navigation-context}
\begin{tabularx}{\linewidth}{@{}>{\raggedright\arraybackslash}p{.21\linewidth}>{\raggedright\arraybackslash}p{.39\linewidth}Y@{}}
\toprule
\textbf{Context type} & \textbf{Evidence and decision} & \textbf{Representative approaches}\\
\midrule
\textbf{Route demonstrations} & Ordered views or landmarks, sometimes with actions. Follow a demonstrated path or sequence of subgoals. & Zero-Shot Visual Imitation~\citep{pathak2018zeroshot}; RPF~\citep{kumar2018rpf}; sparse visual memory~\citep{yoo2020sparsepath}\\
\addlinespace[5pt]
\textbf{Environment observations} & Scene previews or earlier visits. Locate objects and connecting passages for new goals. & SPTM~\citep{savinov2018sptm}; NOLO~\citep{zhou2024nolo}; ReLIC~\citep{elawady2024relic}; NavProbe~\citep{liu2026navprobe}; SparseNav~\citep{arxiv260926408}\\
\addlinespace[5pt]
\textbf{Instruction and decision examples} & Instruction--landmark, instruction--subtask, or image--decision pairs. Interpret a new instruction or select a move. & LM-Nav~\citep{shah2022lmnav}; A$^2$Nav~\citep{chen2023a2nav}; Select2Plan~\citep{buoso2024select2plan}\\
\addlinespace[5pt]
\textbf{Outcome feedback} & Outcomes, failure notes, terrain response, or corrective dialogue. Revise guidance or traversal costs. & VLM-GroNav~\citep{elnoor2024vlmgronav}; CMMR-VLN~\citep{li2026cmmrvln}; HAM-VLN~\citep{liu2026hamvln}; Talk2Escape~\citep{li2026talk2escape}\\
\bottomrule
\end{tabularx}
\par\phantomsection\label{floatend:tab:navigation-context}
\end{table}

\phantomsection\label{discuss:tab:navigation-context}%

\paragraph{Route demonstrations specify where to go in sequence.}
A tour can teach a particular passage through an environment: pass the reception desk, traverse a corridor, and stop by the stairs. The ordered reference remains relevant even when execution deviates from the original camera poses. Zero-Shot Visual Imitation~\citep{pathak2018zeroshot} treats demonstration images as successive goals for a pretrained goal-conditioned policy, including navigation with a TurtleBot in unseen offices. RPF~\citep{kumar2018rpf} instead encodes views and recorded actions into a path memory that guides execution under motion noise and scene changes. Sparse visual memory~\citep{yoo2020sparsepath} reduces the retained views while preserving the action sequence, examining how much visual evidence path following needs. These approaches differ in whether the reference specifies visual subgoals or an action-bearing route. Their shared mechanism aligns present progress with the demonstrated route.

\paragraph{Environment observations reveal where places and objects are.}
A preview or earlier visit can reveal the building without prescribing the next route. SPTM~\citep{savinov2018sptm} converts an exploration recording into a graph of visual locations, then selects reachable waypoints for different goals using fixed retrieval and control networks. NOLO~\citep{zhou2024nolo} conditions navigation actions on a scene video, while ReLIC~\citep{elawady2024relic} accumulates experience across episodes with changing starts and goals in one home. NavProbe uses a spatial index to find visual and geometric evidence relevant to the current subgoal~\citep{liu2026navprobe}. Its executive consults these records when the available evidence is insufficient, then revises the subgoal before invoking navigation skills. The distinction is how local knowledge becomes available: a graph, a supplied sequence, policy memory, or actively retrieved evidence. A new goal can reuse these facts; a new building requires acquiring them again. CLUE additionally lets exploration revise what the task requires: hypotheses about relevant objects and functions are tested against an online language-embedded map~\citep{arxiv260930428}. Context acquisition then resolves an underspecified objective as well as an unknown route.

The current instruction can also decide which facts are worth acquiring. SparseNav retains traversability geometry but grounds semantic landmarks only when the active sub-instruction makes their location useful~\citep{arxiv260926408}. Its sparse landmark memory and instruction-progress record then guide waypoint selection. This separates maintaining a navigable map from repeatedly labeling every visible object. Together with NavProbe's targeted retrieval, it treats context construction as a decision about missing information. The resulting context is organized around the active instruction and the evidence needed for its next navigation decision.

\paragraph{Instruction and decision examples teach an interpretation.}
Example pairs can specify how a navigation request should be translated into executable choices. LM-Nav~\citep{shah2022lmnav} supplies instruction--landmark examples to a language model; the extracted sequence is grounded in a topological graph and executed by a pretrained navigator. A$^2$Nav~\citep{chen2023a2nav} extends this interpretation to action demands: passing a doorway and entering it require different subtasks even when the landmark is identical. Its prompted parser selects among navigators trained for these demands. Select2Plan~\citep{buoso2024select2plan} moves example conditioning closer to the decision itself by retrieving image--answer pairs for movement or waypoint selection. Thus the example can teach a symbolic decomposition or a visual choice. Unlike a route recording tied to particular places, such a pairing can explain how to interpret a new query using an existing execution interface.

\paragraph{Outcome feedback revises what should be attempted next.}
Execution adds evidence that a successful demonstration may omit: a branch was unhelpful, a stopping decision was premature, or a visually open surface caused slip. CMMR-VLN~\citep{li2026cmmrvln} stores successful routes and notes about the first incorrect decision for retrieval in later navigation. HAM-VLN~\citep{liu2026hamvln} attaches failure notes to visited places during an episode, making them available when those regions become relevant again. VLM-GroNav~\citep{elnoor2024vlmgronav} grounds terrain judgments in physical interaction and updates waypoint selection and local traversal costs. Talk2Escape acquires additional guidance when retained evidence cannot resolve a failure: detected looping or trajectory divergence triggers a landmark-grounded question to a human or oracle~\citep{li2026talk2escape}. The reply corrects navigation context. Feedback thus revises a decision through remembered outcomes, physical measurements, or newly requested clarification. These lessons remain useful while the associated instruction, place, and \mbox{physical conditions still apply.}

The four navigation contexts retain different things: a prescribed route, facts about an environment, an interpretation rule, or a conditional lesson. Their validity therefore changes at different boundaries. A new goal can reuse a map but replace the route; a new building invalidates local locations while leaving an interpretation convention useful; changed terrain can invalidate an earlier traversal estimate. Section~\ref{sec:context-transfer-evaluation} translates these distinctions into controls. Combining navigation with manipulation adds procedural state: after reaching a pickup site, the robot must establish whether the object was acquired. The next subsection examines these dependencies across a complete procedure.

\subsection{Long procedures and irreversible operations}
\label{sec:procedure-demands}

Long procedures change the teaching problem from selecting an action to maintaining a dependency structure. Interaction phases and sequential end-effector goals can expose it explicitly, including in teleoperated RGB-D demonstration interfaces \citep{chen2025manilong}; a context-conditioned policy must retain the corresponding distinctions in its representation. An unfinished step, a completed step, and a failed prerequisite require different continuations even when the current image is similar. Long procedures therefore require correspondence between the teacher's successful sequence and the robot's interrupted or incomplete execution, beyond retaining a longer record. The analysis moves from dependency tracking to the additional constraints \mbox{imposed by irreversible steps.}

\begin{figure}[t]
\centering
\includegraphics[width=\linewidth]{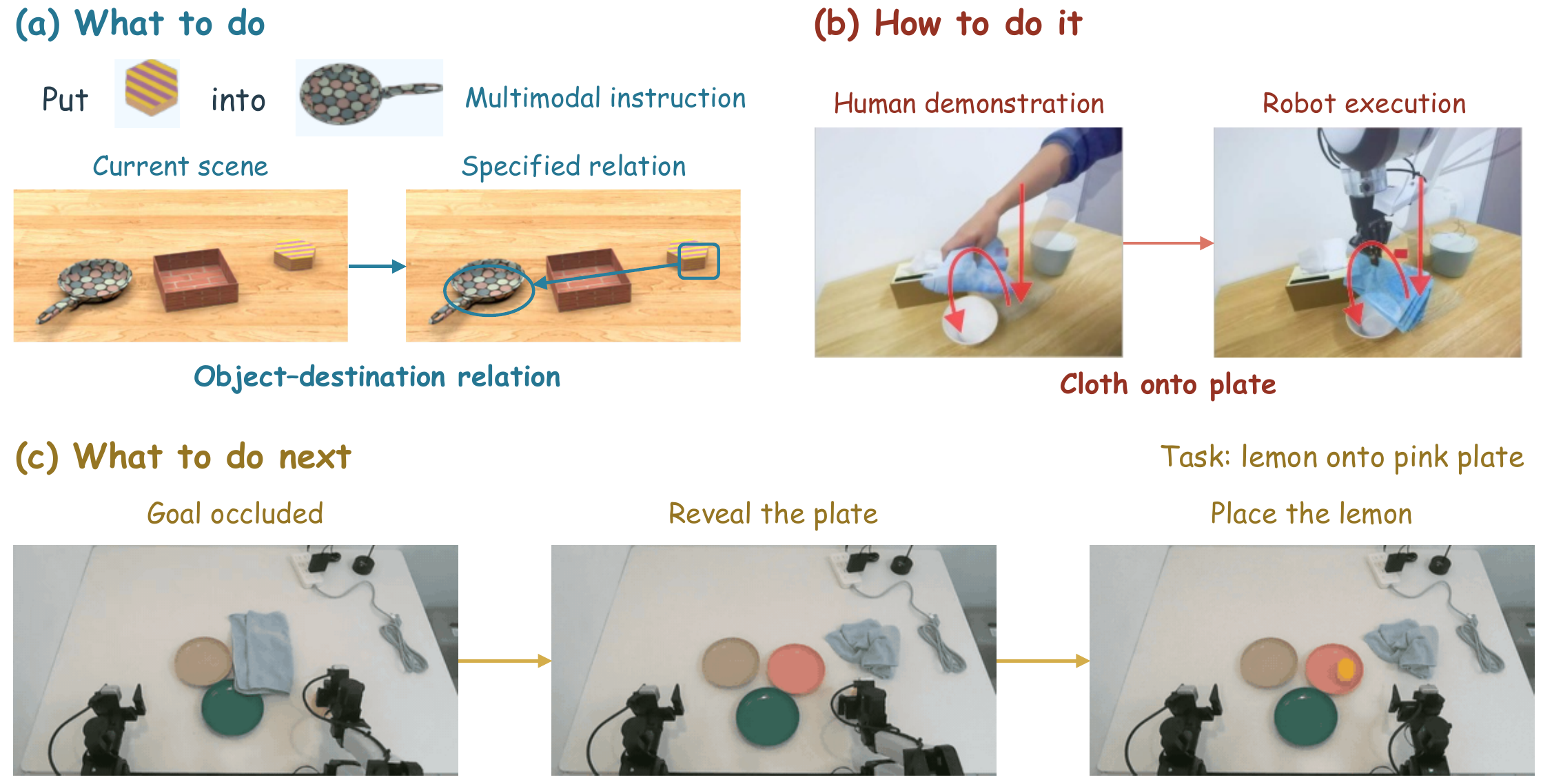}
\caption{Three decision roles of context: (a) task specification selects the object and destination; (b) interaction guidance conveys how the cloth should move onto the plate; (c) execution-state tracking uses the result of removing the towel to advance from uncovering the target to placing the lemon. Representative implementations are VIMA~\citep{jiang2022vima}, UniSkill~\citep{kim2025uniskill}, and GPT-Policy~\citep{cheng2026gptpolicyeval}, respectively.}
\label{fig:application-taxonomy}
\par\phantomsection\label{floatend:fig:application-taxonomy}
\end{figure} 

Irreversible procedures add a constraint on how progress and failure can be established. In laboratory workflows, intermediate products must remain valid while the robot decides how to continue. Benchmarks spanning atomic skills, compositions, and extended workflows expose the interaction of procedural inference and dexterity \citep{arxiv260818618}; household planning with low-level embodiment abstracted away isolates dependencies among steps \citep{arxiv260514504}. Varying the supplied procedure within these settings would test whether contextual teaching preserves dependencies through the available executor.

Long procedures also couple memories with different validity periods. Progress memory records which steps have been completed; physical memory estimates how to perform the remaining operation. Zeva's two-timescale analysis tests reuse of action effects across reset attempts~\citep{chen2026zeva}. In a laboratory workflow, however, a physical reset may be impossible: repeating an insertion and repeating a dose have different consequences. The informative next interaction must therefore respect both the taught dependency structure and the state of the intermediate product.

Reversibility determines how the recovery mechanisms in Section~\ref{sec:recovery} can be used. A placement may permit local correction; dispensing or fastening can require inspection before a retry. The application therefore needs evidence of completed operations, alongside the demonstrated dependency structure, to decide whether a failed step can be resumed.

\surveyanchor{discuss:fig:application-taxonomy}
The distinctions in Figure~\ref{fig:application-taxonomy} concern what information is missing from the next decision. In (a), \emph{task specification} identifies the required object--destination relation among alternatives already present in the scene. In (b), \emph{interaction guidance} conveys the motion relation that should survive the change from a human hand to a robot gripper. In (c), \emph{execution-state tracking} establishes that the plate has been uncovered, making lemon placement the next feasible step~\citep{cheng2026gptpolicyeval}. These examples instantiate the information roles in Section~\ref{sec:foundations}: panel (b) combines a specified interaction with cross-embodiment correspondence, while panel (c) updates execution state under an unchanged goal. They are complementary information requirements within a task, and can be supplied together through language, demonstrations, and interaction history.

A new procedure can require either a new arrangement of available operations or a motion the current policy has not acquired. Assembly adaptation through fast weights~\citep{jiang2026robottt} and GEN-1.5's brush-to-dustpan example after five minutes of fine-tuning~\citep{generalist2026gen15} place new evidence into parameters. They provide a useful comparison with fixed-weight teaching: context can select, order, and parameterize available operations, including new motions within the executor's capabilities; further policy learning can expand those capabilities. This separation localizes long-procedure failures to task interpretation or motor competence.

Long-procedure evaluation must separate adherence to a new ordering from adaptation to changed physical response. Body and dynamics adaptation isolates the second demand: how a familiar requirement is realized when the physical system changes.

\subsection{Adaptation to body and dynamics}
\label{sec:body-demands}

Body adaptation isolates physical response while keeping the requested task specified. Locomotion makes this separation explicit; manipulation introduces analogous uncertainty through payloads and contact. Their combination in mobile manipulation then requires procedure and progress memory alongside the physical estimate. A direction or speed command can already specify the goal; recent action consequences reveal how the current body and terrain respond. Learned adaptation modules address this physical uncertainty \citep{kumar2021rma}, while longer-context experience broadens the setting toward different morphologies, actuator changes, and evidence retained across attempts \citep{liu2025locoformer}. A broader range of bodies expands the response variations that the policy must infer from its interaction history.

Locomotion also separates a change in the requested behavior from a change in the body that realizes it. Example-conditioned foot-contact generation can specify a gait-like procedure for a pretrained controller~\citep{tang2023saytap}. Active tracking across wheeled, quadrupedal, and aerial platforms instead uses a shared velocity interface, with body-specific realization delegated to lower-level control and visual history identifying the moving target~\citep{wu2026adatracker}. The shared interface supports cross-platform execution by assigning target tracking to visual history and body-specific motion to lower-level control. The contextual contribution depends on whether the evidence specifies a gait, tracks a target, or identifies an unfamiliar response.

Manipulation introduces analogous uncertainty through a new gripper, payload, or deformable object. RopeFormer retains action--response history across trials on unseen ropes, adapting rotation and whipping with fixed policy weights~\citep{arxiv260923432}. MetaPusher instead updates a dynamics model during pushing and revises the plan for the unfamiliar object~\citep{arxiv260921122}. The goal can remain fixed in both cases: interaction reveals how to realize it under changed physical conditions. Combining this evidence with demonstrations requires distinguishing a changed task requirement from a changed response.

Histories of a stable command under different body or terrain responses provide the training relationship for this adaptation. Mobile manipulation combines this requirement with procedural teaching and the spatial knowledge discussed in Section~\ref{sec:navigation-demands}. In a delivery procedure, the example specifies the destination and handling order, lifting an object changes the payload, and execution memory records whether pickup has completed. The contextual learner must retain the procedure and progress while updating the physical estimate as the load changes. Evaluating the combination requires both adherence to the demonstrated handling and reliable transport under the changed response.

Procedural teaching and response adaptation become especially valuable when the same capabilities serve recurring tasks. Their practical benefit then depends on which acquired knowledge can be reused and which conditions must be checked again.

\subsection{Task changeover, teaching, and reuse}
\label{sec:business-applications}

Frequent procedural change over a reusable motor repertoire gives contextual teaching a concrete application. Two kitting orders can use the same parts and grasps but require different placement sequences or handling conventions. A demonstration is useful when it communicates that difference beyond the bill of materials; a shorter instruction can suffice for a fully specified count or destination. Across batches, the chosen sequence persists while completed-step memory resets. A changed tool requires a new physical estimate while the order may remain valid. This separates three parts of changeover: specifying what changes, acquiring the evidence needed to validate it, and retaining the result for subsequent batches.

The recurring change determines the economical teaching channel. Counts or destinations may be stated directly; a fold can require a sequential example; a changed material may require a test interaction. Precision and irreversibility determine the verification needed before accepting the new procedure (Section~\ref{sec:recovery}). Changeover therefore measures the complete acquisition process, including the evidence needed to validate it.

The relevant endpoint is repeatable execution at the required quality. Reusing a task specification can reduce preparation for later batches, while changed tools or materials create a new validation cost. This distinguishes the value of contextual teaching at changeover from the value of a retained procedure during production. Section~\ref{sec:changeover-cost} measures both through teaching, interaction, and operating effort.

\paragraph{Teaching and reuse interfaces.}
\label{sec:industrial-teaching}

Teaching can specify a task for existing competence or revise the resources used in later execution (Table~\ref{tab:industrial-teaching}).

\begin{table}[!tp]
\centering\surveytable
\caption{Teaching and reuse interfaces relevant to production. GEN-1.5, S1, and GLOW condition execution with fixed weights; physical feedback can instead revise programs or train policies.}
\label{tab:industrial-teaching}
\begin{tabularx}{\linewidth}{@{}p{.18\linewidth}Y Y Y@{}}
\toprule
\textbf{System} & \textbf{Task evidence} & \textbf{Adaptation} & \textbf{Context or artifact}\\
\midrule
\tablegroup{4}{Demonstrations condition robot execution}
GEN-1.5~\citep{generalist2026gen15} & Sensors; trajectories & Physical prompting & Demonstration; history\\
\addlinespace[2pt]
S1~\citep{skild2026s1} & Video demonstration & Video conditioning & Demonstration\\
\addlinespace[2pt]
GLOW~\citep{knowin2026glow} & Single human video & Contextual action generation & Encoded skill context\\
\addlinespace[2pt]
HOST~\citep{chen2026host} & Human video; robot progress & Predict; decode & Demonstration\\
\addlinespace[2pt]
\tablegroup{4}{Physical feedback revises procedures and learning experiments}
ASPIRE~\citep{lu2026aspire} & Task; execution traces & Diagnose; repair & Repair guidance\\
\addlinespace[2pt]
ENPIRE~\citep{xiao2026enpire} & Task; physical trial outcomes & Revise code; train policy & Programs; policies\\
\bottomrule
\end{tabularx}
\par\phantomsection\label{floatend:tab:industrial-teaching}
\end{table}

\surveyanchor{discuss:tab:industrial-teaching}
Sensorimotor and video prompts specify tasks without retraining~\citep{generalist2026gen15,skild2026s1}; HOST instead predicts task evolution before decoding actions~\citep{chen2026host}. GLOW reports watering and box-storage examples that reuse encoded skill context across changed objects or positions~\citep{knowin2026glow}. The qualitative cases illustrate reuse of supplied teaching, while aggregate benchmark scores summarize broader capability. Context interventions isolate the demonstration's contribution (Section~\ref{sec:context-transfer-evaluation}).

Outcome-driven reuse changes what is retained. ASPIRE preserves repair guidance~\citep{lu2026aspire}; ENPIRE can revise a control program or train a policy~\citep{xiao2026enpire}. Prompt preparation, physical trials, and training incur different changeover costs. Their value depends on whether the change requires another procedure, correction of a recurring failure, or a motion outside existing competence. The next chapter evaluates these contributions to task acquisition and reuse.

\begin{takeaway}
\takepoint Applications determine the missing information: an object relation, spatial knowledge, procedural state, or physical response. Each requires teaching or feedback that resolves the corresponding uncertainty.
\takepoint Unclear requirements call for teaching; uncertain execution conditions call for observation or interaction; missing motor competence calls for an expanded executor or additional skill learning. \mbox{These deficits require different remedies.}
\takepoint Task changes create value for new teaching; repeated operations create value for retained procedures. Acquisition and verification costs determine that value, especially when actions are irreversible or time-critical.
\end{takeaway}

\FloatBarrier
\suppressfloats[t]
\section{Evaluation: When Does Context Help?}
\label{sec:evaluation}

Evaluation traces the contribution of context from changed evidence to physical success. Familiar scenes, executor capabilities, and retry opportunities provide the comparison conditions needed to isolate that contribution. Does changed evidence produce the appropriate change in behavior? Does that change survive physical execution and transfer? Does retaining the lesson improve a later attempt enough to justify its acquisition cost? Context interventions, intermediate diagnostics, and retention tests separate these questions. Reported comparisons ground the analysis; proposed controls test the learning mechanisms behind them.

\begin{figure}[tp]
\centering
\includegraphics[width=\linewidth]{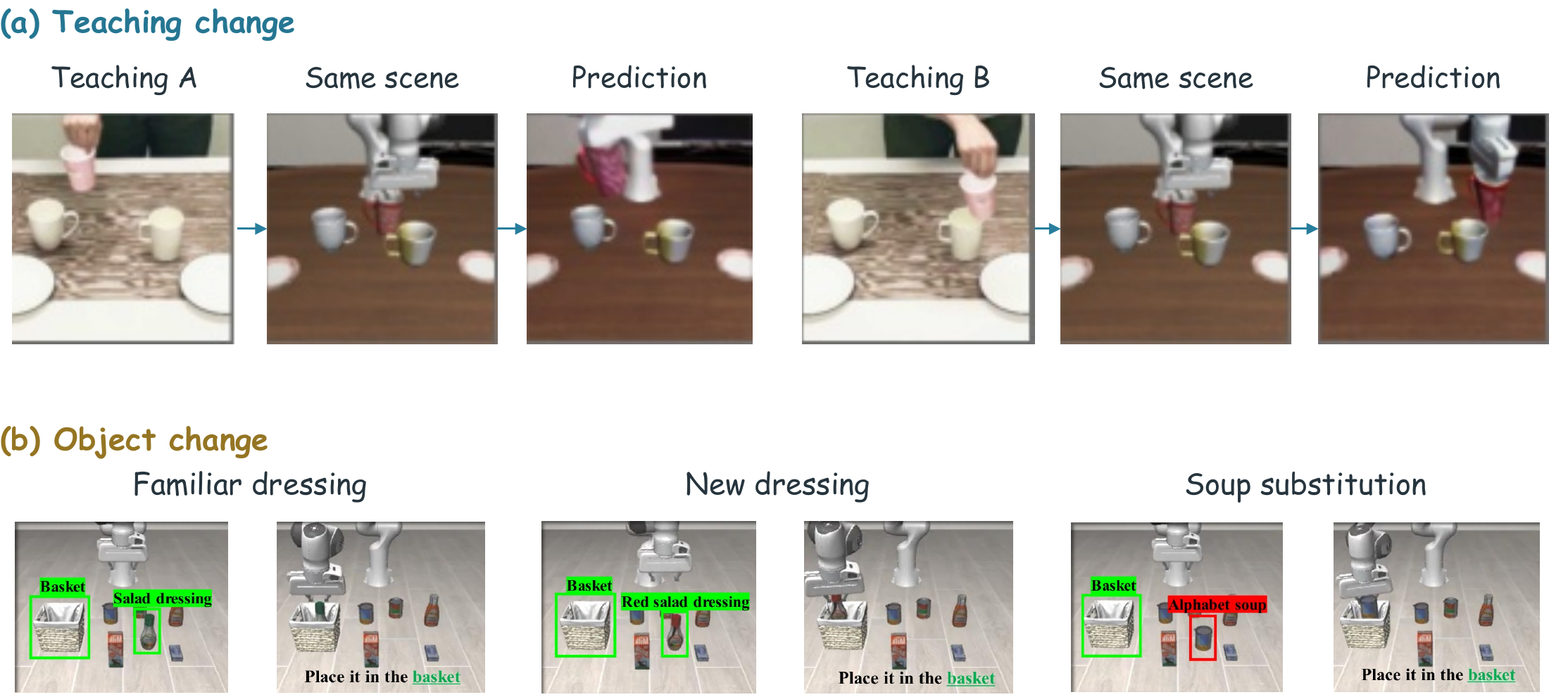}
\caption{Two tests of contextual transfer. (a) Changing the demonstration while holding the robot scene fixed changes UniSkill\textquotesingle{}s predicted continuation~\citep{kim2025uniskill}. (b) LIBERO-PRO changes objects to test target selection~\citep{zhou2025liberopro}. Together, these axes motivate tests that vary teaching and object identity independently, measuring both the inferred requirement and its physical realization.}
\label{fig:evaluation-protocol}
\par\phantomsection\label{floatend:fig:evaluation-protocol}
\end{figure}


\subsection{Context use and transfer across tasks and environments}
\label{sec:context-transfer-evaluation}

Context interventions test sensitivity to teaching; changes in tasks, objects, environments, or bodies locate its transfer boundary. Varying evidence quantity, retention distance, and distractors tests \mbox{access to the required information.}

\paragraph{Sensitivity to task-defining context.}
\begin{table}[tp]
\centering
\surveytable
\renewcommand{\arraystretch}{1.0}
\caption{Proposed evaluation controls and resources that illustrate the corresponding capabilities. Comparisons hold other task conditions fixed; cross-attempt tests separate physical resets from memory resets.}
\label{tab:benchmarks}
\begin{tabularx}{\linewidth}{@{}>{\raggedright\arraybackslash}p{0.20\linewidth}YYY@{}}
\toprule
\textbf{Capability} & \textbf{Intervention} & \textbf{Readout} & \textbf{Example resources} \\
\midrule
Context use & Change teaching & Task-consistent actions & LIBERO-CF~\citep{arxiv260217659}; LMAct~\citep{ruoss2025lmact} \\
\addlinespace[2pt]
Relation transfer & Hold out relations & Relation adherence & VIMA~\citep{jiang2022vima}; MOSAIC~\citep{mandi2021mosaic}; Imitator Game~\citep{zhou2026imitator} \\
\addlinespace[2pt]
Evidence retention & Retain or clear history & Recall; memory updates & RoboMME~\citep{dai2026robomme}; MEMOBench~\citep{sun2026memobench} \\
\addlinespace[2pt]
Cross-attempt reuse & Separate reset types & Gain; interference & RopeFormer~\citep{arxiv260923432}; Zeva-Ego~\citep{arxiv260924411} \\
\addlinespace[2pt]
Physical transfer & Change dynamics & Adaptation; goal fidelity & MemMimic~\citep{gao2026gmp}; PACE-Bench~\citep{arxiv260814441} \\
\addlinespace[2pt]
Human-data transfer & Vary human data; fix robot data & Trained-policy success & H2RBench~\citep{arxiv260924778} \\
\addlinespace[2pt]
Scene robustness & Change scene & Target; outcome & LIBERO~\citep{liu2023libero}; LIBERO-PRO~\citep{zhou2025liberopro} \\
\addlinespace[2pt]
Execution steering & Change command & Redirection; success & ReSteer~\citep{arxiv260317300} \\
\addlinespace[2pt]
Physical execution & Supply target or prerequisites & Target selection; success & RoboSemanticBench~\citep{arxiv260602277}; Behavior-Skill~\citep{arxiv260830536} \\
\addlinespace[2pt]
Route teaching & Change route; retain goal & Path; landmark-order adherence & RPF~\citep{kumar2018rpf}; sparse visual memory~\citep{yoo2020sparsepath} \\
\addlinespace[2pt]
Spatial adaptation & Change preview; reset memory & Success; route efficiency & NOLO~\citep{zhou2024nolo}; ReLIC~\citep{elawady2024relic} \\
\addlinespace[2pt]
Example interpretation & Change examples; retain query & Parsed subtasks; selected moves & A$^2$Nav~\citep{chen2023a2nav}; Select2Plan~\citep{buoso2024select2plan} \\
\addlinespace[2pt]
Outcome reuse & Remove notes; retain retries & Repeated errors; path cost & CMMR-VLN~\citep{li2026cmmrvln}; HAM-VLN~\citep{liu2026hamvln} \\
\addlinespace[2pt]
Traversability & Withhold terrain feedback & Success; physical disturbance & VLM-GroNav~\citep{elnoor2024vlmgronav} \\
\bottomrule
\end{tabularx}
\par\phantomsection\label{floatend:tab:benchmarks}
\end{table}

Demonstrations specifying different feasible behaviors in the same robot scene isolate the influence of teaching. In a kitting task, changing the demonstrated part count should change the selected count; changing the demonstrator while preserving that count should preserve the requirement. Together, meaning-changing and meaning-preserving interventions test what information the learner uses. Success without context provides the prior-only comparison. A fully specified target tests whether the required behavior is executable.

Instruction benchmarks illustrate why these controls matter. LIBERO varies spatial, object, goal, and procedural factors \citep{liu2023libero}, yet task-index embeddings can achieve high performance when the scene reveals task identity \citep{jiang2026benchmarkaudit}. LIBERO-CF supplies alternative feasible instructions within familiar layouts and finds that tested VLAs often continue the scene-associated task \citep{arxiv260217659}. For demonstration ICL, replacing a video must therefore change the intended order or relation, rather than merely perturb its appearance. Robustness tests supply the complementary invariant: preserve the required procedure across viewpoint, object appearance, or instruction rephrasing \citep{zhou2025liberopro,fei2025liberoplus}. RoboStressBench separates material, viewpoint, lighting, and geometry stresses to diagnose upstream visual recognition, reasoning, and planning~\citep{wu2026robostressbench}. VidPair-Halluc instead pairs similar backgrounds with different foreground semantics~\citep{huang2026vidpairhalluc}. Applied to demonstration evaluation, this design would test whether inferred task content follows the demonstrated event rather than scene similarity; a subsequent rollout tests physical realization.

Show-Harness's video-order comparison provides task information left unspecified by the no-video instruction \citep{chen2026showharness}. An equally informative textual order is the relevant control for a claim about video's particular advantage. Comparing language-only, demonstration-only, and joint context with matched task information separates modality from information content. Holding the instruction fixed while changing the demonstrated convention tests learning from examples; holding the demonstration fixed while changing a correction tests use of feedback. Discovering an unfamiliar action convention proved harder: without supplied conventions, the system succeeds in one of 20 episodes and infers 23.3\% of mappings correctly. Interpreting a demonstrated procedure and discovering an unfamiliar action interface thus require separate evidence.

\surveyanchor{discuss:fig:evaluation-protocol}
Sensitivity to teaching and robustness of execution require different observations. A changed predicted continuation can expose a changed task interpretation; a rollout establishes whether the robot realizes it. UniSkill's decoded futures and LIBERO-PRO's object substitutions illustrate these complementary views (Figure~\ref{fig:evaluation-protocol}). The former diagnoses an inferred skill within skill- and agent-based execution; the latter tests target selection and physical behavior. Evaluation must relate both to a controlled change \mbox{in the supplied evidence.}

\surveyanchor{discuss:tab:benchmarks}
Table~\ref{tab:benchmarks} pairs evaluation questions with resources that make the relevant variation possible. The rows concern different dependencies, so their metrics are interpreted within the stated task setting. In particular, the human-data row measures transfer into trained policy weights, whereas contextual tests intervene on teaching or retained evidence at deployment. This distinction connects evaluation to the data \mbox{roles in Figure~\ref{fig:data-roles}.}

\paragraph{Object substitution and relation transfer.}
Task novelty has to be defined at the level of the relation being learned. Task suites make adaptation to held-out tasks measurable~\citep{james2019rlbench,yu2019metaworld}, but withheld names can combine several kinds of change. New object combinations and task templates probe different generalization in VIMA~\citep{jiang2022vima}; the Imitator Game separates changes in demonstrated motion from changes in functional intent~\citep{zhou2026imitator}. A familiar procedure on a new body tests realization, whereas a new ordering of available actions tests contextual composition. A useful transfer claim identifies both the unseen relation and the motor competence available to realize it.

Object substitution supplies a complementary test to changing the demonstrated requirement. Following Table~\ref{tab:object-transfer}, retain the requirement while replacing only the manipulated object, only the receiving object, and then both. Within each condition, compare the correct demonstration, no demonstration, and a demonstration specifying another feasible behavior in the same scene. Comparing object conditions measures substitution scope; comparing demonstration conditions measures dependence on teaching. To isolate the demonstration\textquotesingle{}s contribution, keep the instruction fixed or hold its task information constant. Report task interpretation and physical completion separately, verifying the latter through robot execution.

The choice of substitutions determines which transfer capability is tested. FUNCTO distinguishes instance and category changes~\citep{tang2025functo}, whereas object-pair transfer exposes changes at both ends of a relation~\citep{thompson2026parttransfer}. Functionally compatible substitutes with different appearances, paired with visually similar but incompatible distractors, can test whether correspondence follows the taught interaction. Training exposure, demonstration count, human correspondence or segmentation assistance, neural updates, and retry opportunities affect the interpretation of this comparison. Undisclosed pretraining leaves training novelty uncertain even when \mbox{context dependence is established.}

Withholding a relation while preserving familiar objects and executable actions isolates relation novelty; substituting objects while retaining the relation tests its physical transfer. Cross-robot imitation and paired human-video interfaces then extend the change to embodiment and observation~\citep{mandi2021mosaic,gu2026roboreel}. Whether the demonstration remains available is a further axis: executing with a supplied reference and retaining its lesson after that reference is removed are evaluated separately in Section~\ref{sec:agentic-evaluation}.

Capture conditions also affect the transfer axis. RoboReel links 2,000 human recordings from four views to simulated trajectories in ten environments, separating clean and distracted recordings from evaluation-only distraction and pose-defined tasks \citep{gu2026roboreel}. These variations test whether the same demonstrated requirement survives a changed observation interface. Reporting the neural update rule alongside these variations identifies whether transfer uses context or task-specific policy fitting.

The real-to-sim benchmark H2RBench tests how human demonstrations support policy training~\citep{arxiv260924778}. It reconstructs four real manipulation tasks in simulation, varies human-video quantity at a fixed robot-demonstration budget, and compares relative simulation and real-robot performance. Its shared training protocol tests human-data transfer; frozen-policy context interventions test whether a newly supplied example changes execution. Both matter when a system combines human-video pretraining with contextual adaptation.

\paragraph{Navigation across goals and environments.}
The four context types in Table~\ref{tab:navigation-context} require different interventions. Changing a route demonstration while keeping its destination fixed tests adherence to the taught path or landmark order. Replacing a scene preview and clearing memory between goals test spatial evidence and within-environment reuse, respectively; NOLO's context-removal experiment~\citep{zhou2024nolo} and ReLIC's cross-episode attention comparison~\citep{elawady2024relic} examine these contributions. Changing instruction--decision examples while retaining the current query tests the demonstrated interpretation. Removing failure notes while retaining backtracking tests whether a remembered outcome changes later decisions, as in HAM-VLN's reflection ablation~\citep{liu2026hamvln}. A new building then tests acquisition after environment-specific locations cease to apply.

Success weighted by path length (SPL) combines reaching the goal with route efficiency~\citep{anderson2018naveval}; a prescribed route additionally requires an adherence measure, since a shortcut can reach the same destination. Success across successive goals measures environment reuse, while repeated wrong turns and stopping errors expose the contribution of feedback. Goal visibility, stopping distance, action discretization, and access to maps or localization define the comparison. Terrain adaptation adds physical cost: VLM-GroNav~\citep{elnoor2024vlmgronav} reports trajectory length and IMU energy density alongside success. Exploration and retries spent acquiring context must be included when relating these gains to total interaction.

Simulation, physical navigation, and recorded-scene decisions provide different evidence about execution. NOLO's unseen-layout transfer is evaluated in simulation; its real maze uses separate training and context recordings from that maze. Select2Plan~\citep{buoso2024select2plan} combines simulated object navigation with waypoint assessment on recorded rover scenes, whereas LM-Nav~\citep{shah2022lmnav} executes routes on a physical mobile robot. HAM-VLN's reported comparisons use 100-episode subsets of unseen simulation splits. Together, these settings span simulated transfer, recorded-scene decisions, and physical route execution.

Instruction adherence needs alternatives that the scene alone cannot resolve. RoboFollow pairs shared scenes with several feasible behaviors and separates stage-level intent from execution scores~\citep{arxiv260925636}. Its layout and semantic recombination tests distinguish choosing the requested object or procedure from physically completing it. This provides an operational counterpart to the context interventions above: if only one behavior is plausible in a scene, high success can conceal weak use of teaching. RoboFollow tests instruction grounding. Testing unfamiliar rule learning requires a held-out rule family and examples \mbox{that distinguish its candidate rules.}

\paragraph{Evidence quantity, retention, and distractors.}
The amount of teaching evidence, retention distance, and distractor load test different properties. LMAct varies the number of expert episodes and includes replay controls, showing that more examples often provide little benefit to the tested models \citep{ruoss2025lmact}. A flat curve can reflect early saturation, weak context use, or inadequate control. Comparing an ambiguous task with a fully specified version helps distinguish these causes. Memory benchmarks make the missing evidence more explicit through object, spatial, sequential, or imitation demands \citep{cherepanov2025mikasa,dai2026robomme}. The necessary history depends on the hidden information: AMAGO reports a MetaWorld setting whose task can be \mbox{identified from short history \citep{grigsby2023amago}.}

Checkpoint-based evaluation localizes memory failures. MEMOBench distinguishes Storage, Update, and Compression demands across 30 history-dependent tasks \citep{sun2026memobench}. These test whether decisive events survive retention, revision, and compression. Supplying a missing decisive record directly at the policy input, in its expected format, can help distinguish storage or retrieval failure from failure to use available evidence. Intermediate interventions then test physical realization (Section~\ref{sec:physical-evaluation}). WorldLines extends state validity to household traces: Memory QA and a downstream planning probe test whether remembered state changes support current answers and action preconditions~\citep{arxiv260618847}. EvoEmpirBench isolates dynamic spatial reasoning in locally observable mazes and match-2 games; Agent-ExpVer distills and revises experience across tasks~\citep{arxiv250912718}. These settings separate state recall and revision from physical execution, complementing robot rollouts with more targeted memory tests.

\subsection{Sources of improvement and comparable evidence}
\label{sec:agentic-evaluation}
\suppressfloats[t]

Fourteen reported comparisons (Table~\ref{tab:reported-evidence}) connect context use, physical transfer, and later learning to observable outcomes. Their findings motivate controls that attribute gains to supplied evidence, parameter updates, or retained artifacts.

\begin{table}[!b]
\centering\surveytable
\caption{Reported comparisons of context use and transfer. $\to$ denotes baseline-to-variant change; $\uparrow$/$\downarrow$ denote increase/decrease; $>$ denotes ordering. SR: success rate (\%); pp: percentage points; fractions: successes/total; qual.: qualitative; BC: behavior cloning. Comparisons are within study; weights stay fixed during deployment (LMPC trains successors between sessions). Full protocols and source locations: Supplement~S1.}
\label{tab:reported-evidence}
\begin{tabularx}{\linewidth}{@{}>{\raggedright\arraybackslash}p{.14\linewidth}>{\raggedright\arraybackslash}p{.13\linewidth}>{\raggedright\arraybackslash}p{.23\linewidth}>{\raggedright\arraybackslash}p{.24\linewidth}Y@{}}
\toprule
\textbf{Study} & \textbf{Target} & \textbf{Contrast} & \textbf{Readout} & \textbf{Setting}\\
\midrule
\tablegroup{5}{I. Training and task evidence}
ICRT~\citep{fu2024icrt} & Training data & DROID-only / multi-task & DROID-only: no progress & Shared test tasks\\
\addlinespace[3pt]
BPP~\citep{patel2026bpp} & Prompt & Goal image $\to$ demo & Fidelity $\uparrow$ (qual.) & Unseen drawings; matched start\\
\addlinespace[3pt]
LMAct~\citep{ruoss2025lmact} & Context size & $0\to512$ episodes & Often little gain & Games; simulated control\\
\addlinespace[3pt]
\mbox{Show-Harness} \citep{chen2026showharness} & API semantics & Arbitrary names; no conventions & $1/20$ successes & Fixed harness\\
\addlinespace[3pt]
NOLO~\citep{zhou2024nolo} & Scene context & No video $\to$ video & SR: $33.58\to43.65$ & Habitat; 3 seeds\\
\tablegroup{5}{II. Transfer and physical realization}
MT3~\citep{dreczkowski2025mt3} & Action reuse & BC $\to$ retrieval;\newline whole $\to$ phases & SR $\uparrow$; demos $\downarrow$ & New instances; align + interact\\
\addlinespace[3pt]
Part-based transfer~\citep{thompson2026parttransfer} & Warp granularity & Whole object $\to$ parts & $11/27\to23/27$ & Mug--rack placement; assisted parts\\
\addlinespace[3pt]
Demo-JEPA~\citep{he2026demojepa} & Action interface & Action head / planning & Known, real: head $>$ plan;
unseen: plan $>$ head & Sim.\ + real; task-group means\\
\addlinespace[3pt]
RAPID~\citep{arxiv260930249} & Verification & Scene variants: off $\to$ on & SR: $53.2\to75.9$ & 8 tasks; 50 scenes/task; 3 runs\\
\tablegroup{5}{III. Retained experience and subsequent learning}
Zeva~\citep{chen2026zeva} & Persistent memory & Off $\to$ on & SR: $+10$--$20$ pp & 5 real tasks; brief trace retained\\
\addlinespace[3pt]
RTCF~\citep{fan2026rtcf} & Memory correction & Off $\to$ on & SR: $86.4\to88.4$ & Fixed bank; 2,000 episodes/variant\\
\addlinespace[3pt]
TraceFlow~\citep{arxiv260920646} & Trace guidance & Base $\to$ guided & Ordered: $21/50\to39/50$;
sim.\ aggregate: unchanged & 1 real packing task; simulation suite\\
\addlinespace[3pt]
FARE~\citep{arxiv260918016} & History revision & Base $\to$ selective;
always-search ablation & SR: $91.5\to93.2$;
always-search: $91.0$ & RoboTwin Easy; fixed checkpoint\\
\addlinespace[3pt]
LMPC~\citep{liang2024lmpc} & Successor training & Base $\to$ successor & SR: $39.4\to66.3$ & Held-out teaching tasks\\
\bottomrule
\end{tabularx}
\par\phantomsection\label{floatend:tab:reported-evidence}
\end{table}

\paragraph{What reported comparisons establish.}
\surveyanchor{discuss:tab:reported-evidence}
The first group shows why evidence quantity and usefulness must be separated. ICRT's training-data comparison links context use to task alternatives within a scene; LMAct shows that adding expert episodes alone often yields little benefit~\citep{fu2024icrt,ruoss2025lmact}. BPP's drawing comparison favors demonstrations that specify a trajectory, whereas Show-Harness's action-convention comparison exposes information needed to ground unfamiliar commands~\citep{patel2026bpp,chen2026showharness}. NOLO\textquotesingle{}s context-removal comparison extends this analysis to scene previews~\citep{zhou2024nolo}. Together, these results connect improvement to the \mbox{information supplied by context.}

The second group connects transfer to the execution interface. MT3's low-demonstration advantage relies on alignment followed by interaction replay~\citep{dreczkowski2025mt3}; part-wise warping improves the tested transfers with supplied part correspondence~\citep{thompson2026parttransfer}. Demo-JEPA's comparison with an action head shows that the preferred realization can change with task familiarity~\citep{he2026demojepa}. RAPID's scene-variant ablation adds a complementary question: whether the acquired program is verified beyond the demonstrated configuration~\citep{arxiv260930249}. Transfer therefore depends jointly on the represented relation, correspondence assistance, motor competence, and conditions used to validate the result.

The third group distinguishes using retained experience from acquiring a better learner. Zeva tests access to earlier interaction; RTCF and TraceFlow test action correction from successful or failed traces; FARE tests selective revision of the history conditioning prediction~\citep{chen2026zeva,fan2026rtcf,arxiv260920646,arxiv260918016}. Their scope differs: TraceFlow's ordering gains coexist with unresolved simulation deficits, while FARE's always-search variant underperforms its base policy. LMPC instead trains a successor between sessions and measures its response to new teaching~\citep{liang2024lmpc}. Cumulative success over retries measures yet another quantity, as in Zeva's evolution milestones. These comparisons locate gains in a specific use of experience; the interventions below test whether that use explains the outcome.

\paragraph{Controlled attribution.}
Table~\ref{tab:adaptation-comparison} separates interventions on the training checkpoint, deployment evidence, adaptive parameters, and retained artifacts. Evidence-acquisition controls additionally test how \mbox{informative observations are obtained.}

\begin{table}[!t]
\centering\surveytable
\caption{Attribution controls expressed as target--intervention--readout. $\leftrightarrow$ denotes paired test conditions; $\times$ crosses checkpoints with context conditions. Match tasks, physical resets, base execution resources, and attempt budgets. Measure both behavioral change and task completion.}
\label{tab:adaptation-comparison}
\begin{tabularx}{\linewidth}{@{}>{\raggedright\arraybackslash}p{.24\linewidth}>{\raggedright\arraybackslash}p{.40\linewidth}Y@{}}
\toprule
\textbf{Target} & \textbf{Intervention} & \textbf{Readout}\\
\midrule
Training checkpoint & Checkpoint $\times$ context & Learned context use\\
\addlinespace[3pt]
Task teaching & Original $\leftrightarrow$ replacement & Requirement adherence\\
\addlinespace[3pt]
Retained history & Retain $\leftrightarrow$ clear & Use of prior evidence\\
\addlinespace[3pt]
Neural parameters & Adapt $\leftrightarrow$ restore & Adaptation gain\\
\addlinespace[3pt]
Reusable guidance & Retain $\leftrightarrow$ withhold & Lesson transfer\\
\addlinespace[3pt]
Skill library & Expanded $\leftrightarrow$ base & Use of new skills\\
\addlinespace[3pt]
Active observation & Enabled $\leftrightarrow$ disabled & Acquisition benefit\\
\bottomrule
\end{tabularx}
\par\phantomsection\label{floatend:tab:adaptation-comparison}
\end{table}

\surveyanchor{discuss:tab:adaptation-comparison}
The first control compares checkpoints under the same context interventions; the remaining controls change what is available during execution. Repeated-task and transfer trials then establish whether the identified dependence improves subsequent behavior.

\paragraph{Acquiring the ability to use context.}
Contextual training is useful when it improves the model's response to task-defining evidence, beyond improving execution generally. Explicitly paired examples offer one route~\citep{duan2017oneshot,mandi2021mosaic}; multimodal prompts, sensorimotor streams, and corrective histories expose other dependencies reviewed in Section~\ref{sec:context-training}~\citep{jiang2022vima,fu2024icrt,jiang2026robottt}. Each checkpoint can be evaluated on the same tasks with informative context, absent context, and context specifying a different feasible behavior. The within-checkpoint comparison measures the effect of evidence; comparing that effect across checkpoints tests whether training improved its use. Performance with a fully specified task separately checks the available execution competence.

\paragraph{Using new teaching and retained observations.}

Supplied teaching and retained history require different interventions. Changing an example with the checkpoint, scene, and executor fixed tests whether its task requirement affects behavior; video-conditioned policies and predictive interfaces provide concrete settings for this comparison~\citep{jain2024vid2robot,zhou2026zerowam}. Clearing history at a matched state instead tests access to earlier evidence. Dynamics-revealing transitions~\citep{kumar2021rma,arxiv260923432} and remembered task events~\citep{shi2025memoryvla,torne2026mem} resolve different unknowns. Memory-focused settings distinguish these requirements~\citep{dai2026robomme,sun2026memobench}; evaluating the action selected under each history tests whether the recalled evidence is used appropriately. Counterfactual Memory Audit compares alternative histories that yield identical current inputs~\citep{zhang2026cma}. It controls policy-sampling randomness and tests the resulting actions under both histories. This distinguishes an action change caused by memory from a justified choice and its physical benefit.

\paragraph{Adapting neural parameters.}
Parameter adaptation requires a matched-input comparison and a distinct reset control. Meta-imitation changes an initialization through a demonstration loss~\citep{finn2017mil,yu2018domainadaptive}; fast-weight approaches restrict the updated subset and its timing~\citep{jiang2026robottt,arxiv260706988}. With demonstrations and test instances fixed, restoring pre-adaptation weights separately from removing context identifies the contribution of weight changes. Optimization and interaction costs characterize the resources required. This comparison isolates an optimization-based contribution; fixed-parameter ICL holds deployed neural weights \mbox{unchanged in both conditions.}

\paragraph{Reusing guidance and executable skills.}
Artifact ablations distinguish a retained lesson from an expanded execution repertoire. Withholding guidance tests whether acquired knowledge improves fresh inference, as motivated by correction memory and repair guidance~\citep{zha2023droc,lu2026aspire}. Restoring the pre-acquisition library instead tests newly acquired routines while preserving the original skills~\citep{tziafas2024lrll,arxiv260619419}. A geometric archive can be treated similarly~\citep{dreczkowski2025mt3}. Fresh task instances, the base executor, and attempt budgets are held fixed so that the contrast concerns reuse rather than additional opportunities to solve the task. Table~\ref{tab:agentic-mechanisms} identifies the artifact associated with each branch.

\paragraph{Acquiring new evidence.}
Active sensing changes the evidence available to the learner, complementing interventions that hold the evidence fixed. VA-Bench gives MLLM agents RGB demonstrations, camera control, metric target commands, and execution feedback through a fixed controller~\citep{arxiv260919554}. Its active-versus-passive viewpoint comparison isolates evidence acquisition, while held-out geometry and long-horizon composition test whether that evidence supports transferable execution. These interventions distinguish obtaining a better observation from interpreting the same observation more effectively.

\begin{figure}[!tp]
\centering
\includegraphics[width=\linewidth]{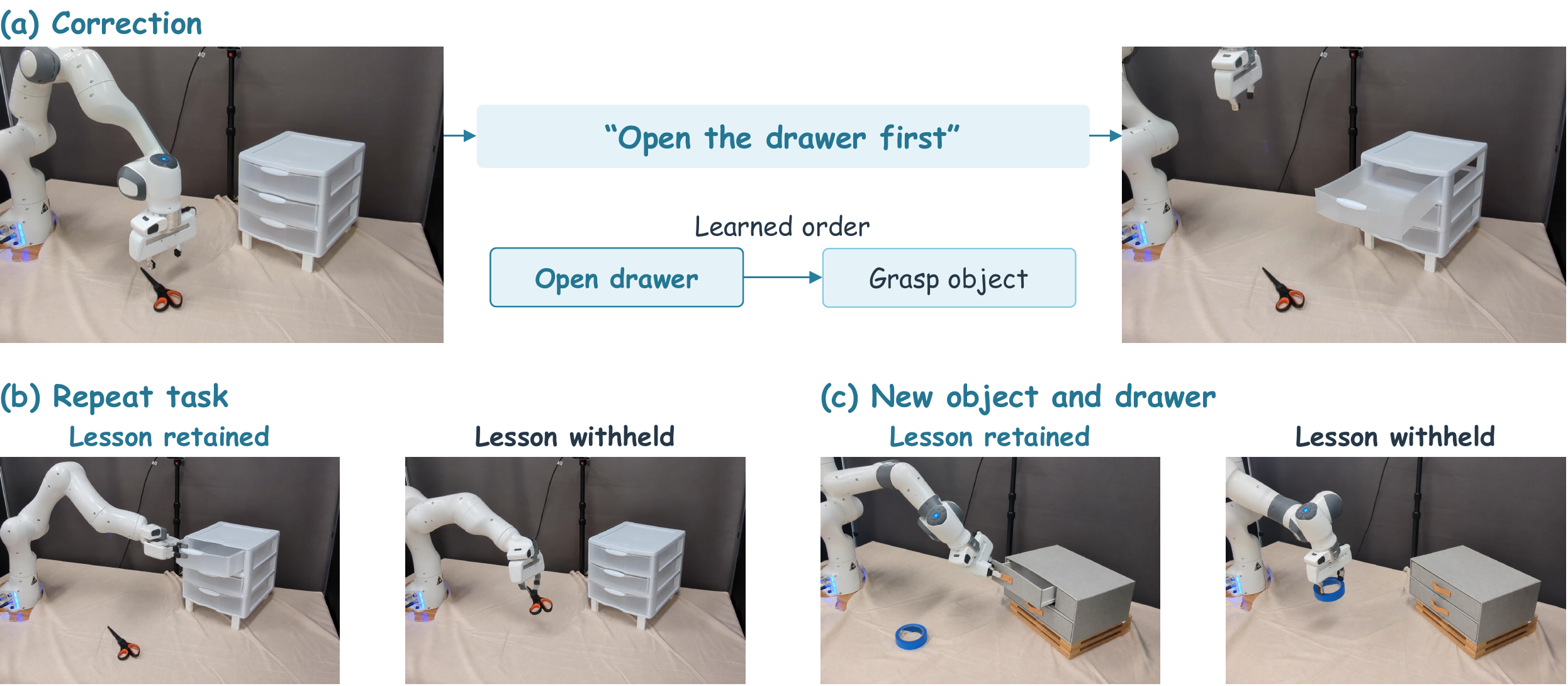}
\caption{Testing the use of a learned ordering constraint. (a) Feedback establishes that the drawer must be opened before grasping the object. (b,c) Conceptual illustrations contrast following that order with premature grasping on repeated and new instances. Panel (a) follows the DROC example~\citep{zha2023droc}; panels (b,c) propose candidate outcomes for retention tests. Success and further human corrections provide the readouts.}
\label{fig:adaptation-sources}
\par\phantomsection\label{floatend:fig:adaptation-sources}
\end{figure}

\paragraph{Matched retention and reuse tests.}
\surveyanchor{discuss:fig:adaptation-sources}
Figure~\ref{fig:adaptation-sources} separates acquisition, retention, and transfer through a drawer-placement example. In the recorded episode, the robot initially reaches for scissors; human feedback establishes that it must open the drawer first~\citep{zha2023droc}. Panel (a) identifies the acquired ordering constraint. Panels (b,c) propose comparing access to that constraint with its removal: repeating the task tests retention, while replacing the object and drawer tests transfer. The illustrated continuations make the predicted effect observable before completion---opening first versus premature grasping. Repeated trials measuring success and further human corrections would establish whether that difference improves execution.

The retention comparison holds task instructions, physical initialization, base competence, and resource budgets fixed; access to the selected lesson is the changed factor. Independent scene resets remove physical progress, and matched retry budgets control additional opportunities for success. Selective resets identify dependence conditional on the components left intact; redundant records can preserve the same information after one carrier is removed. Training exposure remains a separate axis when pretraining data are undisclosed. Figure~\ref{fig:adaptation-state} identifies the possible carriers and timescales, while Table~\ref{tab:adaptation-comparison} specifies interventions on each. This links a particular acquired fact to its availability and subsequent behavioral effect.

Acquiring an artifact consumes experience before it can save experience. Separating play from downstream testing and comparing against additional retries, as in RATs, exposes this trade-off~\citep{arxiv260619419}. Separating synthesis feedback from fresh validation serves the same purpose for program and oversight revision~\citep{kumar2026aor,arxiv260816590}. A successful acquisition trial establishes that a candidate worked once; a retention benefit on later instances establishes useful reuse. Relating that gain to the trials and corrections spent acquiring it determines whether the retained knowledge reduces total learning effort.

Adherence to teaching, physical success, and acquisition cost characterize reuse within a shared task setting. Success at a fixed attempt budget or attempts to a common criterion provide comparable outcomes, including unsuccessful trials; repeated matched trials quantify variation. The next section traces these gains through the stages of physical execution.

\subsection{Physical realization and learning across attempts}
\label{sec:physical-evaluation}

Channel and intermediate interventions trace contextual information into physical outcomes. Independent physical and memory resets then test whether retained evidence improves later attempts and recovery.

Channel interventions complement memory resets. With the task, checkpoint, and recorded history matched, removing language, visual, proprioceptive, or contact inputs tests which evidence supports adaptation. The comparison must retain the policy's valid input format and distinguish a missing measurement from a measured zero. Goal completion and demonstration adherence should also be reported separately when contact is specified: a handle-grasp requirement calls for both successful transport and \mbox{preservation of handle contact.}

Intermediate interventions localize failure along the four control interfaces. For a contextual policy, supply the relevant demonstration segment; for geometric transfer, supply the correct correspondence and reference; for predictive control, supply a task-consistent future; for an agent, supply the correct executable specification. Each intervention must preserve the controller's expected format. Recovery after the substitution locates a bottleneck in the replaced stage or its input. Failure in both conditions calls for tests of downstream execution and interface compatibility. The contrast is most informative at the earliest decision where inferred and required behavior diverge. Target-selection assessment and restored-precondition tests provide related diagnostics~\citep{arxiv260602277,arxiv260830536}. Both conditions require measuring prescribed paths, order, or contacts alongside the terminal state.

Physical adaptation requires evidence about the changed condition. MemMimic distinguishes event recall from adaptation to initially unknown physical response, while PACE-Bench varies physical conditions with a stable goal and interface \citep{gao2026gmp,arxiv260814441}. For synthesized teaching videos, the video-generation benchmark H2R-Bench evaluates goal, event, contact, and embodiment consistency before robot execution \citep{arxiv260813049}. These tests are useful when tied to the specific information that the contextual policy must recover. Probe-then-commit audits further separate whether a belief covers the action-relevant truth from whether the executor predicts failure accurately~\citep{arxiv260930608}. Their offline ground-truth checks show why sharper beliefs alone need not justify physical commitment.

Independent physical and memory resets separate the sources of improvement across attempts. Resetting the scene removes task progress, whereas resetting memory removes acquired evidence. Retaining each separately distinguishes physical continuation, contextual adaptation, and additional chances from repeated execution.

For trial-memory adaptation, the reuse gain at attempt $n$ is $\Delta_n^{\mathrm{reuse}}=S_n^{\mathrm{keep}}-S_n^{\mathrm{reset}}$. Here $S_n$ is the same predefined success or adherence rate at attempt $n$; the superscripts $\mathrm{keep}$ and $\mathrm{reset}$ specify retained and cleared memory. Their difference $\Delta_n^{\mathrm{reuse}}$ measures the benefit of retaining evidence from earlier attempts. The comparison holds task distribution, physical initialization, executor, other retained artifacts, earlier attempt count, and execution budget constant; only access to evidence from those attempts changes. Per-attempt rates isolate the benefit of retained evidence, whereas cumulative success also benefits from independent retries. A subsequent tool or dynamics change tests whether the learner revises a lesson whose physical conditions have become obsolete.

Recovery evaluation connects the detected failure to the acquired evidence, the revised decision, and the outcome of that revision. End-to-end success includes all attempted tasks; conditional repair success concerns the subset of detected, eligible failures. Distinguishing autonomous repair, human correction, and external reset makes their respective contributions visible. Later attempts establish whether the correction remains useful, while newly taught valid procedures test whether the failure detector follows the current task reference. Section~\ref{sec:recovery} describes the corresponding mechanisms.

RoboRecover makes the starting deviation reproducible by replaying action prefixes, then evaluating continuation under the original goal~\citep{arxiv260928952}. Its 2,000 scenarios distinguish the stage to resume from the deviation to correct, across RoboTwin and LIBERO. Initializing candidate policy history at the recovery point isolates continuation from the reconstructed off-nominal state. A contextual-recovery extension should compare retained and cleared prefixes at that same reconstructed state, preserving the continuation budget. This connects reproducible physical deviations to the selective memory resets above.

\subsection{Teaching effort, response delay, and reuse cost}
\label{sec:changeover-cost}

Contextual adaptation has three operational timescales: acquiring the task, responding to fresh evidence, and reusing the result. Task acquisition includes instruction preparation, demonstration recording, exploration, and validation. Extra search can improve execution with unchanged task information; varying relevant context independently of search budget identifies the contribution of learning \citep{arxiv250617811,arxiv260816885}. SAIL's reconstruction and simulated trials illustrate preparation costs beyond model inference \citep{sato2026sail}.

Acquisition is measured end to end by \emph{changeover time}: the interval from receiving a specification to satisfying a stated quality and operating criterion. A common criterion permits comparison of contextual teaching, fine-tuning, trajectory reuse, and programming. Elapsed time, human labor, robot interaction, and resets capture different costs: ENPIRE's program refinement requires physical trials and verification even with fixed neural weights \citep{xiao2026enpire}. Repeatable task performance therefore provides a more informative endpoint for changeover than a first successful attempt.

During execution, the relevant delay runs from informative feedback to the affected command. Model inference time, command rate, and servo frequency describe different parts of this interval. Cached demonstration features can be reused, while a new contact observation must reach control before the corresponding motion is committed. SimpleMemVLA reports latency falling from 1.02 to 0.68 seconds with identical outputs through scheduling and shared-prefix processing \citep{yin2026simplememvla}. This scheduling gain reduces response delay while preserving the decision itself. State-dependent steering tests such as ReSteer make the arrival point \mbox{of feedback explicit \citep{arxiv260317300}.}

Repeated batches test whether acquisition costs are repaid. Accepted output and intervention time across repeated tasks quantify this return; changing a tool or requirement then reveals the cost of obsolete knowledge. These outcomes motivate the next chapter: broader contextual learning must expand what an episode can teach while preserving the ability to revise what no longer applies.

\begin{takeaway}
\takepoint Context interventions establish dependence on task-defining evidence beyond familiar-scene recognition. Physical execution and held-out relation tests then establish how far that requirement transfers.
\takepoint Common task outcomes expose the contributions of training, supplied context, adaptive neural state, and retained knowledge. Acquisition and reuse carry different resource costs.
\takepoint Teaching effort, feedback-to-action delay, and cost per accepted execution connect contextual learning to deployment value across repeated attempts.
\end{takeaway}

\FloatBarrier
\suppressfloats[t]
\section{Future Directions: What Limits Transfer and Improvement?}
\label{sec:discussion}
\label{sec:frontier}
\label{sec:outlook}

\begin{figure}[t]
\centering
\includegraphics[width=\linewidth]{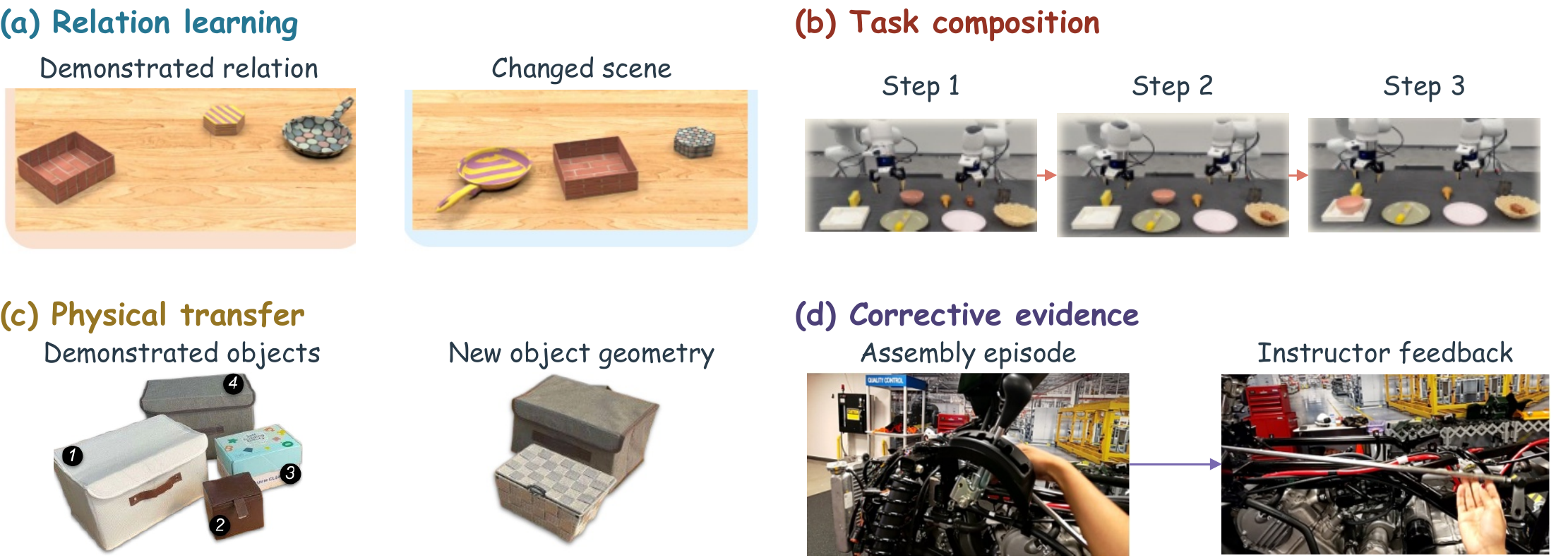}
\caption{Four research directions: (a) infer taught relations, (b) compose task steps, (c) preserve function across physical changes, and (d) use corrective feedback. Together they extend contextual learning toward interpreting, transferring, and retaining task requirements.}
\label{fig:learning-roadmap}
\par\phantomsection\label{floatend:fig:learning-roadmap}
\end{figure}

Broader contextual learning has four connected objectives: infer unfamiliar rules, preserve binding physical constraints through abstraction, revise retained corrections when conditions change, and improve how subsequent tasks are learned. The directions below connect these objectives to training, representation, and experience acquisition, with behavioral tests that distinguish progress on each.


\subsection{From context dependence to a reusable learning rule}
\label{sec:reusable-learning-rule}

The learning distinctions introduced in Section~\ref{sec:learning-horizons} separate familiar-task selection, composition, and inference of an unfamiliar rule. Building on the learning mechanisms in Section~\ref{sec:context-learning-mechanisms}, we ask which training relationships would extend this behavior to unfamiliar requirements.

\begin{table}[H]
\centering\surveytable
\caption{Proposed tests of behavioral transfer in a placement task. Keep available objects and motor actions matched; vary what teaching must supply. Novelty is relative to the controlled training exposure.}
\label{tab:transfer-levels}
\begin{tabularx}{\linewidth}{@{}>{\raggedright\arraybackslash}p{.23\linewidth}>{\raggedright\arraybackslash}p{.34\linewidth}Y@{}}
\toprule
\textbf{Learning claim} & \textbf{What is already familiar} & \textbf{What the example must establish}\\
\midrule
Familiar-task selection & Object-to-location assignments & Choose a known left- or right-bin routine\\
\addlinespace[3pt]
Compositional transfer & Placement relations; order constraints & Combine familiar color and order rules\\
\addlinespace[3pt]
Rule inference & Objects; executable placements & Infer an unfamiliar size-to-slot rule\\
\bottomrule
\end{tabularx}
\par\phantomsection\label{floatend:tab:transfer-levels}
\end{table}

\surveyanchor{discuss:tab:transfer-levels}
Table~\ref{tab:transfer-levels} distinguishes selecting a familiar task, composing known relations, and inferring a new rule. An example can identify a routine without teaching a new relation: demonstration benefits under incorrect labels in some NLP settings expose the role of task-format recognition~\citep{min2022demonstrations}. Conversely, function-learning studies show inference of unseen functions, and controlled regression connects task diversity to more general learning behavior~\citep{garg2022functions,raventos2023diversity}. The illustrative robotic tests keep available objects and motor skills fixed while varying what teaching must establish: a familiar bin choice, a new composition of known constraints, or a size-to-slot mapping from a family excluded during training. Rule inference requires enough examples to distinguish the candidate mappings. BPP's continuous drawing tasks and the Imitator Game's functional substitutions provide concrete task variations from which to design these tests~\citep{patel2026bpp,zhou2026imitator}.

The acquisition routes in Section~\ref{sec:icl-acquisition-routes} motivate a testable hypothesis: coverage of task relations and coverage of executable motions make distinct contributions to contextual transfer. Broad pretraining may support unfamiliar requirements, while grounded pairs establish precise action correspondence. Varying pretraining scale and explicit pairing independently, with the executor and task information fixed, would test their contributions. Motion coverage, relation diversity, interaction experience, and model size are distinct factors; agentic systems also expose library size and acquisition cost~\citep{arxiv260817209}. Relation-level outcomes and continuous errors are needed alongside all-steps success, because thresholded metrics can make \mbox{smooth component gains appear abrupt~\citep{schaeffer2023mirage}.}

Cross-modal transfer supplies a complementary test. GPT-Policy provides a concrete interface in which a general VLM receives teaching and acts through fixed robot tools~\citep{cheng2026gptpolicyeval}. Expressing the same unfamiliar rule through text, demonstrations, or both would test whether its inferred meaning transfers across formats. Matching task information and motor capabilities isolates rule inference from access to additional details, such as a procedural step visible in one format but absent from another.

\surveyanchor{discuss:fig:learning-roadmap}
The task examples in Figure~\ref{fig:learning-roadmap} connect these research questions to observable behavior. Panel (a) asks whether a taught relation determines actions in a new scene; (b) extends it to a sequence of dependent steps; (c) changes the contact geometry while preserving the interaction's function; and (d) uses human feedback to revise the next action. The following sections develop the corresponding teaching, transfer, and retention requirements. Section~\ref{sec:icl-self-improvement} then considers whether experience improves how \mbox{subsequent requirements are learned.}

\subsection{Instruction tuning for more complex contextual learning}
\label{sec:embodied-instruction-tuning}
\suppressfloats[t]
\begin{figure}[!htbp]
\centering
\includegraphics[width=\linewidth]{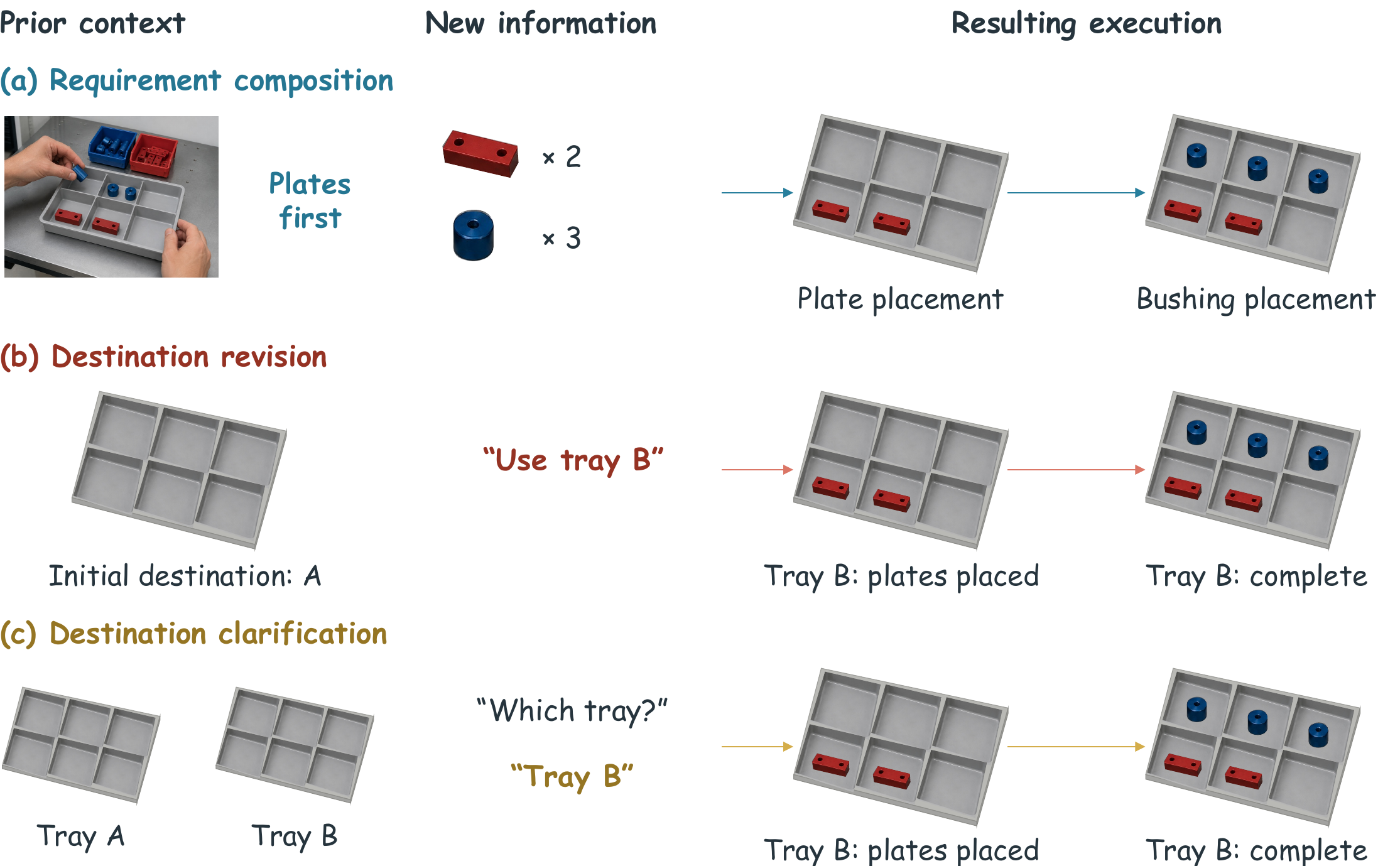}
\caption{Illustrative kitting tasks for three contextual learning problems: (a) combine demonstrated order with requested quantities, (b) change the destination while retaining the order, and (c) resolve an ambiguous destination before acting. Intermediate tray states reveal adherence to teaching beyond the final inventory.}
\label{fig:instruction-training}
\par\phantomsection\label{floatend:fig:instruction-training}
\end{figure}

MetaICL and Flan provide language-model precedents for example-based learning and instruction tuning~\citep{min2022metaicl,chung2022flan}. For ICL for robots, the training problem is to make new teaching select among feasible continuations within a controlled motor repertoire. The three cases below progress from combining requirements to revising them and resolving missing information.

\surveyanchor{discuss:fig:instruction-training}
Figure~\ref{fig:instruction-training} illustrates these cases through separate kitting episodes with comparable parts and motor skills. Paired, continuous, and corrective records (Section~\ref{sec:context-training}) provide possible training formats; the intermediate tray states reveal which requirements survive execution.

\paragraph{Requirement binding and composition.}
Panel (a) combines a demonstrated plate-first order with a request for two plates and three bushings. Reversing the demonstrated order should reverse intermediate tray states while preserving the final inventory. The example separates \emph{compositional complexity}, which varies skill order and dependencies, from \emph{constraint complexity}, which adds jointly binding counts, contacts, or dwell times. Measuring each requirement and their joint satisfaction tests whether a new composition was followed through the method's intermediate representation.

\paragraph{Selective revision of a procedure.}
Panel (b) changes the destination to tray B before placement, retaining the order and quantities. DROC's language guidance, RACER's corrective supervision, and RoboTTT's separation of failed context from corrective targets offer components for learning such revisions~\citep{zha2023droc,dai2024racer,jiang2026robottt}. Training should reward the targeted change together with unaffected requirements. Varying correction timing, scope, and precedence then tests \emph{revision complexity}: reaching tray B while losing the taught count is an overbroad revision. Intervening tasks test whether the revised procedure remains available later.

\paragraph{Decision-directed information acquisition.}
Panel (c) leaves both trays feasible without specifying which to use. Asking for the destination resolves this ambiguity before placement, building on question-based teaching and uncertainty-aware assistance~\citep{cakmak2012questions,ren2023knowno}. The acquired answer must preserve the already taught order and quantity. Other uncertainties call for different interventions: a view or probe can reveal execution conditions (Section~\ref{sec:recovery}), whereas a missing grasp skill requires an additional controller or motor learning. Behavioral Exploration offers a related action-based route: a policy uses interaction history to select expert-like behaviors that explore beyond previous attempts~\citep{wagenmaker2025exploration}. Clarification and exploration target different unknowns; training should connect each request or action to the information it supplies.

These cases motivate comparing action pretraining, demonstration-conditioned training, and mixed instruction--demonstration--correction training under a common backbone and experience budget. Withheld compositions and corrections test requirement adherence, physical completion, and later reuse. The next section examines which requirements should remain invariant when objects and execution conditions change.

\subsection{Preserving taught relations across physical change}
\label{sec:physical-relation-transfer}

Cross-object transfer in Section~\ref{sec:context-mechanisms} establishes useful relations when the corresponding parts and admissible substitutions can be identified. The remaining challenge is to infer which demonstrated details are binding and when no valid substitute exists. A handleless vessel may serve a pouring goal but cannot preserve an explicitly required handle grasp. Similarly, a pause that lets adhesive set is part of the process, whereas an incidental pause may be shortened. These cases require selective invariance: retain the taught requirement while changing only details that its meaning permits. The part-decomposition study supplies cross-category part equivalencies~\citep{thompson2026parttransfer}; learning to infer and validate such equivalencies would extend its transfer scope.

The crossed comparisons in Section~\ref{sec:context-transfer-evaluation} locate these limits by separating changed requirements from changed objects. Beyond successful substitution, a learner must recognize when the available objects cannot satisfy the demonstrated relation and acquire the missing information. Contrasting demonstrations or a clarification can establish whether a pause or contact is required; interaction can reveal whether a geometrically compatible replacement has unfamiliar resistance or dynamics~\citep{liu2025locoformer,chen2026zeva}. Correspondence and physical feedback therefore answer different questions: where an interaction should occur, and whether that interaction \mbox{realizes the intended effect.}

The available executor sets the boundary of this transfer. If the revised requirement can be expressed through existing actions or skills, contextual inference can select a new realization. If the necessary motion lies outside that repertoire, further motor learning contributes a separate stage, as in human-video-initialized WHIRL \citep{bahl2022whirl}. Such a hybrid process can preserve a context-inferred task relation while \mbox{learning its physical realization.}

Navigation introduces a related separation between reusable strategy and local knowledge. An exploration strategy can transfer across buildings, while a remembered object location is tied to one layout. ReLIC~\citep{elawady2024relic} illustrates the value of retaining such local experience across goals. Broader transfer requires identifying which spatial facts remain valid when the environment changes, alongside preserving the \mbox{task relations considered above.}

\subsection{Retaining teaching while revising obsolete experience}
\label{sec:last-mile}
\suppressfloats[t]

\begin{figure}[!htbp]
\centering
\includegraphics[width=\linewidth]{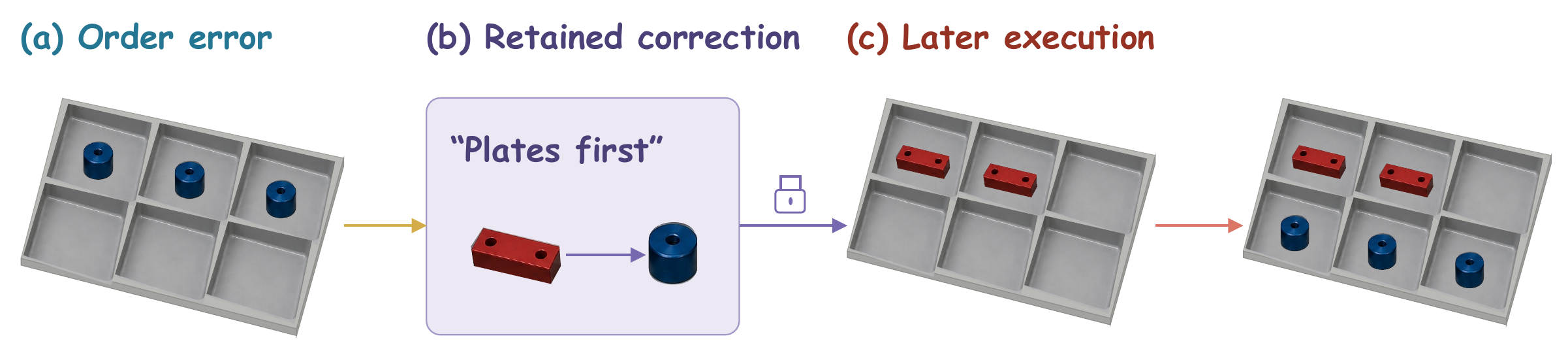}
\caption{Illustrative reuse of corrective context: (a) an order error, (b) plate-first feedback, and (c) execution in a new layout with fixed policy weights. The intermediate tray makes retention of the taught order observable before \mbox{the kit is complete.}}
\label{fig:research-agenda}
\par\phantomsection\label{floatend:fig:research-agenda}
\end{figure}

Physical transfer preserves a requirement across changes in its realization; temporal reuse must also accommodate new teaching. Video outlines and lower-cost planning latents retain different details~\citep{chen2026showharness,huang2026fastthinkact}. A later correction can make an initially incidental contact or pause decisive, after a task-specific summary has discarded it. Delayed corrections can test what new teaching compact summaries and retrievable source segments support at matched storage and access costs. Preserving current task meaning and enabling future reinterpretation impose different requirements on retained evidence.

Consolidation into weights adds training after contextual acquisition. The $\pi_{0.7}$ coaching example and the RECAP training framework for $\pi^*_{0.6}$ connect interaction or correction to subsequent training~\citep{arxiv260415483,arxiv251114759}. Immediate context-driven behavior and persistence after consolidation or reset characterize different stages of acquisition. Retained programs, records, or policy weights can help later tasks, but obsolete corrections can also \mbox{interfere with new teaching.}

\surveyanchor{discuss:fig:research-agenda}
A taught requirement can remain useful after the scene that elicited it has changed. In the kitting example, retained plate-first feedback supplies information absent from the new layout, illustrating how a fixed policy could combine an earlier correction with current geometry (Figure~\ref{fig:research-agenda}). Keeping, omitting, or replacing that correction tests whether the later decision depends on the acquired lesson. Retention must also survive the representation changes used to make such lessons cheaper to store and apply.

Across successive deployments, the criterion also applies to model upgrades and changing environments \citep{run2026nonstationary}. Replacing the interpreter or action interface raises two questions: whether retained teaching still governs behavior, and whether revised requirements can override it. An intervening task can leave the earlier procedure available for later reuse; a new instruction for that procedure can supersede it. Section~\ref{sec:changeover-cost} connects this selective retention to reduced teaching and validation effort.

\subsection{Physical recursive self-improvement}
\label{sec:icl-self-improvement}

Physical feedback can improve the current solution or the process used to acquire future solutions. Retained context, program revision, and neural updates provide different means of doing so, with their learning benefit determined by what the successor acquires from new experience. Improved teachability connects these updates to the choice of experience and the cost of acquiring the next task. In the survey's six-horizon framework, S5 names this subsequent-learning objective and S6 extends it to \mbox{knowledge exchanged between robots.}

\begin{table}[!htbp]
\centering\surveytable
\caption{Feedback-driven improvement through update target, operation, and reusable output. Targets can overlap; successor comparisons measure their contribution to subsequent learning. ENPIRE and REVOLVE denote their program and policy branches, respectively.}
\label{tab:physical-self-improvement}
\begin{tabularx}{\linewidth}{@{}>{\raggedright\arraybackslash}p{.17\linewidth}>{\raggedright\arraybackslash}p{.16\linewidth}>{\raggedright\arraybackslash}p{.22\linewidth}Y@{}}
\toprule
\textbf{Update target} & \textbf{Operation} & \textbf{Reusable output} & \textbf{Examples}\\
\midrule
Interaction context & Retain & Action--effect memory & RopeFormer~\citep{arxiv260923432}; Zeva-Ego~\citep{arxiv260924411}\\
\addlinespace[3pt]
Executable artifacts & Revise & Skills; programs & RoboRSI~\citep{noematrix2026roborsi}; ENPIRE~\citep{xiao2026enpire}; HarnessPAI~\citep{arxiv260929166}\\
\addlinespace[3pt]
Neural components & Fit; retrain & Dynamics; policy & MetaPusher~\citep{arxiv260921122}; REVOLVE~\citep{arxiv260914633}\\
\addlinespace[3pt]
Acquisition process & Refine & Teaching; training procedure & LMPC~\citep{liang2024lmpc}; Eureka~\citep{ma2023eureka}; DrEureka~\citep{ma2024dreureka}\\
\bottomrule
\end{tabularx}
\par\phantomsection\label{floatend:tab:physical-self-improvement}
\end{table}

\surveyanchor{discuss:tab:physical-self-improvement}
Table~\ref{tab:physical-self-improvement} separates where experience is stored from how it changes subsequent learning. RopeFormer and Zeva-Ego use retained action--response history to adapt a fixed policy to encountered conditions~\citep{arxiv260923432,arxiv260924411}. Programs retain executable solutions, and neural updates change predictive or motor competence. Any of these can contribute to physical recursive self-improvement when experience improves acquisition or verification on subsequent tasks~\citep{duan2026rsisurvey}. The decisive comparison is the later learning process, not the storage format of the change.

Program improvement can supply data for subsequent model improvement. HarnessPAI uses validated program rollouts to fine-tune the action model; its published sequence ends at policy refinement, before a further round of harness evolution around the improved checkpoint~\citep{arxiv260929166}. TraceFlow instead grows an external trace bank with weights fixed, and its reported stacking gains peak before the final round~\citep{arxiv260920646}. These examples motivate measuring the successor's acquisition rate on new tasks and the persistence of gains across rounds, beyond demonstrating a stronger current solver.

\paragraph{Learning to become more teachable.}
LMPC links accumulated teaching to improved subsequent acquisition~\citep{liang2024lmpc}. Within a session, a fixed language model converts instructions and corrections into reward code for model-predictive control. Across sessions, successful interactions train it to predict teaching exchanges; the rollout variant searches those exchanges for a shorter route to success. Test-task success rises from 39.4\% to 66.3\%, while mean turns in successful sessions fall from 2.4 to 1.9. This connects experience to a successor's response to new teaching.

For action policies and VLM agents, the corresponding comparison holds new tasks, motor interfaces, and interaction budgets fixed. Evaluating each successor with and without informative context separates improved use of teaching from stronger prior execution. Familiar tasks and fully specified targets measure execution competence; withheld requirements and matched teaching measure acquisition. This extends the controls in Section~\ref{sec:agentic-evaluation} across successive versions of a learner.

\paragraph{Improving execution, supervision, and acquisition.}
Retained guidance and context-building code can improve selection or interpretation of future evidence~\citep{lu2026aspire,wang2026shaper}. Autonomous practice can improve resetting, execution, or a successor policy~\citep{sharma2023medal,bousmalis2023robocat}. RoboRSI's successive skill revisions exemplify repertoire growth; its release-level cumulative coverage records tasks solved at least once~\citep{noematrix2026roborsi}. REVOLVE updates two distinct components: failure and correction records retrain the manipulation policy, while mismatch memory refines the supervising agent's judgments~\citep{arxiv260914633}. Selective resets and successor comparisons attribute gains to the retained solution, control, or supervision.

Predictive components offer further update locations. Q-Planning trains a value function on deployment experience while retaining the action policy~\citep{arxiv260821204}; SILVR trains a visual planner on its executions~\citep{luo2025silvr}; joint refinement changes both policies and the models that assess or train them~\citep{arxiv260212063,arxiv260206508,bi2026motus2}. These loops improve the control and training signals available to contextual learning. Their contribution to teachability is measured by how the resulting successor responds to new evidence, alongside its execution success.

Changing the acquisition procedure makes the recursive objective explicit. Eureka revises reward code from policy-training feedback; DrEureka also configures domain randomization for transfer~\citep{ma2023eureka,ma2024dreureka}. ENPIRE extends agent-directed experimentation to physical environments, revising programs, data use, and learning algorithms through established interfaces~\citep{xiao2026enpire}. Retaining effective learning choices together with their applicable conditions offers a route to improving future acquisition. Successor comparisons measure this benefit alongside repertoire growth. Evidence from simulated practice and physical experiments characterizes the operating conditions of this improvement.

\paragraph{Making the next experience informative.}
An improving learner can also change what it asks, observes, or tests. EnvGen, OMNI-EPIC, and SimWorld Studio use learner performance to shape subsequent simulated experience~\citep{zala2024envgen,faldor2024omniepic,arxiv260509423}. For ICL, teaching--execution pairs near interpretation failures could target conflicting examples, unfamiliar compositions, or corrections with a limited scope. Physical trials then test whether the inferred requirement survives contact and embodiment changes.

A failed insertion makes the distinction concrete. An ambiguous target calls for clarification; an uncertain contact response calls for a diagnostic motion. Retaining when each intervention resolved uncertainty could reduce unnecessary demonstrations and trials on later assemblies. The improvement loop then links requirement inference, evidence acquisition, physical verification, and reuse of an acquisition strategy.

\paragraph{Collective knowledge evolution.}
S6 extends the learning cycle to exchanges between robots (Figure~\ref{fig:cover}). An arm can demonstrate a placement relation, a humanoid realize it under different kinematic constraints, and the receiver return a verified execution or correction. The exchanged lesson preserves task order and object relations while allowing grasps and balance control to change.

Shared progress and skill representations preserve task content across appearances~\citep{zakka2021xirl,xu2023xskill,kim2025uniskill}; multi-robot training supplies compatible competence~\citep{oxe2023,doshi2024crossformer}; morphology interpolation and kinematic constraints support physical realization~\citep{liu2022revolver,gupta2026kinematic}. Collective knowledge evolution adds a reciprocal loop: each execution helps revise knowledge another robot can use. Measuring transfer in both directions, with the teaching and practice required by each body, tests whether the exchange improves the group's subsequent learning.

Across individual and collective learning cycles, requirement adherence, correction efficiency, and transfer measure the benefit to subsequent acquisition. Interaction, intervention, and training costs measure the investment required to obtain it.

\begin{takeaway}
\takepoint Diversity in what examples teach, motion coverage, and model scale address different limits. Learning unfamiliar combinations and relation families requires more than recognizing new instances.
\takepoint Composition, selective correction, and information acquisition expand what an episode can teach over \mbox{an available motor repertoire.}
\takepoint Retained corrections should change later decisions through context and remain applicable when layouts change. Across longer learning cycles, the central outcome is a successor that learns more effectively from new teaching.
\end{takeaway}

\FloatBarrier
\suppressfloats[t]
\section{Conclusion}
\label{sec:conclusion}

ICL for robots addresses a gap between possessing reusable competence and knowing how to apply it in the present task. Teaching and interaction resolve that gap with deployed neural parameters fixed. Four interfaces carry this information into action: contextual action inference, geometric reference transfer, prediction of consequences, and skill or program execution. Their benefits depend on compatible representations, applicable interaction geometry, realizable predictions, and sufficient executor capabilities, respectively. Correspondence and memory connect the evidence across situations and time; grounding establishes whether it remains usable.

The central synthesis is that contextual learning depends on preserving task-relevant distinctions from evidence to execution. Loss during selection, interpretation, realization, or reuse calls for different data, representations, or control capabilities. Across manipulation and navigation, the corresponding tests ask whether changed teaching appropriately redirects behavior, its requirement survives physical change, and retained experience helps later attempts. Acquisition costs and motor competence determine the value and scope of these gains.

The next challenge is to make these dependencies transferable: infer unfamiliar combinations of requirements, revise only the constraints affected by feedback, and discard experience whose conditions no longer hold. Physical recursive self-improvement extends this agenda to a successor that acquires new tasks more effectively; collective knowledge evolution asks whether verified exchanges improve learning across embodiments. Both are assessed by what later learners acquire and the total teaching, interaction, and training effort required.
\paragraph{Image attribution.}
Figures combine original vectors, AI-assisted illustrations, and cited task imagery. AI-assisted scenes appear in Figures~\ref{fig:cover}, \ref{fig:mechanism-families}, \ref{fig:application-map}, \ref{fig:adaptation-sources}, \ref{fig:instruction-training}, and~\ref{fig:research-agenda}; the S6 robot illustrations in Figures~\ref{fig:cover} and~\ref{fig:stage-landscape} are original vector drawings. Source motion and task annotations are preserved.

Method imagery comes from Vid2Robot, SemAnCorr, Zero-WAM, and UniSkill~\citep{jain2024vid2robot,dong2026semancorr,zhou2026zerowam,kim2025uniskill}. Additional examples come from ICRT, MemoryVLA, WAM-TTT, DROC, MT3, and CorrectVLA~\citep{fu2024icrt,shi2025memoryvla,arxiv260706988,zha2023droc,dreczkowski2025mt3,arxiv260829967}; research examples use VIMA, Instant Policy, and HoloAssist~\citep{jiang2022vima,vosylius2024instantpolicy,holoassist2023}. Acquisition examples use DROID, UMI, Ego2Robot, and MimicGen~\citep{khazatsky2024droid,chi2024umi,wang2026ego2robot,mandlekar2023mimicgen}. Other recordings come from Galaxea Open-World~\citep{galaxea2025}, LIBERO-PRO~\citep{zhou2025liberopro}, \href{https://github.com/cheng-haha/GPT-Policy-Eval/blob/main/assets/hidden-goal.gif}{GPT-Policy}~\citep{cheng2026gptpolicyeval}, and \href{https://huggingface.co/datasets/dannyXSC/HumanAndRobot}{HumanAndRobot}.

Vid2Robot imagery is adapted under \href{https://creativecommons.org/licenses/by-sa/4.0/}{CC BY-SA 4.0}; SemAnCorr, Zero-WAM, UniSkill, LIBERO-PRO, VIMA, Instant Policy, HoloAssist, and DROID imagery under \href{https://creativecommons.org/licenses/by/4.0/}{CC BY 4.0}; and Galaxea imagery under \href{https://creativecommons.org/licenses/by-nc-sa/4.0/}{CC BY-NC-SA 4.0}. Editable sources include image provenance and institution-mark sources.

\phantomsection
\addcontentsline{toc}{section}{References}
\begingroup
\flushbottom
\setlength{\bibsep}{0.2pt plus .2pt minus .1pt}
\bibliographystyle{survey-unsrtnat}
\bibliography{references,feishu-arxiv,broad-intake,recount-references,frontier-integration,related-survey-additions}
\par
\endgroup
\appendix
\FloatBarrier
\suppressfloats[t]
\Needspace*{14\baselineskip}
\section{Complete Reference Index}
\label{app:study-index}
\setcounter{table}{0}
\renewcommand{\thetable}{\thesection\arabic{table}}
\renewcommand{\theHtable}{\thesection\arabic{table}}

\surveyanchor{discuss:tab:corpus-index}
Table~\ref{tab:corpus-index} links every reference to Figure~\ref{fig:corpus-trends}. Full titles, dates, primary sources, and classification details accompany the reference data.

\begingroup
\fontsize{9}{10.2}\selectfont\setlength{\tabcolsep}{3pt}
\renewcommand{\arraystretch}{1.02}
\setlength{\LTleft}{\fill}\setlength{\LTright}{\fill}
\setlength{\LTpre}{8pt plus 4pt}\setlength{\LTpost}{6pt}
\begin{longtable}{@{}p{.068\linewidth}p{.065\linewidth}p{.066\linewidth}>{\raggedright\arraybackslash}p{\dimexpr.801\linewidth-6\tabcolsep\relax}@{}}
\caption{Reference index for Figure~\ref{fig:corpus-trends}, grouped by first public year. Method families: C, context-conditioned policies; G, geometric transfer; W, world-model-based control; S, skill/agent execution. Other groups record data, evaluation, foundations, and training references.}\label{tab:corpus-index}\\
\toprule
\textbf{Year} & \textbf{Family} & \textbf{Count} & \textbf{References}\\\midrule
\endfirsthead
\caption[]{Complete reference assignments (continued).}\\
\toprule
\textbf{Year} & \textbf{Family} & \textbf{Count} & \textbf{References}\\\midrule
\endhead
\bottomrule
\endfoot
\midrule
\textbf{Total} & & \textbf{412} & \\
\bottomrule
\endlastfoot

2016 & C & 1 & \citep{duan2016rl2} \\
\addlinespace[1.5pt]
2017 & C & 4 & \citep{duan2017oneshot,mishra2017snail,finn2017mil,yu2017uposi} \\
 & S & 1 & \citep{xu2017ntp} \\
\addlinespace[1.5pt]
2018 & C & 4 & \citep{yu2018domainadaptive,james2018tec,pathak2018zeroshot,kumar2018rpf} \\
 & W & 2 & \citep{srinivas2018upn,ebert2018retrying} \\
 & S & 1 & \citep{savinov2018sptm} \\
\addlinespace[1.5pt]
2019 & C & 4 & \citep{rakelly2019pearl,zhou2019wtl,bonardi2019humans,zintgraf2019varibad} \\
\addlinespace[1.5pt]
2020 & C & 5 & \citep{dasari2020tosil,li2020focal,matsushima2020uncertainty,zhang2020metacure,yoo2020sparsepath} \\
 & G & 1 & \citep{argus2020flowcontrol} \\
\addlinespace[1.5pt]
2021 & C & 3 & \citep{mandi2021mosaic,kumar2021rma,pari2021vinn} \\
\addlinespace[1.5pt]
2022 & C & 6 & \citep{xu2022promptdt,laskin2022ad,jiang2022vima,jang2022bcz,yuan2022corro,brohan2022rt1} \\
 & G & 2 & \citep{valassakis2022dome,simeonov2023rndf} \\
 & W & 1 & \citep{mu2022domino} \\
 & S & 5 & \citep{liang2022codeaspolicies,singh2022progprompt,ahn2022saycan,huang2022monologue,shah2022lmnav} \\
\addlinespace[1.5pt]
2023 & C & 6 & \citep{oxe2023,bousmalis2023robocat,grigsby2023amago,bharadhwaj2023roboagent,zhao2023act,brohan2023rt2} \\
 & G & 4 & \citep{dipalo2023dinobot,vecerik2023robotap,vitiello2023pose,biza2023warping} \\
 & W & 1 & \citep{ni2023metadiffuser} \\
 & S & 9 & \citep{xu2023xskill,zha2023droc,liu2023reflect,ren2023knowno,tang2023saytap,chang2023awda,huang2023voxposer,wang2023mimicplay,chen2023a2nav} \\
\addlinespace[1.5pt]
2024 & C & 16 & \citep{fu2024icrt,jain2024vid2robot,dipalo2024keypoint,vosylius2024instantpolicy,kedia2024rhyme,kim2024openvla,black2024pi0,doshi2024crossformer,yin2024roboprompt,sridhar2024regent,fang2024kalm,elawady2024relic,octo2024,hpt2024,liu2024rdt,zhou2024nolo} \\
 & G & 4 & \citep{heppert2024ditto,zhang2024imop,liu2024magic,bahety2024screwmimic} \\
 & S & 11 & \citep{tziafas2024lrll,ahn2024autort,dai2024racer,liang2024lmpc,zhu2024orion,singh2024malmm,curtis2025proc3s,sarch2024ical,liang2024introplan,buoso2024select2plan,elnoor2024vlmgronav} \\
\addlinespace[1.5pt]
2025 & C & 18 & \citep{sridhar2025ricl,liu2025locoformer,shi2025memoryvla,sridhar2025memer,galaxea2025,nvidia2025gr00t,park2025demodiffusion,chen2025vivla,haldar2025pointpolicy,qu2025spatialvla,oh2025rip,torne2025ptp,shah2025mimicdroid,arxiv251114759,arxiv251222414,vuong2025actiontokenizer,arxiv250617811,wagenmaker2025exploration} \\
 & G & 8 & \citep{dreczkowski2025mt3,wang2025odil,chen2025manilong,allu2025hrt1,tang2025functo,defarias2025gift,wichitwechkarn2025annotationfree,li2025emp} \\
 & W & 2 & \citep{goswami2025osviwm,luo2025silvr} \\
 & S & 5 & \citep{merwe2025icpi,kim2025uniskill,arxiv250918597,jain2025transitions,li2025iclhf} \\
\addlinespace[1.5pt]
2026 & C & 56 & \citep{jiang2026robottt,yang2026icivla,wang2026icwm,torne2026mem,yin2026simplememvla,patel2026bpp,ding2026contextflow,qian2026synthicl,chen2026zeva,wang2026instantfold,palma2026bicicle,wu2026adatracker,generalist2026gen15,skild2026s1,gao2026gmp,huang2026mint,she2026matchingpolicy,arxiv260805738,xu2026stellavla,oh2026vlaff,nguyen2026iclr,son2026seetraceact,choi2026ponderpounce,fan2026rtcf} \\
 &  &  & \citep{li2026rememvla,cherepanov2026muvla,yang2026eventvla,shah2026halo,arxiv260826821,arxiv260821204,yu2026walloss05,xu2026bimanualscaling,xiaomi2026robotics1,arxiv260313528,arxiv260415483,fateh2026histat,mai2026crvlaforce,arxiv260825798,sato2026sail,huang2026fastthinkact,arxiv260212063,arxiv260206508,arxiv260919796,arxiv260920648,arxiv260920659,arxiv260924411,arxiv260923432,pala2026membodied} \\
 &  &  & \citep{wang2026tpflow,yi2026arms,knowin2026glow,arxiv260930134,arxiv260920646,arxiv260930092,arxiv260920820,arxiv260930828} \\
 & G & 5 & \citep{dong2026semancorr,gupta2026kinematic,zhu2026sparsedense,thompson2026parttransfer,arxiv260930404} \\
 & W & 16 & \citep{arxiv260706988,he2026demojepa,chen2026host,zhou2026zerowam,arxiv260718840,park2026recap,shi2026memoryvlapp,core2026realignment,arxiv260816885,bi2026motus2,arxiv260919824,arxiv260919315,arxiv260921740,arxiv260921122,arxiv260911561,arxiv260918016} \\
 & S & 59 & \citep{chen2026showharness,lu2026aspire,chen2026architect,wang2026shaper,arxiv260619419,arxiv260817209,xiao2026enpire,arxiv260816590,cheng2026gptpolicyeval,arxiv260829967,arxiv260322435,physcap2026,arxiv260829537,arxiv260708448,chen2026steerable,liu2026guava,galanti2026physicalagency,ju2026embodiskill,xie2026uniskillrepo,kumar2026aor,hu2026orchestrating,cui2026roboclaw,arxiv260831167,arxiv260819891} \\
 &  &  & \citep{arxiv260601247,arxiv260817129,arxiv260511951,chen2026volo,arxiv260514504,arxiv260920791,arxiv260920388,arxiv260919906,arxiv260919512,arxiv260919554,arxiv260914633,noematrix2026roborsi,chen2026eta,huang2026roboharness,li2026cmmrvln,liu2026hamvln,arxiv260922966,huang2026affordanceharness,li2026roboharnessk1,li2026racap,you2026embodiedswe,liu2026navprobe,li2026talk2escape,zhang2026waa} \\
 &  &  & \citep{arxiv260930249,arxiv260804933,arxiv260928798,arxiv260926408,arxiv260929166,arxiv260911308,arxiv260930428,arxiv260930233,arxiv260931337,arxiv260930594,arxiv260931112} \\
\addlinespace[1.5pt]
\midrule
\multicolumn{4}{@{}l}{\textbf{Data resources}}\\*
 & & 44 & \citep{walke2023bridge,khazatsky2024droid,fang2023rh20t,agibot2025,chi2024umi,hoi4d2022,hot3d2025,egodex2025,ohkawa2026yubi,egoscale2026,grauman2021ego4d,damen2020epic100,grauman2023egoexo4d,goyal2017something,assembly1012022,holoassist2023,mandlekar2023mimicgen,robocasa2024,arxiv260724744,wang2026ego2robot,simpleai2026hifiumi,zhaxizhuoma2024fastumi,ha2024umilegs,luo2026omniumi} \\
 & &  & \citep{bridge2021,agibot2026release,agibot2026corrections,robomind2024,robomind2025v2,robocoin2025,li2026aceego,punamiya2026egoverse,aoe2026openaoe,cao2026acedata,qian2026robotok,lin2026simdex,zhou2025yoto,zhou2026bidemosyn,robosplat2025,arxiv260509423,arxiv260304356,zala2024envgen,faldor2024omniepic,arxiv260921229} \\
\midrule
\multicolumn{4}{@{}l}{\textbf{Evaluation}}\\*
 & & 39 & \citep{agia2024sentinel,xu2025faildetect,arxiv241104549,duan2024aha,arxiv260818618,zhou2026imitator,ruoss2025lmact,dai2026robomme,sun2026memobench,zhou2025liberopro,arxiv260217659,arxiv260317300,arxiv260602277,arxiv260830536,fei2025generallevel,james2019rlbench,liu2023libero,gu2023maniskill2,chen2025robotwin2,arxiv260518746,arxiv260623085,arxiv260818701,arxiv260814441,jiang2026benchmarkaudit} \\
 & &  & \citep{fei2025liberoplus,yu2019metaworld,gu2026roboreel,cherepanov2025mikasa,arxiv260813049,anderson2018naveval,arxiv260924778,arxiv250912718,arxiv260618847,wu2026robostressbench,huang2026vidpairhalluc,zhang2026cma,arxiv260925636,arxiv260928952,arxiv260930608} \\
\midrule
\multicolumn{4}{@{}l}{\textbf{Foundations and surveys}}\\*
 & & 40 & \citep{chan2022distribution,vonoswald2022gradient,min2022metaicl,brown2020fewshot,argall2009survey,ravichandar2020survey,dong2024iclsurvey,moeini2025icrlsurvey,ma2026humanvideosurvey,run2026nonstationary,hou2026worldmodel,wang2026vladata,domae2026embodimentgap,fikes1971strips,brooks1985subsumption,khatib1987operational,kaelbling2011hpn,marzinotto2014bt,lesort2019continual,hospedales2022metalearning,hu2025agentmemory,survey2026wam,li2025embodiedwm,chaumette2006visualservo} \\
 & &  & \citep{vaswani2017attention,reddy2023abrupt,dehaan2019causal,raventos2023diversity,cakmak2012questions,xie2021bayesian,chen2024parallel,min2022demonstrations,garg2022functions,shen2023gradient,schaeffer2023mirage,duan2026rsisurvey,jena2026weightsskills,lu2026wamsurvey,li2026demonstrationiclsurvey,shao2025vlasurvey} \\
\midrule
\multicolumn{4}{@{}l}{\textbf{Training and representations}}\\*
 & & 29 & \citep{sermanet2017tcn,zakka2021xirl,ma2023eureka,ma2024dreureka,liu2022revolver,levine2015visuomotor,kalashnikov2018qtopt,sharma2023medal,bahl2022whirl,torabi2018bco,feng2026regrind,yuan2024robopoint,mtopt2021,openai2026astratraining,ghasemipour2025selfimproving,arxiv260519242,arxiv260401985,arxiv250706219,ross2011dagger,laskey2017dart,hoque2021thrifty,arxiv260421741,arxiv260101075,arxiv260618960} \\
 & &  & \citep{arxiv260819059,arxiv260900619,arxiv260900950,chung2022flan,arxiv260930889} \\
\end{longtable}
\par\phantomsection\label{floatend:tab:corpus-index}
\endgroup

\FloatBarrier
\suppressfloats[t]
\section{Method and Resource Comparisons}
\label{app:comparisons}
\setlength{\intextsep}{3pt plus 1pt minus 1pt}
\setcounter{table}{0}
\renewcommand{\thetable}{\thesection\arabic{table}}
\renewcommand{\theHtable}{\thesection\arabic{table}}

The following tables provide per-study detail for the main-text classifications. They follow the progression from demonstration correspondence to retained experience and the data and comparisons \mbox{needed to evaluate it.}

\begingroup
\surveytable
\fontsize{9.5}{11.5}\selectfont
\renewcommand{\arraystretch}{1.15}
\setlength{\LTleft}{\fill}\setlength{\LTright}{\fill}
\setlength{\LTpre}{8pt}\setlength{\LTpost}{6pt}
\begin{longtable}{@{}>{\raggedright\arraybackslash}p{0.26\linewidth}*{2}{>{\raggedright\arraybackslash}p{\dimexpr0.37\linewidth-2\tabcolsep\relax}}@{}}
\caption{Correspondence mechanisms compared by the entities they relate and their use in control. The same matched points can condition action inference or specify a reference for a separate executor.}\label{tab:correspondence-interfaces}\\
\toprule
\textbf{Method} & \textbf{Correspondence} & \textbf{Control interface}\\
\midrule
\endfirsthead
\caption[]{Correspondence interfaces (continued).}\\
\toprule
\textbf{Method} & \textbf{Correspondence} & \textbf{Control interface}\\
\midrule
\endhead
\bottomrule
\endfoot

\rowcolor{panel}[0pt][0pt]\multicolumn{3}{@{}l@{}}{\strut\textbf{Correspondence conditions or corrects action generation}}\\*
MatchingPolicy~\citep{she2026matchingpolicy} & Demonstration--query 3D points & Graph-conditioned gripper motion\\
RTCF~\citep{fan2026rtcf} & Current and recorded progress & Bounded correction\\
\addlinespace[3pt]
\rowcolor{panel}[0pt][0pt]\multicolumn{3}{@{}l@{}}{\strut\textbf{Correspondence specifies an executable motion or constraint}}\\*
RoboTAP~\citep{vecerik2023robotap} & Stage-specific points & Visual-servo reference tracking\\
FUNCTO~\citep{tang2025functo} & Grasp; function; center & Function-preserving tool motion\\
MAGIC~\citep{liu2024magic} & Shape; contact geometry & Retarget; simulate\\
SemAnCorr~\citep{dong2026semancorr} & Surfaces; local frames & Grasp; relative motion\\
Sparse Meets Dense~\citep{zhu2026sparsedense} & Changing contact pairs & Constraint; pose updates\\
\end{longtable}
\par\phantomsection\label{floatend:tab:correspondence-interfaces}
\endgroup

\FloatBarrier
\surveyanchor{discuss:tab:correspondence-interfaces}
The same matched geometry can have different computational roles. Table~\ref{tab:correspondence-interfaces} separates correspondence that conditions action inference from correspondence that supplies an executable motion reference.

\begingroup
\surveytable
\fontsize{9.5}{11.5}\selectfont
\renewcommand{\arraystretch}{1.15}
\setlength{\LTleft}{\fill}\setlength{\LTright}{\fill}
\setlength{\LTpre}{8pt}\setlength{\LTpost}{6pt}
\begin{longtable}{@{}>{\raggedright\arraybackslash}p{.24\linewidth}>{\raggedright\arraybackslash}p{.32\linewidth}>{\raggedright\arraybackslash}p{\dimexpr.44\linewidth-4\tabcolsep\relax}@{}}
\caption{Memory mechanisms compared by representation, access, and update regime. Groups (a)--(d) retain evidence with fixed neural parameters; group (e) adapts weights. A fixed test bank supports \mbox{retrieval without adding records.}}\label{tab:context-memory}\\
\toprule
\textbf{Method} & \textbf{Retained evidence} & \textbf{Access and update} \\
\midrule
\endfirsthead
\caption[]{Memory mechanisms (continued).}\\
\toprule
\textbf{Method} & \textbf{Retained evidence} & \textbf{Access and update} \\
\midrule
\endhead
\bottomrule
\endfoot
\rowcolor{panel}[0pt][0pt]\multicolumn{3}{@{}l@{}}{\strut\textbf{(a) Demonstration records}}\\*
KAT~\citep{dipalo2024keypoint} & Keypoint--action examples & Load before rollout \\
ICRT~\citep{fu2024icrt} & Demonstrations; own trajectory & Read prefix; append execution \\
RICL~\citep{sridhar2025ricl} & Retrieved demonstrations & Query-dependent selection \\
ReCAP~\citep{park2026recap} & Trajectory retrieval pool & Extend pool; retrieve each step \\
RTCF~\citep{fan2026rtcf} & Successful trajectories & Phase retrieval; fixed test bank \\
TraceFlow~\citep{arxiv260920646} & Success and failure traces & Progress-aligned guidance; rollout admission \\
\addlinespace[3pt]
\rowcolor{panel}[0pt][0pt]\multicolumn{3}{@{}l@{}}{\strut\textbf{(b) Interaction histories and inferred state}}\\*
RMA~\citep{kumar2021rma} & Recent state--action history & Online dynamics estimation \\
AMAGO~\citep{grigsby2023amago} & Trial history and feedback & Append across episodes \\
ReLIC (navigation)~\citep{elawady2024relic} & Multi-episode visual history & Attend across trials \\
LocoFormer~\citep{liu2025locoformer} & Recurrent rollout state & Carry across trials \\
ICWM~\citep{wang2026icwm} & Action--effect probes & Probe before execution \\
\addlinespace[3pt]
\rowcolor{panel}[0pt][0pt]\multicolumn{3}{@{}l@{}}{\strut\textbf{(c) Execution memory}}\\*
MemoryVLA~\citep{shi2025memoryvla} & Perceptual; cognitive tokens & Retrieve; merge \\
MemER~\citep{sridhar2025memer} & Historical visual keyframes & Retrieve; predict subtask \\
MEM~\citep{torne2026mem} & Recent views; semantic history & Recent and long-term \\
ReMem-VLA~\citep{li2026rememvla} & Recurrent memory queries & Frame and chunk updates \\
MemoryVLA++~\citep{shi2026memoryvlapp} & Memory bank; imagined future & Retrieve; predict \\
$\mu$VLA~\citep{cherepanov2026muvla} & Recurrent memory tokens & Carry state; reset per episode \\
EventVLA~\citep{yang2026eventvla} & Anchors; event keyframes & Event-triggered writes \\
SimpleMemVLA~\citep{yin2026simplememvla} & Timestamped video window & Resample per decision \\
AGM~\citep{arxiv260829537} & Verified subgoal achievements & Completion-verified update \\
Workspace Models~\citep{arxiv260920820} & Latent history tokens & Saliency-supervised compression \\
Mimir; OCC4M~\citep{arxiv260804933,arxiv260928798} & Task state; persistent entities & Goal binding; feedback-supported updates \\
2AM~\citep{arxiv260911308} & Agent-side interaction history & Compile current intent into 2D cues \\

\addlinespace[3pt]
\rowcolor{panel}[0pt][0pt]\multicolumn{3}{@{}l@{}}{\strut\textbf{(d) Reusable guidance}}\\*
DROC~\citep{zha2023droc} & Distilled correction knowledge & Feedback update; retrieval \\
\addlinespace[3pt]
\rowcolor{panel}[0pt][0pt]\multicolumn{3}{@{}l@{}}{\strut\textbf{(e) Parameter adaptation through fast-weight memory}}\\*
RoboTTT~\citep{jiang2026robottt} & Video and rollout history & Sequential weight updates \\
WAM-TTT~\citep{arxiv260706988} & Human demonstration video & Adapt; freeze \\
\end{longtable}
\par\phantomsection\label{floatend:tab:context-memory}
\endgroup

\FloatBarrier
\surveyanchor{discuss:tab:context-memory}
Table~\ref{tab:context-memory} compares what each system retains, when it accesses that evidence, and how the retained state changes. The comparison links decision-relevant information to its storage and update mechanism.

\begingroup
\surveytable
\fontsize{9.5}{11.5}\selectfont
\renewcommand{\arraystretch}{1.15}
\setlength{\LTleft}{\fill}\setlength{\LTright}{\fill}
\setlength{\LTpre}{8pt}\setlength{\LTpost}{6pt}
\begin{longtable}{@{}>{\raggedright\arraybackslash}p{.35\linewidth}*{5}{>{\centering\arraybackslash}p{\dimexpr.13\linewidth-2\tabcolsep\relax}}@{}}
\caption{Artifacts revised under fixed neural weights. Bullets mark updates; dashes mark undocumented updates for the specified branch. Plans specify procedures, guidance stores lessons, and skills are callable routines. Programs include controller settings; oversight includes context construction and verification. ZS denotes the zero-shot \mbox{video branch (Section~\ref{sec:agentic-adaptation}).}}\label{tab:agentic-mechanisms}\\
\toprule
\textbf{System and branch} & \textbf{Plan} & \textbf{Guidance} & \textbf{Skills} & \textbf{Program} & \textbf{Oversight} \\
\midrule
\endfirsthead
\caption[]{Artifacts revised under fixed neural weights (continued).}\\
\toprule
\textbf{System and branch} & \textbf{Plan} & \textbf{Guidance} & \textbf{Skills} & \textbf{Program} & \textbf{Oversight} \\
\midrule
\endhead
\bottomrule
\endfoot
\rowcolor{panel}[0pt][0pt]\multicolumn{6}{@{}l@{}}{\strut\textbf{(a) Constructing and revising the current procedure}}\\*
Inner Monologue~\citep{huang2022monologue} & $\bullet$ & -- & -- & -- & -- \\
Code as Policies~\citep{liang2022codeaspolicies} & -- & -- & $\bullet$ & $\bullet$ & -- \\
ProgPrompt~\citep{singh2022progprompt} & -- & -- & -- & $\bullet$ & -- \\
VoxPoser (fixed dynamics)~\citep{huang2023voxposer} & $\bullet$ & -- & -- & $\bullet$ & -- \\
PRoC3S~\citep{curtis2025proc3s} & -- & -- & -- & $\bullet$ & -- \\
MALMM~\citep{singh2024malmm} & $\bullet$ & -- & -- & $\bullet$ & -- \\
Show-Harness (video, ZS)~\citep{chen2026showharness} & $\bullet$ & -- & -- & -- & -- \\
GPT-Policy~\citep{cheng2026gptpolicyeval} & -- & -- & -- & $\bullet$ & -- \\
ICPI~\citep{merwe2025icpi} & -- & -- & -- & $\bullet$ & -- \\
AOR~\citep{kumar2026aor} & -- & -- & -- & $\bullet$ & -- \\
ENPIRE (code branch)~\citep{xiao2026enpire} & -- & -- & -- & $\bullet$ & -- \\
\addlinespace[3pt]
\rowcolor{panel}[0pt][0pt]\multicolumn{6}{@{}l@{}}{\strut\textbf{(b) Acquiring examples, guidance, and reusable behavior}}\\*
DROC~\citep{zha2023droc} & $\bullet$ & $\bullet$ & -- & $\bullet$ & -- \\
ICAL (TEACh, retrieval)~\citep{sarch2024ical} & $\bullet$ & $\bullet$ & -- & $\bullet$ & -- \\
LRLL~\citep{tziafas2024lrll} & -- & -- & $\bullet$ & $\bullet$ & $\bullet$ \\
LYRA~\citep{arxiv250918597} & -- & -- & $\bullet$ & $\bullet$ & -- \\
Uni-Skill (deployment)~\citep{xie2026uniskillrepo} & -- & $\bullet$ & $\bullet$ & $\bullet$ & -- \\
CaP-Agent0 (CaP-X)~\citep{arxiv260322435} & -- & -- & $\bullet$ & $\bullet$ & -- \\
RATs~\citep{arxiv260619419} & $\bullet$ & $\bullet$ & $\bullet$ & $\bullet$ & -- \\
RoboRSI (code skills)~\citep{noematrix2026roborsi} & $\bullet$ & -- & $\bullet$ & $\bullet$ & -- \\
ASPIRE~\citep{lu2026aspire} & -- & $\bullet$ & -- & $\bullet$ & -- \\
ARCHITECT~\citep{chen2026architect} & -- & $\bullet$ & -- & $\bullet$ & -- \\
Teach and Grow (pilot)~\citep{arxiv260817209} & -- & -- & $\bullet$ & -- & -- \\
Harness VLA (memory)~\citep{arxiv260708448} & $\bullet$ & $\bullet$ & -- & -- & -- \\
Hi-VLA (episode summary)~\citep{hu2026orchestrating} & -- & $\bullet$ & -- & -- & -- \\
\addlinespace[3pt]
\rowcolor{panel}[0pt][0pt]\multicolumn{6}{@{}l@{}}{\strut\textbf{(c) Revising context construction and execution oversight}}\\*
SHAPER~\citep{wang2026shaper} & -- & $\bullet$ & -- & -- & $\bullet$ \\
Zetta~\citep{arxiv260816590} & -- & $\bullet$ & $\bullet$ & -- & $\bullet$ \\
\end{longtable}
\par\phantomsection\label{floatend:tab:agentic-mechanisms}
\endgroup

\FloatBarrier
\surveyanchor{discuss:tab:agentic-mechanisms}
Table~\ref{tab:agentic-mechanisms} distinguishes revisions to procedures, retained guidance, and callable skills. These artifacts can change together: RoboRSI retains both revised plans and executable routines.


\begingroup
\surveytable
\fontsize{9.5}{11.5}\selectfont
\renewcommand{\arraystretch}{1.0}
\setlength{\LTleft}{\fill}\setlength{\LTright}{\fill}
\setlength{\LTpre}{8pt}\setlength{\LTpost}{6pt}
\begin{longtable}{@{}>{\raggedright\arraybackslash}p{.30\linewidth}>{\raggedright\arraybackslash}p{.18\linewidth}*{2}{>{\raggedright\arraybackslash}p{\dimexpr.26\linewidth-3\tabcolsep\relax}}@{}}
\caption{Data resources compared by demonstrator, motion supervision, and role in contextual control. Motion may be measured, estimated, retargeted, simulated, or synthesized; human $\to$ robot denotes robot imagery synthesized from human footage. The final column identifies potential uses of the available records in contextual learning. Est., Sim., and Synth. denote estimated, simulated, and synthesized.}\label{tab:data}\\
\toprule
\textbf{Source} & \textbf{Visual record} & \textbf{Motion supervision} & \textbf{Available learning evidence} \\
\midrule
\endfirsthead
\caption[]{Data resources (continued).}\\
\toprule
\textbf{Source} & \textbf{Visual record} & \textbf{Motion supervision} & \textbf{Available learning evidence} \\
\midrule
\endhead
\bottomrule
\endfoot
\rowcolor{panel}[0pt][0pt]\multicolumn{4}{@{}l@{}}{\strut\textbf{Robot and cross-domain recordings}}\\*
Open X-Emb.~\citep{oxe2023} & Robot & Robot control & Robot action targets \\
DROID~\citep{khazatsky2024droid} & Robot & Robot control & Robot action targets \\
AgiBot World~\citep{agibot2025} & Robot & Robot control & Stages; corrections \\
BridgeData V2~\citep{walke2023bridge} & Robot & Robot control & Robot action targets \\
RoboSet~\citep{bharadhwaj2023roboagent} & Robot & Robot control & Task groups \\
PrimeBot, XR-2~\citep{xu2026bimanualscaling} & Human + robot & Robot; gripper & Separate corpora \\
RH20T~\citep{fang2023rh20t} & Human + robot & Robot control & Human--robot pairs \\
ICRT data~\citep{fu2024icrt} & Robot & Robot control & Robot demonstrations \\
MimicPlay~\citep{wang2023mimicplay} & Human + robot & Robot control & Separate play \\
\rowcolor{panel}[0pt][0pt]\multicolumn{4}{@{}l@{}}{\strut\textbf{Instrumented human recordings}}\\*
UMI~\citep{chi2024umi} & Human & Gripper motion & Motion correspondence \\
YUBI~\citep{ohkawa2026yubi} & Human & Gripper motion & Motion correspondence \\
HiFi-UMI~\citep{simpleai2026hifiumi} & Human & Gripper motion & Motion correspondence \\
ACE-Data-0~\citep{cao2026acedata} & Human & Human motion & Contact; procedure \\
\rowcolor{panel}[0pt][0pt]\multicolumn{4}{@{}l@{}}{\strut\textbf{Human activity video}}\\*
Ego4D~\citep{grauman2021ego4d} & Human & No robot actions & Behavior pretraining \\
Ego-Exo4D~\citep{grauman2023egoexo4d} & Human & No robot actions & Cross-view \\
EPIC-KITCHENS-100~\citep{damen2020epic100} & Human & No robot actions & Behavior pretraining \\
Something-Something~\citep{goyal2017something} & Human & No robot actions & Action clips \\
EgoVerse~\citep{punamiya2026egoverse} & Human & Est. hand motion & Behavior pretraining \\
Open-AoE~\citep{aoe2026openaoe} & Human & Est. hand motion & Action segments \\
\rowcolor{panel}[0pt][0pt]\multicolumn{4}{@{}l@{}}{\strut\textbf{Synthesis}}\\*
MimicGen~\citep{mandlekar2023mimicgen} & Robot & Sim. control & Query synthesis \\*
HumanGen~\citep{zhou2026zerowam} & Human + robot & Robot control & Synth. support \\*
Ego2Robot~\citep{wang2026ego2robot} & Human $\to$ robot & Retargeted motion & Synth. query \\
\end{longtable}
\par\phantomsection\label{floatend:tab:data}
\endgroup

\FloatBarrier
\surveyanchor{discuss:tab:data}
Table~\ref{tab:data} connects data provenance to action correspondence. Measured robot commands, estimated human motion, and synthesized robot imagery offer different supervision even when they depict the same task.

\FloatBarrier
\suppressfloats[t]
\section{Related Survey Perspectives}
\label{app:related-surveys}
\setcounter{table}{0}
\renewcommand{\thetable}{\thesection\arabic{table}}
\renewcommand{\theHtable}{\thesection\arabic{table}}

\surveyanchor{discuss:tab:related-surveys}
Table~\ref{tab:related-surveys} compares related surveys by scope and organizing principle. ICL for robots connects their perspectives through the evidence and interfaces that change execution.

\begingroup
\surveytable
\fontsize{9.5}{11.5}\selectfont
\renewcommand{\arraystretch}{1.24}
\setlength{\LTleft}{\fill}\setlength{\LTright}{\fill}
\begin{longtable}{@{}>{\raggedright\arraybackslash}p{.28\linewidth}>{\raggedright\arraybackslash}p{.30\linewidth}>{\raggedright\arraybackslash}p{\dimexpr.42\linewidth-4\tabcolsep\relax}@{}}
\caption{Related surveys by scope, organizing axes, and analytical emphasis.}
\label{tab:related-surveys}\\
\toprule
\textbf{Review focus} & \textbf{Organizing axes} & \textbf{Analytical emphasis}\\
\midrule
\endfirsthead
\caption[]{Related surveys (continued).}\\
\toprule
\textbf{Review focus} & \textbf{Organizing axes} & \textbf{Analytical emphasis}\\
\midrule
\endhead
\bottomrule
\endfoot
\rowcolor{panel}[0pt][0pt]\multicolumn{3}{@{}l@{}}{\strut\textbf{Demonstration and contextual adaptation}}\\*
Learning from demonstration (2009)~\citep{argall2009survey} & Demonstrator, problem space, policy derivation & Acquiring examples and deriving policies\\
\addlinespace[3pt]
Learning from demonstration (2020)~\citep{ravichandar2020survey} & Teaching interface; policy, reward, or plan & Matching teaching and learning to the task\\
\addlinespace[3pt]
In-context learning~\citep{dong2024iclsurvey} & Training, prompting, mechanisms & Examples as language-model context\\
\addlinespace[3pt]
In-context RL~\citep{moeini2025icrlsurvey} & Pretraining; context; models & History-based adaptation with fixed weights\\
\addlinespace[3pt]
Nonstationary ICRL~\citep{run2026nonstationary} & Shift type; timing; evidence & Validity of past evidence after changes\\
\addlinespace[3pt]
Demonstration-driven manipulation ICL~\citep{li2026demonstrationiclsurvey} & Context, inference target, adaptation, transfer & How examples and priors support manipulation\\
\rowcolor{panel}[0pt][0pt]\multicolumn{3}{@{}l@{}}{\strut\textbf{Data and embodiment}}\\*
Human-video learning~\citep{ma2026humanvideosurvey} & Task, observation, and action transfer & Acquiring skills from video-derived evidence\\
\addlinespace[3pt]
VLM-based VLA models~\citep{shao2025vlasurvey} & Monolithic/hierarchical models & From multimodal features to actions\\
\addlinespace[3pt]
VLA data infrastructure~\citep{wang2026vladata} & Data, benchmarks, data engines & Data and evaluation for robot learning\\
\addlinespace[3pt]
The embodiment gap~\citep{domae2026embodimentgap} & Shared structure; adaptation stage & Reuse and robot-specific requirements\\
\rowcolor{panel}[0pt][0pt]\multicolumn{3}{@{}l@{}}{\strut\textbf{Predictive control and reusable skills}}\\*
World models for robot learning~\citep{hou2026worldmodel} & Policy coupling, simulation, representations & Prediction for learning, planning, and evaluation\\
\addlinespace[3pt]
World Action Models~\citep{survey2026wam} & Cascaded/joint models; conditioning, decoding & Coupling future-state prediction and action generation\\
\addlinespace[3pt]
World-action learning and control~\citep{lu2026wamsurvey} & Transition models, architectures, training & Predictive control across robot applications\\
\addlinespace[3pt]
Weights or Skills?~\citep{jena2026weightsskills} & Parameters; skills; revision & Capability representation and reuse\\
\midrule
\textbf{ICL for robots (this survey)} & Context-to-action interfaces and transfer assumptions & Teaching, physical realization, and experience reuse across tasks\\
\end{longtable}
\par\phantomsection\label{floatend:tab:related-surveys}
\endgroup

\end{document}